%% file: neurips_2026.tex
\documentclass{article}

\usepackage[main, final]{neurips_2026}

\usepackage[utf8]{inputenc}
\usepackage[T1]{fontenc}
\usepackage{url}
\usepackage{booktabs}
\usepackage{amsfonts}
\usepackage{nicefrac}
\usepackage{microtype}
\usepackage[table]{xcolor}
\usepackage{graphicx}
\usepackage{amsmath}
\usepackage{subcaption}
\usepackage{pdfpages}
\usepackage{adjustbox}
\usepackage{multirow}
\usepackage{enumitem}
\usepackage{makecell}
\IfFileExists{fontawesome5.sty}{\usepackage{fontawesome5}}{}
\usepackage{algorithm}
\usepackage{algpseudocode}
\usepackage{wrapfig}
\usepackage{placeins}

\usepackage{hyperref}
\usepackage{etoc}
\hypersetup{
    colorlinks=true,
    linkcolor=blue,
    citecolor=blue,
    urlcolor=blue,
    filecolor=blue
}

\definecolor{linkpink}{HTML}{D000A0}
\newcommand{\pinklink}[2]{%
    \href{#1}{\textcolor{linkpink}{\nolinkurl{#2}}}%
}

\definecolor{deltagreen}{HTML}{2E9B5B}
\definecolor{deltared}{HTML}{D2453C}
\definecolor{revgreen}{HTML}{1B7A3D}
\definecolor{tbdblue}{HTML}{1F5FBF}

\newcommand{\pmstd}[1]{%
    \ensuremath{{\scriptstyle\,\pm\,#1}}%
}

\title{Beyond Flat Walks: Compositional Abstraction for Autoregressive Graph Generation}

\author{%
  Kaiwen Bian$^{*,1,2}$ \quad
  Andrew H. Yang$^{*,1}$ \quad 
  Yusu Wang$^{\dagger,1}$  \quad \\
  \textbf{Ali Parviz$^{\dagger,1}$ \quad
  Gal Mishne$^{\dagger,1}$} \\
  $^{1}$University of California San Diego \quad
  $^{2}$Salk Institute for Biological Studies \\
  $^{*}$Equal contribution. \quad
  $^{\dagger}$Equal supervision. \\
  \texttt{\{kbian,\,any012,\,yusuwang,\,alparviz,\,gmishne\}@ucsd.edu}
}

\begin{document}
\etocdepthtag{main}

\maketitle

\begin{center}
\small
Website: \pinklink{https://kbian.org/MOSAIC-website/}{kbian.org/MOSAIC-website}\\
Code: \pinklink{https://github.com/KevinBian107/MOSAIC}{github.com/KevinBian107/MOSAIC}
\end{center}

\begin{abstract}
Autoregressive graph generators typically linearize graphs through flat node-edge walks, which can obscure higher-order structure in large and complex graphs. We introduce a hierarchical graph abstraction and coarse-to-fine generation pipeline that separates coarsening, tokenization, and decoding. This exposes three design axes: community coarsening, hierarchical tokenization, and order randomization. Across these axes, traversal-based hierarchy consistently improves distributional and structural fidelity, while its V.U.N.\ cost depends strongly on the coarsening and randomization policy. In particular, the fidelity--V.U.N.\ trade-off observed under within-community randomization is substantially reduced when traversal order is randomized across the full hierarchy. With a fixed backbone, training protocol, and matched training time across five molecular and five non-molecular benchmarks, hierarchical serialization yields consistent distributional and structural gains on all molecular datasets, while the non-molecular results further show that these gains depend on how well the supplied hierarchy aligns with structure in the target distribution. Together, these results characterize when and how hierarchical serialization benefits autoregressive graph generation.
\end{abstract}

\section{Introduction}\label{sec:intro}

Recent progress in generative modeling has been driven by scalable architectures such as transformers and diffusion models, together with the availability of increasingly large-scale datasets. These advances have transformed generation across multiple modalities, including images~\citep{rombach2022high}, natural language~\citep{gpt3,touvron2023llama,touvron2023llama2}, and audio~\citep{dhariwal2020jukebox,huang2024audiogpt}. Motivated by the importance of graphs in scientific applications, recent work has extended diffusion-based generation to graph-structured data, achieving strong performance on molecular and general graph synthesis tasks~\citep{jo2022scoregraph,vignac2023digress, mathiasen2024reducing,ingraham2019generative, brockschmidt2018generative}. However, most diffusion-based graph models rely on dense adjacency representations and iterative denoising procedures, resulting in quadratic memory complexity and additional computational overhead from structural encodings such as spectral features, limiting their scalability to large graphs.

\begin{figure}[!t]
  \centering
  \includegraphics[width=\linewidth]{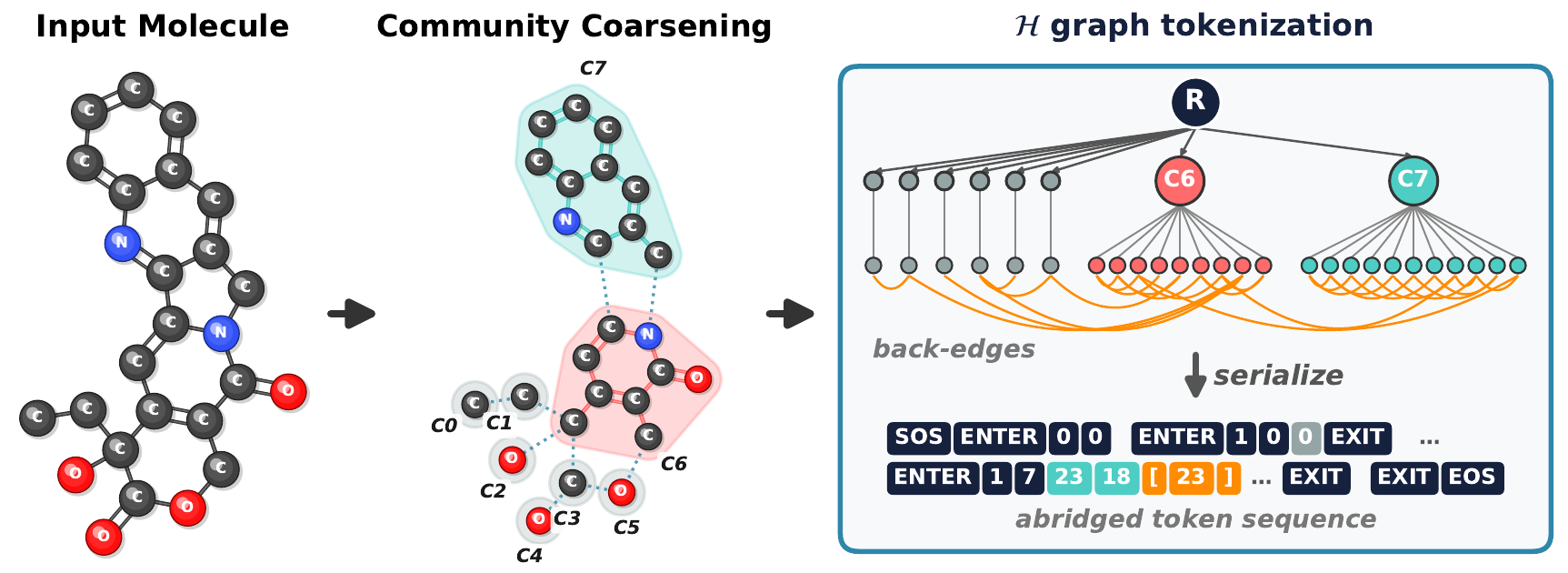}
  \caption{\textbf{Encoding pipeline, illustrated on camptothecin.} Motif coarsening gives two non-singleton communities (C6, C7) and six singleton communities (C0--C5), defining $\mathcal{H}$. \emph{HDT} serializes the hierarchy depth-first: \texttt{[ENTER]}/\texttt{[EXIT]} delimit communities, and back-edge brackets connect each atom to previously emitted neighbors within or across communities (orange arcs). The sequence is abridged, retains original atom indices, and omits atom- and bond-type tokens.}
  \label{fig:main_pipeline_encode}
\end{figure}

Autoregressive graph generation provides an alternative paradigm in which graphs are constructed sequentially through node and edge generation decisions~\citep{liao2019efficient,you2018graphrnn}. Early approaches relied on handcrafted sequentializations and recurrent architectures, enabling structurally valid generation on small graphs but struggling to capture long-range dependencies and global topology, which led to weaker performance compared to recent diffusion-based methods~\mbox{\citep{vignac2023digress,jo2023graph, lim2020scaffold}}. More recent work has explored adapting autoregressive transformers to graph generation by representing graphs as token sequences compatible with causal language modeling~\citep{chen2025autograph}. Existing tokenization strategies typically linearize graphs through random walks~\citep{bojchevski2018netgan} or depth-first traversals~\citep{goyal2020graphgen}. Although these representations preserve local connectivity, they often fail to capture the hierarchical and compositional structure of graphs, such as motifs, rings, and communities~\citep{jin2018jtvae,newman2004modularity}, making it increasingly difficult to model large and structurally complex graphs using flat token sequences alone.

In this work, we introduce a hierarchical graph abstraction, $\mathcal{H}$, and a coarse-to-fine framework for autoregressive graph generation (Figures~\ref{fig:main_pipeline_encode} and~\ref{fig:main_pipeline_decode}). The framework first coarsens a graph into a hierarchy of communities, then serializes that hierarchy into a token sequence, and finally decodes the sequence exactly back into graph space. This formulation exposes three design axes: \emph{how the graph is coarsened}, \emph{how the hierarchy is tokenized}, and \emph{how ordering is randomized during training}. We study domain-agnostic and chemistry-aware coarsening strategies together with several hierarchical tokenizers. Our main tokenizer, Hierarchical Depth Traversal (\emph{HDT}), traverses the hierarchy depth-first while representing both intra- and inter-community connectivity through standard back edges. We also consider Hierarchical SENT (\emph{HSENT}), which represents communities and cross-community connectivity explicitly, and \emph{HDTC}, a chemistry-aware variant of HDT that augments community blocks with semantic type labels. All representations are lossless and compatible with standard causal transformers.

Our contributions are threefold and concern the serialization rather than a new model architecture: we keep a standard causal transformer throughout and compare how graphs are structured and presented to it. First, we introduce a generic hierarchical framework for autoregressive graph generation that decomposes generation into graph coarsening, hierarchical tokenization, and exactly invertible decoding, making it applicable across molecular, biological, and synthetic graph domains. Second, we propose simple and scalable coarse-to-fine hierarchical tokenization schemes that expose global community structure before local connectivity, and show that they substantially improve distributional and structural fidelity over flat tokenization across five molecular benchmarks and non-molecular datasets when meaningful hierarchy is present, while in some settings introducing a trade-off with V.U.N. Third, and critically, we study the three design axes exposed by this framework (coarsening strategy, hierarchical serialization scheme, and order-randomization strategy) showing how the quality of the recovered hierarchy, the way it is represented, and the extent of order randomization together determine the gains, with hierarchy-wide randomization largely removing the V.U.N.\ cost of more restricted hierarchical orderings.

The remainder of the paper is structured as follows: we review related work in Section~\ref{sec:background}, formalize the pipeline in Section~\ref{sec:methods}, report empirical results in Section~\ref{sec:results}, and conclude in Section~\ref{sec:conclusion}.

\vspace{-5pt}

\section{Related work}\label{sec:background}

\noindent \textbf{Graph generation and tokenization.} Existing graph generation methods broadly fall into two categories: autoregressive models, which construct graphs sequentially by adding nodes and edges~\citep{you2018graphrnn,jangsimple}, and one-shot approaches that generate the entire graph simultaneously while preserving permutation equivariance, including graph VAEs~\citep{kipf2016variational}, GANs~\citep{martinkus2022spectre}, and more recently diffusion-based models~\citep{ho2020denoising,songscore,lipmanflow}. Within the autoregressive setting, prior work such as GraphRNN~\citep{you2018graphrnn}, GRAN~\citep{liao2019efficient}, GraphGen~\citep{goyal2020graphgen}, and AutoGraph/SENT~\citep{chen2025autograph} typically linearizes graphs using traversals such as BFS, DFS, or random walks. While these approaches are lossless and computationally efficient, they flatten the graph into a single sequential view, leaving higher-order structures such as motifs and communities implicit in the token stream. Our work builds on this sequence-based autoregressive paradigm and investigates whether hierarchical tokenization can expose global graph structure while retaining exact graph reconstruction. This direction is complementary to diffusion and score-based graph generators~\citep{niu2020scoregraph,jo2022scoregraph,vignac2023digress,kong2023autoregressivediffusion}, which operate under a fundamentally different modeling framework.

Recent efforts have also explored latent diffusion architectures for graph generation~\citep{rombach2022high,peebles2023scalable,gainski2026scalable, esser2024scaling,joshiall}, motivated by the efficiency gains of performing diffusion in compressed latent spaces. However, existing graph latent diffusion methods remain limited in scalability and reconstruction fidelity. Prior work either focuses on small molecular graphs~\citep{nguyen2024glad}, relies on dense attention or adjacency representations with quadratic complexity~\citep{nguyen2024glad,zhou2024unifying}, or targets specialized downstream tasks such as molecular generation, link prediction, or regression~\citep{fu2024hyperbolic,ketatalift,sghaier2026probabilistic,parviz2026nonlinear,huang2025large,10.1093/bib/bbae142,10.1145/3627673.3679547,zhou2024unifying,10508504}. In contrast, our approach introduces a hierarchical autoregressive representation in which the graph structure remains losslessly recoverable while avoiding explicit dense adjacency modeling.

\noindent
\begin{minipage}[b]{0.70\textwidth}
\noindent \textbf{Hierarchical representations.} Hierarchical structure is pervasive in graph domains such as chemistry and biology, where low-level components organize into higher-order motifs and functional substructures~\citep{jin2018jtvae,maziarz2022moler,dobson2003ddprotein}. Prior work has leveraged this property through hierarchical and multiscale architectures for both representation learning and generation, including JT-VAE~\citep{jin2018jtvae}, HiGen~\citep{karami2024higen}, and domain-specific serializations such as T-SMILES~\citep{wu2024tsmiles}. Unlike our setting, however, T-SMILES operates over SMILES strings rather than native graphs. Motif-aware generators such as MoLeR~\citep{maziarz2022moler} likewise build the structural prior into the model. More broadly, these approaches rely on specialized generative architectures that are difficult to scale with modern transformer-based modeling~\citep{chen2025autograph}. In contrast, our work introduces hierarchy directly at the tokenization level while retaining a standard autoregressive transformer backbone, so the structural prior is a property of the input encoding rather than of the model. Its reach beyond chemistry and biology is set by the coarsening step, for which a domain-specific detector can be substituted where interpretable units exist and a domain-agnostic one used where they do not (Section~\ref{sec:results:limitations}).

\section{Methods}\label{sec:methods}

We formulate autoregressive graph generation as next-token prediction over a hierarchical sequence representation of the input graph. Our pipeline (Figures~\ref{fig:main_pipeline_encode} and~\ref{fig:main_pipeline_decode}) proceeds in three stages: \emph{coarsening}, which builds a partition tree $\mathcal{H}$ from the input graph $G$; \emph{tokenization}, which serializes $G$ using $\mathcal{H}$ into a flat token sequence in coarse-before-fine order; and \emph{decoding}, which losslessly reconstructs the graph from its complete encoding.
\end{minipage}\hfill
\begin{minipage}[b]{0.26\textwidth}
\centering
\includegraphics[width=\linewidth]{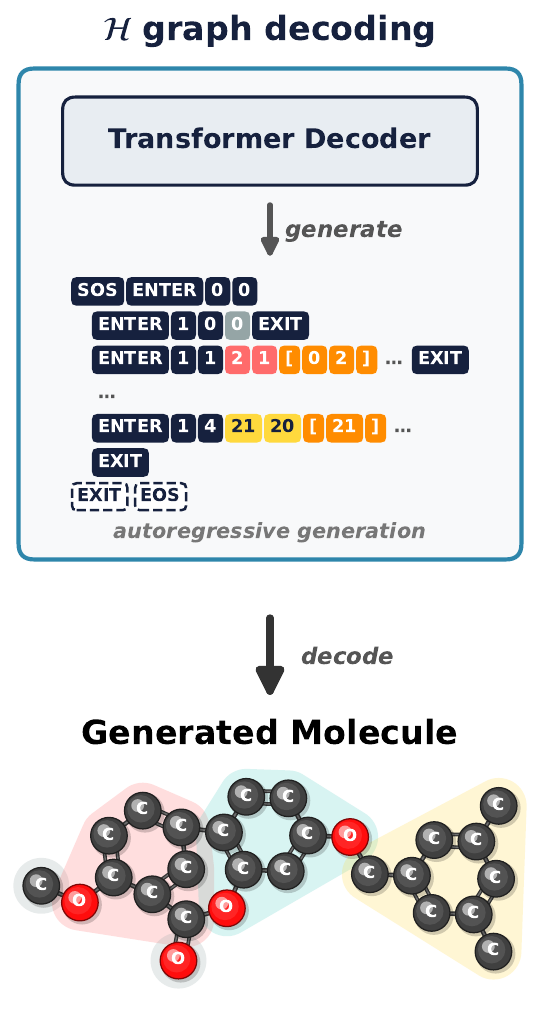}
\captionsetup{type=figure,font=small,justification=raggedright,singlelinecheck=false}
\caption{\textbf{Decoding pipeline illustration}. A causal transformer generates an \emph{HDT} sequence; community nesting and back-edge brackets specify the graph structure. The schematic shows an abridged sequence and molecular output, with atom- and bond-type tokens omitted.}
  \label{fig:main_pipeline_decode}
\end{minipage}

\subsection{Autoregressive setting}
We serialize a graph $G$ into a token sequence
$\mathbf{t}=(t_1,\dots,t_L)\in\mathcal{V}^L$ over a finite
vocabulary $\mathcal{V}$ and model it with a causal transformer
parameterized by $\theta$. Thus, once the graph has been serialized, the learning problem is ordinary next-token prediction:
\[
  p_\theta(\mathbf{t})=\prod_{i=1}^{L} p_\theta\!\left(t_i\mid \mathbf{t}_{<i}\right),
  \qquad \mathbf{t}_{<i}=(t_1,\dots,t_{i-1}).
\]
Equivalently, for a single training example we minimize the negative log-likelihood
\[
  \mathcal{L}(\theta;\mathbf{t})=-\sum_{i=1}^{L}\log p_\theta\!\left(t_i\mid \mathbf{t}_{<i}\right).
\]
The transformer architecture and training objective are standard; our contribution lies in how the graph is serialized into $\mathbf{t}$. We represent each input token by the sum of its token and position embeddings. A causal attention mask ensures that the prediction of $t_i$ depends only on the preceding tokens $\mathbf{t}_{<i}$. The next subsections define the hierarchical representation and tokenization scheme that produce sequences in a coarse-before-fine order.

\subsection{Hierarchical graph as a partition tree}\label{sec:methods:hgraph}

\begin{wrapfigure}[20]{r}{0.42935\textwidth}
\vspace{-0.7cm}
\begin{adjustbox}{width=\linewidth}
\begin{minipage}[t]{0.56\textwidth}
\vspace{0pt}
\begin{algorithm}[H]
\caption{\emph{HDT} tokenization (topology shown)}\label{alg:hdt}
\small
\algrenewcommand\alglinenumber[1]{#1:}
\algrenewcommand\algorithmicindent{0.85em}
\begin{algorithmic}[1]
\Require \parbox[t]{\dimexpr\linewidth-4em\relax}{Graph $G$; partition tree $\mathcal{H}$ with root $\mathcal{R}$;\\ R/FR traversal policy}
\Ensure Token sequence $\mathbf{t}$
\State $\mathbf{t}\leftarrow[\texttt{[SOS]}]$;\, $V_{\text{seen}}\leftarrow\emptyset$
\State $\rho\leftarrow\emptyset$ \Comment{node-ID map}
\State \textsc{Traverse}$(\mathcal{R},\,\ell=0)$
\State $\mathbf{t}.\text{append}(\texttt{[EOS]})$;\, \Return $\mathbf{t}$
\Procedure{Traverse}{$C$, $\ell$}
    \State $\mathbf{t}.\text{append}(\texttt{[ENTER]},\,\ell,\,\text{id}(C))$
    \If{$C$ is a leaf of $\mathcal{H}$}
        \For{$v\in C$ in traversal order}
            \State $\rho(v)\leftarrow |V_{\text{seen}}|$; $\mathbf{t}.\text{append}(\rho(v))$
            \State $B\leftarrow N_G(v)\cap V_{\text{seen}}$
            \If{$B\neq\emptyset$}
                \State $B\leftarrow\text{sorted}(\{\rho(u):u\in B\})$
                \State $\mathbf{t}.\text{append}(\texttt{[LEDGE]},\,B,\,\texttt{[REDGE]})$
            \EndIf
            \State $V_{\text{seen}}\leftarrow V_{\text{seen}}\cup\{v\}$
        \EndFor
    \Else \Comment{internal node}
        \For{child $C'$ of $C$ in traversal order}
            \State \textsc{Traverse}$(C',\,\ell+1)$
        \EndFor
    \EndIf
    \State $\mathbf{t}.\text{append}(\texttt{[EXIT]})$
\EndProcedure
\end{algorithmic}
\end{algorithm}
\end{minipage}
\end{adjustbox}
\end{wrapfigure}

Let $G=(V,E)$ be a finite simple undirected graph with $n=|V|$ nodes and $m=|E|$ edges, optionally carrying discrete node and edge attributes. A partition $\sigma:V\to[k]$, where $[k]=\{1,\ldots,k\}$, assigns each node to one of $k$ communities $C_1,\dots,C_k$ with $C_i=\sigma^{-1}(i)$. Each community induces the subgraph $G[C_i]=(C_i,E[C_i])$ where $E[C_i]=\{(u,v)\in E: u,v\in C_i\}$ collects intra-community edges, and the remaining edges $B_{ij}=\{(u,v)\in E: u\in C_i,\, v\in C_j\}$ for $i\neq j$ are cross-community edges. We define the \emph{hierarchical graph} $\mathcal{H}$ as a rooted partition tree: the root corresponds to $V$ and encloses the top-level partition, with one terminal child when no split is made. For recursive HAC, the same coarsening rule is reapplied to induced subgraphs $G[C]$. The threshold $\tau$ is the minimum community size at which splitting is attempted; recursion stops when $|C|<\tau$ or no split into at least two nonempty child communities is found. The leaves of $\mathcal{H}$ form a partition of $V$ at the finest scale, while internal nodes encode coarser-scale community structure. We use $\tau=4$ by default, with block-model threshold interventions reported in Appendix~\ref{app:sbm_full}; empirical depth distributions are reported in Appendix~\ref{app:hierarchy_depths}, and a visualization of $\mathcal{H}$ on cholesterol is shown in Appendix~\ref{app:tokenization_fig}.

\subsection{Coarsening}\label{sec:methods:coarsen}

The coarsening step determines the partition tree $\mathcal{H}$. We consider one \emph{domain-agnostic} strategy that applies to every benchmark and one \emph{chemistry-aware} strategy used only on molecular graphs; a second, typed chemistry-aware strategy is used only by \emph{HDTC} and is described with that tokenizer (Appendix~\ref{app:grammar:hdtc}). The domain-agnostic option is hierarchical agglomerative clustering (HAC) with average linkage on pairwise dissimilarities derived from edge affinities. A finite non-edge penalty discourages merging nonadjacent nodes, and the cut maximizes weighted modularity over the candidate community counts~\citep{newman2004modularity}.

For \emph{molecular graphs}, the natural unit is the ring system, motivated by the observation that molecular validity and function are organized around rings and functional groups. \emph{Motif Community} (MC) detects ring systems via SMARTS substructure queries~\citep{weininger1988smiles}, merges overlapping rings into a single motif community, and absorbs leftover atoms into adjacent motif communities when possible. HAC derives a recursive hierarchy from graph structure, whereas MC uses a fixed catalog of molecular substructures and supplies one community level below the root.
HAC communities need not correspond to recognizable motifs,
while MC's granularity depends on the motifs present.
Section~\ref{sec:results:design} compares their generation
performance. We name a configuration \emph{tokenizer}-\emph{coarsening} (so \emph{HDT}-MC is the traversal tokenizer with Motif Community coarsening, \emph{HDT}-HAC the same with agglomerative clustering) and append a suffix for the order-randomization policy defined in Section~\ref{sec:methods:randomize}.

\subsection{Tokenization}\label{sec:methods:tokenize}

Given $G$ and $\mathcal{H}$, we serialize the graph into a one-dimensional token sequence $\mathbf{t}\in\mathcal{V}^L$ over a discrete vocabulary $\mathcal{V}$. All hierarchical tokenizers we consider follow a \emph{coarse-before-fine} principle. They emit higher-level community structure before the node-level detail within each community. The purpose of introducing hierarchy is to expose large-scale organization early in the sequence, so that later local decisions are made in that context. A fine-to-coarse order would weaken this benefit, because the model would have to generate low-level structure before seeing the higher-level organization meant to guide it. The design axis is then \emph{how explicit the hierarchy is in the token stream} (Table~\ref{tab:tokenizer_characteristics}). Two more explicit alternatives, the declarative \emph{HSENT} and the typed molecular \emph{HDTC}, are defined in Appendix~\ref{app:grammar} and evaluated in Appendix~\ref{app:legacy_variants}; the tokenizer we introduce here and use throughout the paper, and the strongest of the three in our experiments, is the traversal-based \emph{HDT}.

\emph{HDT} performs a single depth-first traversal of $\mathcal{H}$ (Algorithm~\ref{alg:hdt}). Each tree node opens a community block with \texttt{[ENTER]}, followed by its recursion depth $\ell$ and community identifier, and closes it with \texttt{[EXIT]}. Internal blocks contain their child communities; leaf blocks emit their original graph nodes in the order specified by the randomization policy. When a node $v$ is emitted, \emph{HDT} lists its previously emitted neighbors $N_G(v)\cap V_{\mathrm{seen}}$ in a back-edge bracket, where $N_G(v)$ is the full neighbor set. The map $\rho$ assigns global serialization IDs in first-visit order; node tokens and targets use these IDs. Community IDs are local to each sibling set, with root ID $0$. Targets may lie in the same or another community. This is the central design choice: \emph{cross-community edges are emitted as ordinary back edges, syntactically indistinguishable from intra-community ones}. Cross-community connectivity therefore uses the same edge syntax as within-community connectivity, while nested traversal of $\mathcal{H}$ expresses the community structure. The resulting sequence length is $O(n+m)$.
For molecular graphs, atom-type tokens follow node identifiers and bond-type tokens follow edge targets. Algorithm~\ref{alg:hdt} shows the traversal structure and omits these attribute tokens.

\input{tables/tokenizer_characteristics}

\subsection{Order randomization}
\label{sec:methods:randomize}

A graph admits many valid serializations. For HDT, we augment the training data as follows:
we resample traversal choices at each presentation while holding the graph, its attributes, community memberships, and partition hierarchy $\mathcal{H}$ fixed. \emph{Within-Community Randomization} (\emph{-R}) preserves community order and randomizes traversal within each terminal community. \emph{Full Hierarchy-Wide Randomization} (\emph{-FR}) also samples an adjacency-constrained depth-first traversal between communities at each hierarchy level. Communities with a traversal parent are entered through
a node adjacent to that parent, and traversal within terminal communities is randomized. Both policies assign node identifiers in first-visit order. Table~\ref{tab:randomization_settings} summarizes the policies; Appendix~\ref{app:order_randomization} specifies the traversal, identifier assignment, and handling of disconnected components, and gives a worked example.

\input{tables/randomization_settings}

\subsection{Decoding}\label{sec:methods:decode}

Decoding inverts each tokenization by parsing community boundaries, node records and edge records, then unifying the recovered intra- and inter-community edge sets to reconstruct $G$, including its encoded node and edge attributes. All three schemes are \emph{lossless} (exactly invertible up to node relabeling) for complete, untruncated encodings:
\[
\mathcal{H}=\operatorname{coarsen}(G),\qquad\operatorname{decode}(\operatorname{tokenize}(G,\mathcal{H}))\cong G,
\]
where the isomorphism preserves the encoded attributes. The leaves form an exhaustive, disjoint partition of $V$, and every node and edge is recorded, so the coarsen-tokenize-decode pipeline is information preserving. HDT and HSENT record each edge once; legacy HDTC additionally repeats cross-community edges in its super-graph record, and decoding coalesces these duplicates. Round-trip tests check this reconstruction property on representative graph fixtures, including labeled graphs.

Losslessness is a property of the encoder-decoder pair on its encodings, not a guarantee that arbitrary model-generated sequences describe valid graphs. Validity is measured separately for every method, including SENT; Appendix~\ref{app:validity} explains the distinction and the treatment of malformed generated sequences.

\FloatBarrier

\section{Results}\label{sec:results}

\subsection{Experimental setup}\label{sec:results:setup}

\noindent\textbf{Model and training.}
Our main comparisons use the same GPT2-XS backbone ($\sim\!11$M parameters); a larger training regime is examined in Figure~\ref{fig:regime_scaling} and Appendix~\ref{app:regime_delta}. Architecture settings are listed in Table~\ref{tab:appendix_arch}, and training schedules in Tables~\ref{tab:appendix_train_inventory_mol} and~\ref{tab:appendix_train_inventory_nonmol}.
SENT's update budgets are calibrated using measured training time rather than matched epochs or steps, accounting for the longer hierarchical sequences (Appendix~\ref{app:token_length}). The calibration targets comparable training budgets, with achieved runtimes and update counts reported separately.

\noindent\textbf{Benchmarks.}
We evaluate five molecular benchmarks: GuacaMol~\citep{brown2019guacamol}, MOSES~\citep{polykovskiy2020moses}, COCONUT 2.0~\citep{chandrasekhar2025coconut2}, a disjoint split of GEOM-Drugs~\citep{axelrod2022geom}, and and Macrocycle, derived from Macrocycle-DB~\citep{jiang2026macrocycle}. Their training sets range from $27{,}392$ to $499{,}968$ examples, with unconditional generation evaluated against each benchmark's held-out reference set (Table~\ref{tab:appendix_dataset_stats}). Beyond molecules, we evaluate Proteins (D\&D)~\citep{dobson2003ddprotein}, Planar~\citep{martinkus2022spectre}, and the three synthetic block-model families defined below.

\noindent\textbf{Synthetic block models.}
In a stochastic block model (SBM), nodes belong to planted communities, and each edge is sampled independently with a probability determined by its endpoints' memberships. Our SBM benchmark~\citep{martinkus2022spectre} has $2$--$5$ communities of $20$--$40$ nodes, with within- and between-community edge probabilities $0.30$ and $0.005$. Sparse SBM (SSBM) retains these probabilities but uses $6$--$10$ smaller communities of $12$--$20$ nodes, producing lower mean degree. Recursive SBM (RSBM) introduces two planted levels: each of $2$--$5$ coarse communities contains $2$--$5$ fine communities of $8$--$9$ nodes. Its contrast parameter $R$ is the ratio of within-fine to between-fine edge probabilities inside a coarse community. These probabilities preserve the expected internal edge count of an SBM community of the same size; edges between coarse communities retain probability $0.005$.
We use $R=4$ here. All three families have $128/32/40$ train/validation/test graphs. Connectivity filtering, including the distinction between original and connected RSBM populations, and complete generator settings are specified in Appendix~\ref{app:block_datasets}. Recorded community memberships provide a \emph{planted-hierarchy} (PH) control, which supplies the generator's partition during training; generation remains unconditional.

\noindent\textbf{Metrics.}
Unconditional molecular evaluation combines the following benchmark and structural metrics: Validity, Uniqueness, Novelty, and their combined V.U.N.\ score; Fr\'echet ChemNet Distance (FCD); Similarity to Nearest Neighbor (SNN); fragment and scaffold similarity (Frag / Scaff); Internal Diversity (IntDiv); motif discrepancies over functional-group, SMARTS, ring, and BRICS features; and PolyGraph Discrepancy (PGD). Non-molecular benchmarks report generation-quality scores using a dataset-specific structural validity test, together with MMD on degree, clustering, orbit, and spectral statistics and the derived Ratio. We emphasize V.U.N., using its decomposition to distinguish invalid generation from repetition and training-set memorization. PGD uses the official \texttt{MoleculePGD} interval estimator, and motif discrepancies use fixed, equal arms, both defined in Appendix~\ref{app:metrics}. Generation results report means and sample standard deviations over three generation seeds from one trained checkpoint per configuration. These SDs describe generation and, where applicable, reference-sampling variability, rather than variation across independently trained models; we interpret small mean differences cautiously.

We report standard unconditional evaluations and, where available, novel-only evaluations. Standard evaluation follows conventional benchmark scoring, while novel-only evaluation recomputes metrics on valid generations absent from the training set (Appendices~\ref{app:novel_only} and~\ref{app:novel_graphs}). RSBM is reported under the standard protocol only.
Both are compared with a \emph{training-sample} reference formed
from an independently permuted prefix of the training set,
up to the evaluation size.
This reference is used only for calibration and is excluded
from all rankings.

\subsection{Unconditional molecular generation}\label{sec:results:unconditional}

\input{tables/main_molecular_all}

\begin{figure}[t]
\centering
\includegraphics[width=\linewidth]{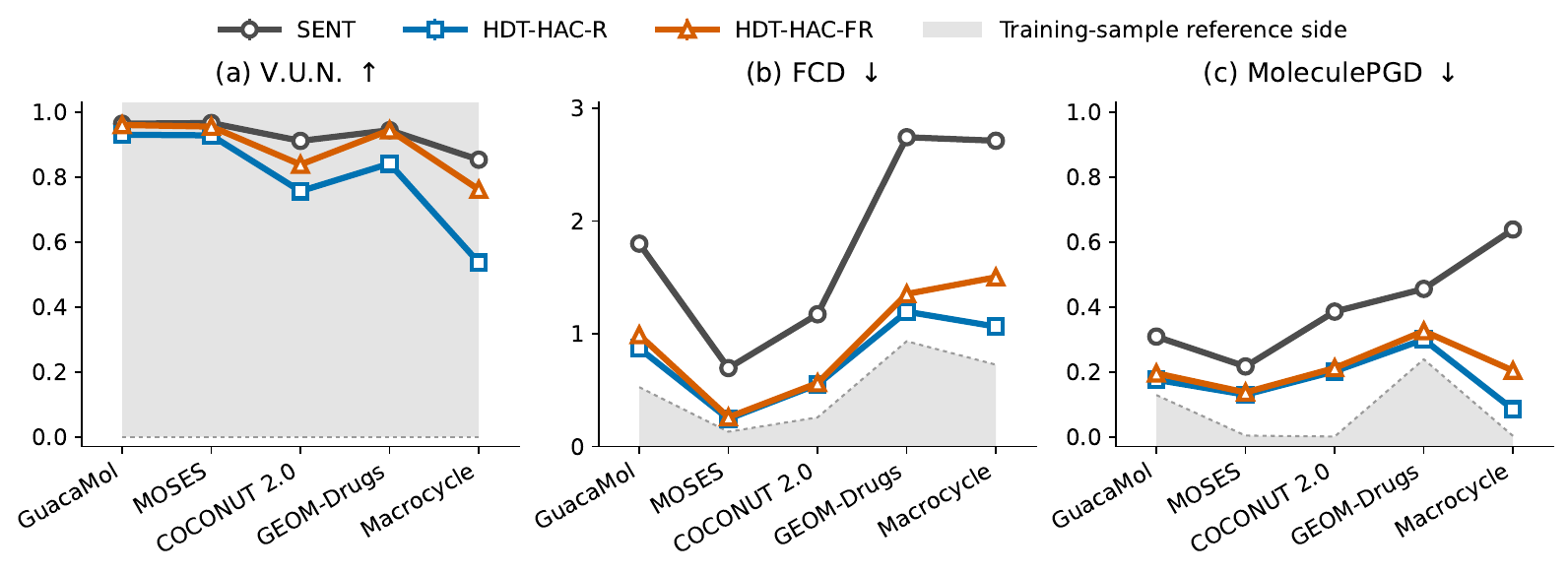}
\caption{\textbf{Hierarchy improves distributional fidelity across all five molecular benchmarks.}
Markers are means over thee generation seeds; error bars are sample SD. Gray marks the
metric-improving side of the  training-sample reference, which for V.U.N.\ indicates
direction rather than an attainable ceiling. Panel (a): the V.U.N.\ lead SENT holds over
\emph{-R} and that \emph{-FR} largely closes. Panels (b, c): the distributional gap, large on
every benchmark. Remaining variants and metrics are in Appendix~\ref{app:unconditional_full}.}
\label{fig:uncond_summary}
\end{figure}

Hierarchical serializations improve distributional fidelity across the molecular benchmarks, while their V.U.N.\ depends on the randomization policy. We first examine the distributional improvements, then assess how hierarchy-wide randomization changes the accompanying V.U.N.\ deficit.

\noindent\textbf{Hierarchy improves distributional fidelity.} The gap between the flat and hierarchical serializations is largest, and most consistent, on the metrics that compare distributions rather than count successes. Against SENT, \emph{HDT}-HAC-R lowers FCD by between $51.6\%$ and $64.6\%$ and \texttt{MoleculePGD} by between $34.3\%$ and $86.6\%$ on all five benchmarks, and improves SNN, fragment and scaffold similarity on all five as well (Table~\ref{tab:main_molecular_all}, Appendix~\ref{app:unconditional_full}). These gains are large relative to the variation across generation seeds, whereas internal diversity changes only marginally, suggesting that the improved fidelity is not obtained simply by collapsing the generated distribution. All four hierarchical configurations have lower FCD and \texttt{MoleculePGD} than SENT on every benchmark. These improvements persist under novel-only evaluation (Figure~\ref{fig:novel_summary},
Appendix~\ref{app:novel_only}), so they cannot be explained solely by reproducing training molecules.

\noindent\textbf{Hierarchy-wide randomization reduces the V.U.N.\ deficit.}

The main cost of community-wise randomization is in V.U.N., driven primarily by reduced novelty on some benchmarks. For \emph{HDT}-HAC-R, the deficit relative to SENT ranges from $3.6\%$ to $37.2\%$, with the largest drop on Macrocycle. Hierarchy-wide randomization substantially narrows this gap: across the fully randomized configurations, the largest V.U.N.\ deficit is $10.85\%$, and for \emph{HDT}-MC-FR the largest deficit across benchmarks is only $2.98\%$. MC-FR also improves FCD, \texttt{MoleculePGD}, SNN, and fragment similarity over SENT on all five benchmarks, while achieving slightly higher mean V.U.N.\ on GuacaMol. These results show that the hierarchy-wide FR policy can substantially reduce the V.U.N.\ deficit while retaining lower FCD and \texttt{MoleculePGD}. The comparison concerns the complete policy, including its entry constraints and community-ID assignment (Section~\ref{sec:methods:randomize}).

\begin{figure}[t]
\centering
\includegraphics[width=0.86\linewidth]{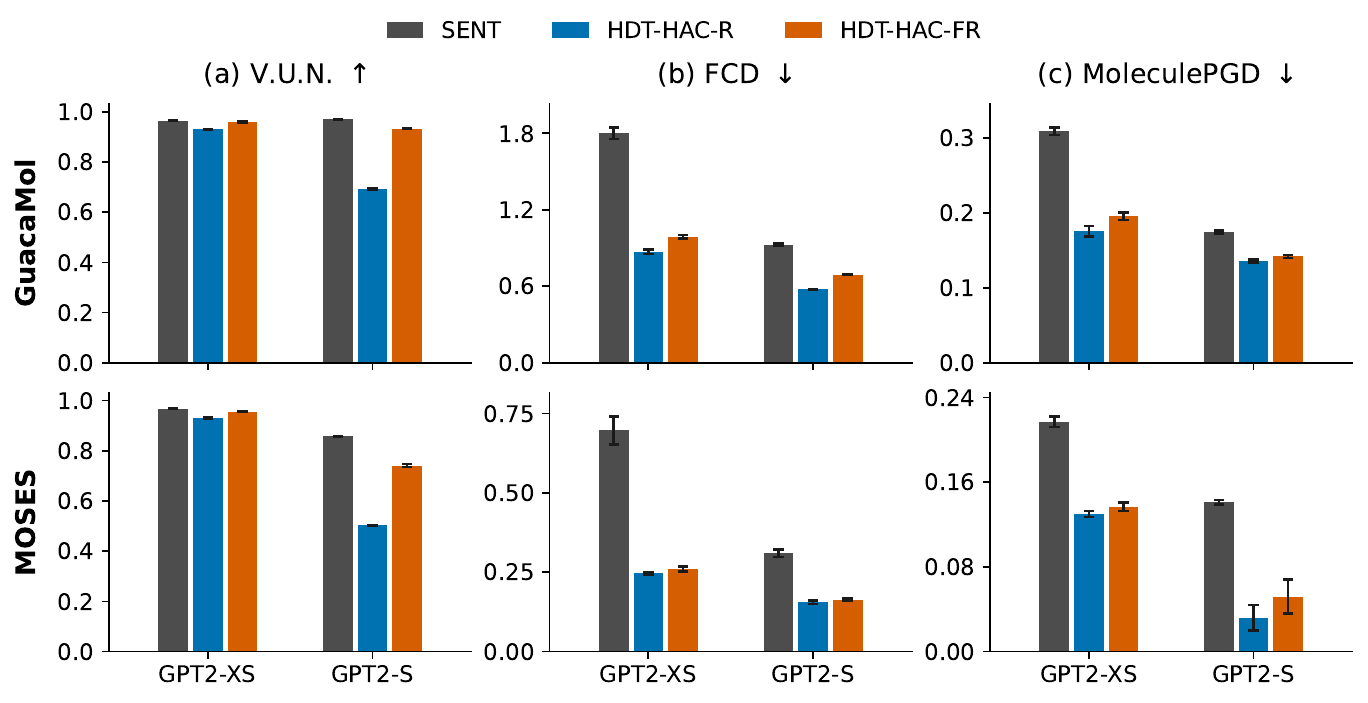}
\caption{\textbf{Training-regime scaling on GuacaMol (top) and MOSES (bottom).} Bars are means
over three generation seeds; error bars are sample SD. Model size, data volume, and schedule
change together (Tables~\ref{tab:appendix_regime_guacamol} and~\ref{tab:appendix_regime_moses}).}
\label{fig:regime_scaling}
\end{figure}

\noindent\textbf{Training-regime scaling.}
 On both GuacaMol and MOSES, \emph{HDT}-HAC-R at GPT2-XS has lower FCD and comparable or lower \texttt{MoleculePGD} than SENT at GPT2-S (Figure~\ref{fig:regime_scaling}). Within each regime, both variants have lower FCD and \texttt{MoleculePGD} than SENT, whereas SENT has higher V.U.N and novelty. In the larger GPT2-S regime, both variants still achieve lower FCD and \texttt{MoleculePGD}, but at the cost of reduced novelty and V.U.N., suggesting potential overfitting. For both HAC and MC, FR has smaller declines than the R version. SENT's mean V.U.N.\ increases slightly on GuacaMol and decreases on MOSES. These are joint training-regime comparisons: backbone size, training population and schedule change together, GuacaMol GPT2-S SENT retains different context and dropout settings, and novelty is evaluated against each regime's own training population. They therefore do not isolate model size, data efficiency or memorization (Tables~\ref{tab:appendix_regime_guacamol} and~\ref{tab:appendix_regime_moses}; Appendix~\ref{app:regime_delta}).

\input{tables/main_autograph_comparison}

\subsection{Motif-conditioned molecular generation}\label{sec:results:conditional}
We prompt the unconditionally trained SENT, \emph{HDT}-HAC-R and \emph{HDT}-HAC-FR checkpoints with an isolated chemical motif, without conditional fine-tuning. Unlike unconditional evaluation, this task uses motif-specific training-derived reference pools and retains the first $500$ chemically valid outputs per motif and seed, including duplicates and prompt copies (Appendix~\ref{app:conditional_generation}). The unconditional fidelity advantage of hierarchy does not carry over. On all five benchmarks, SENT has the lowest mean values on all four motif discrepancies, the highest mean fragment similarity, and the highest mean V.U.N. \emph{HDT}-HAC-R often repeats completions with mean uniqueness around $0.378$--$0.456$ on GuacaMol, COCONUT 2.0 and GEOM-Drugs. FR narrows this gap and has the highest mean validity on four benchmarks and scaffold similarity on all five, but still trails SENT on the motif discrepancies and V.U.N. These results characterize zero-shot motif prompting rather than demonstrating a conditional-generation advantage for hierarchical serialization. We will further explore conditional generation in the hierarchical setting in future work. Full results are in Tables~\ref{tab:appendix_conditional_motif_guacamol}--\ref{tab:appendix_conditional_motif_macrocycle_db}.

\subsection{Beyond molecules}\label{sec:results:nonmolecular}

The graph benchmarks show that hierarchical serialization can help outside chemistry, but its benefit depends on the supplied hierarchy. On Proteins, \emph{HDT}-HAC-R improves V.U.N.\ by $62.8\%$ over SENT. On SSBM, it improves V.U.N.\ by $59.3\%$ and reduces Degree MMD, Orbit MMD, and Ratio by more than $90\%$. The effect of randomization depends on the dataset and metric. On Proteins, FR raises V.U.N.\ further to $0.960$ but has higher means on all four graph MMDs than R; on SSBM, FR's V.U.N.\ is $0.485$, below SENT's $0.527$. The HAC-R gains are consistent with our molecularly tuned coarsening transferring to smaller, sparser communities, but these datasets do not constitute a controlled density or community-size ablation.

The block models expose two distinct failure modes (Table~\ref{tab:appendix_block_vun}). On SBM, default \emph{HDT}-HAC-R reaches validity $0.926$, but uniqueness and novelty fall to approximately $0.41$: its V.U.N.\ is only $0.276$, compared with SENT's $0.669$. \emph{HDT}-HAC-FR instead maintains uniqueness and novelty near $1$ but reaches V.U.N.\ $0.445$ because of lower validity. The same contrast appears on the original RSBM population. High validity alone therefore does not establish successful generation.

Hierarchy diagnostics suggest that the main failure on SBM is over-refinement rather than poor recovery of the coarse partition. Default HAC recovers the top-level communities well, but recursively subdivides them into leaves that are much smaller than the planted blocks (Table~\ref{tab:appendix_block_alignment}).  Increasing the stopping threshold substantially improves V.U.N. for R/FR,  while supplying the planted hierarchy (PH) yields a further gain. 
A similar pattern appears on connected RSBM: with a better-aligned stopping threshold, HAC-R nearly matches PH-R in V.U.N.$^\dagger$, whereas the corresponding FR configuration remains substantially weaker than PH-FR ($\dagger$ denotes compatibility with the original RSBM validity gate, as detailed in Appendix~\ref{app:block_datasets}). These results are consistent with a benefit from hierarchy alignment, whose effect also depends on the randomization policy. Experiment settings and population differences are reported in Appendix~\ref{app:sbm_full}.

\subsection{Coarsening and tokenization ablations}\label{sec:results:design}\label{sec:results:serialization}

\noindent\textbf{Coarsening.}
The relative performance of HAC and motif-based coarsening depends on the metric and randomization policy.
Under within-community randomization, \emph{HDT}-HAC-R has lower \texttt{MoleculePGD} and higher SNN and scaffold similarity than \emph{HDT}-MC-R on all five molecular benchmarks, and lower FCD on four.
 Under hierarchy-wide randomization, neither coarsener dominates across metrics, although MC-FR has higher mean V.U.N.\ on four of five datasets.
Both coarsenings improve FCD and \texttt{MoleculePGD} over SENT across all five benchmarks.
The block-model experiments further suggest that alignment with the data's community structure matters.
On connected RSBM ($R=4$), PH-FR reaches V.U.N.$^\dagger$ $0.744$, compared with $0.083$ for HAC-FR at $\tau=10$, a difference of $66.2$ percentage points.
The tokenizer and randomization policy are unchanged, although the context lengths differ (Appendix~\ref{app:sbm_full}).

\noindent\textbf{Hierarchy explicitness.}
On GuacaMol and MOSES, both \emph{HSENT} configurations have lower validity and V.U.N.\ and higher \texttt{MoleculePGD} than their corresponding \emph{HDT}-R configurations;  \emph{HDTC}-R also has lower validity and V.U.N.\ and higher \texttt{MoleculePGD} than both \emph{HDT}-R configurations (Appendix~\ref{app:unconditional_full}, Table~\ref{tab:appendix_legacy_variants}).
Both \emph{HSENT} configurations additionally have higher FCD on GuacaMol.
The ranking varies across metrics: on MOSES, \emph{HSENT}-HAC-R achieves lower FCD than \emph{HDT}-HAC-R.
\emph{HSENT} sequences also average roughly $2.9\times$ SENT's length, compared with $1.9\times$ for \emph{HDT} (Appendix~\ref{app:token_length}).
These earlier runs predate the current protocol, so they provide supporting evidence for traversal-based serialization rather than a controlled isolation of hierarchy explicitness.
Appendix~\ref{app:traversal_vs_declaration} discusses possible explanations.

\subsection{Limitations}\label{sec:results:limitations}
Two limitations qualify our results. First, hierarchical serialization does not guarantee improved distributional fidelity, as demonstrated by the Planar benchmark, highlighting the importance of choosing an appropriate coarsening strategy for each dataset. Second, hierarchical serialization increases sequence length: \emph{HDT} uses roughly twice as many tokens as SENT in the measured benchmarks (Appendix~\ref{app:token_length}).
Longer sequences increase attention cost and context requirements; actual training times are reported in Tables~\ref{tab:appendix_train_inventory_mol} and~\ref{tab:appendix_train_inventory_nonmol}.
SENT's update budgets are calibrated using measured training time, while achieved runtimes vary across configurations. Appendix~\ref{app:limitations} discusses these limitations further.

\section{Conclusion}\label{sec:conclusion}
We introduced a hierarchical framework for autoregressive graph generation that separates coarsening, tokenization, and decoding, enabling lossless graph reconstruction while retaining a standard causal transformer. In the main regime, all four HDT configurations lower FCD and \texttt{MoleculePGD} over SENT across five molecular benchmarks under standard and novel-only evaluation, sometimes at a V.U.N.\ cost. Under standard molecular evaluation, hierarchy-wide FR substantially reduces within-community R's V.U.N.\ deficit, with remaining deficits depending on coarsening and dataset. \emph{HDT}-HAC-R improves V.U.N.\ over SENT on Proteins and SSBM, while R versus FR performance depends on dataset and metric. SBM and RSBM threshold and planted-hierarchy controls are consistent with hierarchy-alignment benefits that depend on policy and configuration. The reported Planar configurations do not improve graph MMDs over SENT. These findings support hierarchical serialization when partitions capture useful structure, with coarsening, tokenization and traversal assessed jointly through fidelity and V.U.N.

\clearpage
\bibliography{references}
\bibliographystyle{abbrvnat}


\clearpage
\appendix
\etocdepthtag{appendix}

\begingroup
\small
\setlength{\parskip}{0pt}
\hypersetup{linktoc=all}
\etocsettagdepth{main}{none}
\etocsettagdepth{appendix}{subsection}
\etocsetnexttocdepth{subsection}
\etocsettocstyle{\section*{Appendix contents}\label{app:contents}}{}
\tableofcontents
\endgroup
\clearpage

\section{Funding}
This work was supported in part by the National Science Foundation under grant CIF-2403452, CCF-2112665  and CCF-2217058. 

\section{Limitations}\label{app:limitations}

\noindent \textbf{Sequence length and compute cost.}
Hierarchy carries a computational cost. \emph{HDT}'s lossless coarse-before-fine encoding uses roughly $2\times$ as many tokens as SENT on small molecules and proportionally more on large graphs. Doubling sequence length increases the quadratic attention component fourfold. Total runtime and memory also depend on the rest of the model, padding, context limits and batching. We therefore calibrate update budgets using measured training time rather than equating steps or epochs. Tables~\ref{tab:appendix_train_inventory_mol} and~\ref{tab:appendix_train_inventory_nonmol} report achieved runtimes and update counts. Calibration does not guarantee equal achieved runtime.

\noindent \textbf{ Novelty, validity, and augmentation.}
 A deterministic linearization admits only one serialization per graph. SENT instead resamples its walk at each training presentation, exposing the model to multiple serializations of each graph. We use order randomization to expose the model to different serializations (Section~\ref{sec:methods:randomize}). Within-community randomization leaves a V.U.N.\ deficit relative to SENT ranging from $3.6\%$ on GuacaMol to $37.2\%$ on Macrocycle. Hierarchy-wide randomization reduces most of this deficit (Section~\ref{sec:results:unconditional}). On Macrocycle, \emph{HDT}-HAC-FR still has lower V.U.N.\ than SENT ($0.762$ versus $0.854$). This remaining deficit mainly reflects lower validity rather than novelty. We report this as an observed limitation rather than attributing it to a mechanism.

\noindent \textbf{Scope and hierarchy transfer.}
The framework requires a coarsening, and its value depends on whether the resulting communities are meaningful in the target domain. On the reported Planar benchmark, both hierarchical configurations are worse than the flat walk on every graph MMD (Section~\ref{sec:results:nonmolecular}). This establishes a limitation of these configurations, not the absence of useful community structure in all planar graphs. Conversely, Proteins shows that the benefit does not require a chemical catalog. Where a domain does define interpretable units, a domain-specific detector may be substituted for the motif detector without further modification. The choice of coarsening depends on the meaning of the nodes and edges in each dataset, so it may need to change across domains. We compute the partitions separately for each dataset. This preprocessing is inexpensive.

\section{ Tokenization details}
\subsection{Traversal vs. declaration of hierarchy}\label{app:traversal_vs_declaration}

This appendix expands the declarative-versus-traversal comparison summarized in Section~\ref{sec:results:serialization}. The two tokenizers expose the same hierarchy in qualitatively different ways: \emph{HSENT} lists communities and then separately specifies the inter-community wiring, whereas \emph{HDT} linearizes the hierarchy through a single DFS traversal so cross-community edges are handled within the same sequential process as every other edge.

One possible explanation for the observed gap is that the declarative grammar introduces an additional consistency requirement. The earlier-protocol comparison does not isolate this mechanism. In \emph{HDT}, \texttt{[ENTER]}/\texttt{[EXIT]} nesting specifies community membership. Each cross-community edge is emitted once as a back edge, using the same syntax as an intra-community edge. There is no separate coarse wiring record to reconcile with the node-level edges. \emph{HSENT} and \emph{HDTC} instead emit an explicit cross-community section. Their grammars do not enforce agreement between this section and the community blocks, so sampled records can conflict (Section~\ref{sec:methods:decode}).

The evidence for the gap itself is in Table~\ref{tab:appendix_legacy_variants}, on the two benchmarks where the declarative and typed variants were evaluated. On GuacaMol, the SENT reference is unchanged between the earlier and current protocols. However, the legacy runs do not isolate the effect of the tokenizer. Both \emph{HSENT} configurations and \emph{HDTC}-R have larger validity and V.U.N.\ deficits relative to SENT than any \emph{HDT} configuration in Table~\ref{tab:main_molecular_all}. \emph{HSENT}-MC-R also has higher FCD and \texttt{MoleculePGD} than SENT. \emph{HSENT} sequences run roughly $3\times$ SENT's length against \emph{HDT}'s $2\times$ (Table~\ref{tab:appendix_token_length}a). At those lengths, the quadratic attention component is about $2.25\times$ larger for HSENT than HDT; this is not a measurement of total compute.

Two caveats qualify this conclusion. The comparison is limited to these particular tokenizers and this training setup, and therefore does not establish that traversal is preferable to declaration in general. What the evidence does support is the narrower practical conclusion drawn in Section~\ref{sec:results:serialization}: these legacy declarative and typed variants use longer sequences without a consistent advantage over HDT. They are weaker on the emphasized V.U.N./MoleculePGD trade-off, while individual fidelity metrics have exceptions, including the lower MOSES FCD of HSENT-HAC-R.

\subsection{Tokenization schemes illustrated}\label{app:tokenization_fig}

\begin{figure}[h]
\centering
\includegraphics[width=\linewidth]{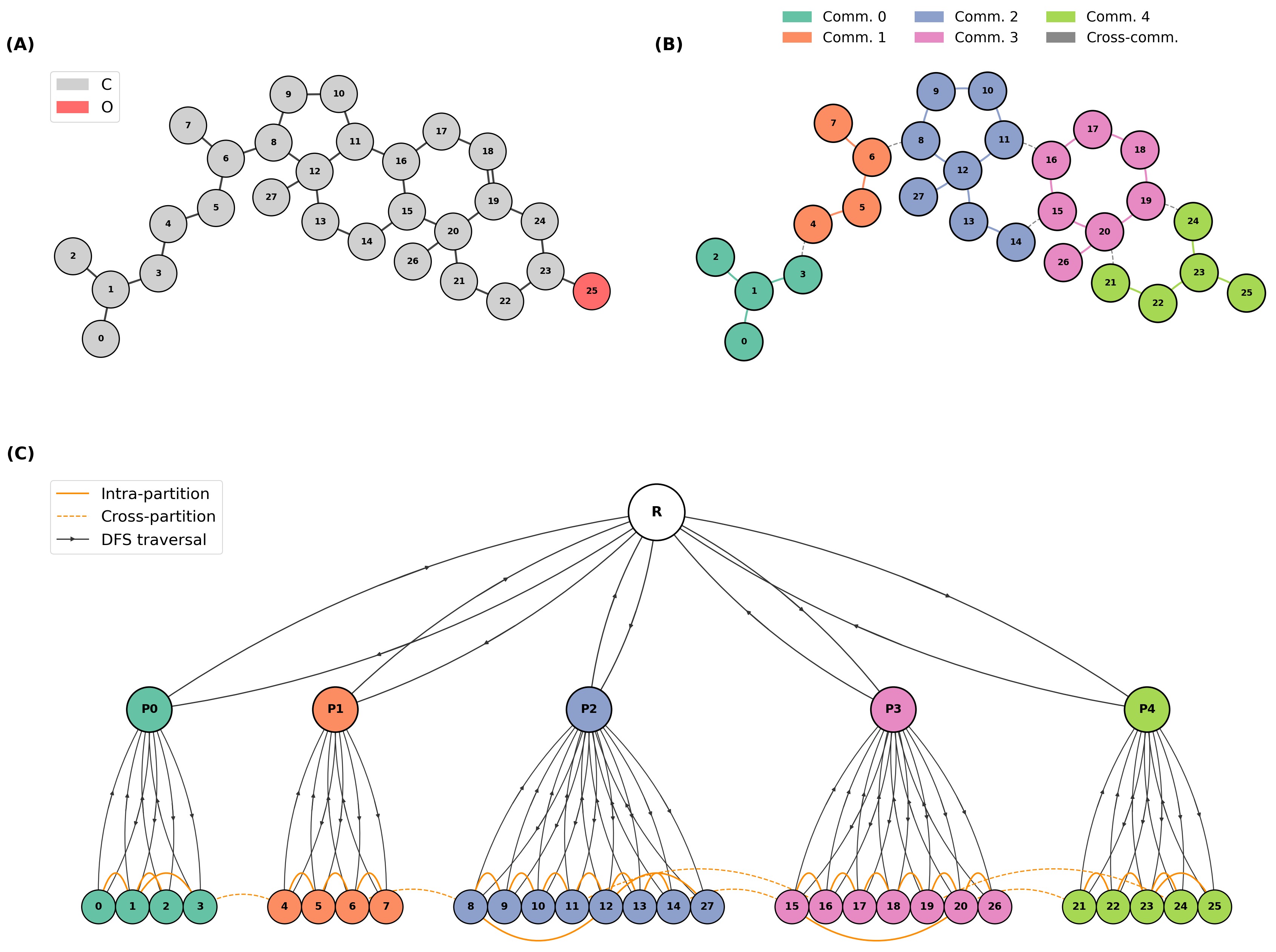}
\caption{\textbf{Hierarchical tokenization schemes on cholesterol.} \textbf{(A)} The molecule.
\textbf{(B)} \emph{HSENT}: community blocks (colored) and explicit cross-community wiring.
\textbf{(C)} \emph{HDT}: a schematic traversal of the same top-level partition; deeper
subdivisions are omitted. Solid orange curves mark intra-community edges, dashed curves
mark cross-community edges, and gray arrows indicate traversal structure. Original atom indices
are retained for correspondence across panels.}
\label{fig:appendix_tokenization}
\end{figure}

Figure~\ref{fig:appendix_tokenization} compares \emph{HSENT} and \emph{HDT} on cholesterol. \emph{HSENT} lists community blocks and a separate cross-community wiring section; \emph{HDT} traverses the hierarchy and uses back-edge brackets for both intra- and inter-community edges. The schematic shows the top-level partition and omits its deeper subdivisions.

\subsection{Order randomization details}
\label{app:order_randomization}

At each training presentation, we resample traversal choices while
keeping the graph, its attributes, and partition hierarchy
$\mathcal{H}$ fixed.

\paragraph{Within-community randomization (\emph{-R}).}
Communities retain their stored order and identifiers.
Within each terminal community, we sample a BFS root and shuffle
neighbor order. If the leaf is disconnected, nodes outside the
root's component are appended in shuffled order.

\paragraph{Hierarchy-wide randomization (\emph{-FR}).}
At each hierarchy level, we sample a DFS forest over the adjacency
graph of sibling communities. A community with a traversal parent
is entered through a node adjacent to that parent; this constraint
is propagated through nested communities to the leaf BFS root.
Disconnected components start new traversals.
Community identifiers follow encounter order within each sibling set.
The sampled DFS forest determines visitation order, not the
containment hierarchy $\mathcal{H}$.

\paragraph{Identifiers and reconstruction.}
Both policies assign node identifiers globally in first-visit order.
Each node lists all previously emitted neighbors as back-edge targets,
including those in other communities.
Thus, a complete, untruncated encoding emits each node and edge once
and reconstructs the input graph up to relabeling.
Table~\ref{tab:randomization_example} illustrates the ordering difference.

\input{tables/appendix_randomization_example}

\subsection{Tokenization grammar: notation and reference frames}\label{app:grammar}

This appendix presents the grammars of \emph{HSENT}, \emph{HDT}, and \emph{HDTC} using shared notation. They differ in how explicitly they encode hierarchy. We also explain the scope of their lossless reconstruction guarantee. Appendix~\ref{app:legacy_variants} compares their performance on GuacaMol and MOSES.

\subsubsection{Notation and partition tree}\label{app:grammar:notation}

Terminal symbols in \texttt{TYPEWRITER} are special vocabulary tokens; non-terminals are italicized. Across all three grammars, $\ell$ denotes the recursion depth level emitted at each \texttt{[ENTER]} token, $\textit{atoms}$ in the \emph{HDTC} grammar expands to a sequence of $\textit{atom}$ non-terminals as defined for \emph{HDT}, $v$ is a node identifier, and $\textit{targets}$ is a list of previously visited node identifiers. The displayed grammars suppress attribute tokens: labeled encodings also emit the atom type after each node and the bond type with each edge record.

\noindent\textbf{Partition tree.} The top-level partition $\sigma:V\to[k]$ assigns each node to one of $k$ communities,  with induced subgraph $G[C_i]=(C_i,E[C_i])$ on community $C_i=\sigma^{-1}(i)$ and cross-community edge sets $B_{ij}\subseteq E$ between communities $i$ and $j$. The hierarchical graph $\mathcal{H}$ is then a rooted partition tree: the root corresponds to $V$, and coarsening recursively partitions $G[C]$. Splitting is attempted when $|C|\geq\tau$; a community becomes a leaf below this size or when no proper split is found. We use $\tau=4$ by default, with block-model threshold interventions reported in Appendix~\ref{app:sbm_full}.

\subsubsection{\emph{HSENT}: declarative hierarchy}\label{app:grammar:hsent}

\emph{HSENT} \emph{declares} the hierarchy rather than traversing it: it writes a catalog of community blocks followed by a separate bipartite section that wires the communities together. Each community block stores the community's global node identifiers followed by a SENT-style~\citep{chen2025autograph} walk over its intra-community edges; the bipartite section then encodes every cross-community edge as an explicit pair $(i,j)$ of community indices and an atom-pair list. \emph{HSENT} therefore differs from the flat baseline SENT in exactly two respects, the community catalog and the bipartite section,  making these additional hierarchy records explicit.

Here $n_C=|C|$ and $\mathbf{g}_C$ is an ordered catalog mapping local indices to global node IDs.
Community IDs are unique within each sibling set; each bipartite section uses that scope.
A leaf walk and the endpoint list $\textit{pairs}$ use the corresponding local node indices.
 The star denotes repetition, the vertical bar denotes a choice, and a question mark denotes an optional group.
\begin{align*}
  \textit{seq} &::= \;\texttt{[SOS]}\;\;\textit{hier}\;\;\texttt{[EOS]} \\
  \textit{hier} &::= \; k\;\;\textit{part}^{*}\;\;\textit{bip}^{*} \\
  \textit{part} &::= \;\texttt{[LCOM]}\;\;\text{id}\;\;n_C\;\;\mathbf{g}_C\;\;\texttt{[SEP]}\;\;(\textit{walk}_C\mid\textit{hier})\;\;\texttt{[RCOM]} \\
  \textit{bip} &::= \;\texttt{[LBIP]}\;\;i\;\;j\;\;|B_{ij}|\;\;\textit{pairs}\;\;\texttt{[RBIP]}
\end{align*}

Every internal block emits the child count, child catalogs, and wiring between its children.
An edge is recorded at the lowest hierarchy node whose children separate its endpoints.
The recursive catalog cost is $O(m+\sum_C |C|)$, where the sum runs over catalog-emitting
community blocks; for bounded depth this is $O(n+m)$. Nonempty wiring blocks are bounded
by the number of cross-community edges, so they do not require a separate quadratic token term.

\begin{algorithm}[h]
\caption{\emph{HSENT} tokenization}\label{alg:hsent}
\begin{algorithmic}[1]
\Require Graph $G=(V,E)$; partition tree $\mathcal{H}$ with root $\mathcal{R}$
\Ensure Token sequence $\mathbf{t}$
\State $\mathbf{t}\leftarrow[\texttt{[SOS]}]$
\State \textsc{EmitHierarchy}$(\mathcal{R})$
\State $\mathbf{t}.\text{append}(\texttt{[EOS]})$;\quad \Return $\mathbf{t}$
\Statex
\Procedure{EmitHierarchy}{$C$}
\State Let $C_1,\dots,C_k$ be the children of $C$ (or $C$ itself if terminal)
\State $\mathbf{t}.\text{append}(k)$
\For{$i=1,\dots,k$} \Comment{community catalog}
    \State \textsc{EmitPart}$(C_i,\,i)$
\EndFor
\For{each pair $(i,j)$ with $i<j$ and $B_{ij}\neq\emptyset$} \Comment{bipartite wiring}
    \State $\mathbf{t}.\text{append}(\texttt{[LBIP]},\,i,\,j,\,|B_{ij}|)$
    \For{$(u,v)\in B_{ij}$ with $u\in C_i,\,v\in C_j$, in canonical order}
        \State $a_u\leftarrow$ local index of $u$ in $C_i$;\quad $a_v\leftarrow$ local index of $v$ in $C_j$
        \State $\mathbf{t}.\text{append}(a_u,\,a_v)$
    \EndFor
    \State $\mathbf{t}.\text{append}(\texttt{[RBIP]})$
\EndFor
\EndProcedure
\Statex
\Procedure{EmitPart}{$C$, $\text{id}$}
    \State $\mathbf{t}.\text{append}(\texttt{[LCOM]},\,\text{id},\,|C|,\,\mathbf{g}_C)$ \Comment{$\mathbf{g}_C$:  global node IDs in $C$}
    \State $\mathbf{t}.\text{append}(\texttt{[SEP]})$
    \If{$C$ is a leaf of $\mathcal{H}$} \Comment{SENT walk over $G[C]$}
        \State $\mathbf{t}.\text{append}(\textsc{SentWalk}(G[C]))$
    \Else \Comment{child catalogs and their wiring}
        \State \textsc{EmitHierarchy}$(C)$
    \EndIf
    \State $\mathbf{t}.\text{append}(\texttt{[RCOM]})$
\EndProcedure
\end{algorithmic}
\end{algorithm}

\subsubsection{\emph{HDT}: traversal-based hierarchy}\label{app:grammar:hdt}

\emph{HDT} encodes hierarchy implicitly through DFS nesting and emits no separate bipartite block.

\begin{align*}
  \textit{seq} &::= \;\texttt{[SOS]}\;\;\textit{hier}\;\;\texttt{[EOS]} \\
  \textit{hier} &::= \;\texttt{[ENTER]}\;\;\ell\;\;\text{id}\;\;(\textit{hier}\mid\textit{atom})^{*}\;\;\texttt{[EXIT]} \\
  \textit{atom} &::= \;v\;\;(\texttt{[LEDGE]}\;\;\textit{targets}\;\;\texttt{[REDGE]})?
\end{align*}

When DFS visits atom $v\in C_j$ and $v$ neighbors a previously visited atom $u\in C_i$ with $i\neq j$, the cross-community edge $(u,v)$ is emitted as a back edge to $u$ inside $v$'s edge bracket. This makes  intra- and inter-community edges syntactically identical and gives $O(n+m)$ tokens (about $33\%$ shorter than \emph{HSENT} on MOSES). For the randomized configurations, node identifiers and back-edge targets use serialization identifiers assigned globally in first-visit order. R retains stored community identifiers, whereas FR assigns them by encounter order within each sibling set
(Appendix~\ref{app:order_randomization}). The full pseudocode is given as Algorithm~\ref{alg:hdt} in the body of the paper (Section~\ref{sec:methods:tokenize}).

\subsubsection{\emph{HDTC}: typed molecular hierarchy}\label{app:grammar:hdtc}

\emph{HDTC} is the one configuration that brings its own coarsening. \emph{Motif plus Functional Group} (MC+FG) extends the Motif Community coarsening of Section~\ref{sec:methods:coarsen} with a functional-group detector, resolving overlaps by a fixed priority: rings first, then multi-atom functional groups, then single-atom functional groups. Every resulting community therefore carries a type label recording whether it is a ring ($R$), a functional group ($F$), or a singleton ($S$), and it is this typed hierarchy that \emph{HDTC} consumes.

 For a community $C$, the type token is $\tau_C\in\{\texttt{TYPE\_RING},\texttt{TYPE\_FUNC},\texttt{TYPE\_SINGLETON}\}$, abbreviated $R$, $F$ and $S$. The tokenizer opens each community block with this token before listing its atoms, and closes with a single trailing super-graph block enumerating the inter-community bonds. Like \emph{HDT}, \emph{HDTC} uses a traversal-based ordering. It additionally encodes community types and a separate cross-community wiring record. The legacy comparison in Section~\ref{sec:results:design} shows no consistent advantage over HDT.

\begin{align*}
  \textit{seq} &::= \;\texttt{[SOS]}\;\;\textit{comm}^{*}\;\;\textit{super}?\;\;\texttt{[EOS]} \\
  \textit{comm} &::= \;\texttt{[COMM\_START]}\;\;\tau_C\;\;\text{id}\;\;\textit{atoms}\;\;\texttt{[COMM\_END]} \\
  \textit{super} &::= \;\texttt{[SUPER\_START]}\;\;(c_s\;c_t\;a_s\;a_t)^{*}\;\;\texttt{[SUPER\_END]}
\end{align*}

 Each typed block gives the community ID and the atoms in traversal order. Legacy HDTC uses graph-global atom IDs and a global seen set: back-edge brackets record all previously emitted neighbors, including those in other communities. In each super-graph tuple $(c_s,c_t,a_s,a_t)$, $c_s,c_t$ identify the endpoint communities and $a_s,a_t$ are global atom IDs. The super-graph therefore repeats cross-community bonds, which decoding coalesces.  With $c$ cross-community edges, the total cost is $O(n+m+c)$ tokens.

\begin{algorithm}[h]
\caption{\emph{HDTC} tokenization (molecular benchmarks)}\label{alg:hdtc}
\begin{algorithmic}[1]
\Require Molecular graph $G=(V,E)$; MC+FG partition $\sigma:V\to[k]$ with type labels  $\tau_C$
\Ensure Token sequence $\mathbf{t}$
\State $\mathbf{t}\leftarrow[\texttt{[SOS]}]$; $V_{\mathrm{seen}}\leftarrow\emptyset$
\State Let $C_1,\dots,C_k$ be the communities under $\sigma$, each with type $\tau_{C_i}$
\For{$i=1,\dots,k$} \Comment{typed community blocks}
    \State $\mathbf{t}.\text{append}(\texttt{[COMM\_START]},\,\tau_{C_i},\,i)$
    \For{$v\in C_i$ in traversal order}
        \State $\mathbf{t}.\text{append}(v)$ \Comment{global atom ID}
        \State $B\leftarrow N_G(v)\cap V_{\mathrm{seen}}$
        \If{$B\neq\emptyset$}
            \State $\mathbf{t}.\text{append}(\texttt{[LEDGE]},\,\text{sorted}(B),\,\texttt{[REDGE]})$
        \EndIf
        \State $V_{\mathrm{seen}}\leftarrow V_{\mathrm{seen}}\cup\{v\}$
    \EndFor
    \State $\mathbf{t}.\text{append}(\texttt{[COMM\_END]})$
\EndFor
\If{$\bigcup_{i<j} B_{ij}\neq\emptyset$} \Comment{super-graph block: cross-community bonds}
    \State $\mathbf{t}.\text{append}(\texttt{[SUPER\_START]})$
    \For{each $(u,v)\in E$ with $\sigma(u)\neq\sigma(v)$, in canonical order}
        \State $c_s\leftarrow\sigma(u)$;\quad $c_t\leftarrow\sigma(v)$
        \State $a_s\leftarrow u$;\quad $a_t\leftarrow v$ \Comment{global atom IDs}
        \State $\mathbf{t}.\text{append}(c_s,\,c_t,\,a_s,\,a_t)$
    \EndFor
    \State $\mathbf{t}.\text{append}(\texttt{[SUPER\_END]})$
\EndIf
\State $\mathbf{t}.\text{append}(\texttt{[EOS]})$;\quad \Return $\mathbf{t}$
\end{algorithmic}
\end{algorithm}

\subsubsection{What invertibility does and does not guarantee}\label{app:validity}

All three schemes support lossless reconstruction from their complete, untruncated encodings (Section~\ref{sec:methods:decode}). This guarantee does not extend to arbitrary sequences sampled from the transformer. Samples can contain unbalanced \texttt{[ENTER]}/\texttt{[EXIT]} nesting, references to nodes that were never emitted, or truncated blocks. The parser is permissive, so even malformed or incomplete sequences may yield a graph. After the evaluation protocol's normalization, we assess that graph using the dataset-specific structural or RDKit checks. A syntax violation therefore does not automatically make a sample invalid.

The same limitation applies to \emph{SENT}. Its complete encodings support lossless reconstruction, but sampled sequences can be malformed or decode to disconnected or chemically invalid graphs. For every method, validity must therefore be measured empirically.

As discussed in Appendix~\ref{app:traversal_vs_declaration}, \emph{HDT} has no separate coarse wiring record that can conflict with its node-level edges. \emph{HSENT} and \emph{HDTC} can produce such conflicts because they encode cross-community connectivity separately. This is a possible explanation for their weaker performance, but the experiments do not isolate this mechanism.

\subsection{Coarsening visualizations across datasets}\label{app:coarsening_viz}

\begin{figure}[h]
\centering
\includegraphics[width=\linewidth]{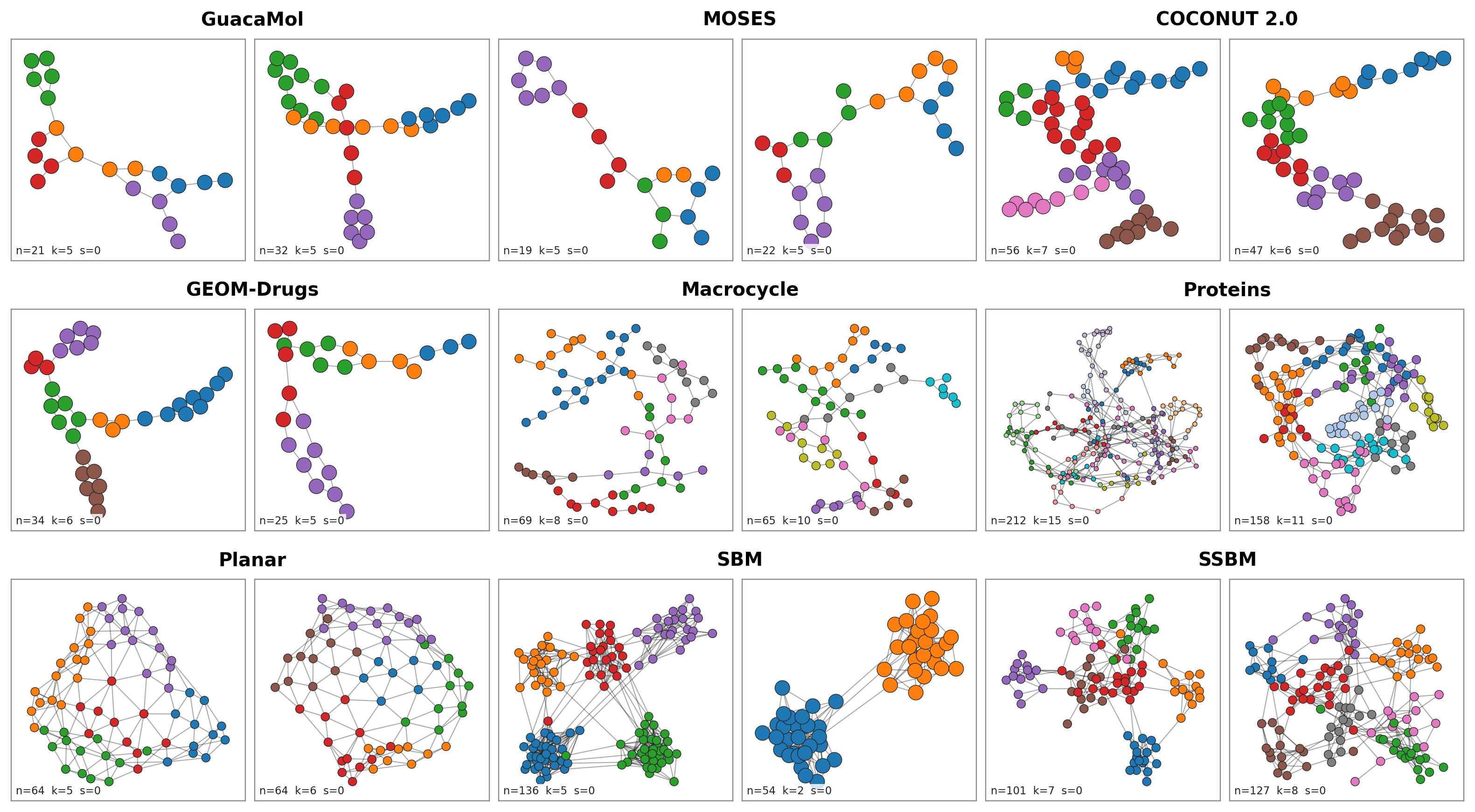}
\caption{\textbf{Production HAC coarsening across the benchmarks.} Two held-out examples per dataset use the average-linkage HAC implementation and modularity-selected cut used in training, with stopping threshold $\tau=4$. Molecular affinities use kekulized bond orders; other graphs use unit edge weights. Colors distinguish top-level communities within each panel, with no shared meaning across graphs. Labels give the node count $n$, community count $k$, and singleton count $s$.}
\label{fig:coarsen_hac}
\end{figure}

Figure~\ref{fig:coarsen_hac} shows HAC partitions for representative graphs from each benchmark. Node colors indicate community membership. These examples illustrate the coarsening used in the quantitative comparisons of Section~\ref{sec:results:design}.

\section{ Training, evaluation, and data details}
\subsection{Training populations and evaluation reference sets}\label{app:dataset_geometry}

\noindent\textbf{Dataset sources and software.}
For COCONUT 2.0~\citep{chandrasekhar2025coconut2}, we use the September 2024
snapshot archived at \url{https://doi.org/10.5281/zenodo.13692394},
specifically \texttt{coconut-09-2024.csv.zip}. This archive is distributed
under CC BY 4.0.
For Macrocycle, we use the Macrocycle-DB~\citep{jiang2026macrocycle}
\texttt{All.csv} export accessed on 15 August 2026 from
\url{https://macro-db.dpbio.tech/Downloads/All.csv}.
The downloaded export contains $50{,}653$ records, rather than the
$45{,}525$ reported in the database publication.
Both datasets are processed using canonical, non-isomeric,
implicit-hydrogen molecular identities and filtered for compatibility
with our lossless graph codec.
The training and evaluation environments pin RDKit~\citep{rdkit}
version \texttt{2024.03.6}, distributed under the BSD 3-Clause license.

Table~\ref{tab:appendix_dataset_stats} lists the training populations, held-out reference sets, and evaluation draw sizes for the five molecular benchmarks. Standard and novel-only evaluation use the same held-out references. They differ only in which generated molecules are scored. Each \texttt{MoleculePGD} draw uses half the evaluation-arm size. The training-sample reference uses an independently permuted prefix of the exact training population, with its size matched to the evaluation population. This controls sample size when interpreting the training-sample reference in Figure~\ref{fig:uncond_summary}.

\input{tables/appendix_dataset_stats}

\subsection{Hierarchy depths under recursive coarsening}\label{app:hierarchy_depths}

HAC is applied recursively: a split is attempted for communities of at least $\tau=4$
nodes, and recursion stops below this size or when no proper split is found. \emph{HDTC}'s hierarchy is two-level by construction (community plus super graph) and
therefore not part of this analysis. Table~\ref{tab:appendix_hierarchy_depths} reports the
observed depth distribution from $200$ randomly sampled graphs per benchmark. Its depth counts levels below the top-level communities; add one to obtain maximum leaf depth under the root-depth-zero convention of the algorithms and Table~\ref{tab:appendix_block_alignment}. Depth grows
with mean graph size: MOSES has mean depth $1.13$, whereas the larger Macrocycle graphs have
mean depth $1.86$; the maximum observed depth is $2$ and $3$, respectively.

\input{tables/appendix_hierarchy_depths}

\subsection{Token sequence lengths}\label{app:token_length}

Sequence length affects the quadratic attention component, but total compute also depends on model operations, padding, context limits and batching.  Table~\ref{tab:appendix_token_length}a reports mean lengths from the molecular sample audit and historical full-population audits described in its caption: \emph{HSENT} is roughly $3\times$ SENT and \emph{HDT} roughly $2\times$, motivating training-time calibration of update budgets rather than equal-step training. The achieved runtimes are reported separately.  Table~\ref{tab:appendix_token_length}b records the main context caps; configuration-specific block-model and scaling settings are referenced in its caption. Comparing a configuration's cap with its length distribution indicates its truncation headroom.

\input{tables/appendix_token_length}

\subsection{Training inventory}\label{app:hyperparameters}

The main comparisons use GPT2-XS ($\sim\!11$M parameters) across ten benchmarks.
Table~\ref{tab:appendix_arch} lists the defaults for this regime;
Tables~\ref{tab:appendix_train_inventory_mol} and~\ref{tab:appendix_train_inventory_nonmol}
report configuration-specific schedules, contexts, batch sizes and achieved runtimes.
Warmup and decay are scaled to each run's calibrated update budget
(Section~\ref{sec:results:setup}); calibration targets comparable training time rather
than equal updates and does not guarantee equal achieved runtime.
Presentations are optimizer steps times effective batch size.
Legacy tokenizers use the context limits in Table~\ref{tab:appendix_token_length};
block-model interventions are detailed in Appendix~\ref{app:sbm_full},
the larger GPT2-S regime in Appendix~\ref{app:regime_delta},
and conditional generation in Appendix~\ref{app:conditional_generation}.

\noindent\textbf{Hardware and timing.} Models are trained from scratch on NVIDIA H200 NVL GPUs
(143\,GB). The main GPT2-XS runs use one GPU without gradient accumulation.
Their achieved runtimes are given in Table~\ref{tab:main_molecular_all} and
Tables~\ref{tab:appendix_train_inventory_mol} and~\ref{tab:appendix_train_inventory_nonmol}.
Scaling training times are stated on a one-GPU basis in Appendix~\ref{app:regime_delta}.
The bf16-mixed precision setting describes training; evaluation uses its recorded sampling configuration.

\input{tables/appendix_architecture}

\input{tables/appendix_train_inventory_molecular}

\input{tables/appendix_train_inventory_nonmolecular}

\subsection*{Evaluation protocols}
\addcontentsline{toc}{subsection}{\protect\numberline{}Evaluation protocols}

\begin{itemize}
    \item \textbf{Standard.} Molecular metrics are computed on every valid generated
    molecule against the held-out reference arm of
    Table~\ref{tab:appendix_dataset_stats}. For graph benchmarks, the four MMDs and Ratio
    are computed on decoded graphs and dataset-specific structural validity is reported
    separately, so the distributional rows remain defined even when validity is low. This
    is the conventional protocol and the one summarized in the body.

    \item \textbf{Novel-only.} For molecules, the distributional and structural metrics
    are recomputed on valid generations whose canonical SMILES is absent from the exact
    training population. For graphs, a sample must pass the dataset-specific validity test
    and be exactly non-isomorphic to every training graph. We decode a prespecified
    oversampled pool and take the first fixed number of qualifying graphs per seed: $184$
    for Proteins, $40$ for Planar, and $1{,}000$ for SBM and SSBM. Graph tables report the
    raw valid-novel yield and uniqueness of the selected population; validity and novelty
    are one by construction and V.U.N.\ is redundant. All metrics use the same held-out
    reference arms as the standard protocol (Appendices~\ref{app:unconditional_full}
    and~\ref{app:novel_graphs}).

    \item \textbf{Training-sample reference.} Both protocols are read against an
    independently permuted prefix of the exact training population, up to the evaluation
    size. When fewer training objects are available, as for the 128-graph SBM and SSBM
    splits, all available training graphs are used in a fresh order for each seed. This is
    a calibration rather than a target and is excluded from every ranking and
    change-relative-to-SENT panel. Because training objects have zero novelty by
    construction, their V.U.N.\ is zero.
\end{itemize}

\subsection{Generation details}\label{app:generation_details}

Table~\ref{tab:appendix_generation_details} separates generation-attempt budgets from
metric-specific draw sizes. Molecular attempt budgets equal the reference counts in
Table~\ref{tab:appendix_dataset_stats}, but invalid attempts reduce the valid generated
population available to distributional metrics. Motif discrepancies use fixed $1000$-valid-molecule
arms, and MoleculePGD uses its specified subsampling protocol.
Proteins and Planar use $184$ and $40$ attempts per seed, respectively;
SBM, SSBM and RSBM use $1000$, with independently generated $1000$-graph reference arms
rather than the $40$-graph dataset test split.
Molecular unconditional decoding uses top-$k=5$ and temperature $1.0$; graph settings
are dataset-specific. Conditional generation instead uses the valid-output stopping rule
and top-$k=10$ in Appendix~\ref{app:conditional_generation}.
Tokenizer-specific sequence limits are in Table~\ref{tab:appendix_token_length}b,
with RSBM settings in Table~\ref{tab:appendix_block_vun}.
\input{tables/appendix_generation_details}

\subsection{Evaluation metrics}\label{app:metrics}

We compare generated samples against reference sets using a hierarchy of metrics drawn from MOSES~\citep{polykovskiy2020moses}, AutoGraph~\citep{chen2025autograph}, and the GraphRNN/SPECTRE convention~\citep{martinkus2022spectre}. The molecular benchmarks (GuacaMol, MOSES, COCONUT 2.0, GEOM-Drugs, Macrocycle) use the molecular suite; the graph benchmarks (Proteins, Planar, SBM, SSBM, RSBM) use graph-statistic MMDs paired with dataset-specific structural validity tests. The categories below measure complementary aspects of generation quality.

\subsubsection*{Molecular metrics}

\begin{itemize}
    \item \textbf{Core generation quality (Validity, Uniqueness, Novelty, V.U.N.).} Baseline filter for decoding competence. \emph{Validity} confirms the model has learned basic chemical valency rules; for the molecular benchmarks Validity is the fraction of generated SMILES that RDKit parses with valid valences. \emph{Uniqueness} checks that the model avoids generating duplicates within a sample, computed as the fraction of distinct canonical SMILES among the valid set. \emph{Novelty} ensures the model is not merely memorizing training data, computed as the fraction of valid generations whose canonical SMILES is absent from the training set. \emph{V.U.N.} is the number of distinct valid canonical molecules absent from the full training population, divided by the number of generation attempts. We compute this joint count for each seed before reporting its mean and sample SD; it is not the product of the marginal V, U and N scores. Low validity indicates chemical decoding failures, whereas low uniqueness or novelty indicates repetition or training-set overlap. As discussed in Section~\ref{sec:results:unconditional}, uniqueness and novelty are sensitive to the strength of the structural prior and to how freely the serialization is randomized, so they are necessary but not sufficient measures of generation quality.

    \item \textbf{Global distribution (FCD, SNN).} These metrics test whether generated molecules ``look like'' real ones in aggregate. \emph{Fr\'echet ChemNet Distance (FCD)}~\citep{preuer2018frechet} is the distance between generated and reference distributions in the hidden space of a pre-trained ChemNet, and captures high-level biological and chemical features. \emph{Nearest neighbor similarity (SNN)} is the average Tanimoto similarity between each generated molecule and its closest reference; low SNN indicates lower similarity to the chosen reference set.

    \item \textbf{Structural identity (Frag, Scaff, IntDiv).} These zoom in on local and global topology. \emph{Fragment similarity (Frag)} is the cosine similarity between BRICS fragment-frequency vectors. \emph{Scaffold similarity (Scaff)} is the cosine similarity between Bemis--Murcko scaffold-frequency vectors, comparing molecular cores after side chains are removed. \emph{Internal diversity (IntDiv)} is the mean pairwise Tanimoto distance within the generated set; together with novelty it detects mode collapse where the model repeats a small set of safe structures.

    \item \textbf{Motif fidelity (FG, SMARTS and Ring MMD$^2$; BRICS L2).}
    Functional-group features use RDKit fragment counters; SMARTS features count curated
    patterns such as carbonyl, amide, ether and halogen; ring features count sizes $3$--$8$
    and aromatic, aliphatic and heterocyclic rings. For these three count-vector distributions
    we report the squared-MMD V-statistic with a unit-bandwidth Gaussian kernel.
    BRICS L2 is the Euclidean distance between normalized BRICS fragment-frequency vectors.
    Lower values indicate closer agreement with the reference features.
    Both arms use exactly $1{,}000$ valid molecules: the first qualifying generations in frozen
    order and $1{,}000$ molecules from the fixed test pool. Fewer valid generations cause the
    metric to fail rather than silently reducing an arm.

    \item \textbf{PolyGraph Discrepancy (\texttt{MoleculePGD}).} PGD~\citep{krimmel2025polygraph}
    approximates the Jensen-Shannon distance between the reference and generated
    distributions by training a binary classifier to tell them apart and deriving a score
    from its held-out class probabilities via log-likelihood; lower means harder to distinguish. We use the official
    \texttt{MoleculePGD} implementation from the public PolyGraph benchmark, which computes
    the discrepancy over molecular descriptors that retain chemical information rather than
    over unlabeled graph topology, at a pinned package version with audited source hashes.
    We report the interval form: ten replicates, each drawing reference and generated
    molecules without replacement, splitting each arm into a descriptor-selection half and a
    held-out half, so the reported value carries resampling uncertainty and a record of
    which descriptor was selected. The official estimator clips finite-sample divergence estimates before taking a square root, including within each interval replicate. A zero score therefore does not establish distributional equality. The interval quantifies resampling variability; we retain full numerical precision for aggregation and round only for display.

\end{itemize}

\subsubsection*{Graph theoretic metrics}

\begin{itemize}
    \item \textbf{Distribution distance MMDs (Degree, Spectral, Clustering, Orbit).} For non-molecular benchmarks (SBM, SSBM, RSBM, Planar, Proteins) we follow the GraphRNN and SPECTRE conventions~\citep{martinkus2022spectre}. We summarize degree, normalized-Laplacian eigenvalues and clustering coefficients by histograms, and ORCA node-orbit counts by their per-graph average vector. We compare these graph-feature distributions with the benchmark Gaussian kernels, using total-variation distance for the reported runs. As in the benchmark convention, rows labeled MMD report the squared-MMD V-statistic. \emph{Degree MMD} compares degree distribution histograms; \emph{Spectral MMD} compares the eigenvalue spectra of the graph Laplacian; \emph{Clustering MMD} compares the distribution of local clustering coefficients; \emph{Orbit MMD} compares counts of small subgraph orbits computed by ORCA. Lower is better on all four.

    \item \textbf{Ratio.} A single dimensionless summary statistic over the four MMDs above, following AutoGraph's evaluation protocol~\citep{chen2025autograph}. For each statistic $i\in\{\mathrm{Deg},\mathrm{Spec},\mathrm{Clust},\mathrm{Orbit}\}$ we compute the ratio of the generated-vs-reference MMD to the train-vs-reference MMD, $r_i = \mathrm{MMD}_i^{\text{gen}} / \mathrm{MMD}_i^{\text{train}}$, and Ratio is the mean of the finite $r_i$. A value near $1$ would indicate that generated samples are statistically as close to the reference as the training set itself; large values indicate the generated distribution drifts away from reference more than train does on average. Lower is better. On Proteins, the denominator uses the full training population, whereas the training-sample column evaluates an independently drawn subset of that population against the same held-out references; its Ratio need not equal one. The published values transcribed in Table~\ref{tab:autograph_comparison} instead use a train-normalized geometric-mean convention. These Ratio columns are not on a common scale, so we do not rank across them.

    \item \textbf{Structural validity.} \emph{Validity} on the graph benchmarks is a dataset-specific structural test rather than a chemical one: a connected graph with at least one edge for Proteins; a non-empty, connected planar graph for Planar; and fitted community structure consistent with the corresponding generator for SBM, SSBM and original RSBM. Connected-RSBM scores measure compatibility with the original RSBM gate, without an additional connectivity requirement (Appendix~\ref{app:block_datasets}). Higher is better.

    \item \textbf{Set-level novelty (Novelty) and V.U.N.}
    Graph identity is tested by exact graph isomorphism, with inexpensive signatures used as
    candidate filters where applicable. Uniqueness counts distinct generated graph classes;
    Novelty counts generations absent from training. Both use the number of attempts as
    denominator; Proteins additionally requires validity for its Novelty count, whereas
    block-model and Planar Novelty are not validity-gated. V.U.N.\ counts distinct valid
    graph classes absent from training, divided by attempts. These are the counting rows
    in Appendix~\ref{app:nonmolecular_full}. Near-zero validity, as on Planar, limits the
    resolution of V.U.N.; the distributional MMD rows remain informative
    (Appendix~\ref{app:planar_full}).

\end{itemize}

\FloatBarrier
\section{Molecular unconditional generation}\label{app:unconditional_full}\label{app:novel_only}

Each dataset subsection pairs the standard and novel-only results for the same
configurations, making it possible to compare distributional fidelity before and after
excluding reproduced training molecules. The paired tables use the following reporting
conventions throughout this section.

\noindent\textbf{Reading the tables.}
The upper panel reports absolute performance as mean\,$\pm$\,sample SD over generation
seeds $42$--$44$ from one trained checkpoint per configuration. Bold and underlining mark
the best and second-best generated model in each row; $^{*}$ identifies our variants.
The training-sample reference is an independently permuted prefix of the training population
(Appendix~\ref{app:metrics}) and is excluded from ranking. It remains unfiltered in the
novel-only tables. The lower panel reports percentage change relative to SENT, with the
sign chosen so that positive is better. Green and red shading are scaled separately within
each row, so color intensity is not comparable across rows. Changes are computed from
unrounded aggregates and may differ from values recomputed from the displayed rounded
means. The legacy configurations in Table~\ref{tab:appendix_legacy_variants} retain their
separate protocol caveat.

\noindent\textbf{Standard evaluation.}
We follow the evaluation protocol of \citet{chen2025autograph}: (i) largest-connected-component
salvage at validity time; (ii) the per-dataset decoding parameters of
Table~\ref{tab:appendix_generation_details}; (iii) dataset-specific extras (GuacaMol
KL\,Div); and (iv) the held-out reference arm of Table~\ref{tab:appendix_dataset_stats} as the
reference for SNN, Frag, Scaff, and PGD. For the labeled-graph SENT tokenizer we apply
AutoGraph's \texttt{LabeledGraph} \texttt{LogitsProcessor} FSM at sampling time on MOSES; on
the other molecular benchmarks we disable it, because the constraint admits out-of-walk-order
position indices that drive the decoder to emit unlabeled phantom nodes, and disabling it
matches the SMILES-LSTM and pre-FSM SENT baselines in the literature.
Table~\ref{tab:autograph_comparison} preserves single-value reporting for the cited methods
only.

\noindent\textbf{Novel-only evaluation.}
The novel-only tables evaluate valid generated molecules absent from the exact training population (Appendix~\ref{app:metrics}). All four hierarchical variants retain lower FCD and \texttt{MoleculePGD} than SENT on all five benchmarks. These gains therefore persist after reproduced training molecules are removed.

Some similarity and motif comparisons change. On GuacaMol, both hierarchy-wide randomization variants have lower SNN and scaffold similarity than SENT under novel-only evaluation, with deficits of up to $17.9\%$ and $15.6\%$. Under standard evaluation, \emph{HDT}-MC-FR had higher SNN than SENT. On COCONUT 2.0, \emph{HDT}-HAC-R has higher values on all four motif discrepancies than SENT under novel-only evaluation, despite having lower values under standard evaluation. Motif-level conclusions on this benchmark therefore depend on the evaluation protocol.

\begin{figure}[htbp]
\centering
\includegraphics[width=\linewidth]{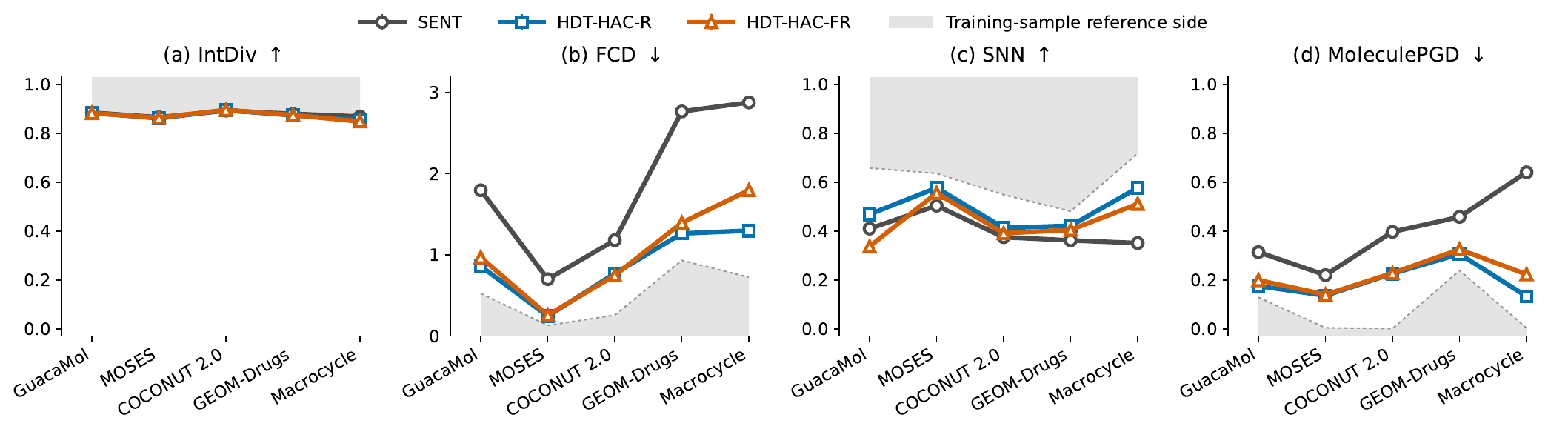}
\caption{\textbf{Novel-only unconditional summary panels.} Metrics are recomputed on the
valid generated molecules absent from the exact training population. Markers report means
over three generation seeds and error bars the sample SD. The gray training-sample
reference is unfiltered and is shown only as an in-distribution reference. Compare
Figure~\ref{fig:uncond_summary}.}
\label{fig:novel_summary}
\end{figure}
\FloatBarrier

\clearpage
\subsection{GuacaMol}\label{app:guacamol_full}

Table~\ref{tab:appendix_guacamol_full} reports GuacaMol. Validity ranges from $0.943$ to $0.976$ across the five generated configurations, and uniqueness is $1.000$ for all five. Distributional metrics show larger differences. FCD ranges from $0.871$ to $1.799$, and \texttt{MoleculePGD} ranges from $0.176$ to $0.309$. \emph{HDT}-HAC-R has the best FCD, SNN, fragment similarity, scaffold similarity, \texttt{MoleculePGD}, and GuacaMol KL score. \emph{HDT}-MC-FR is the only configuration with higher V.U.N.\ than SENT, by $0.58\%$.

\begin{table}[!htbp]
\input{tables/appendix_guacamol_full}
\par\medskip
\input{tables/appendix_guacamol_novel}
\end{table}
\FloatBarrier

\clearpage
\subsection{MOSES}\label{app:moses_full}

Table~\ref{tab:appendix_moses_full} reports MOSES, which is the benchmark on which the
hierarchical serializations are least costly and most uniformly ahead. Every hierarchical
variant except \emph{HDT}-MC-R improves on SENT's validity, all four reduce FCD by more than
$62\%$ and \texttt{MoleculePGD} by more than $35\%$, and all four improve SNN, fragment and
scaffold similarity. The residual V.U.N.\ deficit is the smallest in the paper, between
$0.91\%$ and $3.99\%$, with novelty the main contributor; validity and uniqueness also contribute for some variants. The functional-group MMD is the one
row where SENT leads more often than not, by margins at or below $0.0001$ in absolute terms.

\begin{table}[!htbp]
\input{tables/appendix_moses_full}
\par\medskip
\input{tables/appendix_moses_novel}
\end{table}
\FloatBarrier

\clearpage
\subsection{COCONUT 2.0}\label{app:coconut_full}

Table~\ref{tab:appendix_coconut_full} reports COCONUT 2.0. Every hierarchical variant improves all four motif discrepancies over SENT. SMARTS MMD, ring MMD, and BRICS L2 improve by up to $68.2\%$, $82.9\%$, and $49.9\%$, respectively. These gains come with a V.U.N. cost. \emph{HDT}-HAC-R has $17.0\%$ lower V.U.N.\ than SENT, with novelty of $0.880$ versus $0.979$. Hierarchy-wide randomization reduces this deficit. \emph{HDT}-MC-FR has V.U.N.\ within $2.98\%$ of SENT while reducing FCD by $50.4\%$ and \texttt{MoleculePGD} by $59.5\%$. The latter is the largest \texttt{MoleculePGD} improvement among the variants on this benchmark.

\begin{table}[!htbp]
\input{tables/appendix_coconut_full}
\par\medskip
\input{tables/appendix_coconut_novel}
\end{table}
\FloatBarrier

\clearpage
\subsection{GEOM-Drugs (disjoint)}\label{app:geom_full}

Table~\ref{tab:appendix_geom_drugs_full} reports the audited disjoint split of GEOM-Drugs. Similar distributional gains occur with roughly one eighth of the training population of GuacaMol and MOSES. Relative to SENT, every hierarchical variant reduces FCD by roughly half, \texttt{MoleculePGD} by $28.6\%$--$34.3\%$, and ring MMD by $52.3\%$--$61.7\%$. The randomization policies differ more clearly on V.U.N. The \emph{-R} variants have deficits of $7.7\%$ and $10.9\%$, while both \emph{-FR} variants are within $0.18\%$ of SENT. On this benchmark, hierarchy-wide randomization therefore retains the distributional gains with little V.U.N. loss.

\begin{table}[!htbp]
\input{tables/appendix_geom_drugs_full}
\par\medskip
\input{tables/appendix_geom_drugs_novel}
\end{table}
\FloatBarrier

\clearpage
\subsection{Macrocycle}\label{app:macrocycle_full}

Table~\ref{tab:appendix_macrocycle_full} reports Macrocycle. Relative to SENT, \texttt{MoleculePGD} improves by up to $86.6\%$, FCD by up to $60.8\%$, and SNN by up to $84.7\%$. Every hierarchical variant has roughly eight to fifteen times SENT's scaffold similarity, although SENT's score is low at $0.035\pmstd{0.001}$.

The within-community variants also have substantial V.U.N.\ deficits. \emph{HDT}-MC-R and \emph{HDT}-HAC-R have $35.4\%$ and $37.2\%$ lower V.U.N.\ than SENT. These deficits mainly reflect novelty, which falls to $0.657$ and $0.658$ from SENT's $0.999$. \emph{HDT}-MC-FR reaches novelty $0.987$ and V.U.N.\ within $1.49\%$ of SENT while retaining a $45.9\%$ FCD reduction. \emph{HDT}-HAC-FR still has a $10.9\%$ V.U.N.\ deficit, mainly because of lower validity rather than novelty. Macrocycle therefore shows both substantial fidelity gains and a remaining V.U.N.\ cost.

\begin{table}[!htbp]
\input{tables/appendix_macrocycle_full}
\par\medskip
\input{tables/appendix_macrocycle_novel}
\end{table}
\FloatBarrier

\subsection{Declarative and typed variants}\label{app:legacy_variants}

Table~\ref{tab:appendix_legacy_variants} reports \emph{HSENT}-HAC-R, \emph{HSENT}-MC-R and
\emph{HDTC}-R on GuacaMol and MOSES, the two benchmarks on which they were evaluated.
Appendix~\ref{app:traversal_vs_declaration} draws the conclusion.

\input{tables/appendix_legacy_variants}
\FloatBarrier

\section{Non-molecular unconditional generation}\label{app:nonmolecular_full}\label{app:novel_graphs}

Each dataset subsection places its standard and novel-only results together where both
are available. The block-model generator definitions and hierarchy interventions are
reported separately from the per-dataset distributional comparisons.

\noindent\textbf{Reading the tables.}
Generated-model entries report mean\,$\pm$\,sample SD over generation seeds $42$--$44$
from one trained checkpoint per configuration. Bold and underlining identify the best and
second-best generated model in each row; $^{*}$ marks our variants. Where present, the
training-sample column is an unfiltered calibration reference and is excluded from ranking,
including in the novel-only tables. The upper panel reports absolute performance; the lower
panel reports percentage change relative to SENT, signed so that positive is better.
Green and red shading are scaled separately within each row. Metric arrows indicate the
favorable direction.

\noindent\textbf{Standard and novel-only evaluation.}
Standard distributional metrics are computed on decoded graphs, with structural validity
reported separately (Appendix~\ref{app:metrics}).
For graphs, the novel-only protocol first enforces the dataset-specific structural validity
criterion and exact non-isomorphism to every training graph, then evaluates a fixed
qualifying population while retaining duplicates. The raw yield therefore records how often
a model reaches that population, and selected uniqueness records repetition within it.
Validity and novelty are one by construction in the selected novel-only population, so
those tables report raw qualifying yield and selected uniqueness instead.

\clearpage
\subsection{Proteins}\label{app:protein_full}

Table~\ref{tab:appendix_proteins_full} reports Proteins, the largest graphs we evaluate. This benchmark also requires the largest context limit (Table~\ref{tab:appendix_token_length}b). \emph{HDT}-HAC-R improves every reported metric over SENT. V.U.N.\ improves by $62.8\%$, and the four graph MMDs improve by $78.6\%$--$91.6\%$. Its Ratio is $1.568\pmstd{0.668}$, compared with $1.421\pmstd{0.263}$ for the training-sample reference. These gains show that the approach can help outside molecular graphs without a chemical motif catalog.

Hierarchy-wide randomization raises V.U.N.\ further to $0.960\pmstd{0.006}$. However, it has higher values on all four graph MMDs than \emph{HDT}-HAC-R, and its clustering MMD is effectively unchanged from SENT's. Within-community randomization therefore performs better on structural fidelity in this comparison.

\noindent\textbf{Novel-only evaluation.}
On Proteins, the large Degree, Orbit and Ratio improvements of \emph{HDT}-HAC-R persist
after filtering; \emph{HDT}-HAC-FR also improves those measures but not Clustering MMD
(Table~\ref{tab:appendix_proteins_novel}).

\begin{table}[!htbp]
\input{tables/appendix_proteins_full}
\par\medskip
\input{tables/appendix_proteins_novel}
\end{table}
\FloatBarrier

\clearpage
\subsection{Planar}\label{app:planar_full}

Table~\ref{tab:appendix_planar_full} reports Planar, and it is a negative result. Neither
hierarchical configuration produces a planar graph across $120$ attempts, and both are worse
than SENT on all four graph MMDs and on the derived Ratio. One qualification is essential:
the SENT run yields only $2$ valid planar graphs from $120$ attempts, a mean validity of
$0.0167$, so the validity and V.U.N.\ rows compare two near-zero quantities and carry little
weight on their own. The MMD rows do not depend on the validity filter and they  favor the
flat walk unambiguously, by between $14.7\%$ and $195.1\%$. These results show that the reported coarsening and training configurations do not
improve fidelity on this Planar benchmark. They do not establish that planar graphs lack
community structure or that a different coarsening could not help.

\noindent\textbf{Novel-only evaluation.}
Planar remains a negative result: only $0.7\%$,
$0.1\%$ and $0.3\%$ of the raw SENT, \emph{HDT}-HAC-R and
\emph{HDT}-HAC-FR pools qualify, respectively, and both hierarchical variants are
worse on every MMD (Table~\ref{tab:appendix_planar_novel}).

\begin{table}[!htbp]
\input{tables/appendix_planar_full}
\par\medskip
\input{tables/appendix_planar_novel}
\end{table}
\FloatBarrier

\clearpage
\subsection{Block-model datasets and the planted-hierarchy control}
\label{app:block_datasets}

Each family contains $128/32/40$ train/validation/test graphs.
Table~\ref{tab:appendix_block_generators} specifies the generators;
Table~\ref{tab:appendix_block_populations} summarizes the realized
populations.
SBM is unfiltered for connectivity.
SSBM rejects disconnected graphs and deduplicates the population.
For RSBM, we distinguish the original, unfiltered population from
the connected population used in the threshold and PH experiments.
These families differ in community sizes and sampling conditions,
so they do not constitute a density-only or depth-only ablation.

 RSBM samples $2$--$5$ coarse communities. Community counts and sizes are sampled independently; $U$ denotes the discrete-uniform law over each indicated inclusive integer range.
Coarse community $i$ contains $F_i\sim U\{2,\dots,5\}$ fine
communities with sizes $L_{ij}\sim U\{8,9\}$, giving
 $S_i=\sum_{j=1}^{F_i} L_{ij}$.
Define the numbers of within-fine and between-fine pairs by
\[
A_i=\displaystyle\sum_{j=1}^{F_i}\binom{L_{ij}}{2},
\qquad
B_i=\binom{S_i}{2}-A_i.
\]
Edges are independent Bernoulli draws with probabilities
\[
p_{\mathrm{middle},i}
 =\frac{0.30(A_i+B_i)}{R A_i+B_i},
\qquad
p_{\mathrm{fine},i}=R\,p_{\mathrm{middle},i},
\qquad
p_{\mathrm{outside}}=0.005.
\]
Thus $R=4$ redistributes the expected internal edges toward fine
communities while preserving $0.30\binom{S_i}{2}$ expected edges
inside each coarse community.
It does not preserve SBM's community-size or degree distributions.

\noindent\textbf{PH and validity.}
PH uses recorded planted memberships during training: one level
for SBM and two for RSBM.
Sampling remains unconditional.
PH provides a privileged training-structure control, rather than
a theoretical upper bound on HAC.
Validity is assessed from decoded graph topology, not emitted
community annotations.
For connected RSBM, $V^\dagger$ and V.U.N.$^\dagger$ use the
original unconditional validity gate; generated connectivity is
not an additional validity requirement.

\input{tables/appendix_block_generators}
\input{tables/appendix_block_populations}
\FloatBarrier

\clearpage
\subsection{Block-model hierarchy interventions}
\label{app:sbm_full}

Table~\ref{tab:appendix_block_alignment} measures partition recovery
on training graphs using adjusted Rand index (ARI; $1$ denotes
identical partitions).
SBM's $\tau=44$ was selected to maximize terminal-partition ARI
while retaining the root partition.
Connected RSBM's $\tau=10$ was selected in the initial sweep to
maximize the fraction of nodes in $8$--$9$-node leaves.
Both choices use planted training information.

Table~\ref{tab:appendix_block_vun} separates validity, uniqueness,
novelty,  their joint U.N.\ score, and V.U.N.
The joint scores are measured directly, not obtained by multiplying
the marginal columns.
Default HAC-R exhibits repetition and training-set memorization;
HAC-FR largely avoids these failures but loses validity.
Threshold adjustment improves SBM under both policies.
On connected RSBM, adjusted HAC-R approaches PH-R, while the
corresponding FR configurations remain far apart.

These comparisons include configuration changes beyond the hierarchy.
On SBM, default runs use context $2048$, whereas threshold-adjusted
and PH runs use $1440$; the latter also use the updated runtime
and a node capacity of $200$ rather than $187$.
On connected RSBM, adjusted HAC uses context $1664$ and PH uses
$1600$.
We therefore interpret the results as evidence for hierarchy
alignment, without attributing every gain solely to the partition.

\input{tables/appendix_block_alignment}
\input{tables/appendix_block_vun}
\FloatBarrier

\clearpage
\subsection{SBM}\label{app:sbm_distribution}

On SBM, default HAC-R trades uniqueness and novelty for higher validity, whereas
PH-R retains both while improving V.U.N. over SENT. Default HAC-FR instead loses validity. Table~\ref{tab:appendix_sbm_full} includes PH
as a privileged control using the generator's memberships. Novel-only results are available
for SENT and the two HAC variants, as shown in Table~\ref{tab:appendix_sbm_novel}.
Canonical SBM remains mixed:
both hierarchical variants improve Degree MMD, Orbit MMD and Ratio but worsen Clustering and
Spectral MMD, while \emph{HDT}-HAC-R has selected uniqueness $0.598\pmstd{0.015}$
(Table~\ref{tab:appendix_sbm_novel}).

\begin{table}[!htbp]
\input{tables/appendix_sbm_full}
\par\medskip
\input{tables/appendix_sbm_novel}
\end{table}
\FloatBarrier

\clearpage
\subsection{SSBM}\label{app:ssbm_full}

Inferred coarsening succeeds when blocks are small. Within-community randomization leads;
 hierarchy-wide randomization is the weaker policy here (Table~\ref{tab:appendix_ssbm_full}).
On SSBM, the \emph{HDT}-HAC-R improvements in raw
yield, Degree MMD, Orbit MMD and Ratio persist, whereas hierarchy-wide randomization remains
worse than SENT on most measures (Table~\ref{tab:appendix_ssbm_novel}).

\begin{table}[!htbp]
\input{tables/appendix_ssbm_full}
\par\medskip
\input{tables/appendix_ssbm_novel}
\end{table}
\FloatBarrier

\clearpage
\subsection{Connected RSBM ($R=4$)}\label{app:rsbm_full}

Connected RSBM extends the comparison to two planted levels. Adjusted HAC-R nearly
matches PH-R on V.U.N.$^\dagger$ ($0.854$ versus $0.858$), whereas adjusted HAC-FR
remains far below PH-FR ($0.083$ versus $0.744$; Table~\ref{tab:appendix_rsbm_full}).
This is a comparison of generation scores, not equivalent partition recovery.
PH is a privileged training-structure control, and the daggered scores use the original
validity gate, as described in Appendix~\ref{app:block_datasets}.
RSBM has no novel-only panel in the completed inventory, so the standard evaluation is the only one reported for it.

\input{tables/appendix_rsbm_full}
\FloatBarrier

\section{ Additional experiments}
\subsection{Scaling and baseline controls}\label{app:efficiency}

This appendix reports training-regime scaling and a reproduction of the flat baseline. Appendix~\ref{app:regime_delta} compares each serialization under a larger training regime. Appendix~\ref{app:resource_autograph} compares our SENT reproduction with the published AutoGraph results. Training times and epoch counts for the main comparisons appear in Tables~\ref{tab:main_molecular_all} and~\ref{tab:appendix_train_inventory_mol}.

\subsubsection{Scaling across training regimes}\label{app:regime_delta}

Tables~\ref{tab:appendix_regime_guacamol} and~\ref{tab:appendix_regime_moses} compare SENT,
\emph{HDT}-HAC-R and \emph{HDT}-HAC-FR across the GPT2-XS and GPT2-S regimes on GuacaMol and
MOSES, where a regime is the backbone together with its data volume and schedule. The GPT2-XS
columns repeat the main-matrix runs (Tables~\ref{tab:appendix_guacamol_full},
\ref{tab:appendix_moses_full} and~\ref{tab:appendix_train_inventory_mol}). The final columns
re-express the same numbers as a within-serialization comparison, giving for each metric the
change from the smaller to the larger regime, signed so that a positive value always means the
larger regime is better. The regime columns establish which serialization is better at each
scale; the change columns establish which benefits more from scaling.
Figure~\ref{fig:regime_scaling} plots the V.U.N., FCD and \texttt{MoleculePGD} rows. Backbone
size, data volume and schedule all change together between regimes, so this is an efficiency
observation rather than a controlled estimate of the effect of any one factor.

The GPT2-S runs use $1{,}201{,}664$ GuacaMol and $1{,}584{,}640$ MOSES training molecules,
against $499{,}968$ for GPT2-XS; within each regime, all variants share the training population
and held-out reference pool. GuacaMol uses effective batch $256$, with $800{,}000$ SENT updates
and $438{,}807$ updates for each HAC variant; historical SENT retains context $256$ and
attention dropout $0$, while both HAC variants use context $512$ and dropout $0.5$. MOSES uses
effective batch $512$, $400{,}000$ SENT updates and $203{,}000$ updates per HAC variant, all
with context $512$ and attention dropout $0.5$. AdamW uses learning rate $6\times10^{-4}$,
weight decay $0.1$, momentum coefficients $0.9$ and $0.95$, $1\%$ warmup, and cosine decay over
$90\%$ of the update budget to $0.1$ of the initial learning rate. Training times are stated on
a one-GPU basis; historical GuacaMol times sum logged active segments, and other GPT2-S times
span dataset setup through final checkpoint save, excluding evaluation.

Each GPT2-S standard evaluation contains $10{,}000$ attempts per seed, with top-$k=5$,
temperature $1.0$ and generation batch $128$, from one seed-$42$ trained checkpoint per variant.
\texttt{MoleculePGD} aggregates the three seed-specific official interval means, each using
$10{,}000$ valid generated molecules and $10{,}000$ held-out references, with $5{,}000$ per
distribution per draw and $10$ draws; supplementary generation supplies only this PGD pool, and
the other metrics retain the original standard generations. The motif discrepancies use
$1{,}000$ valid molecules per arm (Appendix~\ref{app:metrics}). Changes are computed from the
printed means, so those in the four-decimal motif rows are coarse.

\input{tables/appendix_regime_guacamol}
\input{tables/appendix_regime_moses}
\FloatBarrier

\subsubsection{Reproduction of the flat baseline}\label{app:resource_autograph}

We check our SENT implementation using the full-data AutoGraph training protocol. Table~\ref{tab:appendix_resource_autograph} compares the reproduction with the published results where available. Validity is higher by $0.071$, novelty is higher by $0.007$, and FCD is lower by $0.243$. Uniqueness is identical at $1.000$. The GuacaMol KL score is the only reported metric on which the reproduction performs worse, by $0.012$.

These results support the quality of our SENT implementation. This check uses the full AutoGraph protocol rather than the controlled settings in Table~\ref{tab:appendix_arch}. Its absolute scores should therefore not be directly compared with those from the main controlled experiments.

\input{tables/appendix_resource_autograph}

\subsection{AutoGraph-comparable metrics}\label{app:autograph_metric_methodology}

Table~\ref{tab:autograph_comparison} includes a small number of metrics and reporting conventions that are not direct outputs of our main evaluation pipeline. For comparability with~\citet{chen2025autograph} we report the GuacaMol \emph{KL score}~\citep{brown2019guacamol}, which measures how closely the distribution of generated molecules matches the reference set in standard molecular descriptors. For MOSES, \emph{Filters}~\citep{polykovskiy2020moses} is the fraction of valid generations passing the benchmark's MCF/PAINS, allowed-atom, and charge checks; we compute it with the official MOSES implementation. One entry does not harmonize and is marked rather than forced. The Proteins \emph{Ratio} we report is the mean-of-ratios form defined in Appendix~\ref{app:metrics}, whereas the transcribed rows use AutoGraph's train-normalized geometric mean:
\[
\displaystyle
\mathrm{Ratio}=\frac{\mathrm{geomean}(\mathrm{Deg,Clust,Orbit,Spec})_{\mathrm{model}}}
{\mathrm{geomean}(\mathrm{Deg,Clust,Orbit,Spec})_{\mathrm{train}}}.
\]
This convention places the training set at $1.00$ by construction; the two Ratio columns are therefore on different scales and we bold no Ratio cell. We do place all FCD results in Table~\ref{tab:autograph_comparison} on the same raw-FCD scale. AutoGraph's published GuacaMol FCD scores are converted via $\mathrm{FCD}_{\mathrm{raw}} = -5\ln(\mathrm{score}/100)$, while  our GuacaMol FCD values are reported directly in the raw form produced by our pipeline. This harmonization is only for scale matching in the comparison table and does not otherwise change the underlying evaluations.

\clearpage
\section{Motif-conditioned molecular generation}
\label{app:conditional_generation}
\input{sections/conditional_motif_protocol}

\clearpage

\section{Unconditional generation galleries}\label{app:galleries}

\begin{figure}[!htbp]
\centering
\includegraphics[width=\linewidth]{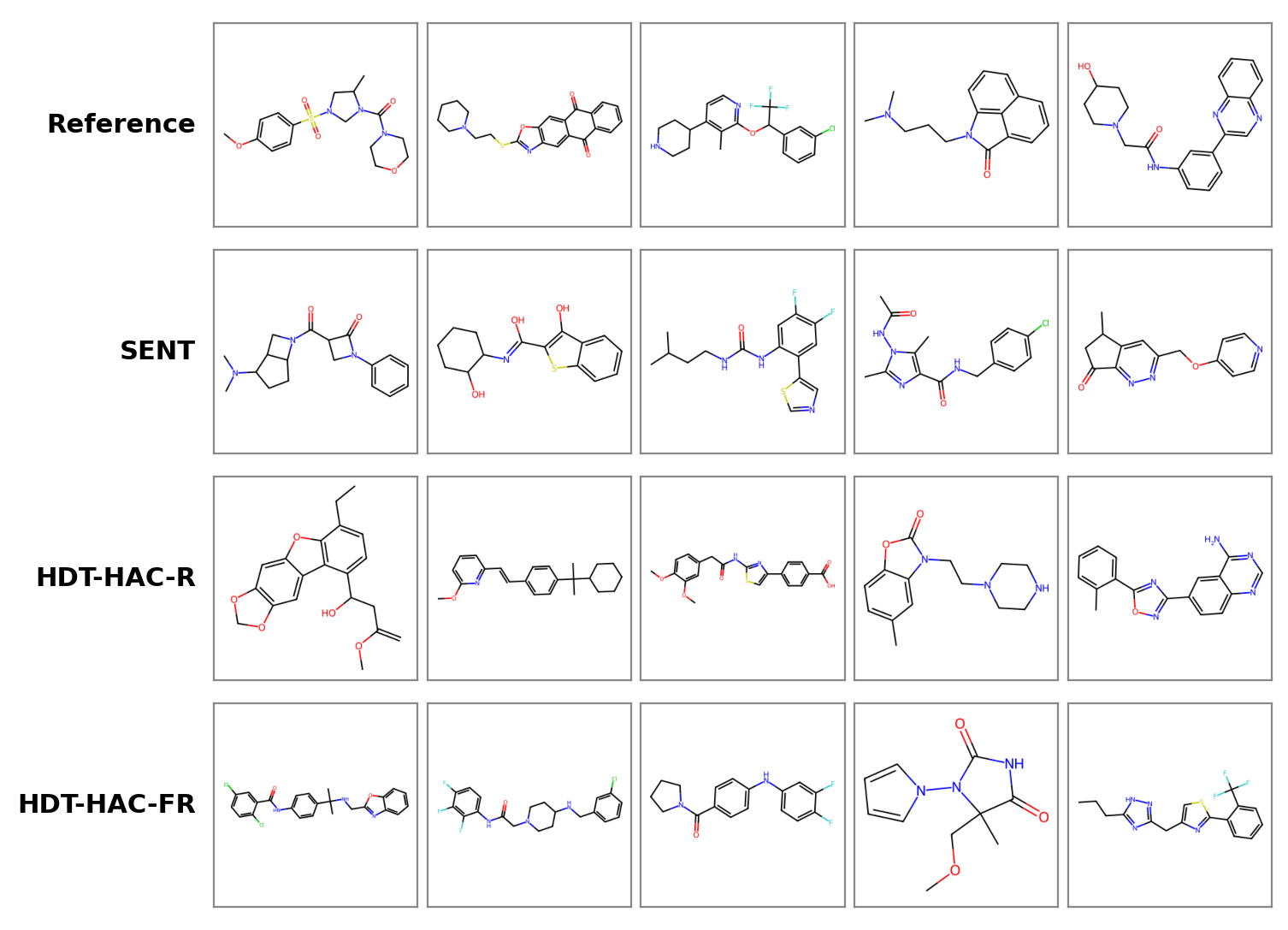}
\caption{\textbf{GuacaMol unconditional generation.} Held-out reference molecules and generations from SENT, HDT-HAC-R, and HDT-HAC-FR.}
\label{fig:gal_guacamol}
\end{figure}

Figures~\ref{fig:gal_guacamol}--\ref{fig:gal_rsbm_connected} qualitatively complement the full
unconditional-generation results in Appendix~\ref{app:unconditional_full}. Each row contains
five deterministic samples drawn without replacement. Generated rows use the seed-$42$
evaluation arm of the corresponding model, while the reference row uses the benchmark's
held-out evaluation arm. The molecular evaluation artifacts contain valid decoded molecules;
their galleries therefore illustrate chemical composition rather than the frequency of valid
decoding, which is reported quantitatively in the benchmark tables.

The block-model galleries distinguish original SBM, SSBM, original RSBM,
and connected RSBM. Generated examples are selected before structural assessment, without
filtering by validity, uniqueness, or novelty. For a common visual comparison, node colors
show a topology-derived HAC partition computed with the default threshold $\tau=4$ in every
row; these colors do not represent planted memberships or the hierarchy emitted by the model.
Columns contain independent examples, and generated connectivity is not an additional
acceptance requirement. Reference borders are neutral. Connected-RSBM borders use compatibility with the original
RSBM gate, as in Table~\ref{tab:appendix_block_vun}, rather than a separately calibrated
connected-population validity test.

\FloatBarrier

\clearpage
\begin{figure}[!htbp]
\centering
\includegraphics[width=\linewidth]{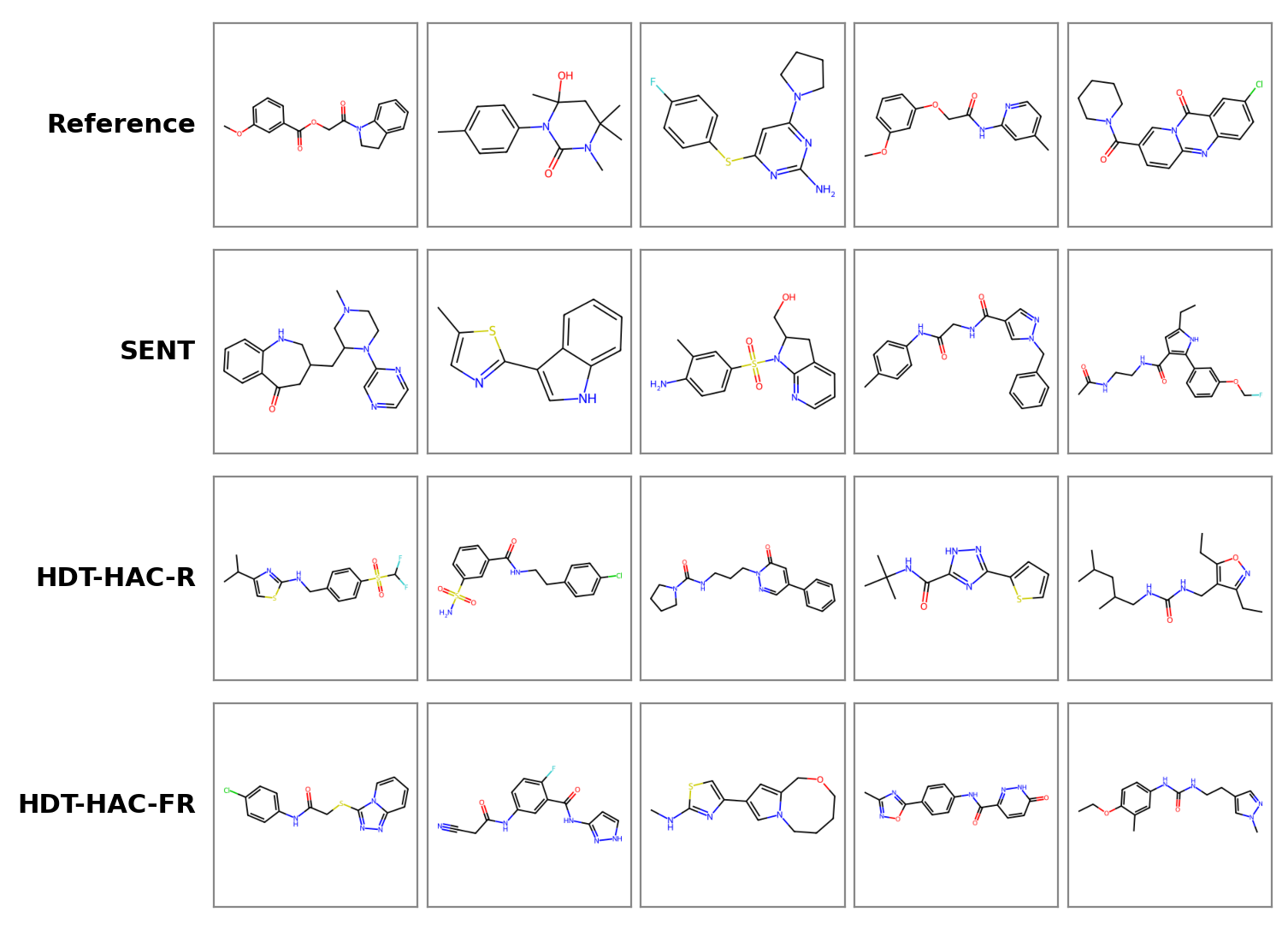}
\caption{\textbf{MOSES unconditional generation.} Held-out reference molecules and generations from SENT, HDT-HAC-R, and HDT-HAC-FR.}
\label{fig:gal_moses}
\par\medskip
\includegraphics[width=\linewidth]{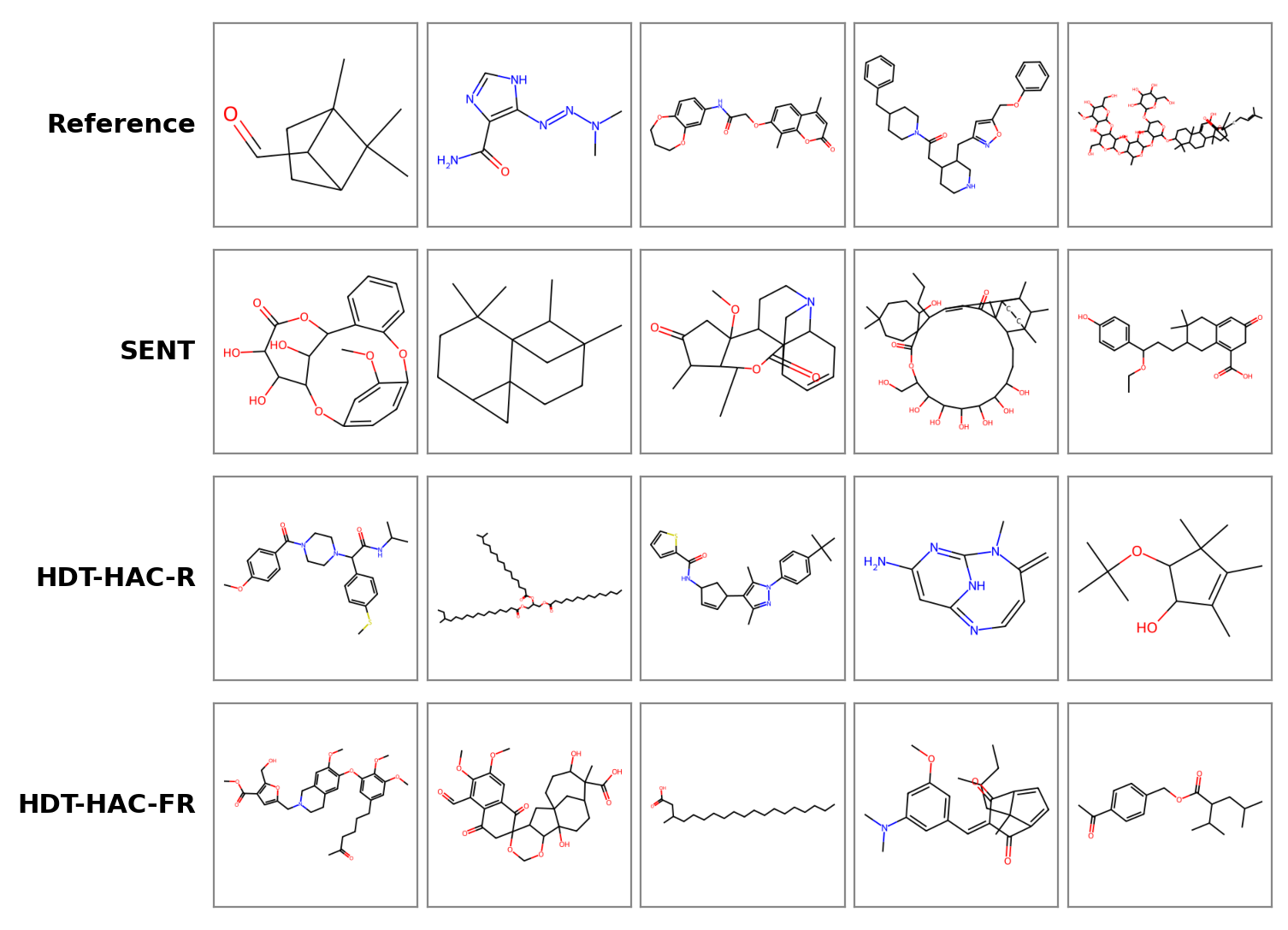}
\caption{\textbf{COCONUT 2.0 unconditional generation.} Held-out reference molecules and generations from SENT, HDT-HAC-R, and HDT-HAC-FR.}
\label{fig:gal_coconut}
\end{figure}
\FloatBarrier

\clearpage
\begin{figure}[!htbp]
\centering
\includegraphics[width=\linewidth]{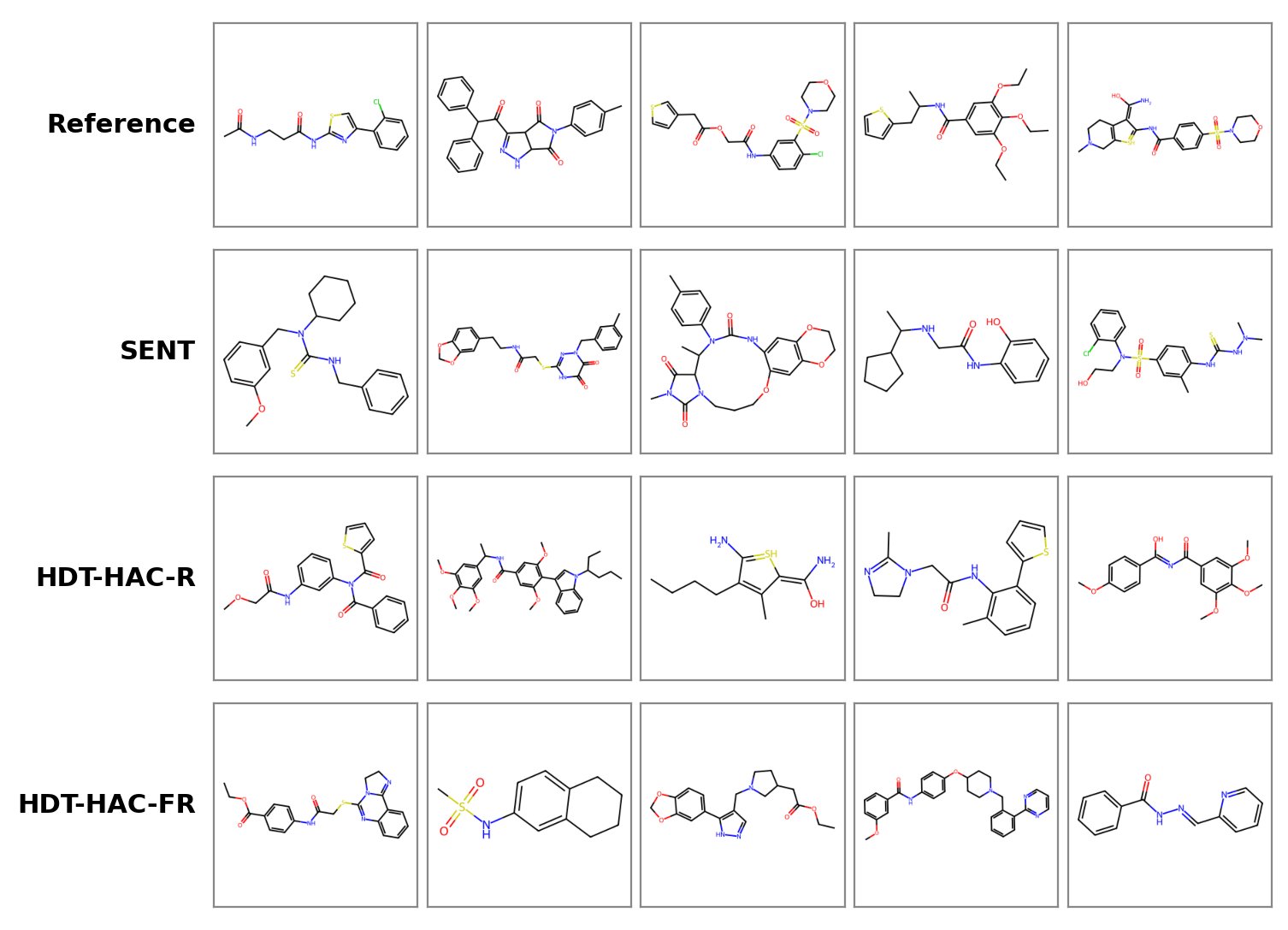}
\caption{\textbf{GEOM-Drugs (disjoint) unconditional generation.} Held-out reference molecules and generations from SENT, HDT-HAC-R, and HDT-HAC-FR.}
\label{fig:gal_geom_drugs}
\par\medskip
\includegraphics[width=\linewidth]{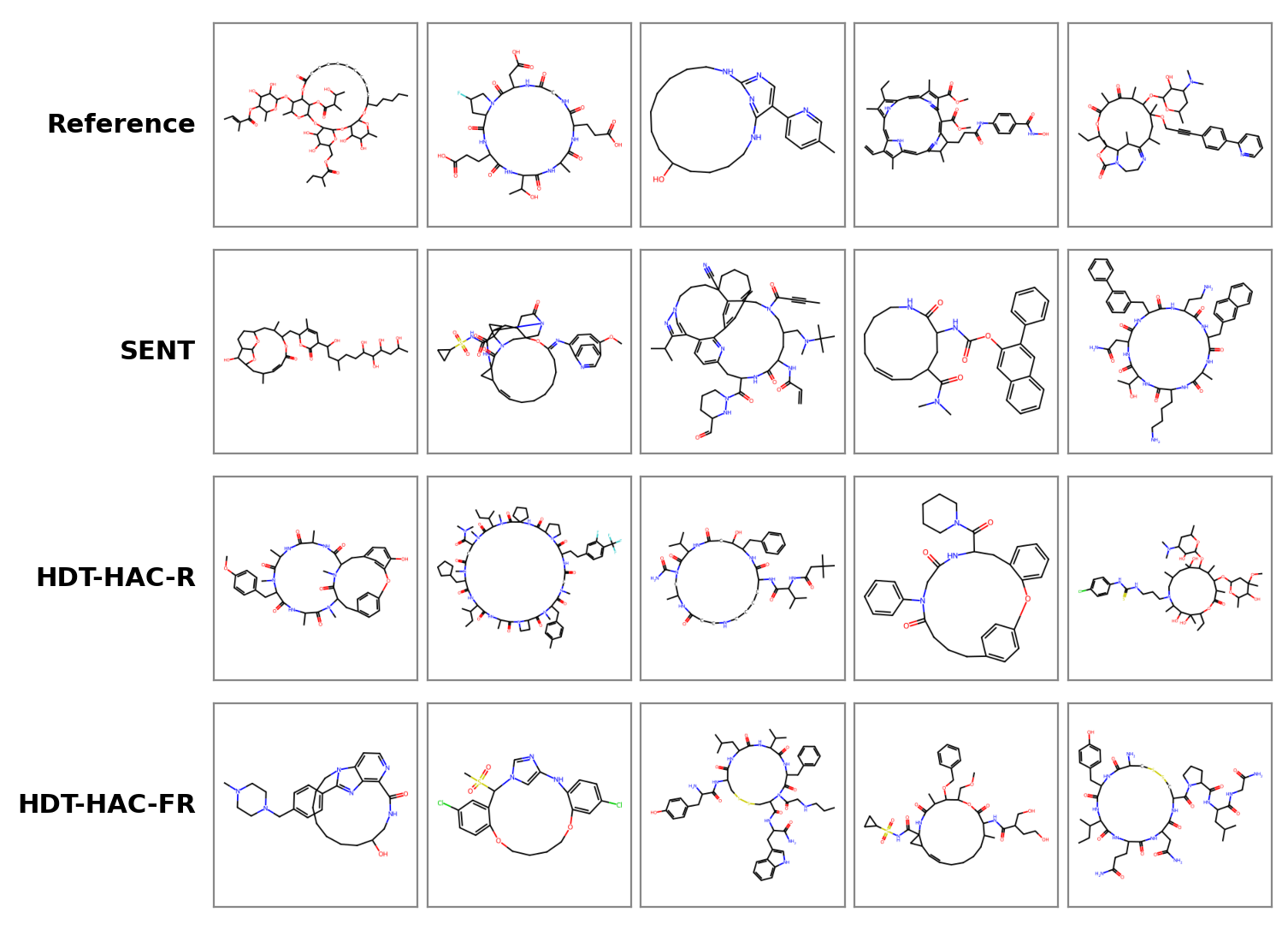}
\caption{\textbf{Macrocycle unconditional generation.} Held-out reference molecules and generations from SENT, HDT-HAC-R, and HDT-HAC-FR.}
\label{fig:gal_macrocycle}
\end{figure}
\FloatBarrier

\clearpage
\begin{figure}[!htbp]
\centering
\includegraphics[width=\linewidth]{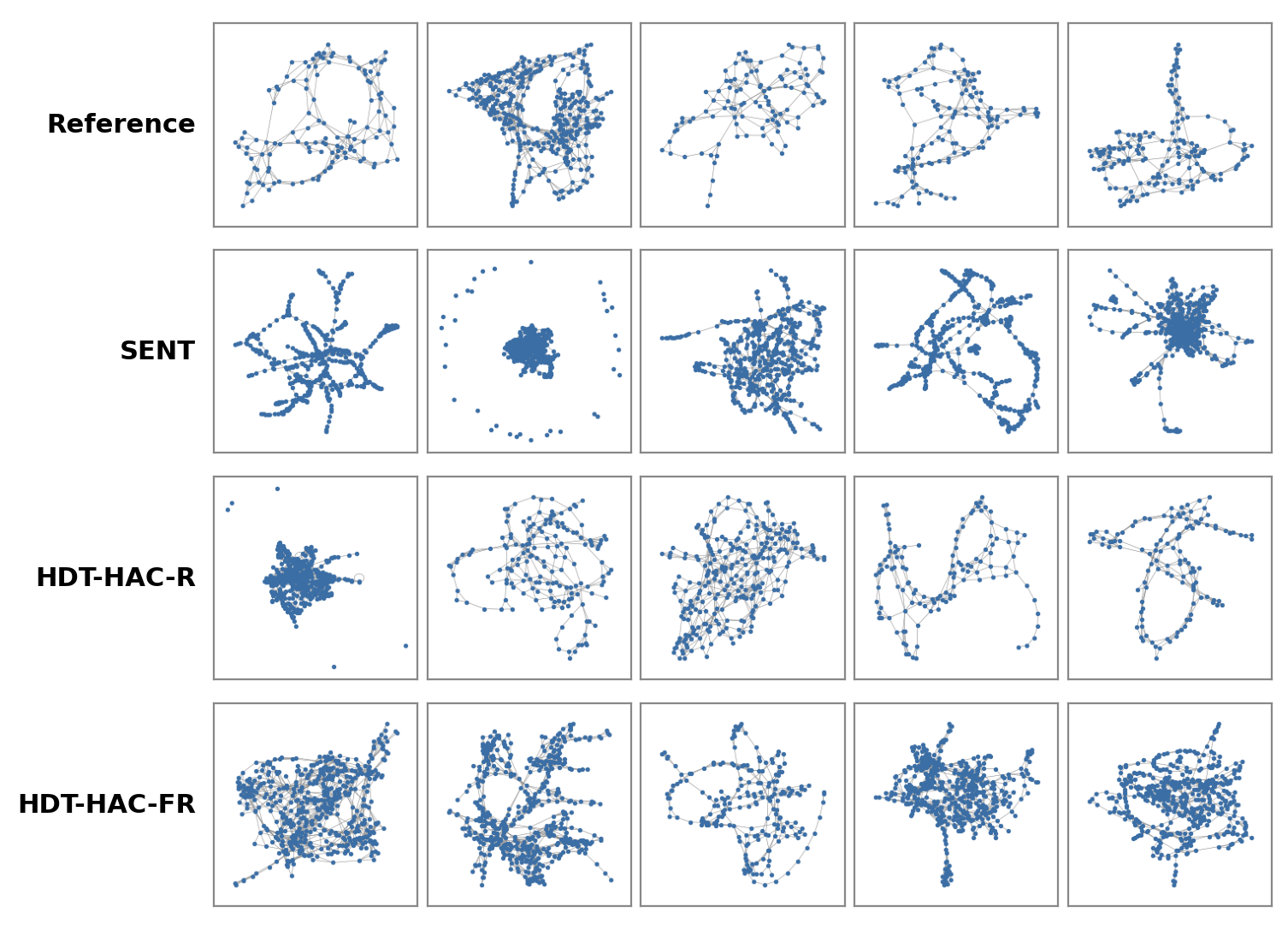}
\caption{\textbf{Proteins unconditional generation.} Held-out reference graphs and generations from SENT, HDT-HAC-R, and HDT-HAC-FR.}
\label{fig:gal_proteins}
\par\medskip
\includegraphics[width=\linewidth]{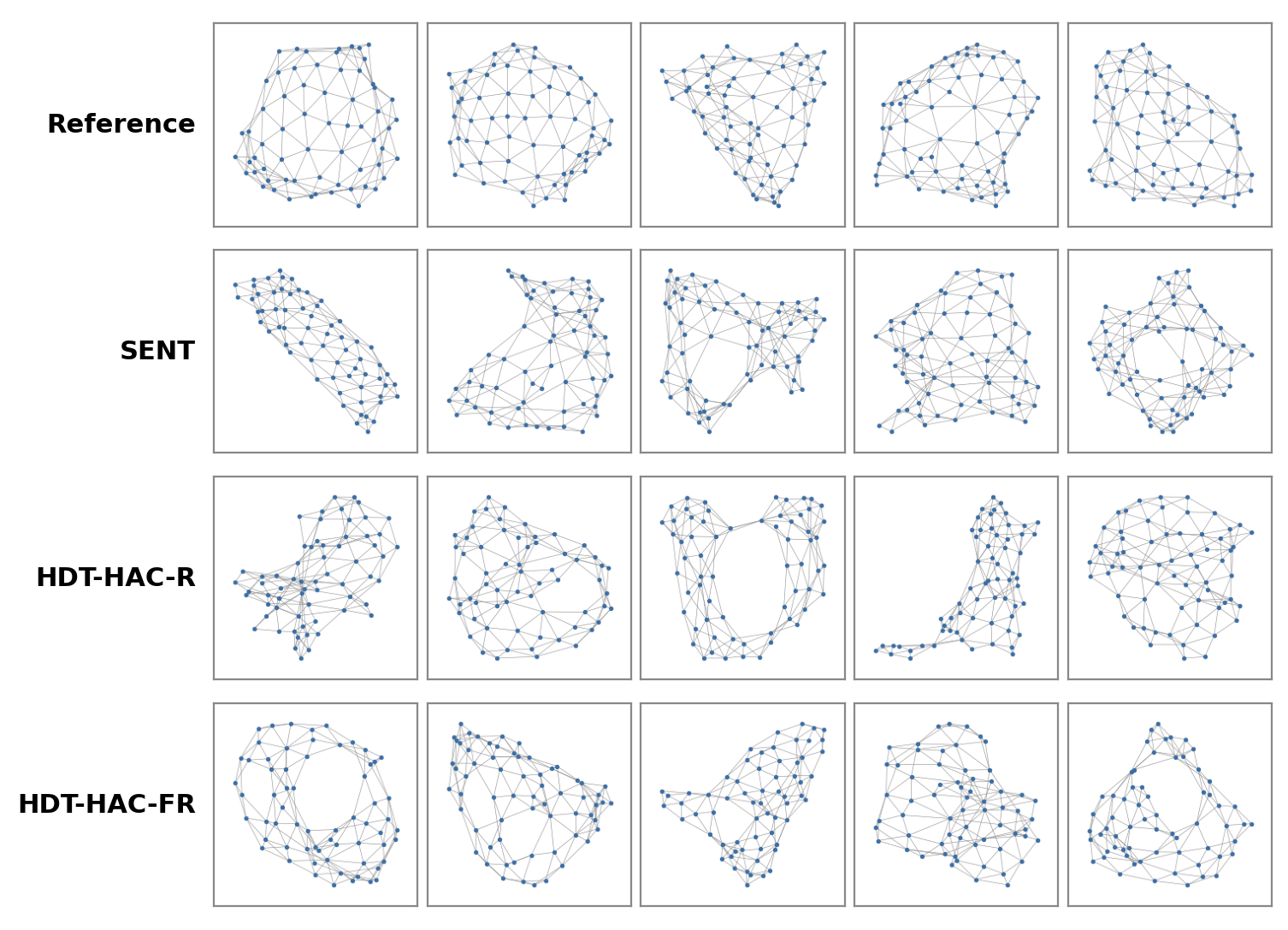}
\caption{\textbf{Planar unconditional generation.} Held-out reference graphs and generations from SENT, HDT-HAC-R, and HDT-HAC-FR.}
\label{fig:gal_planar}
\end{figure}
\FloatBarrier

\clearpage

\begin{figure}[!htbp]
\centering
\includegraphics[width=\linewidth]{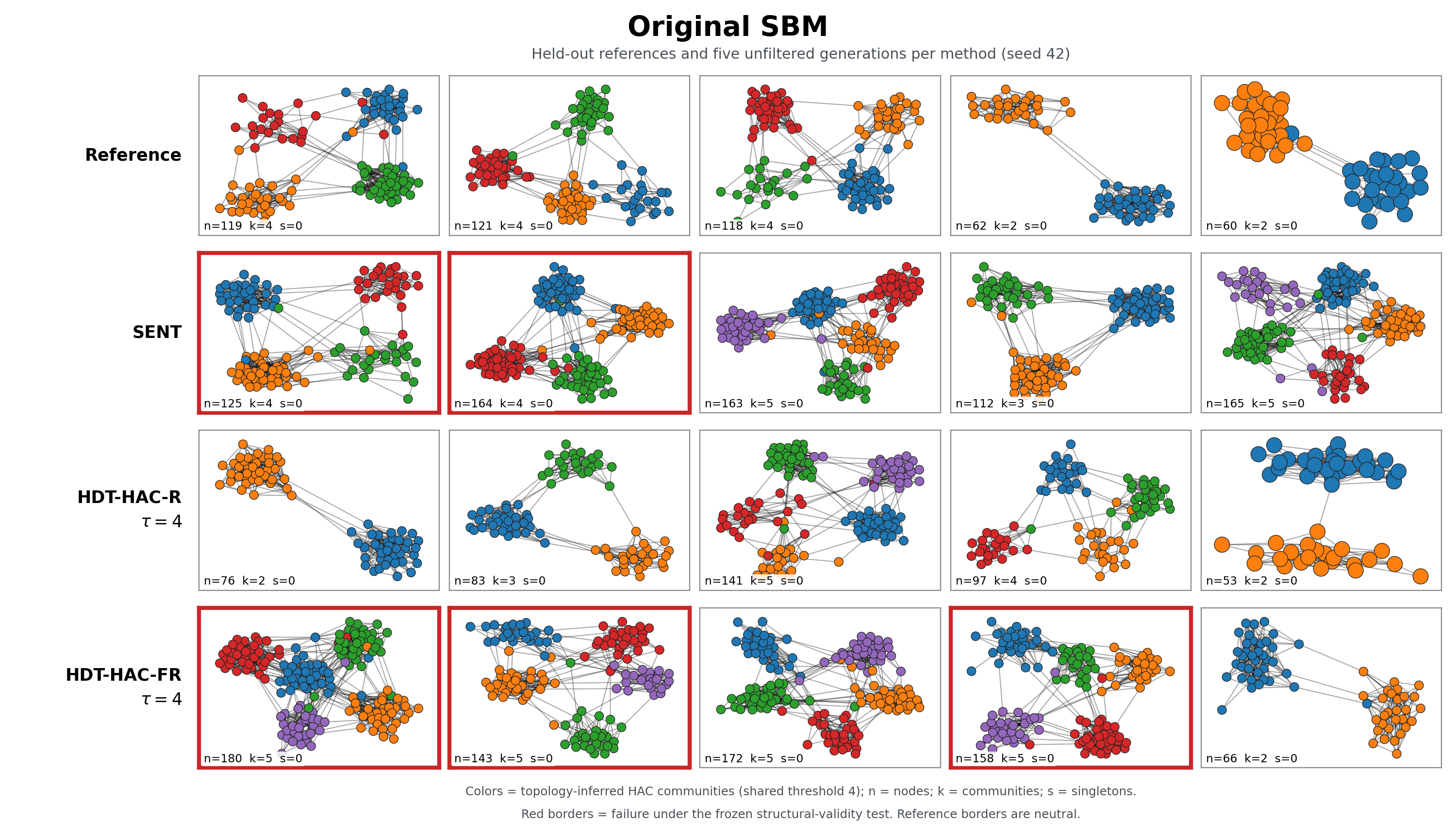}
\caption{\textbf{Original SBM: default coarsening.} Rows show held-out references, SENT, HDT-HAC-R, and HDT-HAC-FR, with five examples per row. The hierarchical models use the default stopping threshold $\tau=4$. Node colors show inferred HAC communities; red borders mark generations that fail the frozen SBM community-structure check.}
\label{fig:gal_sbm}
\par\medskip
\includegraphics[width=\linewidth,height=0.78\textheight,keepaspectratio]{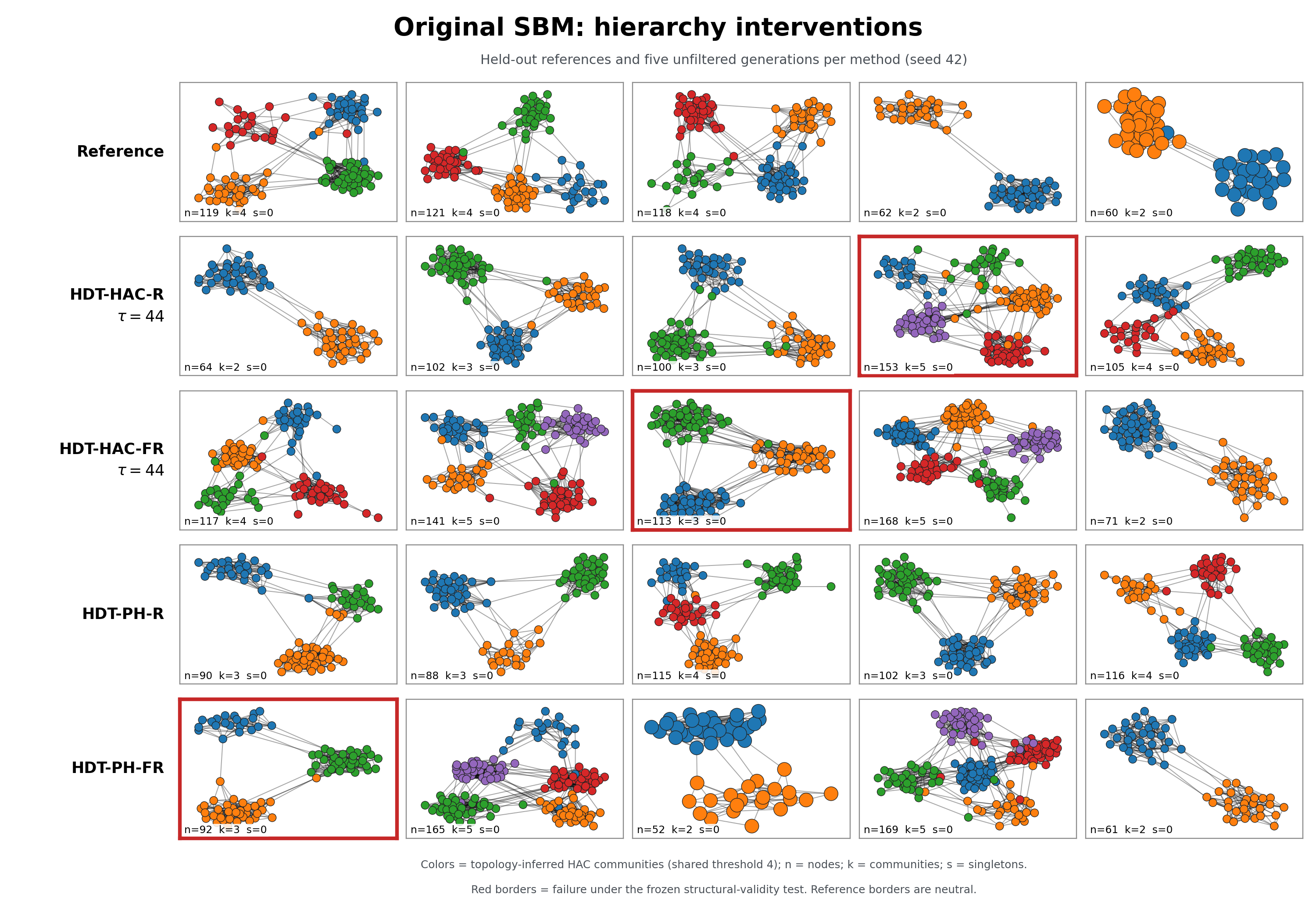}
\caption{\textbf{Original SBM: coarsening interventions.} The reference row matches Figure~\ref{fig:gal_sbm}; generated rows show HDT-HAC-R and HDT-HAC-FR trained with stopping threshold $\tau=44$, followed by HDT-PH-R and HDT-PH-FR trained with the generator's planted memberships. Display colors remain the shared inferred HAC partition, not the training hierarchy. Red borders indicate failure of the same frozen SBM check.}
\label{fig:gal_sbm_interventions}
\end{figure}
\FloatBarrier

\clearpage

\begin{figure}[!htbp]
\centering
\includegraphics[width=\linewidth]{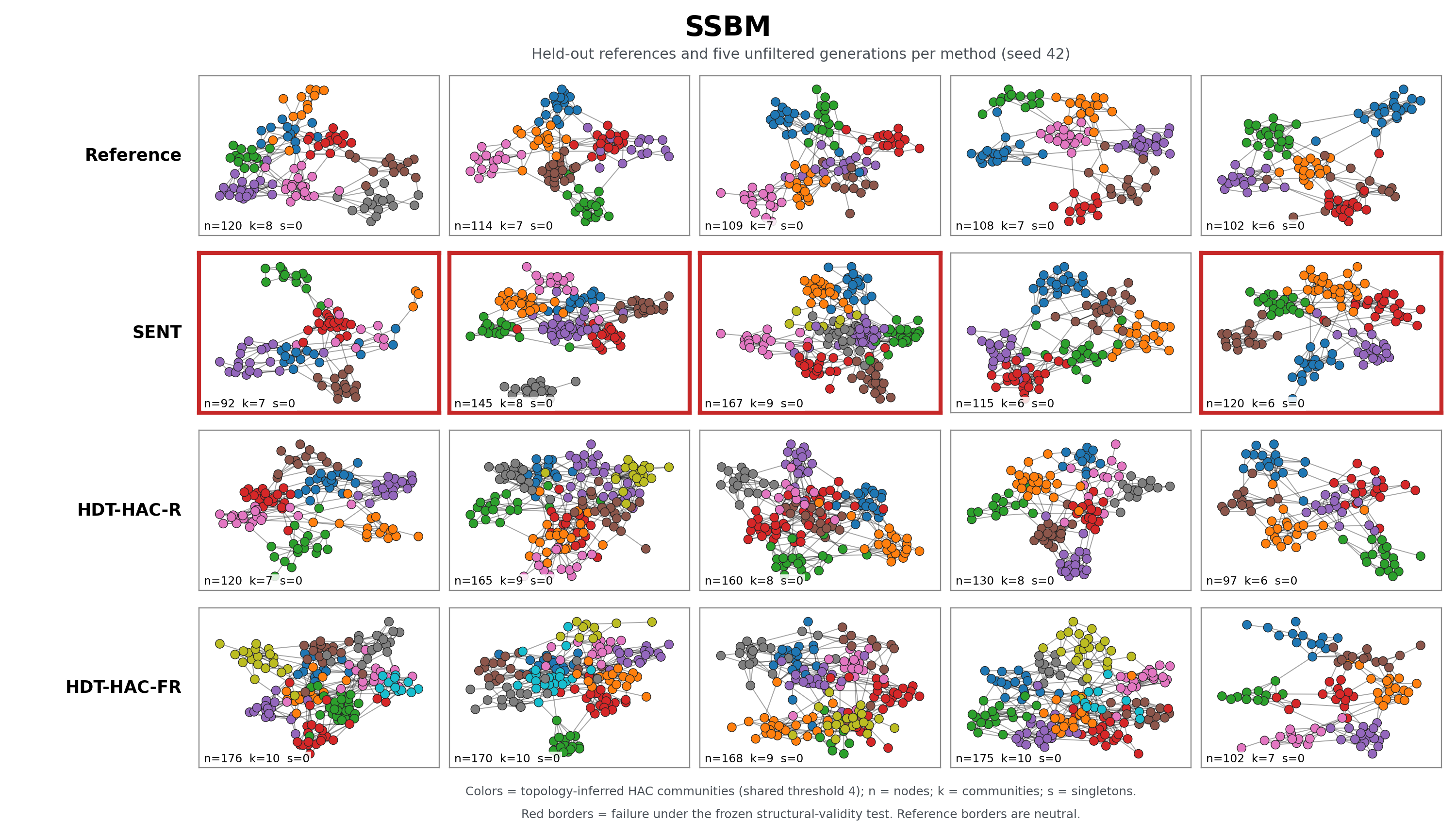}
\caption{\textbf{Sparse SBM (SSBM) generation gallery.} Rows show held-out references, SENT, HDT-HAC-R, and HDT-HAC-FR, with five examples per row. Node colors show inferred HAC communities; red borders mark generations that fail the frozen SSBM community-structure check.}
\label{fig:gal_ssbm}
\par\medskip
\includegraphics[width=\linewidth,height=0.78\textheight,keepaspectratio]{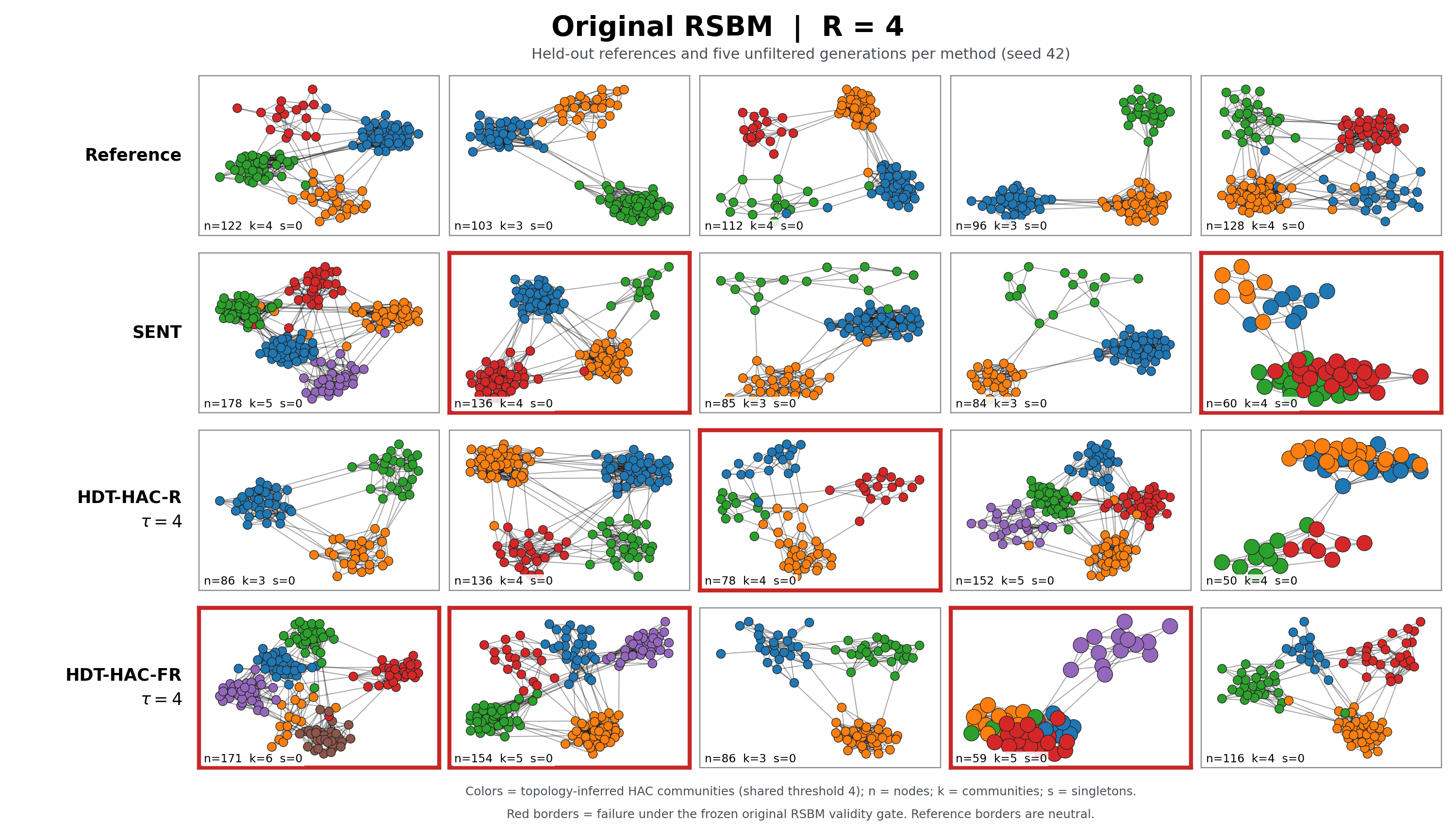}
\caption{\textbf{Original RSBM ($R=4$): default coarsening.} Rows show references from the original population, which retains disconnected graphs, followed by SENT, HDT-HAC-R, and HDT-HAC-FR. The hierarchical models use $\tau=4$. Node colors show inferred HAC communities; red borders mark generations that fail the frozen original-RSBM validity gate.}
\label{fig:gal_rsbm_original}
\end{figure}
\FloatBarrier

\clearpage

\begin{figure}[htbp]
\centering
\includegraphics[width=\linewidth,height=0.78\textheight,keepaspectratio]{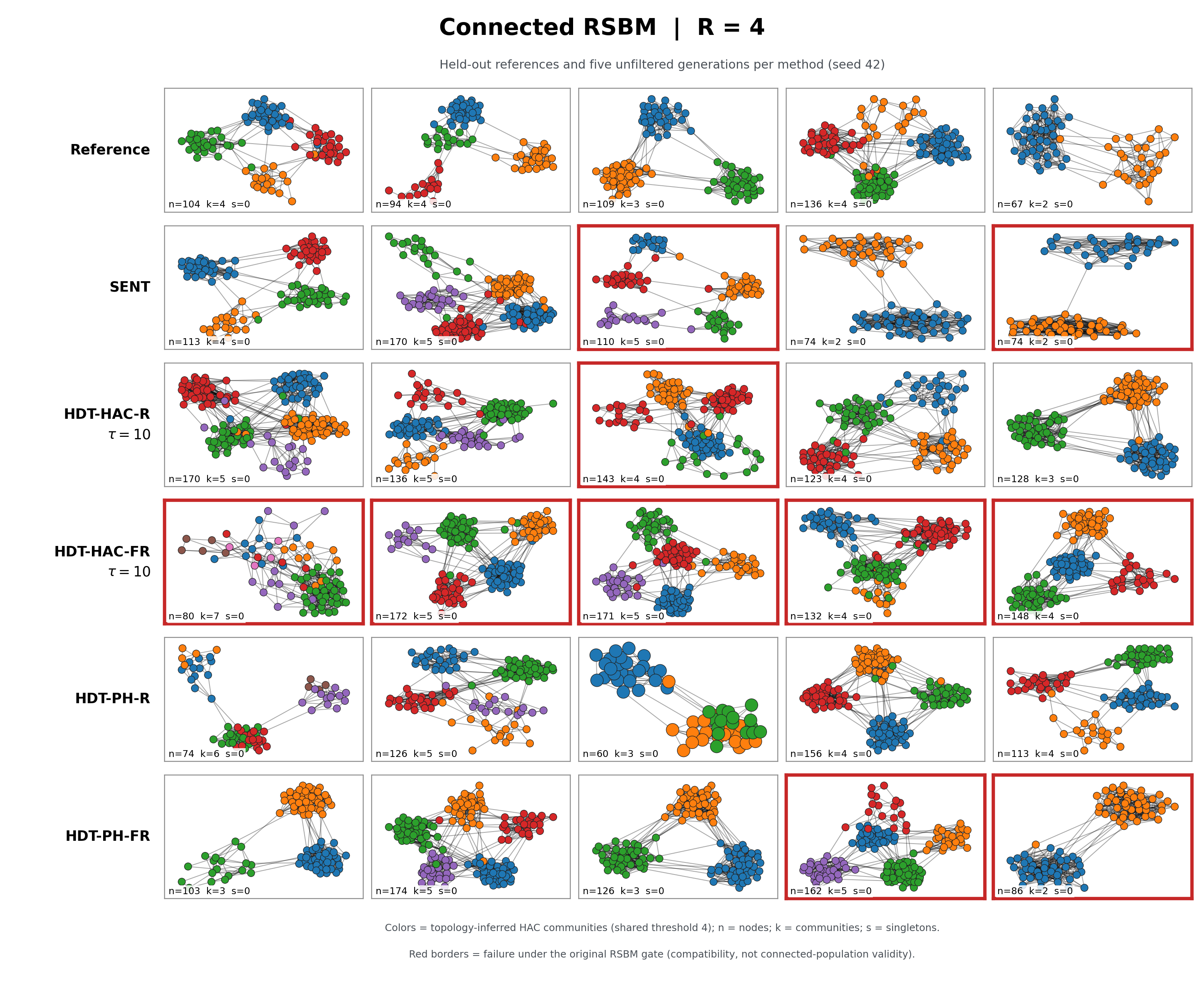}
\caption{\textbf{Connected RSBM ($R=4$): coarsening interventions.} Rows show connected-population references, SENT, HDT-HAC-R and HDT-HAC-FR with $\tau=10$, and HDT-PH-R and HDT-PH-FR trained with planted coarse and fine memberships. Node colors show the shared inferred HAC partition. Red borders indicate failure of compatibility with the original RSBM gate ($V^\dagger$ in Table~\ref{tab:appendix_block_vun}). This population is separate from Figure~\ref{fig:gal_rsbm_original}.}
\label{fig:gal_rsbm_connected}
\end{figure}

\clearpage
\input{checklist.tex}

\end{document}

%% file: tables/tokenizer_characteristics.tex
\begin{table}[!htbp]
\caption{\textbf{Tokenizer characteristics.} MC+FG means Motif plus Functional Group; $R/F/S$ denote ring, functional-group and singleton type labels. The design axis is how explicitly hierarchy appears
in the token stream. $n$, $m$, $c$, $k$ denote number of nodes, edges, cross-community edges, and top-level
communities. For HSENT, the sum counts catalog-emitting community blocks. Length ratios are means over the molecular benchmarks
(Appendix~\ref{app:token_length}); Section~\ref{sec:results:design} reports what each degree of
explicitness is worth.}
\label{tab:tokenizer_characteristics}
\centering
\small
\begin{adjustbox}{max width=\textwidth}
\begin{tabular}{@{}l l l l c c@{}}
\toprule
\textbf{Tokenizer} & \textbf{Hierarchy} & \textbf{Coarsening} & \textbf{Cross-community edges} & \textbf{Token cost} & \textbf{Length vs SENT} \\
\midrule
SENT & flat (single walk) & not used & N/A (no communities) & $O(n+m)$ & $1\times$ \\
\emph{HSENT} & declarative (catalog + wiring) & any & explicit bipartite block $(i,j)$ & $O(m+\sum_C |C|)$ & $\sim 3\times$ \\
\emph{HDT} & traversal-based (DFS of $\mathcal{H}$) & any & DFS back edges (implicit) & $O(n+m)$ & $\sim 2\times$ \\
\emph{HDTC} & traversal-based, typed blocks ($R/F/S$) & MC+FG only & explicit super-graph tuple & $O(n+m+c)$ & $\sim 2\times$ \\
\bottomrule
\end{tabular}
\end{adjustbox}
\end{table}

%% file: tables/randomization_settings.tex
\begin{table}[!htbp]
\caption{\textbf{Order-randomization policies.}
Traversal choices are resampled at each training presentation.
Both policies preserve the graph and its partition hierarchy and assign
node identifiers in first-visit order.
FR additionally changes community order, entry constraints, and community
identifier assignment; its comparison with R therefore evaluates the
complete traversal policy.}
\label{tab:randomization_settings}
\centering
\small
\setlength{\tabcolsep}{4pt}
\begin{adjustbox}{max width=\linewidth}
\begin{tabular}{@{}l p{0.35\linewidth} p{0.46\linewidth}@{}}
\toprule
\textbf{Choice} & \textbf{Within-community (\emph{-R})}
               & \textbf{Hierarchy-wide (\emph{-FR})} \\
\midrule
Community order
& Stored order
& Randomized DFS over sibling-community adjacency graphs \\

Leaf traversal
& Random-root BFS with randomized neighbor order
& Randomized BFS with entry constrained by the traversal parent,
  when one exists \\

Community identifiers
& Stored identifiers
& Encounter order within each sibling set \\

Hierarchy levels
& Randomization within terminal communities
& Community traversal at every available level,
  plus traversal within terminal communities \\
\bottomrule
\end{tabular}
\end{adjustbox}
\end{table}

%% file: tables/main_molecular_all.tex
\begin{table}[t]
\caption{\textbf{Unconditional generation across the five molecular benchmarks.} Comparison of flat
baseline to our two hierarchical configurations fixed in advance: \emph{HDT}-HAC-R (\textbf{HAC-R}),
which randomizes atom order within each community, and \emph{HDT}-HAC-FR (\textbf{HAC-FR}), which randomizes traversal across the hierarchy. The main GPT2-XS protocol uses training-time-calibrated update budgets rather than matched epochs. The achieved runtimes vary and both runtimes and epoch counts are given. Values are means over three generation seeds;
\textbf{bold} = best within a dataset. Standard deviations, 
additional variants,  training-sample reference and novel-only protocol are detailed in 
Appendix~\ref{app:unconditional_full}.}
\label{tab:main_molecular_all}
\centering
\scriptsize
\setlength{\tabcolsep}{2.2pt}
\begin{adjustbox}{max width=\textwidth}
\begin{tabular}{@{}l ccc ccc ccc ccc ccc @{}}
\toprule
 & \multicolumn{3}{c}{GuacaMol} & \multicolumn{3}{c}{MOSES} & \multicolumn{3}{c}{COCONUT 2.0} & \multicolumn{3}{c}{GEOM-Drugs} & \multicolumn{3}{c}{Macrocycle} \\
\cmidrule(lr){2-4} \cmidrule(lr){5-7} \cmidrule(lr){8-10} \cmidrule(lr){11-13} \cmidrule(lr){14-16}
 & SENT & HAC-R & HAC-FR & SENT & HAC-R & HAC-FR & SENT & HAC-R & HAC-FR & SENT & HAC-R & HAC-FR & SENT & HAC-R & HAC-FR \\
\midrule
Time (h) & 4.98 & 6.70 & 5.92 & 3.35 & 4.50 & 1.36 & 4.30 & 4.25 & 4.42 & 0.98 & 1.00 & 1.05 & 1.80 & 1.85 & 1.94 \\
Epochs & 83.2 & 51.2 & 51.2 & 51.2 & 32.0 & 32.0 & 82.3 & 41.8 & 41.8 & 187.8 & 102.0 & 102.0 & 439.3 & 233.6 & 233.6 \\
\midrule
Val $\uparrow$ & $\boldsymbol{0.967}$ & $0.943$ & $0.966$ & $0.977$ & $0.978$ & $\boldsymbol{0.985}$ & $\boldsymbol{0.932}$ & $0.860$ & $0.908$ & $0.949$ & $0.913$ & $\boldsymbol{0.957}$ & $\boldsymbol{0.855}$ & $0.819$ & $0.790$ \\
Uniq $\uparrow$ & $\boldsymbol{1.000}$ & $\boldsymbol{1.000}$ & $\boldsymbol{1.000}$ & $\boldsymbol{1.000}$ & $0.999$ & $\boldsymbol{1.000}$ & $\boldsymbol{1.000}$ & $0.997$ & $0.999$ & $\boldsymbol{1.000}$ & $0.995$ & $0.999$ & $\boldsymbol{1.000}$ & $0.970$ & $0.991$ \\
Nov $\uparrow$ & $\boldsymbol{0.998}$ & $0.987$ & $0.996$ & $\boldsymbol{0.990}$ & $0.950$ & $0.971$ & $\boldsymbol{0.979}$ & $0.880$ & $0.924$ & $\boldsymbol{0.997}$ & $0.925$ & $0.988$ & $\boldsymbol{0.999}$ & $0.658$ & $0.967$ \\
V.U.N. $\uparrow$ & $\boldsymbol{0.965}$ & $0.931$ & $0.961$ & $\boldsymbol{0.967}$ & $0.929$ & $0.956$ & $\boldsymbol{0.912}$ & $0.757$ & $0.839$ & $\boldsymbol{0.945}$ & $0.842$ & $\boldsymbol{0.945}$ & $\boldsymbol{0.854}$ & $0.537$ & $0.762$ \\
FCD $\downarrow$ & $1.799$ & $\boldsymbol{0.871}$ & $0.989$ & $0.696$ & $\boldsymbol{0.246}$ & $0.260$ & $1.173$ & $\boldsymbol{0.551}$ & $0.563$ & $2.742$ & $\boldsymbol{1.196}$ & $1.353$ & $2.711$ & $\boldsymbol{1.064}$ & $1.501$ \\
PGD $\downarrow$ & $0.309$ & $\boldsymbol{0.176}$ & $0.196$ & $0.217$ & $\boldsymbol{0.130}$ & $0.137$ & $0.386$ & $\boldsymbol{0.202}$ & $0.212$ & $0.456$ & $\boldsymbol{0.300}$ & $0.326$ & $0.639$ & $\boldsymbol{0.085}$ & $0.204$ \\
SNN $\uparrow$ & $0.410$ & $\boldsymbol{0.472}$ & $0.445$ & $0.506$ & $\boldsymbol{0.583}$ & $0.560$ & $0.385$ & $\boldsymbol{0.463}$ & $0.428$ & $0.362$ & $\boldsymbol{0.430}$ & $0.407$ & $0.352$ & $\boldsymbol{0.650}$ & $0.522$ \\
Frag $\uparrow$ & $0.994$ & $\boldsymbol{0.999}$ & $0.998$ & $0.998$ & $\boldsymbol{0.999}$ & $\boldsymbol{0.999}$ & $0.991$ & $0.996$ & $\boldsymbol{0.997}$ & $0.991$ & $0.996$ & $\boldsymbol{0.997}$ & $0.963$ & $\boldsymbol{0.994}$ & $0.987$ \\
Scaff $\uparrow$ & $0.890$ & $\boldsymbol{0.932}$ & $0.925$ & $0.962$ & $\boldsymbol{0.981}$ & $0.979$ & $0.861$ & $\boldsymbol{0.913}$ & $0.906$ & $0.839$ & $\boldsymbol{0.863}$ & $0.846$ & $0.035$ & $\boldsymbol{0.537}$ & $0.368$ \\
IntDiv $\uparrow$ & $\boldsymbol{0.886}$ & $0.885$ & $0.884$ & $\boldsymbol{0.867}$ & $0.863$ & $0.864$ & $\boldsymbol{0.896}$ & $0.892$ & $0.892$ & $\boldsymbol{0.879}$ & $0.876$ & $0.876$ & $\boldsymbol{0.869}$ & $0.854$ & $0.851$ \\
FG MMD$^2$ $\downarrow$ & $\boldsymbol{0.0020}$ & $\boldsymbol{0.0020}$ & $\boldsymbol{0.0020}$ & $\boldsymbol{0.0030}$ & $0.0031$ & $\boldsymbol{0.0030}$ & $0.0003$ & $\boldsymbol{0.0002}$ & $\boldsymbol{0.0002}$ & $\boldsymbol{0.0021}$ & $\boldsymbol{0.0021}$ & $\boldsymbol{0.0021}$ & $0.0023$ & $\boldsymbol{0.0020}$ & $0.0021$ \\
BRICS L2 $\downarrow$ & $0.0243$ & $\boldsymbol{0.0184}$ & $0.0193$ & $0.0521$ & $0.0509$ & $\boldsymbol{0.0498}$ & $0.0178$ & $0.0112$ & $\boldsymbol{0.0105}$ & $0.0241$ & $0.0220$ & $\boldsymbol{0.0210}$ & $0.0384$ & $\boldsymbol{0.0184}$ & $0.0238$ \\
\bottomrule
\end{tabular}
\end{adjustbox}
\end{table}

%% file: tables/main_autograph_comparison.tex
\newsavebox{\AutographLeftBox}
\newlength{\AutographLeftHeight}

\begin{lrbox}{\AutographLeftBox}
\begin{minipage}[t]{0.48\linewidth}
\centering
\begin{adjustbox}{width=\linewidth}
\begin{tabular}{@{}l cccccc@{}}
\toprule
\multicolumn{7}{c}{\textbf{MOSES}}\\
\midrule
\textbf{Model} & \textbf{Valid$\uparrow$} & \textbf{Unique$\uparrow$} & \textbf{Novel$\uparrow$} & \textbf{Filters$\uparrow$} & \textbf{FCD$\downarrow$} & \textbf{SNN$\uparrow$}\\
\midrule
VAE - \emph{SMILES}$^*$         & 97.7 & 99.8 & 69.5 & \textbf{99.7} & 0.57 & 0.58\\
JT-VAE - \emph{FRAGMENTS}$^*$      & \textbf{100.0} & \textbf{100.0} & \textbf{99.9} & 97.8 & 1.00 & 0.53\\
GraphINVENT$^*$ & 96.4 & 99.8 & n/a   & 95.0 & 1.22 & 0.54\\
DiGress$^*$     & 85.7 & \textbf{100.0} & 95.0 & 97.1 & 1.19 & 0.52\\
SENT ( GPT2-S)$^*$       & 87.4 & \textbf{100.0} & 85.9 & 98.6 & 0.91 & 0.55\\
SENT ( GPT2-XS)              & $97.7$ & $100.0$ & $99.0$ & $94.2$ & $0.70$ & $0.506$\\
\emph{HDT}-HAC-R ( GPT2-XS)  & $97.8$ & $99.9$ & $95.0$ & $98.1$ & $0.25$ & $0.583$\\
\emph{HDT}-HAC-FR ( GPT2-XS) & $98.5$ & $100.0$ & $97.1$ & $97.5$ & $0.26$ & $0.560$\\
\bottomrule
\end{tabular}
\end{adjustbox}

\begin{adjustbox}{width=\linewidth}
\begin{tabular}{@{}l ccccc@{}}
\toprule
\multicolumn{6}{c}{\textbf{GuacaMol}}\\
\midrule
\textbf{Model} & \textbf{Valid$\uparrow$} & \textbf{Unique$\uparrow$} & \textbf{Novel$\uparrow$} & \textbf{KL score$\uparrow$} & \textbf{FCD$\downarrow$}\\
\midrule
LSTM - \emph{SMILES}$^*$         & 95.9 & \textbf{100.0} & 91.2 & \textbf{99.1} & \textbf{0.46}\\
NAGVAE$^*$       & 92.7 & 95.5 & \textbf{100.0} & 38.4 & 23.55\\
MCTS$^*$         & \textbf{100.0} & \textbf{100.0} & 99.4 & 52.2 & 21.00\\
DiGress$^*$      & 85.2 & \textbf{100.0} & 99.9 & 92.9 & 1.93\\
SENT ( GPT2-S)$^*$   & 91.6 & \textbf{100.0} & 97.7 & 97.5 & 1.17\\
SENT ( GPT2-XS)              & $96.7$ & $100.0$ & $99.8$ & $93.8$ & $1.80$\\
\emph{HDT}-HAC-R ( GPT2-XS)  & $94.3$ & $100.0$ & $98.7$ & $98.1$ & $0.87$\\
\emph{HDT}-HAC-FR ( GPT2-XS) & $96.6$ & $100.0$ & $99.6$ & $96.3$ & $0.99$\\
\bottomrule
\end{tabular}
\end{adjustbox}
\end{minipage}
\end{lrbox}

\setlength{\AutographLeftHeight}{\dimexpr\ht\AutographLeftBox+\dp\AutographLeftBox\relax}

\begin{table*}[t]
    \caption{\textbf{Placement within the graph-generation literature.} Layout follows \citet{chen2025autograph}; methods marked $^*$ are transcribed as published. Included for context, not as a superiority claim: budgets and evaluation lineages differ. Our controlled comparison is Table~\ref{tab:main_molecular_all}. Validity, uniqueness, novelty, Filters and the GuacaMol KL score are percentages; FCD and SNN are unscaled. Harmonization in Appendix~\ref{app:autograph_metric_methodology}.}
    \label{tab:autograph_comparison}
    \centering
    \small
    \setlength{\tabcolsep}{4pt}
    \begin{minipage}[t]{0.48\linewidth}
    \vspace{0pt}
    \usebox{\AutographLeftBox}
    \end{minipage}\hfill
    \begin{minipage}[t][\AutographLeftHeight][t]{0.49\linewidth}
    \vspace{0pt}
    \centering
    \begin{adjustbox}{max totalsize={\linewidth}{\AutographLeftHeight},keepaspectratio}
    \begin{tabular}{@{}l ccccc@{}}
    \toprule
    \multicolumn{6}{c}{\textbf{Proteins}}\\
    \midrule
    \textbf{Model} & \textbf{Deg$\downarrow$} & \textbf{Clus$\downarrow$} & \textbf{Orbit$\downarrow$} & \textbf{Spec$\downarrow$} & \textbf{Ratio$\downarrow$}\\
    \midrule
    Training set    & 0.0003 & 0.0068 & 0.0032 & 0.0005 & 1.00\\
    GraphRNN$^*$    & 0.0040 & 0.1475 & 0.5851 & 0.0152 & 62.10\\
    GRAN$^*$        & 0.0479 & 0.1234 & 0.3458 & 0.0125 & 77.70\\
    SPECTRE$^*$     & 0.0056 & 0.0843 & 0.0267 & 0.0052 & 12.50\\
    EDGE$^*$        & 0.1863 & 0.3406 & 0.6786 & 0.1075 & 274.50\\
    GraphGen$^*$    & 0.0159 & 0.1677 & 0.3789 & 0.0181 & 58.10\\
    BiGG$^*$        & 0.0070 & 0.1150 & 0.4696 & 0.0067 & 50.10\\
    DiGress$^*$     & 0.0041 & 0.0489 & 0.1286 & 0.0018 & 16.20\\
    GruM$^*$        & 0.0019 & 0.0660 & 0.0345 & 0.0030 & 8.20\\
    GEEL$^*$        & 0.2110 & 0.3753 & 0.1768 & 0.1689 & 287.90\\
    ESGG$^*$        & 0.0030 & 0.0309 & \textbf{0.0047} & 0.0013 & 4.70\\
    SENT ( GPT2-S)$^*$       & 0.0004 & 0.0244 & 0.0056 & 0.0013 & 2.30\\
    SENT ( GPT2-XS)            & 0.0039 & 0.0820 & 0.0278 & 0.0047 & 10.72\\
    \emph{HDT}-HAC-R ( GPT2-XS)  & \textbf{0.0003} & \textbf{0.0128} & 0.0060 & \textbf{0.0007} & 1.57\\
    \bottomrule
    \end{tabular}
    \end{adjustbox}
    \end{minipage}
\end{table*}

%% file: tables/appendix_randomization_example.tex
\begin{table}[h]
\caption{\textbf{Randomization on a fixed graph.}
Consider the path $a-b-c-d-e-f$ with communities
$C_1=\{a,b\}$, $C_2=\{c,d\}$, and $C_3=\{e,f\}$.
R preserves community order; FR also changes that order subject to
adjacency and entry constraints.
Bars separate communities. Letters denote original nodes;
emitted node identifiers are reassigned as $0,\ldots,5$ in visitation
order. Delimiters and back-edge brackets are omitted.}
\label{tab:randomization_example}
\centering
\small
\begin{tabular}{@{}l c l@{}}
\toprule
\textbf{Policy} & \textbf{Community order} & \textbf{Node order} \\
\midrule
\emph{-R}
& $C_1,C_2,C_3$
& $(a,b)\mid(c,d)\mid(e,f)$ \\
\emph{-R}
& $C_1,C_2,C_3$
& $(b,a)\mid(d,c)\mid(f,e)$ \\
\emph{-FR}
& $C_2,C_3,C_1$
& $(c,d)\mid(e,f)\mid(b,a)$ \\
\emph{-FR}
& $C_3,C_2,C_1$
& $(f,e)\mid(d,c)\mid(b,a)$ \\
\bottomrule
\end{tabular}
\end{table}

%% file: tables/appendix_dataset_stats.tex
\begin{table}[h]
\caption{\textbf{Molecular training populations and evaluation reference sets.} The
standard and novel-only evaluations both score against the listed held-out reference arm.
Official \texttt{MoleculePGD} uses ten no-replacement draws of the listed per-arm size. The
 training-sample reference column independently permutes the exact training population
for each generation seed before taking the same evaluation-sized prefix, so it measures how
close an in-distribution sample of the training data itself lands on each metric. These
figures are properties of the data and the protocol, and are independent of tokenizer and
model.}
\label{tab:appendix_dataset_stats}
\centering
\small
\setlength{\tabcolsep}{5pt}
\begin{adjustbox}{max width=\textwidth}
\begin{tabular}{@{}l r l r r r@{}}
\toprule
\textbf{Dataset} & \textbf{Training $n$} & \textbf{Reference arm} & \makecell{\textbf{Evaluation}\\$n$} & \makecell{\textbf{PGD}\\\textbf{draw}} & \makecell{\textbf{ Training}\\\textbf{ samples} $n$} \\
\midrule
GuacaMol               & 499{,}968 & Held-out test                    & 10{,}000 & 5{,}000 & 10{,}000 \\
MOSES                  & 499{,}968 & Held-out test                    & 10{,}000 & 5{,}000 & 10{,}000 \\
COCONUT 2.0            & 367{,}104 & Dedicated held-out test          & 10{,}000 & 5{,}000 & 10{,}000 \\
GEOM-Drugs (disjoint)  & 62{,}720  & Audited disjoint held-out test   & 8{,}235  & 4{,}117 & 8{,}235 \\
Macrocycle             & 27{,}392  & Dedicated held-out test          & 3{,}328  & 1{,}664 & 3{,}328 \\
\bottomrule
\end{tabular}
\end{adjustbox}
\end{table}

%% file: tables/appendix_hierarchy_depths.tex
\begin{table}[h]
\caption{Observed hierarchy depth under recursive HAC coarsening. We sample $200$ graphs
without replacement from each frozen training population at $\tau=4$, seed $=42$.
\emph{$k$} counts communities in the top-level partition. \emph{Depth} counts levels
below these top-level communities; add one to obtain maximum leaf depth with root depth
zero, as used in the algorithms and Table~\ref{tab:appendix_block_alignment}.}
\label{tab:appendix_hierarchy_depths}
\centering
\small
\begin{adjustbox}{max width=\textwidth}
\begin{tabular}{@{}l r ccc@{}}
\toprule
\textbf{Dataset} & \textbf{Mean nodes} & \textbf{Mean $k$} & \textbf{Mean depth} & \textbf{Max depth} \\
\midrule
GuacaMol              & $27.9$ & $5.3$ & $1.29$ & $2$ \\
MOSES                 & $21.7$ & $4.8$ & $1.13$ & $2$ \\
COCONUT 2.0           & $37.2$ & $5.9$ & $1.51$ & $3$ \\
GEOM-Drugs (disjoint) & $25.5$ & $5.1$ & $1.21$ & $3$ \\
Macrocycle            & $58.5$ & $8.0$ & $1.86$ & $3$ \\
\bottomrule
\end{tabular}
\end{adjustbox}
\end{table}

%% file: tables/appendix_token_length.tex
\begin{table}[h]
\caption{ \textbf{Token sequence lengths.} \textbf{(a)} Molecular SENT/HDT means use $200$ sampled training graphs. Legacy HSENT/HDTC means use $499{,}968$ graphs per dataset; the historical non-molecular cohorts contain $587$ Proteins and $128$ Planar/SBM training graphs. Each audit uses three deterministic serialization draws per graph. HSENT/HDT family means average the audited coarsenings, including Louvain in the historical audit. \textbf{(b)} Main GPT2-XS context caps, with legacy tokenizer exceptions. The SBM column denotes default SENT/HAC; adjusted HAC and PH use $1440$. SSBM and RSBM contexts are given in Tables~\ref{tab:appendix_train_inventory_nonmol} and~\ref{tab:appendix_block_vun}; GPT2-S settings are in Tables~\ref{tab:appendix_regime_guacamol} and~\ref{tab:appendix_regime_moses}. HSENT/HDTC results are reported only for GuacaMol and MOSES; HDTC is molecular only.}
\label{tab:appendix_token_length}
\centering
\small
\begin{adjustbox}{max width=\textwidth}
\begin{tabular}{@{}l cccccccc@{}}
\toprule
 & \textbf{GuacaMol} & \textbf{MOSES} & \makecell{\textbf{COCONUT}\\\textbf{2.0}} & \makecell{\textbf{GEOM-}\\\textbf{Drugs}} & \textbf{Macrocycle} & \textbf{Proteins} & \textbf{Planar} & \textbf{SBM} \\
\midrule
\multicolumn{9}{@{}l}{\emph{(a) Mean  tokenization length}}\\
SENT           & 121.6 & 94.3 & 161.4 & 111.0 & 255.2 & 964.3 & 258.7 &  656.5 \\
\emph{HSENT}   & 357.3 & 274.8 & \texttt{n/a} & \texttt{n/a} & \texttt{n/a} & \texttt{n/a} & \texttt{n/a} & \texttt{n/a} \\
\emph{HDT}     & 227.6 & 178.2 & 300.6 & 209.4 & 497.0 & 2046.5 & 509.2 &  1081.8 \\
\emph{HDTC}    & 275.3 & 215.8 & \texttt{n/a} & \texttt{n/a} & \texttt{n/a} & \texttt{n/a} & \texttt{n/a} & \texttt{n/a} \\
\midrule
\multicolumn{9}{@{}l}{\emph{(b) Maximum token length}}\\
SENT           & 512 & 512 & 1024 & 1024 & 2048 & 5120 & 2048 &  2048 \\
\emph{HSENT}   & 768 & 512 & \texttt{n/a} & \texttt{n/a} & \texttt{n/a} & \texttt{n/a} & \texttt{n/a} & \texttt{n/a} \\
\emph{HDT}     & 512 & 512 & 1024 & 1024 & 2048 & 5120 & 2048 &  2048 \\
\emph{HDTC}    & 768 & 512 & \texttt{n/a} & \texttt{n/a} & \texttt{n/a} & \texttt{n/a} & \texttt{n/a} & \texttt{n/a} \\
\bottomrule
\end{tabular}
\end{adjustbox}
\end{table}

%% file: tables/appendix_architecture.tex
\begin{table}[h]
\centering
\caption{\textbf{Defaults for the main GPT2-XS regime.} Per-configuration schedules, contexts
and achieved runtimes are in Tables~\ref{tab:appendix_train_inventory_mol}
and~\ref{tab:appendix_train_inventory_nonmol}. Legacy, scaling and block-intervention
settings are described in Appendices~\ref{app:legacy_variants},
\ref{app:regime_delta} and~\ref{app:sbm_full}; conditional generation follows
Appendix~\ref{app:conditional_generation}. Shared defaults do not remove these differences.}
\label{tab:appendix_arch}
\small
\begin{tabular}{@{}p{0.33\linewidth}p{0.63\linewidth}@{}}
\toprule
\textbf{Setting} & \textbf{Default choice} \\
\midrule
Backbone & GPT2-XS (6 layers, hidden 384, 12 heads, $\sim$11M parameters) \\
Optimizer & AdamW, weight decay $0.1$, gradient clipping $1.0$, cosine decay \\
Training precision & bf16 mixed \\
Regularization & attention dropout $0.5$ \\
Molecular batch & 256 per GPU; one GPU; no gradient accumulation \\
Batching & length-aware / bucketed batching disabled \\
Molecular unconditional decoding & top-$k$ $5$; temperature $1.0$ \\
Checkpoint seed & 42 \\
Evaluation seeds & 42, 43, 44 from the same checkpoint \\
Reporting & mean $\pm$ sample SD over the three generation seeds \\
Budget calibration & measured hierarchical training throughput; achieved times vary \\
Coarsening recursion threshold $\tau$
& 4 by default; 44 for adjusted SBM; 10 for adjusted RSBM \\
\bottomrule
\end{tabular}
\end{table}

%% file: tables/appendix_train_inventory_molecular.tex
\begin{table}[h]
\caption{\textbf{Molecular training inventory.} Presentations are optimizer steps times
effective batch size. Runtime is active wall-clock time inferred from each run's resolved
configuration and its final checkpoint timestamps. Update budgets were calibrated using measured training time rather than equal steps.
The achieved runtimes shown here can differ from those calibration targets; both
update counts and runtimes are given. All runs use GPT2-XS on a single GPU with no gradient
accumulation (Table~\ref{tab:appendix_arch}).}
\label{tab:appendix_train_inventory_mol}
\centering
\small
\setlength{\tabcolsep}{5pt}
\begin{adjustbox}{max width=\textwidth}
\begin{tabular}{@{}l l r r r r r r@{}}
\toprule
\textbf{Dataset} & \textbf{Variant} & \textbf{Context} & \textbf{Batch} & \textbf{Steps} & \textbf{Presentations} & \textbf{Epochs} & \textbf{Runtime (h)} \\
\midrule
\multirow{5}{*}{GuacaMol}
 & SENT             & 512 & 256 & 162{,}500 & 41.6M & 83.2 & 4.98 \\
 & \emph{HDT}-MC-R      & 512 & 256 & 100{,}000 & 25.6M & 51.2 & 5.65 \\
 & \emph{HDT}-HAC-R     & 512 & 256 & 100{,}000 & 25.6M & 51.2 & 6.70 \\
 & \emph{HDT}-MC-FR     & 512 & 256 & 100{,}000 & 25.6M & 51.2 & 4.24 \\
 & \emph{HDT}-HAC-FR    & 512 & 256 & 100{,}000 & 25.6M & 51.2 & 5.92 \\
\midrule
\multirow{5}{*}{MOSES}
 & SENT             & 512 & 256 & 100{,}000 & 25.6M & 51.2 & 3.35 \\
 & \emph{HDT}-MC-R      & 512 & 256 & 62{,}500 & 16M & 32.0 & 3.31 \\
 & \emph{HDT}-HAC-R     & 512 & 256 & 62{,}500 & 16M & 32.0 & 4.50 \\
 & \emph{HDT}-MC-FR     & 512 & 256 & 62{,}500 & 16M & 32.0 & 1.41 \\
 & \emph{HDT}-HAC-FR    & 512 & 256 & 62{,}500 & 16M & 32.0 & 1.36 \\
\midrule
\multirow{5}{*}{COCONUT 2.0}
 & SENT             & 1{,}024 & 256 & 118{,}000 & 30.2M & 82.3 & 4.30 \\
 & \emph{HDT}-MC-R      & 1{,}024 & 256 & 60{,}000 & 15.4M & 41.8 & 4.35 \\
 & \emph{HDT}-HAC-R     & 1{,}024 & 256 & 60{,}000 & 15.4M & 41.8 & 4.25 \\
 & \emph{HDT}-MC-FR     & 1{,}024 & 256 & 60{,}000 & 15.4M & 41.8 & 4.40 \\
 & \emph{HDT}-HAC-FR    & 1{,}024 & 256 & 60{,}000 & 15.4M & 41.8 & 4.42 \\
\midrule
\multirow{5}{*}{GEOM-Drugs (disjoint)}
 & SENT             & 1{,}024 & 256 & 46{,}000 & 11.8M & 187.8 & 0.98 \\
 & \emph{HDT}-MC-R      & 1{,}024 & 256 & 25{,}000 & 6.4M & 102.0 & 1.00 \\
 & \emph{HDT}-HAC-R     & 1{,}024 & 256 & 25{,}000 & 6.4M & 102.0 & 1.00 \\
 & \emph{HDT}-MC-FR     & 1{,}024 & 256 & 25{,}000 & 6.4M & 102.0 & 1.02 \\
 & \emph{HDT}-HAC-FR    & 1{,}024 & 256 & 25{,}000 & 6.4M & 102.0 & 1.05 \\
\midrule
\multirow{5}{*}{Macrocycle}
 & SENT             & 2{,}048 & 256 & 47{,}000 & 12M & 439.3 & 1.80 \\
 & \emph{HDT}-MC-R      & 2{,}048 & 256 & 25{,}000 & 6.4M & 233.6 & 2.10 \\
 & \emph{HDT}-HAC-R     & 2{,}048 & 256 & 25{,}000 & 6.4M & 233.6 & 1.85 \\
 & \emph{HDT}-MC-FR     & 2{,}048 & 256 & 25{,}000 & 6.4M & 233.6 & 2.12 \\
 & \emph{HDT}-HAC-FR    & 2{,}048 & 256 & 25{,}000 & 6.4M & 233.6 & 1.94 \\
\bottomrule
\end{tabular}
\end{adjustbox}
\end{table}

%% file: tables/appendix_train_inventory_nonmolecular.tex
\begin{table}[h]
\caption{\textbf{Non-molecular training inventory.} Presentations are steps $\times$ effective batch. SENT update budgets were calibrated using benchmark timings; achieved runtimes vary and can be lower than those of hierarchical configurations. Contexts differ by row because each was set from that configuration's own hierarchy bounds.}
\label{tab:appendix_train_inventory_nonmol}
\centering
\small
\setlength{\tabcolsep}{5pt}
\begin{adjustbox}{max width=\textwidth}
\begin{tabular}{@{}l l r r r r r r@{}}
\toprule
\textbf{Dataset} & \textbf{Variant} & \textbf{Context} & \textbf{Batch} & \textbf{Steps} & \textbf{Presentations} & \textbf{Epochs} & \textbf{Runtime (h)} \\
\midrule
\multirow{3}{*}{Proteins}
 & SENT                  & 5{,}120 & 32 & 955{,}000 & 30.6M & 52061.3 & 24.45 \\
 & \emph{HDT}-HAC-R          & 5{,}120 & 32 & 400{,}000 & 12.8M & 21805.8 & 24.39 \\
 & \emph{HDT}-HAC-FR   & 5{,}120 & 32 & 400{,}000 & 12.8M & 21805.8 & 25.72 \\
\midrule
\multirow{3}{*}{Planar}
 & SENT                  & 2{,}048 & 128 & 13{,}000 & 1.66M & 162.5 & 0.20 \\
 & \emph{HDT}-HAC-R          & 2{,}048 & 128 & 10{,}000 & 1.28M & 125.0 & 0.29 \\
 & \emph{HDT}-HAC-FR   & 2{,}048 & 128 & 10{,}000 & 1.28M & 125.0 & 0.30 \\
\midrule
\multirow{5}{*}{SBM}
 & SENT                      & 2{,}048 & 64 & 458{,}500 & 29.3M & 229250.0 & 17.15 \\
 & \emph{HDT}-HAC-R          & 2{,}048 & 64 & 200{,}000 & 12.8M & 100000.0 & 17.74 \\
 & \emph{HDT}-HAC-FR   & 2{,}048 & 64 & 200{,}000 & 12.8M & 100000.0 & 20.44 \\
 & \emph{HDT}-PH-R           & 1{,}440 & 64 & 200{,}000 & 12.8M & 100000.0 & 4.02 \\
 & \emph{HDT}-PH-FR          & 1{,}440 & 64 & 200{,}000 & 12.8M & 100000.0 & 4.03 \\
\midrule
\multirow{3}{*}{Sparse SBM (SSBM)}
 & SENT                      & 2{,}048 & 64 & 63{,}000 & 4.03M & 31500.0 & 1.56 \\
 & \emph{HDT}-HAC-R          & 2{,}048 & 64 & 20{,}000 & 1.28M & 10000.0 & 1.62 \\
 & \emph{HDT}-HAC-FR   & 2{,}048 & 64 & 20{,}000 & 1.28M & 10000.0 & 1.88 \\
\midrule
\multirow{5}{*}{\makecell[l]{\\Connected RSBM\\($R=4$)}}
 & SENT                      & 1{,}600 & 64 & 245{,}000 & 15.7M & 122500.0 & 4.77 \\
 & \emph{HDT}-HAC-R          & 1{,}664 & 64 & 200{,}000 & 12.8M & 100000.0 & 4.55 \\
 & \emph{HDT}-HAC-FR   & 1{,}664 & 64 & 200{,}000 & 12.8M & 100000.0 & 4.56 \\
 & \emph{HDT}-PH-R           & 1{,}600 & 64 & 200{,}000 & 12.8M & 100000.0 & 4.46 \\
 & \emph{HDT}-PH-FR          & 1{,}600 & 64 & 200{,}000 & 12.8M & 100000.0 & 4.46 \\
\bottomrule
\end{tabular}
\end{adjustbox}
\end{table}

%% file: tables/appendix_generation_details.tex
\begin{table}[h]
\caption{Unconditional generation parameters per dataset. \textbf{Attempts} counts all generation
attempts per seed, including invalid outputs; the valid population available for molecular
metrics can be smaller. \textbf{PGD draw} is the per-arm size of a no-replacement
MoleculePGD draw; \textbf{Motif ref.} is the fixed valid reference-arm size for motif discrepancies.
Molecular reference counts are in Table~\ref{tab:appendix_dataset_stats}; block models
use independent $1000$-graph reference arms. Top-$k$ and temperature are fixed across
variants within each reported unconditional comparison. Context limits are in
Table~\ref{tab:appendix_token_length}b and, for RSBM, Table~\ref{tab:appendix_block_vun}.
Conditional generation uses its separate protocol in Appendix~\ref{app:conditional_generation}.}
\label{tab:appendix_generation_details}
\centering
\small
\begin{adjustbox}{max width=\textwidth}
\begin{tabular}{l rrrrr}
\toprule
\textbf{Dataset} & \textbf{Attempts} & \textbf{Top-$k$} & \textbf{Temperature} & \textbf{PGD draw} & \textbf{Motif ref.} \\
\midrule
GuacaMol              & 10{,}000 & 5  & 1.0 & 5{,}000 & 1{,}000 \\
MOSES                 & 10{,}000 & 5  & 1.0 & 5{,}000 & 1{,}000 \\
COCONUT 2.0           & 10{,}000 & 5  & 1.0 & 5{,}000 & 1{,}000 \\
GEOM-Drugs (disjoint) & 8{,}235  & 5  & 1.0 & 4{,}117 & 1{,}000 \\
Macrocycle            & 3{,}328  & 5  & 1.0 & 1{,}664 & 1{,}000 \\
\midrule
Proteins              & 184 & 40 & 1.0 & N/A & N/A \\
Planar                & 40  & 10 & 1.0 & N/A & N/A \\
SBM                   & 1{,}000 & 60 & 1.0 & N/A & N/A \\
Sparse SBM (SSBM)     & 1{,}000 & 60 & 1.0 & N/A & N/A \\
Original/connected RSBM ($R=4$) & 1{,}000 & 60 & 1.0 & N/A & N/A \\
\bottomrule
\end{tabular}
\end{adjustbox}
\end{table}

%% file: tables/appendix_guacamol_full.tex
\begingroup
\caption{\textbf{GuacaMol: standard unconditional generation.} Mean\,$\pm$\,sample SD; lower panel: improvement over SENT. Shared conventions: Appendix~\ref{app:unconditional_full}.}
\label{tab:appendix_guacamol_full}
\centering
\scriptsize
\setlength{\tabcolsep}{2.2pt}
\renewcommand{\arraystretch}{0.96}
\begin{adjustbox}{max width=\textwidth}
\begin{tabular}{@{}lcccccc@{}}
\toprule
 & \textbf{SENT} & \textbf{HDT-MC-R}$^{*}$ & \textbf{HDT-MC-FR}$^{*}$ & \textbf{HDT-HAC-R}$^{*}$ & \textbf{HDT-HAC-FR}$^{*}$ & \textbf{Training samples} \\
\midrule
\multicolumn{7}{@{}l}{\emph{Absolute performance}} \\
Validity $\uparrow$ & $0.967 \pm 0.001$ & $0.947 \pm 0.001$ & $\boldsymbol{0.976 \pm 0.002}$ & $0.943 \pm 0.001$ & $0.966 \pm 0.002$ & $1.000 \pm 0.000$ \\
Uniqueness $\uparrow$ & $\boldsymbol{1.000 \pm 0.000}$ & \underline{$1.000 \pm 0.000$} & $1.000 \pm 0.000$ & $1.000 \pm 0.000$ & $1.000 \pm 0.000$ & $1.000 \pm 0.000$ \\
Novelty $\uparrow$ & $\boldsymbol{0.998 \pm 0.001}$ & $0.990 \pm 0.001$ & $0.995 \pm 0.001$ & $0.987 \pm 0.001$ & \underline{$0.996 \pm 0.001$} & $0.000 \pm 0.000$ \\
V.U.N. $\uparrow$ & $0.965 \pm 0.000$ & $0.937 \pm 0.001$ & $\boldsymbol{0.971 \pm 0.001}$ & $0.931 \pm 0.001$ & $0.961 \pm 0.003$ & $0.000 \pm 0.000$ \\
FCD $\downarrow$ & $1.799 \pm 0.045$ & \underline{$0.936 \pm 0.015$} & $0.997 \pm 0.014$ & $\boldsymbol{0.871 \pm 0.019}$ & $0.989 \pm 0.013$ & $0.527 \pm 0.004$ \\
SNN $\uparrow$ & $0.410 \pm 0.001$ & \underline{$0.458 \pm 0.002$} & $0.448 \pm 0.000$ & $\boldsymbol{0.472 \pm 0.000}$ & $0.445 \pm 0.001$ & $0.658 \pm 0.001$ \\
Frag $\uparrow$ & $0.994 \pm 0.000$ & \underline{$0.998 \pm 0.000$} & $0.998 \pm 0.000$ & $\boldsymbol{0.999 \pm 0.000}$ & $0.998 \pm 0.000$ & $0.999 \pm 0.000$ \\
Scaff $\uparrow$ & $0.890 \pm 0.007$ & $0.906 \pm 0.004$ & $0.887 \pm 0.002$ & $\boldsymbol{0.932 \pm 0.004}$ & \underline{$0.925 \pm 0.006$} & $0.937 \pm 0.008$ \\
IntDiv $\uparrow$ & $\boldsymbol{0.886 \pm 0.001}$ & $0.884 \pm 0.001$ & $0.884 \pm 0.001$ & \underline{$0.885 \pm 0.001$} & $0.884 \pm 0.001$ & $0.887 \pm 0.001$ \\
MoleculePGD $\downarrow$ & $0.309 \pm 0.005$ & \underline{$0.179 \pm 0.005$} & $0.186 \pm 0.004$ & $\boldsymbol{0.176 \pm 0.007}$ & $0.196 \pm 0.005$ & $0.129 \pm 0.005$ \\
KL score $\uparrow$ & $0.938 \pm 0.010$ & \underline{$0.978 \pm 0.001$} & $0.969 \pm 0.001$ & $\boldsymbol{0.981 \pm 0.001}$ & $0.963 \pm 0.002$ & $0.998 \pm 0.001$ \\
FG MMD$^2$ $\downarrow$ & $\boldsymbol{0.0020 \pm 0.0000}$ & \underline{$0.0020 \pm 0.0000$} & $0.0020 \pm 0.0000$ & $0.0020 \pm 0.0000$ & $0.0020 \pm 0.0000$ & $0.0020 \pm 0.0000$ \\
SMARTS MMD$^2$ $\downarrow$ & $0.0023 \pm 0.0001$ & $\boldsymbol{0.0021 \pm 0.0000}$ & \underline{$0.0021 \pm 0.0001$} & $0.0021 \pm 0.0001$ & $0.0022 \pm 0.0001$ & $0.0019 \pm 0.0000$ \\
Ring MMD$^2$ $\downarrow$ & $0.0036 \pm 0.0003$ & $0.0024 \pm 0.0003$ & $\boldsymbol{0.0021 \pm 0.0005}$ & $0.0024 \pm 0.0003$ & \underline{$0.0022 \pm 0.0003$} & $0.0019 \pm 0.0002$ \\
BRICS L2 $\downarrow$ & $0.0243 \pm 0.0018$ & $0.0191 \pm 0.0013$ & $\boldsymbol{0.0181 \pm 0.0003}$ & \underline{$0.0184 \pm 0.0008$} & $0.0193 \pm 0.0008$ & $0.0184 \pm 0.0005$ \\
\midrule
\multicolumn{7}{@{}l}{\emph{Direction-normalized change relative to SENT}} \\
Validity $\uparrow$ & $0.00\%$ & \cellcolor{deltared!52}$-2.14\%$ & \cellcolor{deltagreen!62}$+0.89\%$ & \cellcolor{deltared!62}$-2.53\%$ & \cellcolor{deltared!5}$-0.19\%$ & \_ \\
Uniqueness $\uparrow$ & $0.00\%$ & \cellcolor{deltared!62}$-0.04\%$ & $0.00\%$ & \cellcolor{deltared!31}$-0.02\%$ & $0.00\%$ & \_ \\
Novelty $\uparrow$ & $0.00\%$ & \cellcolor{deltared!42}$-0.72\%$ & \cellcolor{deltared!18}$-0.30\%$ & \cellcolor{deltared!62}$-1.06\%$ & \cellcolor{deltared!11}$-0.18\%$ & \_ \\
V.U.N. $\uparrow$ & $0.00\%$ & \cellcolor{deltared!50}$-2.87\%$ & \cellcolor{deltagreen!56}$+0.58\%$ & \cellcolor{deltared!62}$-3.58\%$ & \cellcolor{deltared!6}$-0.37\%$ & \_ \\
FCD $\downarrow$ & $0.00\%$ & \cellcolor{deltagreen!58}$+47.94\%$ & \cellcolor{deltagreen!54}$+44.57\%$ & \cellcolor{deltagreen!62}$+51.56\%$ & \cellcolor{deltagreen!54}$+45.03\%$ & \_ \\
SNN $\uparrow$ & $0.00\%$ & \cellcolor{deltagreen!48}$+11.67\%$ & \cellcolor{deltagreen!39}$+9.32\%$ & \cellcolor{deltagreen!62}$+14.95\%$ & \cellcolor{deltagreen!35}$+8.40\%$ & \_ \\
Frag $\uparrow$ & $0.00\%$ & \cellcolor{deltagreen!56}$+0.45\%$ & \cellcolor{deltagreen!58}$+0.47\%$ & \cellcolor{deltagreen!62}$+0.50\%$ & \cellcolor{deltagreen!57}$+0.46\%$ & \_ \\
Scaff $\uparrow$ & $0.00\%$ & \cellcolor{deltagreen!23}$+1.78\%$ & \cellcolor{deltared!2}$-0.35\%$ & \cellcolor{deltagreen!62}$+4.70\%$ & \cellcolor{deltagreen!51}$+3.84\%$ & \_ \\
IntDiv $\uparrow$ & $0.00\%$ & \cellcolor{deltared!43}$-0.24\%$ & \cellcolor{deltared!50}$-0.28\%$ & \cellcolor{deltared!21}$-0.12\%$ & \cellcolor{deltared!46}$-0.26\%$ & \_ \\
MoleculePGD $\downarrow$ & $0.00\%$ & \cellcolor{deltagreen!60}$+42.02\%$ & \cellcolor{deltagreen!57}$+39.80\%$ & \cellcolor{deltagreen!62}$+43.18\%$ & \cellcolor{deltagreen!52}$+36.51\%$ & \_ \\
KL score $\uparrow$ & $0.00\%$ & \cellcolor{deltagreen!57}$+4.21\%$ & \cellcolor{deltagreen!44}$+3.26\%$ & \cellcolor{deltagreen!62}$+4.61\%$ & \cellcolor{deltagreen!36}$+2.70\%$ & \_ \\
FG MMD$^2$ $\downarrow$ & $0.00\%$ & \cellcolor{deltagreen!45}$+0.60\%$ & \cellcolor{deltagreen!23}$+0.31\%$ & \cellcolor{deltagreen!42}$+0.56\%$ & \cellcolor{deltagreen!62}$+0.82\%$ & \_ \\
SMARTS MMD$^2$ $\downarrow$ & $0.00\%$ & \cellcolor{deltagreen!57}$+9.05\%$ & \cellcolor{deltagreen!62}$+9.80\%$ & \cellcolor{deltagreen!53}$+8.43\%$ & \cellcolor{deltagreen!48}$+7.66\%$ & \_ \\
Ring MMD$^2$ $\downarrow$ & $0.00\%$ & \cellcolor{deltagreen!48}$+32.89\%$ & \cellcolor{deltagreen!62}$+42.57\%$ & \cellcolor{deltagreen!47}$+32.59\%$ & \cellcolor{deltagreen!56}$+38.65\%$ & \_ \\
BRICS L2 $\downarrow$ & $0.00\%$ & \cellcolor{deltagreen!53}$+21.47\%$ & \cellcolor{deltagreen!62}$+25.28\%$ & \cellcolor{deltagreen!59}$+24.00\%$ & \cellcolor{deltagreen!50}$+20.28\%$ & \_ \\
\bottomrule
\end{tabular}
\end{adjustbox}
\par\endgroup

%% file: tables/appendix_guacamol_novel.tex
\begingroup
\caption{\textbf{GuacaMol: novel-only unconditional generation.} Mean\,$\pm$\,sample SD; lower panel: improvement over SENT. Shared conventions: Appendix~\ref{app:unconditional_full}.}
\label{tab:appendix_guacamol_novel}
\centering
\scriptsize
\setlength{\tabcolsep}{2.2pt}
\renewcommand{\arraystretch}{0.96}
\begin{adjustbox}{max width=\textwidth}
\begin{tabular}{@{}lcccccc@{}}
\toprule
 & \textbf{SENT} & \textbf{HDT-MC-R}$^{*}$ & \textbf{HDT-MC-FR}$^{*}$ & \textbf{HDT-HAC-R}$^{*}$ & \textbf{HDT-HAC-FR}$^{*}$ & \textbf{Training samples} \\
\midrule
\multicolumn{7}{@{}l}{\emph{Absolute performance}} \\
FCD $\downarrow$ & $1.798 \pm 0.014$ & \underline{$0.941 \pm 0.011$} & $0.982 \pm 0.017$ & $\boldsymbol{0.862 \pm 0.010}$ & $0.970 \pm 0.014$ & $0.527 \pm 0.004$ \\
SNN $\uparrow$ & $0.410 \pm 0.000$ & \underline{$0.456 \pm 0.000$} & $0.339 \pm 0.001$ & $\boldsymbol{0.469 \pm 0.000}$ & $0.337 \pm 0.001$ & $0.658 \pm 0.001$ \\
Frag $\uparrow$ & $0.994 \pm 0.000$ & \underline{$0.998 \pm 0.000$} & $0.998 \pm 0.000$ & $\boldsymbol{0.999 \pm 0.000}$ & $0.998 \pm 0.000$ & $0.999 \pm 0.000$ \\
Scaff $\uparrow$ & $0.894 \pm 0.010$ & \underline{$0.900 \pm 0.005$} & $0.754 \pm 0.003$ & $\boldsymbol{0.932 \pm 0.003}$ & $0.784 \pm 0.004$ & $0.937 \pm 0.008$ \\
IntDiv $\uparrow$ & $\boldsymbol{0.885 \pm 0.001}$ & $0.883 \pm 0.001$ & $0.882 \pm 0.000$ & \underline{$0.884 \pm 0.001$} & $0.883 \pm 0.001$ & $0.887 \pm 0.001$ \\
MoleculePGD $\downarrow$ & $0.314 \pm 0.009$ & \underline{$0.185 \pm 0.004$} & $0.189 \pm 0.003$ & $\boldsymbol{0.175 \pm 0.004}$ & $0.199 \pm 0.003$ & $0.129 \pm 0.005$ \\
KL score $\uparrow$ & $0.943 \pm 0.004$ & \underline{$0.975 \pm 0.001$} & $0.965 \pm 0.001$ & $\boldsymbol{0.977 \pm 0.002}$ & $0.963 \pm 0.002$ & $0.998 \pm 0.001$ \\
FG MMD$^2$ $\downarrow$ & $\boldsymbol{0.0020 \pm 0.0000}$ & \underline{$0.0020 \pm 0.0000$} & $0.0020 \pm 0.0000$ & $0.0020 \pm 0.0000$ & $0.0020 \pm 0.0000$ & $0.0020 \pm 0.0000$ \\
SMARTS MMD$^2$ $\downarrow$ & $0.0024 \pm 0.0001$ & $0.0022 \pm 0.0001$ & $0.0022 \pm 0.0001$ & \underline{$0.0021 \pm 0.0000$} & $\boldsymbol{0.0020 \pm 0.0001}$ & $0.0019 \pm 0.0000$ \\
Ring MMD$^2$ $\downarrow$ & $0.0040 \pm 0.0007$ & $0.0023 \pm 0.0002$ & \underline{$0.0022 \pm 0.0001$} & $0.0024 \pm 0.0003$ & $\boldsymbol{0.0021 \pm 0.0001}$ & $0.0019 \pm 0.0002$ \\
BRICS L2 $\downarrow$ & $0.0249 \pm 0.0016$ & $0.0199 \pm 0.0020$ & $0.0188 \pm 0.0009$ & \underline{$0.0181 \pm 0.0004$} & $\boldsymbol{0.0180 \pm 0.0010}$ & $0.0184 \pm 0.0005$ \\
\midrule
\multicolumn{7}{@{}l}{\emph{Direction-normalized change relative to SENT}} \\
FCD $\downarrow$ & $0.00\%$ & \cellcolor{deltagreen!57}$+47.63\%$ & \cellcolor{deltagreen!54}$+45.37\%$ & \cellcolor{deltagreen!62}$+52.05\%$ & \cellcolor{deltagreen!55}$+46.02\%$ & \_ \\
SNN $\uparrow$ & $0.00\%$ & \cellcolor{deltagreen!48}$+11.08\%$ & \cellcolor{deltared!61}$-17.47\%$ & \cellcolor{deltagreen!62}$+14.29\%$ & \cellcolor{deltared!62}$-17.85\%$ & \_ \\
Frag $\uparrow$ & $0.00\%$ & \cellcolor{deltagreen!56}$+0.40\%$ & \cellcolor{deltagreen!48}$+0.34\%$ & \cellcolor{deltagreen!62}$+0.44\%$ & \cellcolor{deltagreen!49}$+0.35\%$ & \_ \\
Scaff $\uparrow$ & $0.00\%$ & \cellcolor{deltagreen!10}$+0.71\%$ & \cellcolor{deltared!62}$-15.64\%$ & \cellcolor{deltagreen!62}$+4.26\%$ & \cellcolor{deltared!49}$-12.28\%$ & \_ \\
IntDiv $\uparrow$ & $0.00\%$ & \cellcolor{deltared!50}$-0.24\%$ & \cellcolor{deltared!62}$-0.30\%$ & \cellcolor{deltared!27}$-0.13\%$ & \cellcolor{deltared!52}$-0.25\%$ & \_ \\
MoleculePGD $\downarrow$ & $0.00\%$ & \cellcolor{deltagreen!58}$+41.05\%$ & \cellcolor{deltagreen!56}$+39.88\%$ & \cellcolor{deltagreen!62}$+44.12\%$ & \cellcolor{deltagreen!51}$+36.47\%$ & \_ \\
KL score $\uparrow$ & $0.00\%$ & \cellcolor{deltagreen!58}$+3.44\%$ & \cellcolor{deltagreen!41}$+2.41\%$ & \cellcolor{deltagreen!62}$+3.66\%$ & \cellcolor{deltagreen!37}$+2.16\%$ & \_ \\
FG MMD$^2$ $\downarrow$ & $0.00\%$ & \cellcolor{deltagreen!37}$+0.46\%$ & \cellcolor{deltared!62}$-0.34\%$ & \cellcolor{deltagreen!62}$+0.77\%$ & \cellcolor{deltagreen!34}$+0.42\%$ & \_ \\
SMARTS MMD$^2$ $\downarrow$ & $0.00\%$ & \cellcolor{deltagreen!34}$+9.10\%$ & \cellcolor{deltagreen!38}$+10.15\%$ & \cellcolor{deltagreen!52}$+13.97\%$ & \cellcolor{deltagreen!62}$+16.67\%$ & \_ \\
Ring MMD$^2$ $\downarrow$ & $0.00\%$ & \cellcolor{deltagreen!55}$+42.24\%$ & \cellcolor{deltagreen!58}$+45.14\%$ & \cellcolor{deltagreen!53}$+40.76\%$ & \cellcolor{deltagreen!62}$+47.91\%$ & \_ \\
BRICS L2 $\downarrow$ & $0.00\%$ & \cellcolor{deltagreen!45}$+19.99\%$ & \cellcolor{deltagreen!55}$+24.48\%$ & \cellcolor{deltagreen!61}$+27.19\%$ & \cellcolor{deltagreen!62}$+27.60\%$ & \_ \\
\bottomrule
\end{tabular}
\end{adjustbox}
\par\endgroup

%% file: tables/appendix_moses_full.tex
\begingroup
\caption{\textbf{MOSES: standard unconditional generation.} Mean\,$\pm$\,sample SD; lower panel: improvement over SENT. Shared conventions: Appendix~\ref{app:unconditional_full}.}
\label{tab:appendix_moses_full}
\centering
\scriptsize
\setlength{\tabcolsep}{2.2pt}
\renewcommand{\arraystretch}{0.96}
\begin{adjustbox}{max width=\textwidth}
\begin{tabular}{@{}lcccccc@{}}
\toprule
 & \textbf{SENT} & \textbf{HDT-MC-R}$^{*}$ & \textbf{HDT-MC-FR}$^{*}$ & \textbf{HDT-HAC-R}$^{*}$ & \textbf{HDT-HAC-FR}$^{*}$ & \textbf{Training samples} \\
\midrule
\multicolumn{7}{@{}l}{\emph{Absolute performance}} \\
Validity $\uparrow$ & $0.977 \pm 0.001$ & $0.973 \pm 0.002$ & \underline{$0.987 \pm 0.001$} & $0.978 \pm 0.001$ & $0.985 \pm 0.000$ & $1.000 \pm 0.000$ \\
Uniqueness $\uparrow$ & $\boldsymbol{1.000 \pm 0.000}$ & \underline{$1.000 \pm 0.000$} & $1.000 \pm 0.000$ & $0.999 \pm 0.000$ & $1.000 \pm 0.000$ & $1.000 \pm 0.000$ \\
Novelty $\uparrow$ & $\boldsymbol{0.990 \pm 0.002}$ & $0.963 \pm 0.001$ & $0.971 \pm 0.001$ & $0.950 \pm 0.002$ & $0.971 \pm 0.002$ & $0.000 \pm 0.000$ \\
V.U.N. $\uparrow$ & $\boldsymbol{0.967 \pm 0.001}$ & $0.936 \pm 0.002$ & $0.958 \pm 0.002$ & $0.929 \pm 0.003$ & $0.956 \pm 0.002$ & $0.000 \pm 0.000$ \\
FCD $\downarrow$ & $0.696 \pm 0.044$ & $0.256 \pm 0.013$ & $\boldsymbol{0.235 \pm 0.014}$ & \underline{$0.246 \pm 0.003$} & $0.260 \pm 0.008$ & $0.132 \pm 0.001$ \\
SNN $\uparrow$ & $0.506 \pm 0.002$ & \underline{$0.569 \pm 0.001$} & $0.561 \pm 0.002$ & $\boldsymbol{0.583 \pm 0.000}$ & $0.560 \pm 0.003$ & $0.636 \pm 0.001$ \\
Frag $\uparrow$ & $0.998 \pm 0.000$ & $\boldsymbol{0.999 \pm 0.000}$ & \underline{$0.999 \pm 0.000$} & $0.999 \pm 0.000$ & $0.999 \pm 0.000$ & $1.000 \pm 0.000$ \\
Scaff $\uparrow$ & $0.962 \pm 0.004$ & \underline{$0.979 \pm 0.002$} & $0.977 \pm 0.001$ & $\boldsymbol{0.981 \pm 0.001}$ & $0.979 \pm 0.002$ & $0.983 \pm 0.001$ \\
IntDiv $\uparrow$ & $\boldsymbol{0.867 \pm 0.002}$ & $0.862 \pm 0.003$ & \underline{$0.864 \pm 0.000$} & $0.863 \pm 0.001$ & $0.864 \pm 0.002$ & $0.865 \pm 0.000$ \\
MoleculePGD $\downarrow$ & $0.217 \pm 0.005$ & $0.139 \pm 0.007$ & $\boldsymbol{0.107 \pm 0.002}$ & \underline{$0.130 \pm 0.003$} & $0.137 \pm 0.004$ & $0.004 \pm 0.004$ \\
FG MMD$^2$ $\downarrow$ & $\boldsymbol{0.0030 \pm 0.0001}$ & $0.0031 \pm 0.0000$ & $0.0031 \pm 0.0000$ & $0.0031 \pm 0.0001$ & \underline{$0.0030 \pm 0.0000$} & $0.0030 \pm 0.0000$ \\
SMARTS MMD$^2$ $\downarrow$ & $0.0145 \pm 0.0007$ & $0.0132 \pm 0.0009$ & $\boldsymbol{0.0121 \pm 0.0017}$ & $0.0137 \pm 0.0015$ & $0.0126 \pm 0.0003$ & $0.0116 \pm 0.0013$ \\
Ring MMD$^2$ $\downarrow$ & $0.0201 \pm 0.0018$ & $\boldsymbol{0.0141 \pm 0.0022}$ & \underline{$0.0145 \pm 0.0010$} & $0.0182 \pm 0.0031$ & $0.0171 \pm 0.0001$ & $0.0169 \pm 0.0028$ \\
BRICS L2 $\downarrow$ & $0.0521 \pm 0.0020$ & $\boldsymbol{0.0477 \pm 0.0016}$ & $0.0521 \pm 0.0020$ & $0.0509 \pm 0.0018$ & $0.0498 \pm 0.0033$ & $0.0492 \pm 0.0010$ \\
\midrule
\multicolumn{7}{@{}l}{\emph{Direction-normalized change relative to SENT}} \\
Validity $\uparrow$ & $0.00\%$ & \cellcolor{deltared!62}$-0.37\%$ & \cellcolor{deltagreen!60}$+1.10\%$ & \cellcolor{deltagreen!10}$+0.18\%$ & \cellcolor{deltagreen!47}$+0.86\%$ & \_ \\
Uniqueness $\uparrow$ & $0.00\%$ & \cellcolor{deltared!27}$-0.03\%$ & $0.00\%$ & \cellcolor{deltared!62}$-0.07\%$ & \cellcolor{deltared!35}$-0.04\%$ & \_ \\
Novelty $\uparrow$ & $0.00\%$ & \cellcolor{deltared!42}$-2.79\%$ & \cellcolor{deltared!30}$-1.98\%$ & \cellcolor{deltared!62}$-4.10\%$ & \cellcolor{deltared!30}$-2.00\%$ & \_ \\
V.U.N. $\uparrow$ & $0.00\%$ & \cellcolor{deltared!49}$-3.17\%$ & \cellcolor{deltared!14}$-0.91\%$ & \cellcolor{deltared!62}$-3.99\%$ & \cellcolor{deltared!18}$-1.19\%$ & \_ \\
FCD $\downarrow$ & $0.00\%$ & \cellcolor{deltagreen!59}$+63.27\%$ & \cellcolor{deltagreen!62}$+66.20\%$ & \cellcolor{deltagreen!61}$+64.64\%$ & \cellcolor{deltagreen!59}$+62.69\%$ & \_ \\
SNN $\uparrow$ & $0.00\%$ & \cellcolor{deltagreen!51}$+12.45\%$ & \cellcolor{deltagreen!44}$+10.84\%$ & \cellcolor{deltagreen!62}$+15.19\%$ & \cellcolor{deltagreen!44}$+10.67\%$ & \_ \\
Frag $\uparrow$ & $0.00\%$ & \cellcolor{deltagreen!52}$+0.16\%$ & \cellcolor{deltagreen!55}$+0.17\%$ & \cellcolor{deltagreen!59}$+0.18\%$ & \cellcolor{deltagreen!59}$+0.18\%$ & \_ \\
Scaff $\uparrow$ & $0.00\%$ & \cellcolor{deltagreen!57}$+1.85\%$ & \cellcolor{deltagreen!48}$+1.58\%$ & \cellcolor{deltagreen!62}$+2.02\%$ & \cellcolor{deltagreen!55}$+1.80\%$ & \_ \\
IntDiv $\uparrow$ & $0.00\%$ & \cellcolor{deltared!62}$-0.57\%$ & \cellcolor{deltared!28}$-0.26\%$ & \cellcolor{deltared!40}$-0.37\%$ & \cellcolor{deltared!39}$-0.36\%$ & \_ \\
MoleculePGD $\downarrow$ & $0.00\%$ & \cellcolor{deltagreen!44}$+35.88\%$ & \cellcolor{deltagreen!62}$+50.82\%$ & \cellcolor{deltagreen!49}$+40.10\%$ & \cellcolor{deltagreen!45}$+36.62\%$ & \_ \\
FG MMD$^2$ $\downarrow$ & $0.00\%$ & \cellcolor{deltared!53}$-1.92\%$ & \cellcolor{deltared!34}$-1.21\%$ & \cellcolor{deltared!62}$-2.23\%$ & \cellcolor{deltared!3}$-0.09\%$ & \_ \\
SMARTS MMD$^2$ $\downarrow$ & $0.00\%$ & \cellcolor{deltagreen!33}$+8.99\%$ & \cellcolor{deltagreen!62}$+16.75\%$ & \cellcolor{deltagreen!19}$+5.15\%$ & \cellcolor{deltagreen!48}$+12.87\%$ & \_ \\
Ring MMD$^2$ $\downarrow$ & $0.00\%$ & \cellcolor{deltagreen!62}$+30.20\%$ & \cellcolor{deltagreen!58}$+28.10\%$ & \cellcolor{deltagreen!20}$+9.80\%$ & \cellcolor{deltagreen!31}$+15.34\%$ & \_ \\
BRICS L2 $\downarrow$ & $0.00\%$ & \cellcolor{deltagreen!62}$+8.59\%$ & \cellcolor{deltagreen!1}$+0.16\%$ & \cellcolor{deltagreen!17}$+2.31\%$ & \cellcolor{deltagreen!33}$+4.56\%$ & \_ \\
\bottomrule
\end{tabular}
\end{adjustbox}
\par\endgroup

%% file: tables/appendix_moses_novel.tex
\begingroup
\caption{\textbf{MOSES: novel-only unconditional generation.} Mean\,$\pm$\,sample SD; lower panel: improvement over SENT. Shared conventions: Appendix~\ref{app:unconditional_full}.}
\label{tab:appendix_moses_novel}
\centering
\scriptsize
\setlength{\tabcolsep}{2.2pt}
\renewcommand{\arraystretch}{0.96}
\begin{adjustbox}{max width=\textwidth}
\begin{tabular}{@{}lcccccc@{}}
\toprule
 & \textbf{SENT} & \textbf{HDT-MC-R}$^{*}$ & \textbf{HDT-MC-FR}$^{*}$ & \textbf{HDT-HAC-R}$^{*}$ & \textbf{HDT-HAC-FR}$^{*}$ & \textbf{Training samples} \\
\midrule
\multicolumn{7}{@{}l}{\emph{Absolute performance}} \\
FCD $\downarrow$ & $0.702 \pm 0.044$ & $0.257 \pm 0.011$ & $\boldsymbol{0.240 \pm 0.014}$ & \underline{$0.247 \pm 0.004$} & $0.258 \pm 0.008$ & $0.132 \pm 0.001$ \\
SNN $\uparrow$ & $0.504 \pm 0.002$ & \underline{$0.564 \pm 0.002$} & $0.557 \pm 0.001$ & $\boldsymbol{0.577 \pm 0.000}$ & $0.556 \pm 0.003$ & $0.636 \pm 0.001$ \\
Frag $\uparrow$ & $0.998 \pm 0.000$ & $\boldsymbol{0.999 \pm 0.000}$ & \underline{$0.999 \pm 0.000$} & $0.999 \pm 0.000$ & $0.999 \pm 0.000$ & $1.000 \pm 0.000$ \\
Scaff $\uparrow$ & $0.962 \pm 0.003$ & \underline{$0.979 \pm 0.003$} & $0.977 \pm 0.001$ & $\boldsymbol{0.980 \pm 0.001}$ & $0.979 \pm 0.002$ & $0.983 \pm 0.001$ \\
IntDiv $\uparrow$ & $\boldsymbol{0.866 \pm 0.001}$ & \underline{$0.864 \pm 0.001$} & $0.864 \pm 0.001$ & $0.862 \pm 0.001$ & $0.864 \pm 0.001$ & $0.865 \pm 0.000$ \\
MoleculePGD $\downarrow$ & $0.220 \pm 0.010$ & $0.152 \pm 0.013$ & $\boldsymbol{0.113 \pm 0.002}$ & \underline{$0.136 \pm 0.006$} & $0.140 \pm 0.006$ & $0.004 \pm 0.004$ \\
FG MMD$^2$ $\downarrow$ & $0.0031 \pm 0.0001$ & $0.0031 \pm 0.0001$ & $0.0031 \pm 0.0001$ & $0.0032 \pm 0.0001$ & $\boldsymbol{0.0030 \pm 0.0000}$ & $0.0030 \pm 0.0000$ \\
SMARTS MMD$^2$ $\downarrow$ & $0.0146 \pm 0.0007$ & $0.0130 \pm 0.0010$ & $\boldsymbol{0.0121 \pm 0.0016}$ & $0.0144 \pm 0.0019$ & $0.0127 \pm 0.0003$ & $0.0116 \pm 0.0013$ \\
Ring MMD$^2$ $\downarrow$ & $0.0203 \pm 0.0021$ & $\boldsymbol{0.0138 \pm 0.0021}$ & \underline{$0.0151 \pm 0.0011$} & $0.0195 \pm 0.0026$ & $0.0173 \pm 0.0003$ & $0.0169 \pm 0.0028$ \\
BRICS L2 $\downarrow$ & $0.0523 \pm 0.0017$ & $\boldsymbol{0.0479 \pm 0.0022}$ & $0.0524 \pm 0.0021$ & $0.0520 \pm 0.0021$ & \underline{$0.0496 \pm 0.0028$} & $0.0492 \pm 0.0010$ \\
\midrule
\multicolumn{7}{@{}l}{\emph{Direction-normalized change relative to SENT}} \\
FCD $\downarrow$ & $0.00\%$ & \cellcolor{deltagreen!60}$+63.40\%$ & \cellcolor{deltagreen!62}$+65.78\%$ & \cellcolor{deltagreen!61}$+64.82\%$ & \cellcolor{deltagreen!60}$+63.22\%$ & \_ \\
SNN $\uparrow$ & $0.00\%$ & \cellcolor{deltagreen!51}$+11.97\%$ & \cellcolor{deltagreen!45}$+10.51\%$ & \cellcolor{deltagreen!62}$+14.51\%$ & \cellcolor{deltagreen!44}$+10.31\%$ & \_ \\
Frag $\uparrow$ & $0.00\%$ & \cellcolor{deltagreen!52}$+0.16\%$ & \cellcolor{deltagreen!55}$+0.17\%$ & \cellcolor{deltagreen!55}$+0.17\%$ & \cellcolor{deltagreen!59}$+0.18\%$ & \_ \\
Scaff $\uparrow$ & $0.00\%$ & \cellcolor{deltagreen!56}$+1.72\%$ & \cellcolor{deltagreen!52}$+1.58\%$ & \cellcolor{deltagreen!62}$+1.90\%$ & \cellcolor{deltagreen!58}$+1.77\%$ & \_ \\
IntDiv $\uparrow$ & $0.00\%$ & \cellcolor{deltared!34}$-0.25\%$ & \cellcolor{deltared!31}$-0.23\%$ & \cellcolor{deltared!62}$-0.46\%$ & \cellcolor{deltared!28}$-0.21\%$ & \_ \\
MoleculePGD $\downarrow$ & $0.00\%$ & \cellcolor{deltagreen!40}$+31.26\%$ & \cellcolor{deltagreen!62}$+48.60\%$ & \cellcolor{deltagreen!49}$+38.54\%$ & \cellcolor{deltagreen!47}$+36.48\%$ & \_ \\
FG MMD$^2$ $\downarrow$ & $0.00\%$ & \cellcolor{deltared!19}$-1.05\%$ & \cellcolor{deltared!27}$-1.51\%$ & \cellcolor{deltared!62}$-3.44\%$ & \cellcolor{deltagreen!13}$+0.44\%$ & \_ \\
SMARTS MMD$^2$ $\downarrow$ & $0.00\%$ & \cellcolor{deltagreen!38}$+10.51\%$ & \cellcolor{deltagreen!62}$+17.08\%$ & \cellcolor{deltagreen!4}$+1.23\%$ & \cellcolor{deltagreen!47}$+13.07\%$ & \_ \\
Ring MMD$^2$ $\downarrow$ & $0.00\%$ & \cellcolor{deltagreen!62}$+31.96\%$ & \cellcolor{deltagreen!50}$+25.56\%$ & \cellcolor{deltagreen!8}$+3.94\%$ & \cellcolor{deltagreen!28}$+14.63\%$ & \_ \\
BRICS L2 $\downarrow$ & $0.00\%$ & \cellcolor{deltagreen!62}$+8.49\%$ & \cellcolor{deltared!62}$-0.16\%$ & \cellcolor{deltagreen!5}$+0.69\%$ & \cellcolor{deltagreen!38}$+5.23\%$ & \_ \\
\bottomrule
\end{tabular}
\end{adjustbox}
\par\endgroup

%% file: tables/appendix_coconut_full.tex
\begingroup
\caption{\textbf{COCONUT 2.0: standard unconditional generation.} Mean\,$\pm$\,sample SD; lower panel: improvement over SENT. Shared conventions: Appendix~\ref{app:unconditional_full}.}
\label{tab:appendix_coconut_full}
\centering
\scriptsize
\setlength{\tabcolsep}{2.2pt}
\renewcommand{\arraystretch}{0.96}
\begin{adjustbox}{max width=\textwidth}
\begin{tabular}{@{}lcccccc@{}}
\toprule
 & \textbf{SENT} & \textbf{HDT-MC-R}$^{*}$ & \textbf{HDT-MC-FR}$^{*}$ & \textbf{HDT-HAC-R}$^{*}$ & \textbf{HDT-HAC-FR}$^{*}$ & \textbf{Training samples} \\
\midrule
\multicolumn{7}{@{}l}{\emph{Absolute performance}} \\
Validity $\uparrow$ & \underline{$0.932 \pm 0.000$} & $0.919 \pm 0.002$ & $\boldsymbol{0.946 \pm 0.001}$ & $0.860 \pm 0.002$ & $0.908 \pm 0.002$ & $1.000 \pm 0.000$ \\
Uniqueness $\uparrow$ & $\boldsymbol{1.000 \pm 0.000}$ & $0.998 \pm 0.000$ & $0.998 \pm 0.001$ & $0.997 \pm 0.000$ & \underline{$0.999 \pm 0.000$} & $1.000 \pm 0.000$ \\
Novelty $\uparrow$ & $\boldsymbol{0.979 \pm 0.001}$ & $0.927 \pm 0.004$ & \underline{$0.936 \pm 0.003$} & $0.880 \pm 0.002$ & $0.924 \pm 0.001$ & $0.000 \pm 0.000$ \\
V.U.N. $\uparrow$ & $\boldsymbol{0.912 \pm 0.001}$ & $0.851 \pm 0.002$ & \underline{$0.885 \pm 0.003$} & $0.757 \pm 0.002$ & $0.839 \pm 0.003$ & $0.000 \pm 0.000$ \\
FCD $\downarrow$ & $1.173 \pm 0.059$ & \underline{$0.540 \pm 0.008$} & $0.582 \pm 0.010$ & $0.551 \pm 0.008$ & $0.563 \pm 0.012$ & $0.259 \pm 0.005$ \\
SNN $\uparrow$ & $0.385 \pm 0.002$ & $0.457 \pm 0.002$ & \underline{$0.460 \pm 0.002$} & $\boldsymbol{0.463 \pm 0.003}$ & $0.428 \pm 0.001$ & $0.549 \pm 0.001$ \\
Frag $\uparrow$ & $0.991 \pm 0.000$ & \underline{$0.997 \pm 0.000$} & $0.995 \pm 0.000$ & $0.996 \pm 0.001$ & $0.997 \pm 0.000$ & $0.999 \pm 0.000$ \\
Scaff $\uparrow$ & $0.861 \pm 0.006$ & \underline{$0.910 \pm 0.002$} & $0.899 \pm 0.013$ & $\boldsymbol{0.913 \pm 0.002}$ & $0.906 \pm 0.002$ & $0.924 \pm 0.005$ \\
IntDiv $\uparrow$ & $\boldsymbol{0.896 \pm 0.001}$ & \underline{$0.892 \pm 0.001$} & $0.891 \pm 0.002$ & $0.892 \pm 0.001$ & $0.892 \pm 0.001$ & $0.897 \pm 0.001$ \\
MoleculePGD $\downarrow$ & $0.386 \pm 0.004$ & $0.230 \pm 0.009$ & $\boldsymbol{0.156 \pm 0.005}$ & $0.202 \pm 0.008$ & $0.212 \pm 0.004$ & $0.001 \pm 0.001$ \\
FG MMD$^2$ $\downarrow$ & $0.0003 \pm 0.0000$ & $\boldsymbol{0.0002 \pm 0.0000}$ & \underline{$0.0002 \pm 0.0000$} & $0.0002 \pm 0.0000$ & $0.0002 \pm 0.0000$ & $0.0020 \pm 0.0001$ \\
SMARTS MMD$^2$ $\downarrow$ & $0.0011 \pm 0.0001$ & \underline{$0.0005 \pm 0.0000$} & $0.0008 \pm 0.0001$ & $0.0005 \pm 0.0001$ & $\boldsymbol{0.0004 \pm 0.0000}$ & $0.0021 \pm 0.0004$ \\
Ring MMD$^2$ $\downarrow$ & $0.0028 \pm 0.0002$ & $0.0008 \pm 0.0001$ & $\boldsymbol{0.0005 \pm 0.0000}$ & $0.0011 \pm 0.0002$ & $0.0008 \pm 0.0000$ & $0.0019 \pm 0.0001$ \\
BRICS L2 $\downarrow$ & $0.0178 \pm 0.0005$ & \underline{$0.0104 \pm 0.0009$} & $0.0138 \pm 0.0002$ & $0.0112 \pm 0.0014$ & $0.0105 \pm 0.0001$ & $0.0201 \pm 0.0023$ \\
\midrule
\multicolumn{7}{@{}l}{\emph{Direction-normalized change relative to SENT}} \\
Validity $\uparrow$ & $0.00\%$ & \cellcolor{deltared!11}$-1.40\%$ & \cellcolor{deltagreen!62}$+1.49\%$ & \cellcolor{deltared!62}$-7.66\%$ & \cellcolor{deltared!21}$-2.58\%$ & \_ \\
Uniqueness $\uparrow$ & $0.00\%$ & \cellcolor{deltared!38}$-0.18\%$ & \cellcolor{deltared!41}$-0.19\%$ & \cellcolor{deltared!62}$-0.29\%$ & \cellcolor{deltared!24}$-0.11\%$ & \_ \\
Novelty $\uparrow$ & $0.00\%$ & \cellcolor{deltared!33}$-5.30\%$ & \cellcolor{deltared!27}$-4.36\%$ & \cellcolor{deltared!62}$-10.05\%$ & \cellcolor{deltared!34}$-5.56\%$ & \_ \\
V.U.N. $\uparrow$ & $0.00\%$ & \cellcolor{deltared!24}$-6.68\%$ & \cellcolor{deltared!11}$-2.98\%$ & \cellcolor{deltared!62}$-17.01\%$ & \cellcolor{deltared!29}$-8.02\%$ & \_ \\
FCD $\downarrow$ & $0.00\%$ & \cellcolor{deltagreen!57}$+53.97\%$ & \cellcolor{deltagreen!53}$+50.37\%$ & \cellcolor{deltagreen!56}$+53.05\%$ & \cellcolor{deltagreen!55}$+52.01\%$ & \_ \\
SNN $\uparrow$ & $0.00\%$ & \cellcolor{deltagreen!57}$+18.64\%$ & \cellcolor{deltagreen!60}$+19.43\%$ & \cellcolor{deltagreen!62}$+20.14\%$ & \cellcolor{deltagreen!35}$+11.23\%$ & \_ \\
Frag $\uparrow$ & $0.00\%$ & \cellcolor{deltagreen!55}$+0.59\%$ & \cellcolor{deltagreen!36}$+0.38\%$ & \cellcolor{deltagreen!50}$+0.53\%$ & \cellcolor{deltagreen!54}$+0.58\%$ & \_ \\
Scaff $\uparrow$ & $0.00\%$ & \cellcolor{deltagreen!58}$+5.65\%$ & \cellcolor{deltagreen!46}$+4.41\%$ & \cellcolor{deltagreen!62}$+6.00\%$ & \cellcolor{deltagreen!53}$+5.17\%$ & \_ \\
IntDiv $\uparrow$ & $0.00\%$ & \cellcolor{deltared!46}$-0.43\%$ & \cellcolor{deltared!62}$-0.58\%$ & \cellcolor{deltared!52}$-0.49\%$ & \cellcolor{deltared!46}$-0.43\%$ & \_ \\
MoleculePGD $\downarrow$ & $0.00\%$ & \cellcolor{deltagreen!42}$+40.38\%$ & \cellcolor{deltagreen!62}$+59.54\%$ & \cellcolor{deltagreen!50}$+47.75\%$ & \cellcolor{deltagreen!47}$+45.25\%$ & \_ \\
FG MMD$^2$ $\downarrow$ & $0.00\%$ & \cellcolor{deltagreen!46}$+10.97\%$ & \cellcolor{deltagreen!33}$+7.92\%$ & \cellcolor{deltagreen!62}$+14.85\%$ & \cellcolor{deltagreen!47}$+11.29\%$ & \_ \\
SMARTS MMD$^2$ $\downarrow$ & $0.00\%$ & \cellcolor{deltagreen!49}$+53.52\%$ & \cellcolor{deltagreen!28}$+30.87\%$ & \cellcolor{deltagreen!50}$+55.23\%$ & \cellcolor{deltagreen!62}$+68.17\%$ & \_ \\
Ring MMD$^2$ $\downarrow$ & $0.00\%$ & \cellcolor{deltagreen!54}$+72.62\%$ & \cellcolor{deltagreen!62}$+82.87\%$ & \cellcolor{deltagreen!44}$+59.00\%$ & \cellcolor{deltagreen!52}$+69.46\%$ & \_ \\
BRICS L2 $\downarrow$ & $0.00\%$ & \cellcolor{deltagreen!52}$+41.48\%$ & \cellcolor{deltagreen!28}$+22.17\%$ & \cellcolor{deltagreen!46}$+37.28\%$ & \cellcolor{deltagreen!51}$+40.71\%$ & \_ \\
\bottomrule
\end{tabular}
\end{adjustbox}
\par\endgroup

%% file: tables/appendix_coconut_novel.tex
\begingroup
\caption{\textbf{COCONUT 2.0: novel-only unconditional generation.} Mean\,$\pm$\,sample SD; lower panel: improvement over SENT. Shared conventions: Appendix~\ref{app:unconditional_full}.}
\label{tab:appendix_coconut_novel}
\centering
\scriptsize
\setlength{\tabcolsep}{2.2pt}
\renewcommand{\arraystretch}{0.96}
\begin{adjustbox}{max width=\textwidth}
\begin{tabular}{@{}lcccccc@{}}
\toprule
 & \textbf{SENT} & \textbf{HDT-MC-R}$^{*}$ & \textbf{HDT-MC-FR}$^{*}$ & \textbf{HDT-HAC-R}$^{*}$ & \textbf{HDT-HAC-FR}$^{*}$ & \textbf{Training samples} \\
\midrule
\multicolumn{7}{@{}l}{\emph{Absolute performance}} \\
FCD $\downarrow$ & $1.182 \pm 0.040$ & \underline{$0.574 \pm 0.025$} & $\boldsymbol{0.566 \pm 0.013}$ & $0.772 \pm 0.023$ & $0.749 \pm 0.025$ & $0.259 \pm 0.005$ \\
SNN $\uparrow$ & $0.375 \pm 0.001$ & \underline{$0.432 \pm 0.002$} & $\boldsymbol{0.438 \pm 0.001}$ & $0.413 \pm 0.003$ & $0.391 \pm 0.001$ & $0.549 \pm 0.001$ \\
Frag $\uparrow$ & $0.992 \pm 0.000$ & $\boldsymbol{0.997 \pm 0.000}$ & \underline{$0.996 \pm 0.000$} & $0.990 \pm 0.002$ & $0.992 \pm 0.000$ & $0.999 \pm 0.000$ \\
Scaff $\uparrow$ & $0.847 \pm 0.004$ & $0.891 \pm 0.002$ & $0.880 \pm 0.017$ & $\boldsymbol{0.903 \pm 0.002}$ & \underline{$0.899 \pm 0.003$} & $0.924 \pm 0.005$ \\
IntDiv $\uparrow$ & $0.893 \pm 0.001$ & $0.894 \pm 0.001$ & $0.892 \pm 0.001$ & \underline{$0.895 \pm 0.000$} & $\boldsymbol{0.896 \pm 0.001}$ & $0.897 \pm 0.001$ \\
MoleculePGD $\downarrow$ & $0.397 \pm 0.001$ & $0.248 \pm 0.008$ & $\boldsymbol{0.166 \pm 0.007}$ & $0.226 \pm 0.008$ & $0.228 \pm 0.002$ & $0.001 \pm 0.001$ \\
FG MMD$^2$ $\downarrow$ & $0.0021 \pm 0.0000$ & $0.0020 \pm 0.0000$ & $\boldsymbol{0.0002 \pm 0.0000}$ & $0.0021 \pm 0.0000$ & \underline{$0.0003 \pm 0.0000$} & $0.0020 \pm 0.0001$ \\
SMARTS MMD$^2$ $\downarrow$ & $0.0022 \pm 0.0002$ & $0.0021 \pm 0.0001$ & $\boldsymbol{0.0005 \pm 0.0001}$ & $0.0039 \pm 0.0002$ & $0.0018 \pm 0.0000$ & $0.0021 \pm 0.0004$ \\
Ring MMD$^2$ $\downarrow$ & $0.0045 \pm 0.0002$ & $0.0035 \pm 0.0006$ & $\boldsymbol{0.0008 \pm 0.0000}$ & $0.0058 \pm 0.0008$ & \underline{$0.0028 \pm 0.0002$} & $0.0019 \pm 0.0001$ \\
BRICS L2 $\downarrow$ & $0.0244 \pm 0.0005$ & $0.0201 \pm 0.0009$ & $\boldsymbol{0.0120 \pm 0.0003}$ & $0.0285 \pm 0.0006$ & $0.0165 \pm 0.0006$ & $0.0201 \pm 0.0023$ \\
\midrule
\multicolumn{7}{@{}l}{\emph{Direction-normalized change relative to SENT}} \\
FCD $\downarrow$ & $0.00\%$ & \cellcolor{deltagreen!61}$+51.46\%$ & \cellcolor{deltagreen!62}$+52.14\%$ & \cellcolor{deltagreen!41}$+34.65\%$ & \cellcolor{deltagreen!44}$+36.66\%$ & \_ \\
SNN $\uparrow$ & $0.00\%$ & \cellcolor{deltagreen!56}$+15.17\%$ & \cellcolor{deltagreen!62}$+16.73\%$ & \cellcolor{deltagreen!37}$+10.09\%$ & \cellcolor{deltagreen!16}$+4.30\%$ & \_ \\
Frag $\uparrow$ & $0.00\%$ & \cellcolor{deltagreen!62}$+0.44\%$ & \cellcolor{deltagreen!52}$+0.37\%$ & \cellcolor{deltared!62}$-0.22\%$ & \cellcolor{deltared!11}$-0.04\%$ & \_ \\
Scaff $\uparrow$ & $0.00\%$ & \cellcolor{deltagreen!49}$+5.23\%$ & \cellcolor{deltagreen!37}$+3.89\%$ & \cellcolor{deltagreen!62}$+6.59\%$ & \cellcolor{deltagreen!57}$+6.09\%$ & \_ \\
IntDiv $\uparrow$ & $0.00\%$ & \cellcolor{deltagreen!16}$+0.07\%$ & \cellcolor{deltared!62}$-0.11\%$ & \cellcolor{deltagreen!44}$+0.19\%$ & \cellcolor{deltagreen!62}$+0.27\%$ & \_ \\
MoleculePGD $\downarrow$ & $0.00\%$ & \cellcolor{deltagreen!40}$+37.40\%$ & \cellcolor{deltagreen!62}$+58.12\%$ & \cellcolor{deltagreen!46}$+43.12\%$ & \cellcolor{deltagreen!45}$+42.47\%$ & \_ \\
FG MMD$^2$ $\downarrow$ & $0.00\%$ & \cellcolor{deltagreen!2}$+3.38\%$ & \cellcolor{deltagreen!62}$+88.93\%$ & \cellcolor{deltared!62}$-1.63\%$ & \cellcolor{deltagreen!60}$+86.18\%$ & \_ \\
SMARTS MMD$^2$ $\downarrow$ & $0.00\%$ & \cellcolor{deltagreen!1}$+1.72\%$ & \cellcolor{deltagreen!62}$+78.97\%$ & \cellcolor{deltared!62}$-81.88\%$ & \cellcolor{deltagreen!15}$+19.06\%$ & \_ \\
Ring MMD$^2$ $\downarrow$ & $0.00\%$ & \cellcolor{deltagreen!16}$+21.67\%$ & \cellcolor{deltagreen!62}$+82.67\%$ & \cellcolor{deltared!62}$-29.54\%$ & \cellcolor{deltagreen!28}$+37.71\%$ & \_ \\
BRICS L2 $\downarrow$ & $0.00\%$ & \cellcolor{deltagreen!22}$+17.79\%$ & \cellcolor{deltagreen!62}$+50.71\%$ & \cellcolor{deltared!62}$-16.95\%$ & \cellcolor{deltagreen!40}$+32.45\%$ & \_ \\
\bottomrule
\end{tabular}
\end{adjustbox}
\par\endgroup

%% file: tables/appendix_geom_drugs_full.tex
\begingroup
\caption{\textbf{GEOM-Drugs (disjoint): standard unconditional generation.} Mean\,$\pm$\,sample SD; lower panel: improvement over SENT. Shared conventions: Appendix~\ref{app:unconditional_full}.}
\label{tab:appendix_geom_drugs_full}
\centering
\scriptsize
\setlength{\tabcolsep}{2.2pt}
\renewcommand{\arraystretch}{0.96}
\begin{adjustbox}{max width=\textwidth}
\begin{tabular}{@{}lcccccc@{}}
\toprule
 & \textbf{SENT} & \textbf{HDT-MC-R}$^{*}$ & \textbf{HDT-MC-FR}$^{*}$ & \textbf{HDT-HAC-R}$^{*}$ & \textbf{HDT-HAC-FR}$^{*}$ & \textbf{Training samples} \\
\midrule
\multicolumn{7}{@{}l}{\emph{Absolute performance}} \\
Validity $\uparrow$ & $0.949 \pm 0.002$ & $0.929 \pm 0.001$ & $0.955 \pm 0.002$ & $0.913 \pm 0.004$ & \underline{$0.957 \pm 0.002$} & $1.000 \pm 0.000$ \\
Uniqueness $\uparrow$ & $\boldsymbol{1.000 \pm 0.000}$ & $0.995 \pm 0.001$ & \underline{$0.999 \pm 0.000$} & $0.995 \pm 0.001$ & $0.999 \pm 0.000$ & $1.000 \pm 0.000$ \\
Novelty $\uparrow$ & $\boldsymbol{0.997 \pm 0.000}$ & $0.942 \pm 0.003$ & \underline{$0.989 \pm 0.001$} & $0.925 \pm 0.004$ & $0.988 \pm 0.001$ & $0.000 \pm 0.000$ \\
V.U.N. $\uparrow$ & \underline{$0.945 \pm 0.002$} & $0.873 \pm 0.004$ & $0.944 \pm 0.001$ & $0.842 \pm 0.006$ & $0.945 \pm 0.002$ & $0.000 \pm 0.000$ \\
FCD $\downarrow$ & $2.742 \pm 0.052$ & $1.353 \pm 0.027$ & \underline{$1.326 \pm 0.033$} & $\boldsymbol{1.196 \pm 0.018}$ & $1.353 \pm 0.027$ & $0.934 \pm 0.004$ \\
SNN $\uparrow$ & $0.362 \pm 0.002$ & \underline{$0.424 \pm 0.001$} & $0.413 \pm 0.001$ & $\boldsymbol{0.430 \pm 0.002}$ & $0.407 \pm 0.001$ & $0.481 \pm 0.001$ \\
Frag $\uparrow$ & $0.991 \pm 0.000$ & $0.996 \pm 0.000$ & $\boldsymbol{0.997 \pm 0.000}$ & $0.996 \pm 0.000$ & \underline{$0.997 \pm 0.000$} & $0.998 \pm 0.000$ \\
Scaff $\uparrow$ & $0.839 \pm 0.008$ & \underline{$0.857 \pm 0.007$} & $0.856 \pm 0.004$ & $\boldsymbol{0.863 \pm 0.008}$ & $0.846 \pm 0.009$ & $0.869 \pm 0.005$ \\
IntDiv $\uparrow$ & $\boldsymbol{0.879 \pm 0.000}$ & $0.875 \pm 0.001$ & \underline{$0.876 \pm 0.001$} & $0.876 \pm 0.001$ & $0.876 \pm 0.001$ & $0.877 \pm 0.002$ \\
MoleculePGD $\downarrow$ & $0.456 \pm 0.005$ & \underline{$0.314 \pm 0.002$} & $0.316 \pm 0.005$ & $\boldsymbol{0.300 \pm 0.006}$ & $0.326 \pm 0.005$ & $0.239 \pm 0.005$ \\
FG MMD$^2$ $\downarrow$ & $\boldsymbol{0.0021 \pm 0.0000}$ & \underline{$0.0021 \pm 0.0000$} & $0.0021 \pm 0.0000$ & $0.0021 \pm 0.0000$ & $0.0021 \pm 0.0000$ & $0.0020 \pm 0.0000$ \\
SMARTS MMD$^2$ $\downarrow$ & $0.0039 \pm 0.0001$ & \underline{$0.0032 \pm 0.0001$} & $\boldsymbol{0.0031 \pm 0.0004}$ & $0.0032 \pm 0.0002$ & $0.0033 \pm 0.0003$ & $0.0030 \pm 0.0003$ \\
Ring MMD$^2$ $\downarrow$ & $0.0059 \pm 0.0003$ & $0.0028 \pm 0.0004$ & $\boldsymbol{0.0023 \pm 0.0001}$ & \underline{$0.0023 \pm 0.0003$} & $0.0024 \pm 0.0003$ & $0.0021 \pm 0.0003$ \\
BRICS L2 $\downarrow$ & $0.0241 \pm 0.0004$ & $0.0218 \pm 0.0010$ & $0.0221 \pm 0.0018$ & $0.0220 \pm 0.0015$ & $\boldsymbol{0.0210 \pm 0.0013}$ & $0.0199 \pm 0.0004$ \\
\midrule
\multicolumn{7}{@{}l}{\emph{Direction-normalized change relative to SENT}} \\
Validity $\uparrow$ & $0.00\%$ & \cellcolor{deltared!34}$-2.09\%$ & \cellcolor{deltagreen!37}$+0.64\%$ & \cellcolor{deltared!62}$-3.76\%$ & \cellcolor{deltagreen!49}$+0.85\%$ & \_ \\
Uniqueness $\uparrow$ & $0.00\%$ & \cellcolor{deltared!59}$-0.43\%$ & \cellcolor{deltared!6}$-0.04\%$ & \cellcolor{deltared!62}$-0.45\%$ & \cellcolor{deltared!6}$-0.04\%$ & \_ \\
Novelty $\uparrow$ & $0.00\%$ & \cellcolor{deltared!47}$-5.48\%$ & \cellcolor{deltared!7}$-0.78\%$ & \cellcolor{deltared!62}$-7.21\%$ & \cellcolor{deltared!7}$-0.85\%$ & \_ \\
V.U.N. $\uparrow$ & $0.00\%$ & \cellcolor{deltared!44}$-7.70\%$ & \cellcolor{deltared!1}$-0.18\%$ & \cellcolor{deltared!62}$-10.87\%$ & \cellcolor{deltared!0}$-0.03\%$ & \_ \\
FCD $\downarrow$ & $0.00\%$ & \cellcolor{deltagreen!56}$+50.66\%$ & \cellcolor{deltagreen!57}$+51.65\%$ & \cellcolor{deltagreen!62}$+56.39\%$ & \cellcolor{deltagreen!56}$+50.64\%$ & \_ \\
SNN $\uparrow$ & $0.00\%$ & \cellcolor{deltagreen!56}$+16.97\%$ & \cellcolor{deltagreen!46}$+14.02\%$ & \cellcolor{deltagreen!62}$+18.75\%$ & \cellcolor{deltagreen!41}$+12.33\%$ & \_ \\
Frag $\uparrow$ & $0.00\%$ & \cellcolor{deltagreen!55}$+0.48\%$ & \cellcolor{deltagreen!60}$+0.52\%$ & \cellcolor{deltagreen!57}$+0.50\%$ & \cellcolor{deltagreen!62}$+0.54\%$ & \_ \\
Scaff $\uparrow$ & $0.00\%$ & \cellcolor{deltagreen!46}$+2.10\%$ & \cellcolor{deltagreen!45}$+2.06\%$ & \cellcolor{deltagreen!62}$+2.82\%$ & \cellcolor{deltagreen!20}$+0.89\%$ & \_ \\
IntDiv $\uparrow$ & $0.00\%$ & \cellcolor{deltared!62}$-0.51\%$ & \cellcolor{deltared!39}$-0.32\%$ & \cellcolor{deltared!35}$-0.29\%$ & \cellcolor{deltared!40}$-0.33\%$ & \_ \\
MoleculePGD $\downarrow$ & $0.00\%$ & \cellcolor{deltagreen!56}$+31.18\%$ & \cellcolor{deltagreen!56}$+30.88\%$ & \cellcolor{deltagreen!62}$+34.33\%$ & \cellcolor{deltagreen!52}$+28.61\%$ & \_ \\
FG MMD$^2$ $\downarrow$ & $0.00\%$ & \cellcolor{deltagreen!60}$+2.43\%$ & \cellcolor{deltagreen!62}$+2.48\%$ & \cellcolor{deltagreen!62}$+2.50\%$ & \cellcolor{deltagreen!60}$+2.41\%$ & \_ \\
SMARTS MMD$^2$ $\downarrow$ & $0.00\%$ & \cellcolor{deltagreen!56}$+17.98\%$ & \cellcolor{deltagreen!62}$+19.74\%$ & \cellcolor{deltagreen!57}$+18.16\%$ & \cellcolor{deltagreen!45}$+14.33\%$ & \_ \\
Ring MMD$^2$ $\downarrow$ & $0.00\%$ & \cellcolor{deltagreen!53}$+52.26\%$ & \cellcolor{deltagreen!62}$+61.65\%$ & \cellcolor{deltagreen!61}$+61.15\%$ & \cellcolor{deltagreen!59}$+58.54\%$ & \_ \\
BRICS L2 $\downarrow$ & $0.00\%$ & \cellcolor{deltagreen!46}$+9.55\%$ & \cellcolor{deltagreen!40}$+8.33\%$ & \cellcolor{deltagreen!41}$+8.58\%$ & \cellcolor{deltagreen!62}$+12.98\%$ & \_ \\
\bottomrule
\end{tabular}
\end{adjustbox}
\par\endgroup

%% file: tables/appendix_geom_drugs_novel.tex
\begingroup
\caption{\textbf{GEOM-Drugs (disjoint): novel-only unconditional generation.} Mean\,$\pm$\,sample SD; lower panel: improvement over SENT. Shared conventions: Appendix~\ref{app:unconditional_full}.}
\label{tab:appendix_geom_drugs_novel}
\centering
\scriptsize
\setlength{\tabcolsep}{2.2pt}
\renewcommand{\arraystretch}{0.96}
\begin{adjustbox}{max width=\textwidth}
\begin{tabular}{@{}lcccccc@{}}
\toprule
 & \textbf{SENT} & \textbf{HDT-MC-R}$^{*}$ & \textbf{HDT-MC-FR}$^{*}$ & \textbf{HDT-HAC-R}$^{*}$ & \textbf{HDT-HAC-FR}$^{*}$ & \textbf{Training samples} \\
\midrule
\multicolumn{7}{@{}l}{\emph{Absolute performance}} \\
FCD $\downarrow$ & $2.769 \pm 0.065$ & $1.380 \pm 0.029$ & \underline{$1.360 \pm 0.038$} & $\boldsymbol{1.265 \pm 0.023}$ & $1.397 \pm 0.017$ & $0.934 \pm 0.004$ \\
SNN $\uparrow$ & $0.362 \pm 0.002$ & \underline{$0.418 \pm 0.001$} & $0.411 \pm 0.001$ & $\boldsymbol{0.422 \pm 0.001}$ & $0.405 \pm 0.001$ & $0.481 \pm 0.001$ \\
Frag $\uparrow$ & $0.991 \pm 0.001$ & $0.996 \pm 0.000$ & $\boldsymbol{0.997 \pm 0.000}$ & $0.996 \pm 0.000$ & \underline{$0.997 \pm 0.000$} & $0.998 \pm 0.000$ \\
Scaff $\uparrow$ & $0.830 \pm 0.011$ & $0.842 \pm 0.010$ & \underline{$0.847 \pm 0.003$} & $\boldsymbol{0.857 \pm 0.006}$ & $0.836 \pm 0.013$ & $0.869 \pm 0.005$ \\
IntDiv $\uparrow$ & $\boldsymbol{0.880 \pm 0.000}$ & \underline{$0.875 \pm 0.001$} & $0.875 \pm 0.001$ & $0.874 \pm 0.000$ & $0.875 \pm 0.001$ & $0.877 \pm 0.002$ \\
MoleculePGD $\downarrow$ & $0.458 \pm 0.007$ & \underline{$0.311 \pm 0.003$} & $0.315 \pm 0.006$ & $\boldsymbol{0.306 \pm 0.005}$ & $0.325 \pm 0.002$ & $0.239 \pm 0.005$ \\
FG MMD$^2$ $\downarrow$ & $\boldsymbol{0.0021 \pm 0.0000}$ & \underline{$0.0021 \pm 0.0000$} & $0.0021 \pm 0.0000$ & $0.0021 \pm 0.0000$ & $0.0021 \pm 0.0000$ & $0.0020 \pm 0.0000$ \\
SMARTS MMD$^2$ $\downarrow$ & $0.0039 \pm 0.0001$ & $\boldsymbol{0.0032 \pm 0.0000}$ & \underline{$0.0032 \pm 0.0004$} & $0.0034 \pm 0.0004$ & $0.0034 \pm 0.0003$ & $0.0030 \pm 0.0003$ \\
Ring MMD$^2$ $\downarrow$ & $0.0059 \pm 0.0003$ & $0.0029 \pm 0.0004$ & $\boldsymbol{0.0022 \pm 0.0000}$ & \underline{$0.0025 \pm 0.0004$} & $0.0026 \pm 0.0004$ & $0.0021 \pm 0.0003$ \\
BRICS L2 $\downarrow$ & $0.0241 \pm 0.0004$ & $0.0219 \pm 0.0009$ & $0.0220 \pm 0.0018$ & $0.0230 \pm 0.0018$ & $\boldsymbol{0.0210 \pm 0.0012}$ & $0.0199 \pm 0.0004$ \\
\midrule
\multicolumn{7}{@{}l}{\emph{Direction-normalized change relative to SENT}} \\
FCD $\downarrow$ & $0.00\%$ & \cellcolor{deltagreen!57}$+50.15\%$ & \cellcolor{deltagreen!58}$+50.89\%$ & \cellcolor{deltagreen!62}$+54.30\%$ & \cellcolor{deltagreen!57}$+49.54\%$ & \_ \\
SNN $\uparrow$ & $0.00\%$ & \cellcolor{deltagreen!58}$+15.67\%$ & \cellcolor{deltagreen!51}$+13.78\%$ & \cellcolor{deltagreen!62}$+16.63\%$ & \cellcolor{deltagreen!44}$+11.90\%$ & \_ \\
Frag $\uparrow$ & $0.00\%$ & \cellcolor{deltagreen!53}$+0.46\%$ & \cellcolor{deltagreen!60}$+0.52\%$ & \cellcolor{deltagreen!55}$+0.48\%$ & \cellcolor{deltagreen!62}$+0.54\%$ & \_ \\
Scaff $\uparrow$ & $0.00\%$ & \cellcolor{deltagreen!27}$+1.41\%$ & \cellcolor{deltagreen!37}$+1.96\%$ & \cellcolor{deltagreen!62}$+3.25\%$ & \cellcolor{deltagreen!14}$+0.72\%$ & \_ \\
IntDiv $\uparrow$ & $0.00\%$ & \cellcolor{deltared!55}$-0.59\%$ & \cellcolor{deltared!46}$-0.49\%$ & \cellcolor{deltared!62}$-0.66\%$ & \cellcolor{deltared!55}$-0.59\%$ & \_ \\
MoleculePGD $\downarrow$ & $0.00\%$ & \cellcolor{deltagreen!60}$+32.02\%$ & \cellcolor{deltagreen!58}$+31.20\%$ & \cellcolor{deltagreen!62}$+33.07\%$ & \cellcolor{deltagreen!54}$+29.01\%$ & \_ \\
FG MMD$^2$ $\downarrow$ & $0.00\%$ & \cellcolor{deltagreen!62}$+2.37\%$ & \cellcolor{deltagreen!61}$+2.32\%$ & \cellcolor{deltagreen!43}$+1.63\%$ & \cellcolor{deltagreen!61}$+2.33\%$ & \_ \\
SMARTS MMD$^2$ $\downarrow$ & $0.00\%$ & \cellcolor{deltagreen!56}$+16.99\%$ & \cellcolor{deltagreen!62}$+18.74\%$ & \cellcolor{deltagreen!43}$+12.86\%$ & \cellcolor{deltagreen!43}$+13.07\%$ & \_ \\
Ring MMD$^2$ $\downarrow$ & $0.00\%$ & \cellcolor{deltagreen!50}$+50.57\%$ & \cellcolor{deltagreen!62}$+62.58\%$ & \cellcolor{deltagreen!57}$+58.00\%$ & \cellcolor{deltagreen!56}$+56.81\%$ & \_ \\
BRICS L2 $\downarrow$ & $0.00\%$ & \cellcolor{deltagreen!44}$+9.27\%$ & \cellcolor{deltagreen!42}$+8.77\%$ & \cellcolor{deltagreen!23}$+4.71\%$ & \cellcolor{deltagreen!62}$+12.87\%$ & \_ \\
\bottomrule
\end{tabular}
\end{adjustbox}
\par\endgroup

%% file: tables/appendix_macrocycle_full.tex
\begingroup
\caption{\textbf{Macrocycle: standard unconditional generation.} Mean\,$\pm$\,sample SD; lower panel: improvement over SENT. Shared conventions: Appendix~\ref{app:unconditional_full}.}
\label{tab:appendix_macrocycle_full}
\centering
\scriptsize
\setlength{\tabcolsep}{2.2pt}
\renewcommand{\arraystretch}{0.96}
\begin{adjustbox}{max width=\textwidth}
\begin{tabular}{@{}lcccccc@{}}
\toprule
 & \textbf{SENT} & \textbf{HDT-MC-R}$^{*}$ & \textbf{HDT-MC-FR}$^{*}$ & \textbf{HDT-HAC-R}$^{*}$ & \textbf{HDT-HAC-FR}$^{*}$ & \textbf{Training samples} \\
\midrule
\multicolumn{7}{@{}l}{\emph{Absolute performance}} \\
Validity $\uparrow$ & $\boldsymbol{0.855 \pm 0.004}$ & $0.846 \pm 0.003$ & \underline{$0.854 \pm 0.002$} & $0.819 \pm 0.009$ & $0.790 \pm 0.003$ & $1.000 \pm 0.000$ \\
Uniqueness $\uparrow$ & $\boldsymbol{1.000 \pm 0.000}$ & $0.966 \pm 0.000$ & \underline{$0.997 \pm 0.001$} & $0.970 \pm 0.003$ & $0.991 \pm 0.002$ & $1.000 \pm 0.000$ \\
Novelty $\uparrow$ & $\boldsymbol{0.999 \pm 0.000}$ & $0.657 \pm 0.011$ & \underline{$0.987 \pm 0.001$} & $0.658 \pm 0.008$ & $0.967 \pm 0.001$ & $0.000 \pm 0.000$ \\
V.U.N. $\uparrow$ & $\boldsymbol{0.854 \pm 0.004}$ & $0.552 \pm 0.009$ & \underline{$0.842 \pm 0.002$} & $0.537 \pm 0.011$ & $0.762 \pm 0.004$ & $0.000 \pm 0.000$ \\
FCD $\downarrow$ & $2.711 \pm 0.013$ & \underline{$1.238 \pm 0.045$} & $1.466 \pm 0.013$ & $\boldsymbol{1.064 \pm 0.033}$ & $1.501 \pm 0.046$ & $0.726 \pm 0.015$ \\
SNN $\uparrow$ & $0.352 \pm 0.001$ & \underline{$0.633 \pm 0.006$} & $0.491 \pm 0.002$ & $\boldsymbol{0.650 \pm 0.003}$ & $0.522 \pm 0.001$ & $0.718 \pm 0.000$ \\
Frag $\uparrow$ & $0.963 \pm 0.001$ & $\boldsymbol{0.994 \pm 0.001}$ & \underline{$0.994 \pm 0.000$} & $0.994 \pm 0.001$ & $0.987 \pm 0.001$ & $0.997 \pm 0.001$ \\
Scaff $\uparrow$ & $0.035 \pm 0.001$ & \underline{$0.497 \pm 0.003$} & $0.276 \pm 0.003$ & $\boldsymbol{0.537 \pm 0.011}$ & $0.368 \pm 0.005$ & $0.608 \pm 0.005$ \\
IntDiv $\uparrow$ & $\boldsymbol{0.869 \pm 0.000}$ & $0.855 \pm 0.001$ & \underline{$0.858 \pm 0.002$} & $0.854 \pm 0.000$ & $0.851 \pm 0.002$ & $0.863 \pm 0.003$ \\
MoleculePGD $\downarrow$ & $0.639 \pm 0.003$ & \underline{$0.124 \pm 0.012$} & $0.232 \pm 0.002$ & $\boldsymbol{0.085 \pm 0.002}$ & $0.204 \pm 0.008$ & $0.003 \pm 0.001$ \\
FG MMD$^2$ $\downarrow$ & $0.0023 \pm 0.0000$ & \underline{$0.0021 \pm 0.0000$} & $0.0022 \pm 0.0000$ & $\boldsymbol{0.0020 \pm 0.0000}$ & $0.0021 \pm 0.0000$ & $0.0020 \pm 0.0000$ \\
SMARTS MMD$^2$ $\downarrow$ & $0.0031 \pm 0.0001$ & \underline{$0.0022 \pm 0.0001$} & $0.0025 \pm 0.0002$ & $\boldsymbol{0.0021 \pm 0.0001}$ & $0.0023 \pm 0.0001$ & $0.0020 \pm 0.0000$ \\
Ring MMD$^2$ $\downarrow$ & $0.0048 \pm 0.0003$ & \underline{$0.0025 \pm 0.0002$} & $0.0029 \pm 0.0002$ & $\boldsymbol{0.0020 \pm 0.0002}$ & $0.0027 \pm 0.0002$ & $0.0019 \pm 0.0003$ \\
BRICS L2 $\downarrow$ & $0.0384 \pm 0.0008$ & \underline{$0.0186 \pm 0.0011$} & $0.0191 \pm 0.0012$ & $\boldsymbol{0.0184 \pm 0.0010}$ & $0.0238 \pm 0.0018$ & $0.0169 \pm 0.0012$ \\
\midrule
\multicolumn{7}{@{}l}{\emph{Direction-normalized change relative to SENT}} \\
Validity $\uparrow$ & $0.00\%$ & \cellcolor{deltared!9}$-1.10\%$ & \cellcolor{deltared!1}$-0.17\%$ & \cellcolor{deltared!34}$-4.17\%$ & \cellcolor{deltared!62}$-7.62\%$ & \_ \\
Uniqueness $\uparrow$ & $0.00\%$ & \cellcolor{deltared!62}$-3.38\%$ & \cellcolor{deltared!6}$-0.32\%$ & \cellcolor{deltared!54}$-2.94\%$ & \cellcolor{deltared!17}$-0.90\%$ & \_ \\
Novelty $\uparrow$ & $0.00\%$ & \cellcolor{deltared!62}$-34.22\%$ & \cellcolor{deltared!2}$-1.22\%$ & \cellcolor{deltared!62}$-34.16\%$ & \cellcolor{deltared!6}$-3.27\%$ & \_ \\
V.U.N. $\uparrow$ & $0.00\%$ & \cellcolor{deltared!59}$-35.39\%$ & \cellcolor{deltared!2}$-1.49\%$ & \cellcolor{deltared!62}$-37.20\%$ & \cellcolor{deltared!18}$-10.85\%$ & \_ \\
FCD $\downarrow$ & $0.00\%$ & \cellcolor{deltagreen!55}$+54.34\%$ & \cellcolor{deltagreen!47}$+45.91\%$ & \cellcolor{deltagreen!62}$+60.75\%$ & \cellcolor{deltagreen!46}$+44.62\%$ & \_ \\
SNN $\uparrow$ & $0.00\%$ & \cellcolor{deltagreen!58}$+79.74\%$ & \cellcolor{deltagreen!29}$+39.48\%$ & \cellcolor{deltagreen!62}$+84.67\%$ & \cellcolor{deltagreen!35}$+48.11\%$ & \_ \\
Frag $\uparrow$ & $0.00\%$ & \cellcolor{deltagreen!62}$+3.26\%$ & \cellcolor{deltagreen!62}$+3.25\%$ & \cellcolor{deltagreen!61}$+3.23\%$ & \cellcolor{deltagreen!47}$+2.47\%$ & \_ \\
Scaff $\uparrow$ & $0.00\%$ & \cellcolor{deltagreen!57}$+1314.18\%$ & \cellcolor{deltagreen!30}$+684.24\%$ & \cellcolor{deltagreen!62}$+1428.84\%$ & \cellcolor{deltagreen!41}$+947.58\%$ & \_ \\
IntDiv $\uparrow$ & $0.00\%$ & \cellcolor{deltared!48}$-1.59\%$ & \cellcolor{deltared!38}$-1.27\%$ & \cellcolor{deltared!51}$-1.69\%$ & \cellcolor{deltared!62}$-2.07\%$ & \_ \\
MoleculePGD $\downarrow$ & $0.00\%$ & \cellcolor{deltagreen!58}$+80.59\%$ & \cellcolor{deltagreen!46}$+63.79\%$ & \cellcolor{deltagreen!62}$+86.64\%$ & \cellcolor{deltagreen!49}$+68.12\%$ & \_ \\
FG MMD$^2$ $\downarrow$ & $0.00\%$ & \cellcolor{deltagreen!49}$+8.84\%$ & \cellcolor{deltagreen!24}$+4.26\%$ & \cellcolor{deltagreen!62}$+11.19\%$ & \cellcolor{deltagreen!35}$+6.33\%$ & \_ \\
SMARTS MMD$^2$ $\downarrow$ & $0.00\%$ & \cellcolor{deltagreen!56}$+28.05\%$ & \cellcolor{deltagreen!36}$+17.96\%$ & \cellcolor{deltagreen!62}$+31.21\%$ & \cellcolor{deltagreen!48}$+23.99\%$ & \_ \\
Ring MMD$^2$ $\downarrow$ & $0.00\%$ & \cellcolor{deltagreen!52}$+47.77\%$ & \cellcolor{deltagreen!43}$+39.70\%$ & \cellcolor{deltagreen!62}$+57.28\%$ & \cellcolor{deltagreen!47}$+43.16\%$ & \_ \\
BRICS L2 $\downarrow$ & $0.00\%$ & \cellcolor{deltagreen!61}$+51.50\%$ & \cellcolor{deltagreen!60}$+50.17\%$ & \cellcolor{deltagreen!62}$+51.97\%$ & \cellcolor{deltagreen!45}$+38.09\%$ & \_ \\
\bottomrule
\end{tabular}
\end{adjustbox}
\par\endgroup

%% file: tables/appendix_macrocycle_novel.tex
\begingroup
\caption{\textbf{Macrocycle: novel-only unconditional generation.} Mean\,$\pm$\,sample SD; lower panel: improvement over SENT. Shared conventions: Appendix~\ref{app:unconditional_full}.}
\label{tab:appendix_macrocycle_novel}
\centering
\scriptsize
\setlength{\tabcolsep}{2.2pt}
\renewcommand{\arraystretch}{0.96}
\begin{adjustbox}{max width=\textwidth}
\begin{tabular}{@{}lcccccc@{}}
\toprule
 & \textbf{SENT} & \textbf{HDT-MC-R}$^{*}$ & \textbf{HDT-MC-FR}$^{*}$ & \textbf{HDT-HAC-R}$^{*}$ & \textbf{HDT-HAC-FR}$^{*}$ & \textbf{Training samples} \\
\midrule
\multicolumn{7}{@{}l}{\emph{Absolute performance}} \\
FCD $\downarrow$ & $2.880 \pm 0.043$ & $\boldsymbol{1.169 \pm 0.044}$ & $1.779 \pm 0.068$ & \underline{$1.300 \pm 0.039$} & $1.798 \pm 0.057$ & $0.726 \pm 0.015$ \\
SNN $\uparrow$ & $0.351 \pm 0.001$ & \underline{$0.558 \pm 0.003$} & $0.489 \pm 0.003$ & $\boldsymbol{0.577 \pm 0.001}$ & $0.511 \pm 0.002$ & $0.718 \pm 0.000$ \\
Frag $\uparrow$ & $0.962 \pm 0.001$ & $\boldsymbol{0.994 \pm 0.000}$ & \underline{$0.992 \pm 0.000$} & $0.989 \pm 0.001$ & $0.983 \pm 0.002$ & $0.997 \pm 0.001$ \\
Scaff $\uparrow$ & $0.023 \pm 0.005$ & \underline{$0.326 \pm 0.017$} & $0.185 \pm 0.008$ & $\boldsymbol{0.356 \pm 0.024}$ & $0.193 \pm 0.018$ & $0.608 \pm 0.005$ \\
IntDiv $\uparrow$ & $\boldsymbol{0.869 \pm 0.002}$ & \underline{$0.867 \pm 0.001$} & $0.853 \pm 0.003$ & $0.856 \pm 0.001$ & $0.849 \pm 0.001$ & $0.863 \pm 0.003$ \\
MoleculePGD $\downarrow$ & $0.641 \pm 0.004$ & \underline{$0.165 \pm 0.012$} & $0.237 \pm 0.004$ & $\boldsymbol{0.133 \pm 0.017}$ & $0.223 \pm 0.013$ & $0.003 \pm 0.001$ \\
FG MMD$^2$ $\downarrow$ & $0.0023 \pm 0.0000$ & $\boldsymbol{0.0021 \pm 0.0000}$ & $0.0022 \pm 0.0000$ & \underline{$0.0021 \pm 0.0000$} & $0.0022 \pm 0.0000$ & $0.0020 \pm 0.0000$ \\
SMARTS MMD$^2$ $\downarrow$ & $0.0031 \pm 0.0001$ & $\boldsymbol{0.0022 \pm 0.0000}$ & $0.0025 \pm 0.0001$ & \underline{$0.0022 \pm 0.0001$} & $0.0024 \pm 0.0000$ & $0.0020 \pm 0.0000$ \\
Ring MMD$^2$ $\downarrow$ & $0.0048 \pm 0.0003$ & $0.0031 \pm 0.0002$ & $0.0029 \pm 0.0002$ & $\boldsymbol{0.0022 \pm 0.0000}$ & $0.0028 \pm 0.0001$ & $0.0019 \pm 0.0003$ \\
BRICS L2 $\downarrow$ & $0.0384 \pm 0.0008$ & \underline{$0.0200 \pm 0.0001$} & $\boldsymbol{0.0192 \pm 0.0014}$ & $0.0213 \pm 0.0007$ & $0.0254 \pm 0.0008$ & $0.0169 \pm 0.0012$ \\
\midrule
\multicolumn{7}{@{}l}{\emph{Direction-normalized change relative to SENT}} \\
FCD $\downarrow$ & $0.00\%$ & \cellcolor{deltagreen!62}$+59.41\%$ & \cellcolor{deltagreen!40}$+38.23\%$ & \cellcolor{deltagreen!57}$+54.85\%$ & \cellcolor{deltagreen!39}$+37.58\%$ & \_ \\
SNN $\uparrow$ & $0.00\%$ & \cellcolor{deltagreen!57}$+59.00\%$ & \cellcolor{deltagreen!38}$+39.22\%$ & \cellcolor{deltagreen!62}$+64.25\%$ & \cellcolor{deltagreen!44}$+45.61\%$ & \_ \\
Frag $\uparrow$ & $0.00\%$ & \cellcolor{deltagreen!62}$+3.29\%$ & \cellcolor{deltagreen!59}$+3.11\%$ & \cellcolor{deltagreen!52}$+2.77\%$ & \cellcolor{deltagreen!41}$+2.16\%$ & \_ \\
Scaff $\uparrow$ & $0.00\%$ & \cellcolor{deltagreen!56}$+1317.74\%$ & \cellcolor{deltagreen!30}$+703.84\%$ & \cellcolor{deltagreen!62}$+1449.10\%$ & \cellcolor{deltagreen!32}$+741.09\%$ & \_ \\
IntDiv $\uparrow$ & $0.00\%$ & \cellcolor{deltared!7}$-0.25\%$ & \cellcolor{deltared!51}$-1.85\%$ & \cellcolor{deltared!40}$-1.46\%$ & \cellcolor{deltared!62}$-2.26\%$ & \_ \\
MoleculePGD $\downarrow$ & $0.00\%$ & \cellcolor{deltagreen!58}$+74.30\%$ & \cellcolor{deltagreen!49}$+62.95\%$ & \cellcolor{deltagreen!62}$+79.28\%$ & \cellcolor{deltagreen!51}$+65.28\%$ & \_ \\
FG MMD$^2$ $\downarrow$ & $0.00\%$ & \cellcolor{deltagreen!62}$+9.30\%$ & \cellcolor{deltagreen!28}$+4.25\%$ & \cellcolor{deltagreen!53}$+7.99\%$ & \cellcolor{deltagreen!35}$+5.20\%$ & \_ \\
SMARTS MMD$^2$ $\downarrow$ & $0.00\%$ & \cellcolor{deltagreen!58}$+27.46\%$ & \cellcolor{deltagreen!36}$+17.10\%$ & \cellcolor{deltagreen!62}$+29.58\%$ & \cellcolor{deltagreen!45}$+21.62\%$ & \_ \\
Ring MMD$^2$ $\downarrow$ & $0.00\%$ & \cellcolor{deltagreen!40}$+35.25\%$ & \cellcolor{deltagreen!44}$+38.40\%$ & \cellcolor{deltagreen!62}$+53.98\%$ & \cellcolor{deltagreen!48}$+42.06\%$ & \_ \\
BRICS L2 $\downarrow$ & $0.00\%$ & \cellcolor{deltagreen!60}$+47.97\%$ & \cellcolor{deltagreen!62}$+49.93\%$ & \cellcolor{deltagreen!55}$+44.49\%$ & \cellcolor{deltagreen!42}$+33.78\%$ & \_ \\
\bottomrule
\end{tabular}
\end{adjustbox}
\par\endgroup

%% file: tables/appendix_legacy_variants.tex
\begin{table}[h]
\caption{\textbf{Declarative and typed hierarchy on GuacaMol and MOSES.} Reference points for the explicitness axis (Section~\ref{sec:results:design}), not entries in the main matrix. These configurations predate the current protocol and do not isolate hierarchy explicitness. The GuacaMol SENT reference is unchanged, whereas the MOSES reference differs; cross-protocol comparisons are supporting evidence, not controlled causal estimates.}
\label{tab:appendix_legacy_variants}
\centering
\scriptsize
\setlength{\tabcolsep}{2.2pt}
\renewcommand{\arraystretch}{0.96}
\begin{adjustbox}{max width=\textwidth}
\begin{tabular}{@{}lcccccccc@{}}
\toprule
 & \multicolumn{4}{c}{\textbf{GuacaMol}} & \multicolumn{4}{c}{\textbf{MOSES}} \\
\cmidrule(lr){2-5}\cmidrule(lr){6-9}
 & \textbf{SENT} & \makecell{\textbf{HSENT-}\\\textbf{HAC-R}$^{*}$} & \makecell{\textbf{HSENT-}\\\textbf{MC-R}$^{*}$} & \textbf{HDTC-R}$^{*}$
 & \textbf{SENT} & \makecell{\textbf{HSENT-}\\\textbf{HAC-R}$^{*}$} & \makecell{\textbf{HSENT-}\\\textbf{MC-R}$^{*}$} & \textbf{HDTC-R}$^{*}$ \\
\midrule
\multicolumn{9}{@{}l}{\emph{Absolute performance}} \\
Validity $\uparrow$ & $\boldsymbol{0.967 \pm 0.001}$ & \underline{$0.898 \pm 0.004$} & $0.872 \pm 0.002$ & $0.850 \pm 0.003$ & $\boldsymbol{0.977 \pm 0.001}$ & \underline{$0.958 \pm 0.001$} & $0.934 \pm 0.001$ & $0.938 \pm 0.001$ \\
Uniqueness $\uparrow$ & $\boldsymbol{1.000 \pm 0.000}$ & \underline{$1.000 \pm 0.000$} & $1.000 \pm 0.000$ & $1.000 \pm 0.000$ & $\boldsymbol{1.000 \pm 0.000}$ & \underline{$0.998 \pm 0.000$} & $0.996 \pm 0.000$ & $0.998 \pm 0.000$ \\
Novelty $\uparrow$ & $\boldsymbol{0.998 \pm 0.001}$ & $0.989 \pm 0.001$ & \underline{$0.994 \pm 0.001$} & $0.994 \pm 0.000$ & $\boldsymbol{0.990 \pm 0.002}$ & $0.938 \pm 0.002$ & $0.951 \pm 0.001$ & \underline{$0.963 \pm 0.002$} \\
V.U.N. $\uparrow$ & $\boldsymbol{0.965 \pm 0.000}$ & \underline{$0.889 \pm 0.005$} & $0.867 \pm 0.003$ & $0.845 \pm 0.003$ & $\boldsymbol{0.967 \pm 0.001}$ & $0.897 \pm 0.002$ & $0.885 \pm 0.000$ & \underline{$0.903 \pm 0.002$} \\
FCD $\downarrow$ & $1.799 \pm 0.045$ & \underline{$1.371 \pm 0.005$} & $2.578 \pm 0.054$ & $\boldsymbol{1.341 \pm 0.037}$ & $0.696 \pm 0.044$ & $\boldsymbol{0.225 \pm 0.003}$ & $1.221 \pm 0.018$ & \underline{$0.245 \pm 0.013$} \\
SNN $\uparrow$ & $0.410 \pm 0.001$ & $\boldsymbol{0.464 \pm 0.003}$ & \underline{$0.457 \pm 0.002$} & $0.427 \pm 0.002$ & $0.506 \pm 0.002$ & $\boldsymbol{0.584 \pm 0.000}$ & \underline{$0.581 \pm 0.001$} & $0.560 \pm 0.001$ \\
Frag $\uparrow$ & $0.994 \pm 0.000$ & $\boldsymbol{0.996 \pm 0.000}$ & \underline{$0.995 \pm 0.000$} & $0.988 \pm 0.000$ & \underline{$0.998 \pm 0.000$} & $\boldsymbol{0.999 \pm 0.000}$ & $0.996 \pm 0.000$ & $0.996 \pm 0.000$ \\
Scaff $\uparrow$ & $0.890 \pm 0.007$ & \underline{$0.930 \pm 0.006$} & $\boldsymbol{0.955 \pm 0.001}$ & $0.920 \pm 0.010$ & $0.962 \pm 0.004$ & $\boldsymbol{0.991 \pm 0.000}$ & $0.981 \pm 0.000$ & \underline{$0.988 \pm 0.001$} \\
IntDiv $\uparrow$ & $\boldsymbol{0.886 \pm 0.001}$ & $0.884 \pm 0.001$ & $0.875 \pm 0.001$ & \underline{$0.885 \pm 0.001$} & $\boldsymbol{0.867 \pm 0.002}$ & $0.862 \pm 0.001$ & $0.858 \pm 0.001$ & \underline{$0.865 \pm 0.002$} \\
MoleculePGD $\downarrow$ & \underline{$0.309 \pm 0.005$} & $\boldsymbol{0.241 \pm 0.006}$ & $0.473 \pm 0.011$ & $0.323 \pm 0.005$ & $\boldsymbol{0.217 \pm 0.005}$ & \underline{$0.515 \pm 0.006$} & $0.563 \pm 0.012$ & $0.517 \pm 0.003$ \\
KL score $\uparrow$ & $0.938 \pm 0.010$ & \underline{$0.948 \pm 0.004$} & $0.857 \pm 0.006$ & $\boldsymbol{0.962 \pm 0.001}$ & \_ & \_ & \_ & \_ \\
FG MMD$^2$ $\downarrow$ & $\boldsymbol{0.0020 \pm 0.0000}$ & \underline{$0.0020 \pm 0.0000$} & $0.0022 \pm 0.0000$ & $0.0020 \pm 0.0000$ & $\boldsymbol{0.0030 \pm 0.0001}$ & \underline{$0.0031 \pm 0.0001$} & $0.0031 \pm 0.0000$ & $0.0031 \pm 0.0001$ \\
SMARTS MMD$^2$ $\downarrow$ & \underline{$0.0023 \pm 0.0001$} & $0.0026 \pm 0.0001$ & $0.0037 \pm 0.0001$ & $\boldsymbol{0.0021 \pm 0.0001}$ & $0.0145 \pm 0.0007$ & $0.0119 \pm 0.0015$ & $\boldsymbol{0.0087 \pm 0.0007}$ & \underline{$0.0104 \pm 0.0011$} \\
Ring MMD$^2$ $\downarrow$ & $\boldsymbol{0.0036 \pm 0.0003}$ & $0.0045 \pm 0.0002$ & $0.0127 \pm 0.0017$ & \underline{$0.0039 \pm 0.0002$} & $0.0201 \pm 0.0018$ & \underline{$0.0158 \pm 0.0011$} & $0.0161 \pm 0.0010$ & $\boldsymbol{0.0128 \pm 0.0001}$ \\
BRICS L2 $\downarrow$ & $0.0243 \pm 0.0018$ & $\boldsymbol{0.0211 \pm 0.0019}$ & \underline{$0.0242 \pm 0.0021$} & $0.0286 \pm 0.0013$ & $0.0521 \pm 0.0020$ & \underline{$0.0501 \pm 0.0020$} & $\boldsymbol{0.0458 \pm 0.0012}$ & $0.0504 \pm 0.0019$ \\
\midrule
\multicolumn{9}{@{}l}{\emph{Direction-normalized change relative to SENT}} \\
Validity $\uparrow$ & $0.00\%$ & \cellcolor{deltared!37}$-7.14\%$ & \cellcolor{deltared!50}$-9.82\%$ & \cellcolor{deltared!62}$-12.10\%$ & $0.00\%$ & \cellcolor{deltared!27}$-1.94\%$ & \cellcolor{deltared!62}$-4.40\%$ & \cellcolor{deltared!56}$-3.99\%$ \\
Uniqueness $\uparrow$ & $0.00\%$ & $0.00\%$ & $0.00\%$ & $0.00\%$ & $0.00\%$ & \cellcolor{deltared!31}$-0.20\%$ & \cellcolor{deltared!62}$-0.40\%$ & \cellcolor{deltared!31}$-0.20\%$ \\
Novelty $\uparrow$ & $0.00\%$ & \cellcolor{deltared!62}$-0.90\%$ & \cellcolor{deltared!28}$-0.40\%$ & \cellcolor{deltared!28}$-0.40\%$ & $0.00\%$ & \cellcolor{deltared!62}$-5.25\%$ & \cellcolor{deltared!47}$-3.94\%$ & \cellcolor{deltared!32}$-2.73\%$ \\
V.U.N. $\uparrow$ & $0.00\%$ & \cellcolor{deltared!39}$-7.88\%$ & \cellcolor{deltared!51}$-10.16\%$ & \cellcolor{deltared!62}$-12.44\%$ & $0.00\%$ & \cellcolor{deltared!53}$-7.24\%$ & \cellcolor{deltared!62}$-8.48\%$ & \cellcolor{deltared!48}$-6.62\%$ \\
FCD $\downarrow$ & $0.00\%$ & \cellcolor{deltagreen!58}$+23.79\%$ & \cellcolor{deltared!62}$-43.30\%$ & \cellcolor{deltagreen!62}$+25.46\%$ & $0.00\%$ & \cellcolor{deltagreen!62}$+67.67\%$ & \cellcolor{deltared!62}$-75.43\%$ & \cellcolor{deltagreen!59}$+64.80\%$ \\
SNN $\uparrow$ & $0.00\%$ & \cellcolor{deltagreen!62}$+13.17\%$ & \cellcolor{deltagreen!54}$+11.46\%$ & \cellcolor{deltagreen!20}$+4.15\%$ & $0.00\%$ & \cellcolor{deltagreen!62}$+15.42\%$ & \cellcolor{deltagreen!60}$+14.82\%$ & \cellcolor{deltagreen!43}$+10.67\%$ \\
Frag $\uparrow$ & $0.00\%$ & \cellcolor{deltagreen!62}$+0.20\%$ & \cellcolor{deltagreen!31}$+0.10\%$ & \cellcolor{deltared!62}$-0.60\%$ & $0.00\%$ & \cellcolor{deltagreen!62}$+0.10\%$ & \cellcolor{deltared!62}$-0.20\%$ & \cellcolor{deltared!62}$-0.20\%$ \\
Scaff $\uparrow$ & $0.00\%$ & \cellcolor{deltagreen!38}$+4.49\%$ & \cellcolor{deltagreen!62}$+7.30\%$ & \cellcolor{deltagreen!29}$+3.37\%$ & $0.00\%$ & \cellcolor{deltagreen!62}$+3.01\%$ & \cellcolor{deltagreen!41}$+1.98\%$ & \cellcolor{deltagreen!56}$+2.70\%$ \\
IntDiv $\uparrow$ & $0.00\%$ & \cellcolor{deltared!11}$-0.23\%$ & \cellcolor{deltared!62}$-1.24\%$ & \cellcolor{deltared!6}$-0.11\%$ & $0.00\%$ & \cellcolor{deltared!34}$-0.58\%$ & \cellcolor{deltared!62}$-1.04\%$ & \cellcolor{deltared!14}$-0.23\%$ \\
MoleculePGD $\downarrow$ & $0.00\%$ & \cellcolor{deltagreen!62}$+22.01\%$ & \cellcolor{deltared!62}$-53.07\%$ & \cellcolor{deltared!5}$-4.53\%$ & $0.00\%$ & \cellcolor{deltared!53}$-137.33\%$ & \cellcolor{deltared!62}$-159.45\%$ & \cellcolor{deltared!54}$-138.25\%$ \\
KL score $\uparrow$ & $0.00\%$ & \cellcolor{deltagreen!26}$+1.07\%$ & \cellcolor{deltared!62}$-8.64\%$ & \cellcolor{deltagreen!62}$+2.56\%$ & \_ & \_ & \_ & \_ \\
FG MMD$^2$ $\downarrow$ & $0.00\%$ & $0.00\%$ & \cellcolor{deltared!62}$-10.00\%$ & $0.00\%$ & $0.00\%$ & \cellcolor{deltared!62}$-3.33\%$ & \cellcolor{deltared!62}$-3.33\%$ & \cellcolor{deltared!62}$-3.33\%$ \\
SMARTS MMD$^2$ $\downarrow$ & $0.00\%$ & \cellcolor{deltared!13}$-13.04\%$ & \cellcolor{deltared!62}$-60.87\%$ & \cellcolor{deltagreen!62}$+8.70\%$ & $0.00\%$ & \cellcolor{deltagreen!28}$+17.93\%$ & \cellcolor{deltagreen!62}$+40.00\%$ & \cellcolor{deltagreen!44}$+28.28\%$ \\
Ring MMD$^2$ $\downarrow$ & $0.00\%$ & \cellcolor{deltared!6}$-25.00\%$ & \cellcolor{deltared!62}$-252.78\%$ & \cellcolor{deltared!2}$-8.33\%$ & $0.00\%$ & \cellcolor{deltagreen!37}$+21.39\%$ & \cellcolor{deltagreen!34}$+19.90\%$ & \cellcolor{deltagreen!62}$+36.32\%$ \\
BRICS L2 $\downarrow$ & $0.00\%$ & \cellcolor{deltagreen!62}$+13.17\%$ & \cellcolor{deltagreen!2}$+0.41\%$ & \cellcolor{deltared!62}$-17.70\%$ & $0.00\%$ & \cellcolor{deltagreen!20}$+3.84\%$ & \cellcolor{deltagreen!62}$+12.09\%$ & \cellcolor{deltagreen!17}$+3.26\%$ \\
\bottomrule
\end{tabular}
\end{adjustbox}
\end{table}

%% file: tables/appendix_proteins_full.tex
\begingroup
\caption{\textbf{Proteins: standard graph evaluation.} Mean\,$\pm$\,sample SD; lower panel: improvement over SENT. Shared conventions: Appendix~\ref{app:nonmolecular_full}.}
\label{tab:appendix_proteins_full}
\centering
\scriptsize
\setlength{\tabcolsep}{2.2pt}
\renewcommand{\arraystretch}{0.96}
\begin{adjustbox}{max width=\textwidth}
\begin{tabular}{@{}lcccc@{}}
\toprule
 & \textbf{SENT} & \textbf{HDT-HAC-R}$^{*}$ & \textbf{HDT-HAC-FR}$^{*}$ & \textbf{Training samples} \\
\midrule
\multicolumn{5}{@{}l}{\emph{Absolute performance}} \\
Validity $\uparrow$ & $0.540 \pm 0.031$ & \underline{$0.879 \pm 0.019$} & $\boldsymbol{0.960 \pm 0.006}$ & $0.984 \pm 0.005$ \\
Uniqueness $\uparrow$ & $\boldsymbol{1.000 \pm 0.000}$ & $\boldsymbol{1.000 \pm 0.000}$ & $\boldsymbol{1.000 \pm 0.000}$ & $1.000 \pm 0.000$ \\
Novelty $\uparrow$ & $0.540 \pm 0.031$ & \underline{$0.879 \pm 0.019$} & $\boldsymbol{0.960 \pm 0.006}$ & $0.000 \pm 0.000$ \\
V.U.N. $\uparrow$ & $0.540 \pm 0.031$ & \underline{$0.879 \pm 0.019$} & $\boldsymbol{0.960 \pm 0.006}$ & $0.000 \pm 0.000$ \\
Degree MMD $\downarrow$ & $0.0039 \pm 0.0016$ & $\boldsymbol{0.0003 \pm 0.0002}$ & \underline{$0.0005 \pm 0.0001$} & $0.0003 \pm 0.0001$ \\
Clustering MMD $\downarrow$ & \underline{$0.0820 \pm 0.0052$} & $\boldsymbol{0.0128 \pm 0.0005}$ & $0.0827 \pm 0.0042$ & $0.0118 \pm 0.0017$ \\
Orbit MMD $\downarrow$ & $0.0278 \pm 0.0064$ & $\boldsymbol{0.0060 \pm 0.0057}$ & \underline{$0.0131 \pm 0.0038$} & $0.0043 \pm 0.0021$ \\
Spectral MMD $\downarrow$ & $0.0047 \pm 0.0006$ & $\boldsymbol{0.0007 \pm 0.0001}$ & \underline{$0.0040 \pm 0.0001$} & $0.0008 \pm 0.0000$ \\
Ratio $\downarrow$ & $10.718 \pm 1.763$ & $\boldsymbol{1.568 \pm 0.668}$ & \underline{$6.525 \pm 0.316$} & $1.421 \pm 0.263$ \\
\midrule
\multicolumn{5}{@{}l}{\emph{Direction-normalized change relative to SENT}} \\
Validity $\uparrow$ & $0.00\%$ & \cellcolor{deltagreen!81}$+62.75\%$ & \cellcolor{deltagreen!100}$+77.85\%$ & \_ \\
Uniqueness $\uparrow$ & $0.00\%$ & $0.00\%$ & $0.00\%$ & \_ \\
Novelty $\uparrow$ & $0.00\%$ & \cellcolor{deltagreen!81}$+62.75\%$ & \cellcolor{deltagreen!100}$+77.85\%$ & \_ \\
V.U.N. $\uparrow$ & $0.00\%$ & \cellcolor{deltagreen!81}$+62.75\%$ & \cellcolor{deltagreen!100}$+77.85\%$ & \_ \\
Degree MMD $\downarrow$ & $0.00\%$ & \cellcolor{deltagreen!100}$+91.59\%$ & \cellcolor{deltagreen!96}$+87.77\%$ & \_ \\
Clustering MMD $\downarrow$ & $0.00\%$ & \cellcolor{deltagreen!100}$+84.40\%$ & \cellcolor{deltared!100}$-0.82\%$ & \_ \\
Orbit MMD $\downarrow$ & $0.00\%$ & \cellcolor{deltagreen!100}$+78.56\%$ & \cellcolor{deltagreen!67}$+52.81\%$ & \_ \\
Spectral MMD $\downarrow$ & $0.00\%$ & \cellcolor{deltagreen!100}$+84.89\%$ & \cellcolor{deltagreen!17}$+14.74\%$ & \_ \\
Ratio $\downarrow$ & $0.00\%$ & \cellcolor{deltagreen!100}$+85.37\%$ & \cellcolor{deltagreen!46}$+39.12\%$ & \_ \\
\bottomrule
\end{tabular}
\end{adjustbox}
\par\endgroup

%% file: tables/appendix_proteins_novel.tex
\begingroup
\caption{\textbf{Proteins: novel-only graph evaluation.} Mean\,$\pm$\,sample SD; lower panel: improvement over SENT. Shared conventions: Appendix~\ref{app:nonmolecular_full}.}
\label{tab:appendix_proteins_novel}
\centering
\scriptsize
\setlength{\tabcolsep}{2.2pt}
\renewcommand{\arraystretch}{0.96}
\begin{adjustbox}{max width=\textwidth}
\begin{tabular}{@{}l c c c c@{}}
\toprule
& \textbf{SENT} & \textbf{HDT-HAC-R}$^{*}$ & \textbf{HDT-HAC-FR}$^{*}$ & \textbf{Training samples} \\
\midrule
\multicolumn{5}{@{}l}{\emph{Absolute performance}} \\
\addlinespace[1pt]
Raw valid-novel yield $\uparrow$ & $0.612 \pm 0.006$ & $\boldsymbol{0.972 \pm 0.007}$ & \underline{$ 0.962 \pm 0.015 $} & \textemdash \\
Selected uniqueness $\uparrow$ & $\boldsymbol{1.000 \pm 0.000}$ & $\boldsymbol{1.000 \pm 0.000}$ & $\boldsymbol{1.000 \pm 0.000}$ & $1.000 \pm 0.000$ \\
Degree MMD $\downarrow$ & $0.0092 \pm 0.0018$ & $\boldsymbol{0.0004 \pm 0.0003}$ & \underline{$ 0.0005 \pm 0.0002 $} & $0.0003 \pm 0.0001$ \\
Clustering MMD $\downarrow$ & \underline{$ 0.0638 \pm 0.0058 $} & $\boldsymbol{0.0127 \pm 0.0006}$ & $0.0806 \pm 0.0047$ & $0.0118 \pm 0.0017$ \\
Orbit MMD $\downarrow$ & $0.0658 \pm 0.0105$ & $\boldsymbol{0.0070 \pm 0.0067}$ & \underline{$ 0.0138 \pm 0.0051 $} & $0.0043 \pm 0.0021$ \\
Spectral MMD $\downarrow$ & $0.0044 \pm 0.0005$ & $\boldsymbol{0.0007 \pm 0.0001}$ & \underline{$ 0.0040 \pm 0.0000 $} & $0.0008 \pm 0.0000$ \\
Ratio $\downarrow$ & $17.137 \pm 2.621$ & $\boldsymbol{1.680 \pm 0.803}$ & \underline{$ 6.483 \pm 0.475 $} & $1.421 \pm 0.263$ \\
\midrule
\multicolumn{5}{@{}l}{\emph{Direction-normalized change relative to SENT}} \\
\addlinespace[1pt]
Raw valid-novel yield $\uparrow$ & $0.00\%$ & \cellcolor{deltagreen!100}$+58.83\%$ & \cellcolor{deltagreen!97}$+57.13\%$ & \textemdash \\
Selected uniqueness $\uparrow$ & $0.00\%$ & $0.00\%$ & $0.00\%$ & \textemdash \\
Degree MMD $\downarrow$ & $0.00\%$ & \cellcolor{deltagreen!100}$+95.89\%$ & \cellcolor{deltagreen!99}$+94.79\%$ & \textemdash \\
Clustering MMD $\downarrow$ & $0.00\%$ & \cellcolor{deltagreen!100}$+80.10\%$ & \cellcolor{deltared!100}$-26.35\%$ & \textemdash \\
Orbit MMD $\downarrow$ & $0.00\%$ & \cellcolor{deltagreen!100}$+89.44\%$ & \cellcolor{deltagreen!88}$+78.99\%$ & \textemdash \\
Spectral MMD $\downarrow$ & $0.00\%$ & \cellcolor{deltagreen!100}$+84.07\%$ & \cellcolor{deltagreen!12}$+10.35\%$ & \textemdash \\
Ratio $\downarrow$ & $0.00\%$ & \cellcolor{deltagreen!100}$+90.19\%$ & \cellcolor{deltagreen!69}$+62.17\%$ & \textemdash \\
\bottomrule
\end{tabular}
\end{adjustbox}
\par\endgroup

%% file: tables/appendix_planar_full.tex
\begingroup
\caption{\textbf{Planar: standard graph evaluation.} Mean\,$\pm$\,sample SD; lower panel: improvement over SENT. Shared conventions: Appendix~\ref{app:nonmolecular_full}.}
\label{tab:appendix_planar_full}
\centering
\scriptsize
\setlength{\tabcolsep}{2.2pt}
\renewcommand{\arraystretch}{0.96}
\begin{adjustbox}{max width=\textwidth}
\begin{tabular}{@{}lcccc@{}}
\toprule
 & \textbf{SENT} & \textbf{HDT-HAC-R}$^{*}$ & \textbf{HDT-HAC-FR}$^{*}$ & \textbf{Training samples} \\
\midrule
\multicolumn{5}{@{}l}{\emph{Absolute performance}} \\
Validity $\uparrow$ & $\boldsymbol{0.017 \pm 0.029}$ & \underline{$0.000 \pm 0.000$} & $0.000 \pm 0.000$ & $1.000 \pm 0.000$ \\
Uniqueness $\uparrow$ & $\boldsymbol{1.000 \pm 0.000}$ & \underline{$1.000 \pm 0.000$} & $1.000 \pm 0.000$ & $1.000 \pm 0.000$ \\
Novelty $\uparrow$ & $\boldsymbol{1.000 \pm 0.000}$ & \underline{$1.000 \pm 0.000$} & $1.000 \pm 0.000$ & $0.000 \pm 0.000$ \\
V.U.N. $\uparrow$ & $\boldsymbol{0.017 \pm 0.029}$ & \underline{$0.000 \pm 0.000$} & $0.000 \pm 0.000$ & $0.000 \pm 0.000$ \\
Degree MMD $\downarrow$ & $\boldsymbol{0.0016 \pm 0.0012}$ & $0.0034 \pm 0.0005$ & \underline{$0.0031 \pm 0.0007$} & $0.0006 \pm 0.0003$ \\
Clustering MMD $\downarrow$ & $\boldsymbol{0.0723 \pm 0.0117}$ & $0.1545 \pm 0.0034$ & \underline{$0.0829 \pm 0.0133$} & $0.0481 \pm 0.0182$ \\
Orbit MMD $\downarrow$ & $\boldsymbol{0.0114 \pm 0.0019}$ & $0.0338 \pm 0.0011$ & \underline{$0.0235 \pm 0.0084$} & $0.0005 \pm 0.0002$ \\
Spectral MMD $\downarrow$ & $\boldsymbol{0.0135 \pm 0.0027}$ & $0.0214 \pm 0.0020$ & \underline{$0.0161 \pm 0.0001$} & $0.0079 \pm 0.0011$ \\
Ratio $\downarrow$ & $\boldsymbol{6.217 \pm 1.595}$ & $16.333 \pm 0.664$ & \underline{$11.786 \pm 2.628$} & $1.224 \pm 0.321$ \\
\midrule
\multicolumn{5}{@{}l}{\emph{Direction-normalized change relative to SENT}} \\
Validity $\uparrow$ & $0.00\%$ & \cellcolor{deltared!62}$-100.00\%$ & \cellcolor{deltared!62}$-100.00\%$ & \_ \\
Uniqueness $\uparrow$ & $0.00\%$ & $0.00\%$ & $0.00\%$ & \_ \\
Novelty $\uparrow$ & $0.00\%$ & $0.00\%$ & $0.00\%$ & \_ \\
V.U.N. $\uparrow$ & $0.00\%$ & \cellcolor{deltared!62}$-100.00\%$ & \cellcolor{deltared!62}$-100.00\%$ & \_ \\
Degree MMD $\downarrow$ & $0.00\%$ & \cellcolor{deltared!62}$-114.36\%$ & \cellcolor{deltared!52}$-95.73\%$ & \_ \\
Clustering MMD $\downarrow$ & $0.00\%$ & \cellcolor{deltared!62}$-113.74\%$ & \cellcolor{deltared!8}$-14.71\%$ & \_ \\
Orbit MMD $\downarrow$ & $0.00\%$ & \cellcolor{deltared!62}$-195.10\%$ & \cellcolor{deltared!34}$-105.47\%$ & \_ \\
Spectral MMD $\downarrow$ & $0.00\%$ & \cellcolor{deltared!62}$-59.22\%$ & \cellcolor{deltared!21}$-19.98\%$ & \_ \\
Ratio $\downarrow$ & $0.00\%$ & \cellcolor{deltared!62}$-162.72\%$ & \cellcolor{deltared!34}$-89.59\%$ & \_ \\
\bottomrule
\end{tabular}
\end{adjustbox}
\par\endgroup

%% file: tables/appendix_planar_novel.tex
\begingroup
\caption{\textbf{Planar: novel-only graph evaluation.} Mean\,$\pm$\,sample SD; lower panel: improvement over SENT. Yields are near zero for every model, so the MMD rows rather than the counting rows carry the comparison. Shared conventions: Appendix~\ref{app:nonmolecular_full}.}
\label{tab:appendix_planar_novel}
\centering
\scriptsize
\setlength{\tabcolsep}{2.2pt}
\renewcommand{\arraystretch}{0.96}
\begin{adjustbox}{max width=\textwidth}
\begin{tabular}{@{}l c c c c@{}}
\toprule
& \textbf{SENT} & \textbf{HDT-HAC-R}$^{*}$ & \textbf{HDT-HAC-FR}$^{*}$ & \textbf{Training samples} \\
\midrule
\multicolumn{5}{@{}l}{\emph{Absolute performance}} \\
\addlinespace[1pt]
Raw valid-novel yield $\uparrow$ & $\boldsymbol{0.007 \pm 0.000}$ & $0.001 \pm 0.000$ & \underline{$ 0.003 \pm 0.000 $} & \textemdash \\
Selected uniqueness $\uparrow$ & $\boldsymbol{1.000 \pm 0.000}$ & \underline{$ 0.992 \pm 0.014 $} & $\boldsymbol{1.000 \pm 0.000}$ & $1.000 \pm 0.000$ \\
Degree MMD $\downarrow$ & $\boldsymbol{0.0008 \pm 0.0004}$ & $0.1782 \pm 0.0108$ & \underline{$ 0.0289 \pm 0.0098 $} & $0.0006 \pm 0.0003$ \\
Clustering MMD $\downarrow$ & $\boldsymbol{0.0397 \pm 0.0023}$ & $0.2576 \pm 0.0082$ & \underline{$ 0.0746 \pm 0.0249 $} & $0.0481 \pm 0.0182$ \\
Orbit MMD $\downarrow$ & $\boldsymbol{0.0121 \pm 0.0055}$ & $1.5909 \pm 0.0282$ & \underline{$ 0.2543 \pm 0.0964 $} & $0.0005 \pm 0.0002$ \\
Spectral MMD $\downarrow$ & $\boldsymbol{0.0109 \pm 0.0002}$ & $0.2229 \pm 0.0026$ & \underline{$ 0.0387 \pm 0.0120 $} & $0.0079 \pm 0.0011$ \\
Ratio $\downarrow$ & $\boldsymbol{5.562 \pm 2.080}$ & $704.348 \pm 5.976$ & \underline{$ 113.200 \pm 40.068 $} & $1.224 \pm 0.321$ \\
\midrule
\multicolumn{5}{@{}l}{\emph{Direction-normalized change relative to SENT}} \\
\addlinespace[1pt]
Raw valid-novel yield $\uparrow$ & $0.00\%$ & \cellcolor{deltared!100}$-84.27\%$ & \cellcolor{deltared!77}$-65.22\%$ & \textemdash \\
Selected uniqueness $\uparrow$ & $0.00\%$ & \cellcolor{deltared!100}$-0.83\%$ & $0.00\%$ & \textemdash \\
Degree MMD $\downarrow$ & $0.00\%$ & \cellcolor{deltared!100}$-22067.54\%$ & \cellcolor{deltared!16}$-3493.91\%$ & \textemdash \\
Clustering MMD $\downarrow$ & $0.00\%$ & \cellcolor{deltared!100}$-549.04\%$ & \cellcolor{deltared!16}$-88.00\%$ & \textemdash \\
Orbit MMD $\downarrow$ & $0.00\%$ & \cellcolor{deltared!100}$-13065.21\%$ & \cellcolor{deltared!15}$-2004.01\%$ & \textemdash \\
Spectral MMD $\downarrow$ & $0.00\%$ & \cellcolor{deltared!100}$-1938.31\%$ & \cellcolor{deltared!13}$-254.03\%$ & \textemdash \\
Ratio $\downarrow$ & $0.00\%$ & \cellcolor{deltared!100}$-12562.68\%$ & \cellcolor{deltared!15}$-1935.09\%$ & \textemdash \\
\bottomrule
\end{tabular}
\end{adjustbox}
\par\endgroup

%% file: tables/appendix_block_generators.tex
\begin{table}[h]
\caption{\textbf{Block-model generator laws.}  Counts and sizes are drawn independently from the discrete-uniform law $U$ over each inclusive integer range. SBM and SSBM share edge probabilities and differ in how many communities are planted and how large; RSBM builds coarse blocks from fine ones. Each supplies $128/32/40$ graphs.}
\label{tab:appendix_block_generators}
\centering
\small
\setlength{\tabcolsep}{5pt}
\begin{adjustbox}{max width=\textwidth}
\begin{tabular}{@{}l l l l c c@{}}
\toprule
\textbf{Dataset} & \textbf{Coarse count} & \textbf{Coarse size} & \textbf{Fine structure} & $p_{\text{in}}/p_{\text{out}}$ & \textbf{Connected filter} \\
\midrule
SBM  & $U\{2,\dots,5\}$  & $U\{20,\dots,40\}$ & none & $0.30/0.005$ & no \\
SSBM & $U\{6,\dots,10\}$ & $U\{12,\dots,20\}$ & none & $0.30/0.005$ & yes \\
Connected RSBM ($R=4$) & $U\{2,\dots,5\}$ &  $\sum_{j=1}^{F_i} L_{ij}$ & $F_i\!\sim\!U\{2,\dots,5\}$, $L_{ij}\!\sim\!U\{8,9\}$ & two-level & yes \\
\bottomrule
\end{tabular}
\end{adjustbox}
\end{table}

%% file: tables/appendix_block_populations.tex
\begin{table}[h]
\caption{\textbf{Realized properties of the frozen $200$-graph populations.}  For each graph, mean degree is $2m/n$. Means and sample variances are computed across the $200$ graphs; variances use denominator $199$. Degree variance refers to variation in graph-level mean degree. Similar means do not imply matched spread: RSBM carries roughly twice SBM's degree variance and SSBM is markedly sparser.}
\label{tab:appendix_block_populations}
\centering
\small
\setlength{\tabcolsep}{6pt}
\begin{adjustbox}{max width=\textwidth}
\begin{tabular}{@{}l r r r r r r c@{}}
\toprule
\textbf{Dataset} & \makecell{Mean\\nodes} & \makecell{Node\\variance} & \makecell{Mean\\edges} & \makecell{Edge\\variance} & \makecell{Mean\\degree} & \makecell{Degree\\variance} & \textbf{Connected} \\
\midrule
SBM  & $104.01$ & $1397.30$ & $500.41$ & $44771.24$ & $9.449$ & $1.246$ & $196/200$ \\
SSBM & $126.92$ & $550.78$  & $333.77$ & $5726.56$  & $5.230$ & $0.149$ & $200/200$ \\
Connected RSBM ($R=4$) & $104.32$ & $1376.40$ & $517.65$ & $53681.15$ & $9.638$ & $2.543$ & $200/200$ \\
\bottomrule
\end{tabular}
\end{adjustbox}
\end{table}

%% file: tables/appendix_block_alignment.tex
\begin{table}[!htbp]
\centering
\small
\setlength{\tabcolsep}{5pt}
\caption{\textbf{Hierarchy alignment on the training graphs.}
Each population contains $128$ graphs.
Leaf size is averaged over leaves; maximum depth and ARI are
averaged over graphs, with root depth zero.
Top-level ARI compares the first partition with planted coarse
communities.
Leaf ARI compares terminal leaves with planted SBM communities
or RSBM fine communities.
PH has ARI $1$ by construction.
The connected-RSBM $\tau=4$ row is a hierarchy diagnostic,
not an additional trained-model result.}
\label{tab:appendix_block_alignment}
\begin{adjustbox}{max width=\textwidth}
\begin{tabular}{@{}llrrrr@{}}
\toprule
Population & Hierarchy &
\makecell{Mean leaf\\size} &
\makecell{Mean maximum\\depth} &
\makecell{Top-level\\ARI} &
\makecell{Leaf\\ARI} \\
\midrule
SBM & HAC ($\tau=4$)  & 2.247  & 4.977 & 0.942 & 0.065 \\
    & HAC ($\tau=44$) & 30.419 & 1.000 & 0.942 & 0.942 \\
    & PH             & 30.625 & 1.000 & 1.000 & 1.000 \\
\midrule
Connected RSBM ($R=4$)
    & HAC ($\tau=4$)  & 2.119 & 5.570 & 0.846 & 0.194 \\
    & HAC ($\tau=10$) & 5.234 & 2.930 & 0.846 & 0.522 \\
    & PH             & 8.497 & 2.000 & 1.000 & 1.000 \\
\bottomrule
\end{tabular}
\end{adjustbox}
\end{table}

%% file: tables/appendix_block_vun.tex
\begin{table}[!htbp]
\centering
\scriptsize
\setlength{\tabcolsep}{4pt}
\renewcommand{\arraystretch}{1.08}
\caption{\textbf{Block-model V.U.N.\ decomposition.}
All HAC and PH rows are HDT variants.
Values are mean$\pm$sample SD over generation seeds $42$--$44$
from one training seed, with $1000$ attempts per seed,
top-$k=60$, and temperature $1$.
Novelty uses the corresponding full $128$-graph training set.
All hierarchical runs use $200$k updates; SENT uses
$458.5$k, $365$k, and $245$k updates in panels A, B, and C,
respectively, based on training-time calibration.
Contexts are $2048$ for panel-A SENT/default HAC and $1440$
for adjusted HAC/PH; $2816$ throughout panel B; and $1664$
for panel-C HAC versus $1600$ for SENT/PH.
Panel C's validity and V.U.N.\ are compatibility scores
under the original unconditional RSBM gate ($\dagger$).
Panels B and C use different populations and should be
compared within panel.}
\label{tab:appendix_block_vun}
\begin{adjustbox}{max width=\textwidth}
\begin{tabular}{@{}lccccc@{}}
\toprule
Configuration & V & U & N &  U.N. & \textbf{V.U.N.} \\
\midrule
\multicolumn{6}{@{}l}{\textbf{A. Original SBM}} \\
SENT
 & $0.669\pmstd{0.022}$ & $1.000\pmstd{0.000}$
 & $1.000\pmstd{0.000}$ & $1.000\pmstd{0.000}$
 & $0.669\pmstd{0.022}$ \\
HAC-R ($\tau=4$)
 & $0.926\pmstd{0.003}$ & $0.406\pmstd{0.014}$
 & $0.408\pmstd{0.007}$ & $0.290\pmstd{0.013}$
 & $0.276\pmstd{0.009}$ \\
HAC-FR ($\tau=4$)
 & $0.445\pmstd{0.022}$ & $0.999\pmstd{0.001}$
 & $0.999\pmstd{0.001}$ & $0.999\pmstd{0.001}$
 & $0.445\pmstd{0.022}$ \\
HAC-R ($\tau=44$)
 & $0.709\pmstd{0.011}$ & $1.000\pmstd{0.001}$
 & $1.000\pmstd{0.001}$ & $1.000\pmstd{0.001}$
 & $0.709\pmstd{0.011}$ \\
HAC-FR ($\tau=44$)
 & $0.694\pmstd{0.006}$ & $1.000\pmstd{0.000}$
 & $1.000\pmstd{0.000}$ & $1.000\pmstd{0.000}$
 & $0.694\pmstd{0.006}$ \\
PH-R
 & $0.887\pmstd{0.010}$ & $1.000\pmstd{0.000}$
 & $1.000\pmstd{0.000}$ & $1.000\pmstd{0.000}$
 & $0.887\pmstd{0.010}$ \\
PH-FR
 & $0.848\pmstd{0.005}$ & $1.000\pmstd{0.000}$
 & $1.000\pmstd{0.000}$ & $1.000\pmstd{0.000}$
 & $0.848\pmstd{0.005}$ \\
\midrule
\multicolumn{6}{@{}l}{\textbf{B. Original RSBM ($R=4$)}} \\
SENT
 & $0.408\pmstd{0.018}$ & $1.000\pmstd{0.000}$
 & $1.000\pmstd{0.000}$ & $1.000\pmstd{0.000}$
 & $0.408\pmstd{0.018}$ \\
HAC-R ($\tau=4$)
 & $0.929\pmstd{0.002}$ & $0.459\pmstd{0.008}$
 & $0.529\pmstd{0.007}$ & $0.346\pmstd{0.003}$
 & $0.316\pmstd{0.008}$ \\
HAC-FR ($\tau=4$)
 & $0.197\pmstd{0.007}$ & $0.999\pmstd{0.002}$
 & $0.999\pmstd{0.002}$ & $0.999\pmstd{0.002}$
 & $0.197\pmstd{0.007}$ \\
\midrule
\multicolumn{6}{@{}l}{
\textbf{C. Connected RSBM ($R=4$): $V^\dagger$ and V.U.N.$^\dagger$}
} \\
SENT
 & $0.373\pmstd{0.019}$ & $1.000\pmstd{0.000}$
 & $1.000\pmstd{0.000}$ & $1.000\pmstd{0.000}$
 & $0.373\pmstd{0.019}$ \\
HAC-R ($\tau=10$)
 & $0.860\pmstd{0.012}$ & $0.995\pmstd{0.002}$
 & $0.994\pmstd{0.002}$ & $0.994\pmstd{0.002}$
 & $0.854\pmstd{0.011}$ \\
HAC-FR ($\tau=10$)
 & $0.083\pmstd{0.009}$ & $0.998\pmstd{0.001}$
 & $0.998\pmstd{0.001}$ & $0.998\pmstd{0.001}$
 & $0.083\pmstd{0.009}$ \\
PH-R
 & $0.858\pmstd{0.006}$ & $1.000\pmstd{0.000}$
 & $1.000\pmstd{0.000}$ & $1.000\pmstd{0.000}$
 & $0.858\pmstd{0.006}$ \\
PH-FR
 & $0.744\pmstd{0.017}$ & $1.000\pmstd{0.000}$
 & $1.000\pmstd{0.000}$ & $1.000\pmstd{0.000}$
 & $0.744\pmstd{0.017}$ \\
\bottomrule
\end{tabular}
\end{adjustbox}
\end{table}

%% file: tables/appendix_sbm_full.tex
\begingroup
\caption{\textbf{SBM: standard graph evaluation.} Mean\,$\pm$\,sample SD; lower panel: improvement over SENT. \emph{PH} uses generator memberships as a privileged control. Shared conventions: Appendix~\ref{app:nonmolecular_full}.}
\label{tab:appendix_sbm_full}
\centering
\scriptsize
\setlength{\tabcolsep}{2.4pt}
\renewcommand{\arraystretch}{0.96}
\begin{adjustbox}{max width=\textwidth}
\begin{tabular}{@{}lcccccc@{}}
\toprule
 & \textbf{SENT} & \textbf{HDT-HAC-R}$^{*}$ & \textbf{HDT-PH-R} & \textbf{HDT-HAC-FR}$^{*}$ & \textbf{HDT-PH-FR} & \textbf{Training samples} \\
\midrule
\multicolumn{7}{@{}l}{\emph{Absolute performance}} \\
V.U.N. $\uparrow$ & $0.669 \pm$ $0.022$ & $0.276 \pm 0.009$ & $\boldsymbol{0.887 \pm 0.010}$ & $0.445 \pm$ $0.022$ & \underline{$0.848 \pm 0.005$} & $0.000$ \\
Validity $\uparrow$ & $0.669 \pm$ $0.022$ & $\boldsymbol{0.926 \pm 0.003}$ & \underline{$0.887 \pm 0.010$} & $0.445 \pm$ $0.022$ & $0.848 \pm 0.005$ & $0.930$ \\
Uniqueness $\uparrow$ & $\boldsymbol{1.000 \pm 0.000}$ & $0.406 \pm 0.014$ & \underline{$1.000 \pm 0.000$} & $0.999 \pm 0.001$ & $1.000 \pm 0.000$ & $1.000$ \\
Novelty $\uparrow$ & $\boldsymbol{1.000 \pm 0.000}$ & $0.408 \pm 0.007$ & \underline{$1.000 \pm 0.000$} & $0.999 \pm 0.001$ & $1.000 \pm 0.000$ & $0.000$ \\
U.N. $\uparrow$ & $\boldsymbol{1.000 \pm 0.000}$ & $0.290 \pm 0.013$ & \underline{$1.000 \pm 0.000$} & $0.999 \pm 0.001$ & $1.000 \pm 0.000$ & $0.000$ \\
Degree MMD $\downarrow$ & $0.0060 \pm 0.0001$ & $\boldsymbol{0.0009 \pm 0.0001}$ & \underline{$0.0011 \pm 0.0002$} & $0.0054 \pm 0.0008$ & $0.0021 \pm 0.0003$ & $0.0008$ \\
Spectral MMD $\downarrow$ & $0.0016 \pm 0.0002$ & $0.0010 \pm 0.0000$ & $\boldsymbol{0.0004 \pm 0.0001}$ & $0.0043 \pm 0.0002$ & \underline{$0.0007 \pm 0.0001$} & $0.0011$ \\
Clustering MMD $\downarrow$ & $0.0049 \pm 0.0003$ & $0.0093 \pm 0.0001$ & $\boldsymbol{0.0023 \pm 0.0000}$ & $0.0094 \pm 0.0006$ & \underline{$0.0027 \pm 0.0001$} & $0.0091$ \\
Orbit MMD $\downarrow$ & $0.0351 \pm 0.0031$ & \underline{$0.0123 \pm 0.0007$} & $\boldsymbol{0.0113 \pm 0.0012}$ & $0.0347 \pm 0.0022$ & $0.0165 \pm 0.0019$ & $0.0122$ \\
Ratio $\downarrow$ & $3.053 \pm 0.320$ & \underline{$1.000 \pm 0.046$} & $\boldsymbol{0.742 \pm 0.021}$ & $3.579 \pm 0.204$ & $1.223 \pm 0.259$ & $1.000$ \\
\midrule
\multicolumn{7}{@{}l}{\emph{Direction-normalized change relative to SENT}} \\
V.U.N. $\uparrow$ & $0.00\%$ & \cellcolor{deltared!62}$-58.67\%$ & \cellcolor{deltagreen!62}$+32.70\%$ & \cellcolor{deltared!35}$-33.40\%$ & \cellcolor{deltagreen!51}$+26.82\%$ & \_ \\
Validity $\uparrow$ & $0.00\%$ & \cellcolor{deltagreen!62}$+38.53\%$ & \cellcolor{deltagreen!53}$+32.70\%$ & \cellcolor{deltared!62}$-33.40\%$ & \cellcolor{deltagreen!43}$+26.82\%$ & \_ \\
Uniqueness $\uparrow$ & $0.00\%$ & \cellcolor{deltared!62}$-59.37\%$ & $0.00\%$ & \cellcolor{deltared!0}$-0.07\%$ & $0.00\%$ & \_ \\
Novelty $\uparrow$ & $0.00\%$ & \cellcolor{deltared!62}$-59.23\%$ & $0.00\%$ & \cellcolor{deltared!0}$-0.07\%$ & $0.00\%$ & \_ \\
U.N. $\uparrow$ & $0.00\%$ & \cellcolor{deltared!62}$-71.00\%$ & $0.00\%$ & \cellcolor{deltared!0}$-0.07\%$ & $0.00\%$ & \_ \\
Degree MMD $\downarrow$ & $0.00\%$ & \cellcolor{deltagreen!62}$+85.46\%$ & \cellcolor{deltagreen!59}$+80.92\%$ & \cellcolor{deltagreen!8}$+10.53\%$ & \cellcolor{deltagreen!48}$+65.68\%$ & \_ \\
Spectral MMD $\downarrow$ & $0.00\%$ & \cellcolor{deltagreen!31}$+36.39\%$ & \cellcolor{deltagreen!62}$+72.09\%$ & \cellcolor{deltared!62}$-173.27\%$ & \cellcolor{deltagreen!45}$+52.32\%$ & \_ \\
Clustering MMD $\downarrow$ & $0.00\%$ & \cellcolor{deltared!60}$-91.96\%$ & \cellcolor{deltagreen!62}$+51.84\%$ & \cellcolor{deltared!62}$-94.27\%$ & \cellcolor{deltagreen!53}$+44.56\%$ & \_ \\
Orbit MMD $\downarrow$ & $0.00\%$ & \cellcolor{deltagreen!59}$+65.01\%$ & \cellcolor{deltagreen!62}$+67.86\%$ & \cellcolor{deltagreen!1}$+1.35\%$ & \cellcolor{deltagreen!49}$+53.11\%$ & \_ \\
Ratio $\downarrow$ & $0.00\%$ & \cellcolor{deltagreen!55}$+67.25\%$ & \cellcolor{deltagreen!62}$+75.70\%$ & \cellcolor{deltared!62}$-17.24\%$ & \cellcolor{deltagreen!49}$+59.95\%$ & \_ \\
\bottomrule
\end{tabular}
\end{adjustbox}
\par\endgroup

%% file: tables/appendix_sbm_novel.tex
\begingroup
\caption{\textbf{SBM: novel-only graph evaluation.} Mean\,$\pm$\,sample SD; lower panel: improvement over SENT. Shared conventions: Appendix~\ref{app:nonmolecular_full}.}
\label{tab:appendix_sbm_novel}
\centering
\scriptsize
\setlength{\tabcolsep}{2.4pt}
\renewcommand{\arraystretch}{0.96}
\begin{adjustbox}{max width=\textwidth}
\begin{tabular}{@{}lccc@{}}
\toprule
 & \textbf{SENT} & \textbf{HDT-HAC-R}$^{*}$ & \textbf{HDT-HAC-FR}$^{*}$ \\
\midrule
\multicolumn{4}{@{}l}{\emph{Absolute performance}} \\
Raw  valid-novel yield $\uparrow$ & $\boldsymbol{0.669 \pm 0.016}$ & $0.398 \pm 0.008$ & \underline{$0.445 \pm 0.006$} \\
Selected uniqueness $\uparrow$ & $\boldsymbol{1.000 \pm 0.000}$ & $0.598 \pm 0.015$ & \underline{$1.000 \pm 0.000$} \\
Degree MMD $\downarrow$ & $0.0046 \pm 0.0002$ & \underline{$0.0017 \pm 0.0004$} & $\boldsymbol{0.0015 \pm 0.0001}$ \\
Spectral MMD $\downarrow$ & $\boldsymbol{0.0010 \pm 0.0001}$ & \underline{$0.0017 \pm 0.0001$} & $0.0020 \pm 0.0001$ \\
Clustering MMD $\downarrow$ & $\boldsymbol{0.0037 \pm 0.0002}$ & $0.0125 \pm 0.0005$ & \underline{$0.0051 \pm 0.0002$} \\
Orbit MMD $\downarrow$ & $0.0306 \pm 0.0017$ & \underline{$0.0194 \pm 0.0006$} & $\boldsymbol{0.0164 \pm 0.0008}$ \\
Ratio $\downarrow$ & $2.368 \pm 0.166$ & \underline{$1.637 \pm 0.074$} & $\boldsymbol{1.397 \pm 0.023}$ \\
\midrule
\multicolumn{4}{@{}l}{\emph{Direction-normalized change relative to SENT}} \\
Raw  valid-novel yield $\uparrow$ & $0.00\%$ & \cellcolor{deltared!62}$-40.61\%$ & \cellcolor{deltared!51}$-33.61\%$ \\
Selected uniqueness $\uparrow$ & $0.00\%$ & \cellcolor{deltared!62}$-40.17\%$ & $0.00\%$ \\
Degree MMD $\downarrow$ & $0.00\%$ & \cellcolor{deltagreen!58}$+63.27\%$ & \cellcolor{deltagreen!62}$+67.34\%$ \\
Spectral MMD $\downarrow$ & $0.00\%$ & \cellcolor{deltared!41}$-68.32\%$ & \cellcolor{deltared!62}$-102.49\%$ \\
Clustering MMD $\downarrow$ & $0.00\%$ & \cellcolor{deltared!62}$-237.31\%$ & \cellcolor{deltared!10}$-38.51\%$ \\
Orbit MMD $\downarrow$ & $0.00\%$ & \cellcolor{deltagreen!49}$+36.52\%$ & \cellcolor{deltagreen!62}$+46.51\%$ \\
Ratio $\downarrow$ & $0.00\%$ & \cellcolor{deltagreen!47}$+30.88\%$ & \cellcolor{deltagreen!62}$+41.04\%$ \\
\bottomrule
\end{tabular}
\end{adjustbox}
\par\endgroup

%% file: tables/appendix_ssbm_full.tex
\begingroup
\caption{\textbf{Sparse SBM: standard graph evaluation.} Mean\,$\pm$\,sample SD; lower panel: improvement over SENT. Shared conventions: Appendix~\ref{app:nonmolecular_full}.}
\label{tab:appendix_ssbm_full}
\centering
\scriptsize
\setlength{\tabcolsep}{2.4pt}
\renewcommand{\arraystretch}{0.96}
\begin{adjustbox}{max width=\textwidth}
\begin{tabular}{@{}lcccc@{}}
\toprule
 & \textbf{SENT} & \textbf{HDT-HAC-R}$^{*}$ & \textbf{HDT-HAC-FR}$^{*}$ & \textbf{Training samples} \\
\midrule
\multicolumn{5}{@{}l}{\emph{Absolute performance}} \\
V.U.N. $\uparrow$ & \underline{$0.527 \pm 0.002$} & $\boldsymbol{0.840 \pm 0.003}$ & $0.485 \pm 0.012$ & $0.000$ \\
Validity $\uparrow$ & \underline{$0.527 \pm 0.002$} & $\boldsymbol{0.842 \pm 0.004}$ & $0.485 \pm 0.012$ & $0.883$ \\
Uniqueness $\uparrow$ & $\boldsymbol{1.000 \pm 0.000}$ & $1.000 \pm 0.001$ & \underline{$1.000 \pm 0.000$} & $1.000$ \\
Novelty $\uparrow$ & $\boldsymbol{1.000 \pm 0.000}$ & $0.998 \pm 0.001$ & \underline{$1.000 \pm 0.000$} & $0.000$ \\
U.N. $\uparrow$ & $\boldsymbol{1.000 \pm 0.000}$ & $0.998 \pm 0.001$ & \underline{$1.000 \pm 0.000$} & $0.000$ \\
Degree MMD $\downarrow$ & \underline{$0.0018 \pm 0.0001$} & $\boldsymbol{0.0001 \pm 0.0000}$ & $0.0021 \pm 0.0004$ & $0.0000$ \\
Spectral MMD $\downarrow$ & $\boldsymbol{0.0007 \pm 0.0000}$ & \underline{$0.0007 \pm 0.0001$} & $0.0073 \pm 0.0004$ & $0.0011$ \\
Clustering MMD $\downarrow$ & $\boldsymbol{0.0044 \pm 0.0003}$ & \underline{$0.0059 \pm 0.0003$} & $0.0093 \pm 0.0002$ & $0.0090$ \\
Orbit MMD $\downarrow$ & \underline{$0.0517 \pm 0.0103$} & $\boldsymbol{0.0032 \pm 0.0010}$ & $0.1238 \pm 0.0052$ & $0.0032$ \\
Ratio $\downarrow$ & \underline{$15.640 \pm 7.556$} & $\boldsymbol{1.270 \pm 0.269}$ & $26.095 \pm 14.717$ & $1.000$ \\
\midrule
\multicolumn{5}{@{}l}{\emph{Direction-normalized change relative to SENT}} \\
V.U.N. $\uparrow$ & $0.00\%$ & \cellcolor{deltagreen!62}$+59.29\%$ & \cellcolor{deltared!62}$-8.09\%$ & \_ \\
Validity $\uparrow$ & $0.00\%$ & \cellcolor{deltagreen!62}$+59.67\%$ & \cellcolor{deltared!62}$-8.09\%$ & \_ \\
Uniqueness $\uparrow$ & $0.00\%$ & \cellcolor{deltared!62}$-0.03\%$ & $0.00\%$ & \_ \\
Novelty $\uparrow$ & $0.00\%$ & \cellcolor{deltared!62}$-0.20\%$ & $0.00\%$ & \_ \\
U.N. $\uparrow$ & $0.00\%$ & \cellcolor{deltared!62}$-0.20\%$ & $0.00\%$ & \_ \\
Degree MMD $\downarrow$ & $0.00\%$ & \cellcolor{deltagreen!62}$+93.10\%$ & \cellcolor{deltared!62}$-16.00\%$ & \_ \\
Spectral MMD $\downarrow$ & $0.00\%$ & \cellcolor{deltared!0}$-0.88\%$ & \cellcolor{deltared!62}$-1009.91\%$ & \_ \\
Clustering MMD $\downarrow$ & $0.00\%$ & \cellcolor{deltared!19}$-33.26\%$ & \cellcolor{deltared!62}$-109.61\%$ & \_ \\
Orbit MMD $\downarrow$ & $0.00\%$ & \cellcolor{deltagreen!62}$+93.84\%$ & \cellcolor{deltared!62}$-139.36\%$ & \_ \\
Ratio $\downarrow$ & $0.00\%$ & \cellcolor{deltagreen!62}$+91.88\%$ & \cellcolor{deltared!62}$-66.85\%$ & \_ \\
\bottomrule
\end{tabular}
\end{adjustbox}
\par\endgroup

%% file: tables/appendix_ssbm_novel.tex
\begingroup
\caption{\textbf{Sparse SBM: novel-only graph evaluation.} Mean\,$\pm$\,sample SD; lower panel: improvement over SENT. Shared conventions: Appendix~\ref{app:nonmolecular_full}.}
\label{tab:appendix_ssbm_novel}
\centering
\scriptsize
\setlength{\tabcolsep}{2.4pt}
\renewcommand{\arraystretch}{0.96}
\begin{adjustbox}{max width=\textwidth}
\begin{tabular}{@{}lccc@{}}
\toprule
 & \textbf{SENT} & \textbf{HDT-HAC-R}$^{*}$ & \textbf{HDT-HAC-FR}$^{*}$ \\
\midrule
\multicolumn{4}{@{}l}{\emph{Absolute performance}} \\
Raw  valid-novel yield $\uparrow$ & \underline{$0.526 \pm 0.008$} & $\boldsymbol{0.839 \pm 0.002}$ & $0.477 \pm 0.005$ \\
Selected uniqueness $\uparrow$ & $\boldsymbol{1.000 \pm 0.000}$ & $1.000 \pm 0.001$ & \underline{$1.000 \pm 0.000$} \\
Degree MMD $\downarrow$ & \underline{$0.0016 \pm 0.0001$} & $\boldsymbol{0.0001 \pm 0.0000}$ & $0.0025 \pm 0.0002$ \\
Spectral MMD $\downarrow$ & $\boldsymbol{0.0008 \pm 0.0000}$ & \underline{$0.0008 \pm 0.0001$} & $0.0050 \pm 0.0004$ \\
Clustering MMD $\downarrow$ & $\boldsymbol{0.0054 \pm 0.0003}$ & \underline{$0.0063 \pm 0.0006$} & $0.0089 \pm 0.0004$ \\
Orbit MMD $\downarrow$ & \underline{$0.0405 \pm 0.0073$} & $\boldsymbol{0.0029 \pm 0.0008}$ & $0.1063 \pm 0.0100$ \\
Ratio $\downarrow$ & \underline{$13.422 \pm 6.838$} & $\boldsymbol{1.225 \pm 0.309}$ & $25.722 \pm 13.591$ \\
\midrule
\multicolumn{4}{@{}l}{\emph{Direction-normalized change relative to SENT}} \\
Raw  valid-novel yield $\uparrow$ & $0.00\%$ & \cellcolor{deltagreen!62}$+59.51\%$ & \cellcolor{deltared!62}$-9.23\%$ \\
Selected uniqueness $\uparrow$ & $0.00\%$ & \cellcolor{deltared!62}$-0.03\%$ & $0.00\%$ \\
Degree MMD $\downarrow$ & $0.00\%$ & \cellcolor{deltagreen!62}$+93.16\%$ & \cellcolor{deltared!62}$-54.76\%$ \\
Spectral MMD $\downarrow$ & $0.00\%$ & \cellcolor{deltared!1}$-11.17\%$ & \cellcolor{deltared!62}$-557.43\%$ \\
Clustering MMD $\downarrow$ & $0.00\%$ & \cellcolor{deltared!17}$-17.92\%$ & \cellcolor{deltared!62}$-66.29\%$ \\
Orbit MMD $\downarrow$ & $0.00\%$ & \cellcolor{deltagreen!62}$+92.84\%$ & \cellcolor{deltared!62}$-162.52\%$ \\
Ratio $\downarrow$ & $0.00\%$ & \cellcolor{deltagreen!62}$+90.87\%$ & \cellcolor{deltared!62}$-91.63\%$ \\
\bottomrule
\end{tabular}
\end{adjustbox}
\par\endgroup

%% file: tables/appendix_rsbm_full.tex
\begin{table}[!htbp]
\caption{\textbf{Connected RSBM ($R=4$): standard graph evaluation.} Mean\,$\pm$\,sample SD; lower panel: improvement over SENT. \emph{PH} uses generator memberships as a privileged control. $\dagger$: compatibility scores under the original validity gate; no recalibrated gate is available. Shared conventions: Appendix~\ref{app:nonmolecular_full}.}
\label{tab:appendix_rsbm_full}
\centering
\scriptsize
\setlength{\tabcolsep}{2.4pt}
\renewcommand{\arraystretch}{0.96}
\begin{adjustbox}{max width=\textwidth}
\begin{tabular}{@{}lccccc@{}}
\toprule
 & \textbf{SENT} & \textbf{HDT-HAC-R}$^{*}$ & \textbf{HDT-PH-R} & \textbf{HDT-HAC-FR}$^{*}$ & \textbf{HDT-PH-FR} \\
\midrule
\multicolumn{6}{@{}l}{\emph{Absolute performance}} \\
V.U.N.$^{\dagger}$ $\uparrow$ & $0.373 \pm 0.019$ & \underline{$0.854 \pm 0.011$} & $\boldsymbol{0.858 \pm 0.006}$ & $0.083 \pm 0.009$ & $0.744 \pm 0.017$ \\
Validity$^{\dagger}$ $\uparrow$ & $0.373 \pm 0.019$ & $\boldsymbol{0.860 \pm}$ $\boldsymbol{0.012}$ & \underline{$0.858 \pm 0.006$} & $0.083 \pm 0.009$ & $0.744 \pm 0.017$ \\
Uniqueness $\uparrow$ & $\boldsymbol{1.000 \pm 0.000}$ & $0.995 \pm 0.002$ & \underline{$1.000 \pm 0.000$} & $0.998 \pm 0.001$ & $1.000 \pm 0.000$ \\
Novelty $\uparrow$ & $\boldsymbol{1.000 \pm 0.000}$ & $0.994 \pm 0.002$ & \underline{$1.000 \pm 0.000$} & $0.998 \pm 0.001$ & $1.000 \pm 0.000$ \\
U.N. $\uparrow$ & $\boldsymbol{1.000 \pm 0.000}$ & $0.994 \pm 0.002$ & \underline{$1.000 \pm 0.000$} & $0.998 \pm 0.001$ & $1.000 \pm 0.000$ \\
Degree MMD $\downarrow$ & $0.0096 \pm 0.0007$ & $\boldsymbol{0.0003 \pm 0.0001}$ & \underline{$0.0006 \pm 0.0002$} & $0.0122 \pm 0.0007$ & $0.0032 \pm 0.0007$ \\
Spectral MMD $\downarrow$ & $0.0013 \pm 0.0002$ & $\boldsymbol{0.0005 \pm 0.0001}$ & $0.0014 \pm 0.0002$ & $0.0052 \pm 0.0005$ & \underline{$0.0006 \pm 0.0000$} \\
Clustering MMD $\downarrow$ & $0.0031 \pm 0.0001$ & $\boldsymbol{0.0023 \pm 0.0001}$ & \underline{$0.0024 \pm 0.0001$} & $0.0119 \pm 0.0005$ & $0.0036 \pm 0.0001$ \\
Orbit MMD $\downarrow$ & $0.0119 \pm 0.0002$ & $\boldsymbol{0.0069 \pm 0.0005}$ & \underline{$0.0078 \pm 0.0011$} & $0.0411 \pm 0.0011$ & $0.0145 \pm 0.0023$ \\
Ratio $\downarrow$ & $5.958 \pm 1.095$ & $\boldsymbol{0.561 \pm 0.045}$ & \underline{$0.985 \pm 0.183$} & $9.447 \pm 1.368$ & $2.337 \pm 0.110$ \\
\midrule
\multicolumn{6}{@{}l}{\emph{Direction-normalized change relative to SENT}} \\
V.U.N.$^{\dagger}$ $\uparrow$ & $0.00\%$ & \cellcolor{deltagreen!62}$+129.04\%$ & \cellcolor{deltagreen!62}$+130.03\%$ & \cellcolor{deltared!62}$-77.84\%$ & \cellcolor{deltagreen!47}$+99.55\%$ \\
Validity$^{\dagger}$ $\uparrow$ & $0.00\%$ & \cellcolor{deltagreen!62}$+130.56\%$ & \cellcolor{deltagreen!62}$+130.03\%$ & \cellcolor{deltared!62}$-77.84\%$ & \cellcolor{deltagreen!47}$+99.55\%$ \\
Uniqueness $\uparrow$ & $0.00\%$ & \cellcolor{deltared!62}$-0.47\%$ & $0.00\%$ & \cellcolor{deltared!22}$-0.17\%$ & $0.00\%$ \\
Novelty $\uparrow$ & $0.00\%$ & \cellcolor{deltared!62}$-0.60\%$ & $0.00\%$ & \cellcolor{deltared!18}$-0.17\%$ & $0.00\%$ \\
U.N. $\uparrow$ & $0.00\%$ & \cellcolor{deltared!62}$-0.60\%$ & $0.00\%$ & \cellcolor{deltared!18}$-0.17\%$ & $0.00\%$ \\
Degree MMD $\downarrow$ & $0.00\%$ & \cellcolor{deltagreen!62}$+96.62\%$ & \cellcolor{deltagreen!60}$+93.58\%$ & \cellcolor{deltared!62}$-27.48\%$ & \cellcolor{deltagreen!43}$+67.06\%$ \\
Spectral MMD $\downarrow$ & $0.00\%$ & \cellcolor{deltagreen!62}$+62.99\%$ & \cellcolor{deltared!1}$-6.25\%$ & \cellcolor{deltared!62}$-284.83\%$ & \cellcolor{deltagreen!57}$+57.72\%$ \\
Clustering MMD $\downarrow$ & $0.00\%$ & \cellcolor{deltagreen!62}$+25.45\%$ & \cellcolor{deltagreen!56}$+22.88\%$ & \cellcolor{deltared!62}$-285.21\%$ & \cellcolor{deltared!3}$-14.88\%$ \\
Orbit MMD $\downarrow$ & $0.00\%$ & \cellcolor{deltagreen!62}$+42.15\%$ & \cellcolor{deltagreen!51}$+34.63\%$ & \cellcolor{deltared!62}$-245.94\%$ & \cellcolor{deltared!6}$-22.16\%$ \\
Ratio $\downarrow$ & $0.00\%$ & \cellcolor{deltagreen!62}$+90.59\%$ & \cellcolor{deltagreen!57}$+83.47\%$ & \cellcolor{deltared!62}$-58.56\%$ & \cellcolor{deltagreen!42}$+60.77\%$ \\
\bottomrule
\end{tabular}
\end{adjustbox}
\end{table}

%% file: tables/appendix_regime_guacamol.tex
\begin{table}[!ht]
\caption{\textbf{Training-regime scaling on GuacaMol.} SENT, \emph{HDT}-HAC-R and \emph{HDT}-HAC-FR under both regimes, as mean\,$\pm$\,sample SD over generation seeds $42$--$44$; bold marks the best mean within each regime. The final columns give each serialization's change from GPT2-XS to GPT2-S, signed so positive is better. Model size, training population ($499{,}968$ to $1{,}201{,}664$ molecules) and schedule change together, and GPT2-S SENT keeps its historical context $256$ and attention dropout $0$, so this is an efficiency observation rather than a controlled estimate of any one factor.}
\label{tab:appendix_regime_guacamol}
\centering
\scriptsize
\setlength{\tabcolsep}{3.5pt}
\begin{adjustbox}{max width=\textwidth}
\begin{tabular}{@{}l ccc ccc ccc@{}}
\toprule
 & \multicolumn{3}{c}{\textbf{GPT2-XS}} & \multicolumn{3}{c}{\textbf{GPT2-S}} & \multicolumn{3}{c}{\textbf{XS $\rightarrow$ S change}} \\
\cmidrule(lr){2-4}\cmidrule(lr){5-7}\cmidrule(lr){8-10}
 & \textbf{SENT} & \makecell{\emph{\textbf{HDT}}\textbf{-}\\\textbf{HAC-R}} & \makecell{\emph{\textbf{HDT}}\textbf{-}\\\textbf{HAC-FR}} & \textbf{SENT} & \makecell{\emph{\textbf{HDT}}\textbf{-}\\\textbf{HAC-R}} & \makecell{\emph{\textbf{HDT}}\textbf{-}\\\textbf{HAC-FR}} & \textbf{SENT} & \makecell{\emph{\textbf{HDT}}\textbf{-}\\\textbf{HAC-R}} & \makecell{\emph{\textbf{HDT}}\textbf{-}\\\textbf{HAC-FR}} \\
\midrule
Time (1 GPU, h) & $4.98$ & $6.70$ & $5.92$ & $69.88$ & $65.46$ & $67.12$ &  &  &  \\
Epochs & $83.2$ & $51.2$ & $51.2$ & $170.4$ & $93.5$ & $93.5$ &  &  &  \\
\midrule
Val $\uparrow$ & $\boldsymbol{0.967\pmstd{0.001}}$ & $0.943\pmstd{0.001}$ & $0.966\pmstd{0.002}$ & $0.987\pmstd{0.001}$ & $0.985\pmstd{0.001}$ & $\boldsymbol{0.989\pmstd{0.002}}$ & $+2.1\%$ & $+4.5\%$ & $+2.4\%$ \\
Uniq $\uparrow$ & $\boldsymbol{1.000\pmstd{0.000}}$ & $\boldsymbol{1.000\pmstd{0.000}}$ & $\boldsymbol{1.000\pmstd{0.000}}$ & $\boldsymbol{1.000\pmstd{0.000}}$ & $0.999\pmstd{0.001}$ & $1.000\pmstd{0.000}$ & $0.0\%$ & $-0.1\%$ & $0.0\%$ \\
Nov $\uparrow$ & $\boldsymbol{0.998\pmstd{0.001}}$ & $0.987\pmstd{0.001}$ & $0.996\pmstd{0.001}$ & $\boldsymbol{0.984\pmstd{0.000}}$ & $0.704\pmstd{0.002}$ & $0.942\pmstd{0.001}$ & $-1.4\%$ & $-28.7\%$ & $-5.4\%$ \\
V.U.N. $\uparrow$ & $\boldsymbol{0.965\pmstd{0.000}}$ & $0.931\pmstd{0.001}$ & $0.961\pmstd{0.003}$ & $\boldsymbol{0.971\pmstd{0.001}}$ & $0.693\pmstd{0.002}$ & $0.932\pmstd{0.001}$ & $+0.6\%$ & $-25.6\%$ & $-3.0\%$ \\
FCD $\downarrow$ & $1.799\pmstd{0.045}$ & $\boldsymbol{0.871\pmstd{0.019}}$ & $0.989\pmstd{0.013}$ & $0.927\pmstd{0.010}$ & $\boldsymbol{0.577\pmstd{0.002}}$ & $0.691\pmstd{0.004}$ & $+48.5\%$ & $+33.8\%$ & $+30.1\%$ \\
MoleculePGD $\downarrow$ & $0.309\pmstd{0.005}$ & $\boldsymbol{0.176\pmstd{0.007}}$ & $0.196\pmstd{0.005}$ & $0.175\pmstd{0.002}$ & $\boldsymbol{0.136\pmstd{0.002}}$ & $0.142\pmstd{0.002}$ & $+43.4\%$ & $+22.7\%$ & $+27.6\%$ \\
SNN $\uparrow$ & $0.410\pmstd{0.001}$ & $\boldsymbol{0.472\pmstd{0.000}}$ & $0.445\pmstd{0.001}$ & $0.459\pmstd{0.000}$ & $\boldsymbol{0.587\pmstd{0.000}}$ & $0.521\pmstd{0.000}$ & $+12.0\%$ & $+24.4\%$ & $+17.1\%$ \\
Frag $\uparrow$ & $0.994\pmstd{0.000}$ & $\boldsymbol{0.999\pmstd{0.000}}$ & $0.998\pmstd{0.000}$ & $0.997\pmstd{0.000}$ & $\boldsymbol{0.999\pmstd{0.000}}$ & $0.999\pmstd{0.000}$ & $+0.3\%$ & $0.0\%$ & $+0.1\%$ \\
Scaff $\uparrow$ & $0.890\pmstd{0.007}$ & $\boldsymbol{0.932\pmstd{0.004}}$ & $0.925\pmstd{0.006}$ & $0.914\pmstd{0.006}$ & $\boldsymbol{0.917\pmstd{0.009}}$ & $0.912\pmstd{0.008}$ & $+2.7\%$ & $-1.6\%$ & $-1.4\%$ \\
IntDiv $\uparrow$ & $\boldsymbol{0.886\pmstd{0.001}}$ & $0.885\pmstd{0.001}$ & $0.884\pmstd{0.001}$ & $\boldsymbol{0.887\pmstd{0.001}}$ & $0.885\pmstd{0.001}$ & $0.885\pmstd{0.001}$ & $+0.1\%$ & $0.0\%$ & $+0.1\%$ \\
KL score $\uparrow$ & $0.938\pmstd{0.010}$ & $\boldsymbol{0.981\pmstd{0.001}}$ & $0.963\pmstd{0.002}$ & $0.963\pmstd{0.002}$ & $\boldsymbol{0.995\pmstd{0.000}}$ & $0.983\pmstd{0.002}$ & $+2.7\%$ & $+1.4\%$ & $+2.1\%$ \\
FG MMD$^2$ $\downarrow$ & $\boldsymbol{0.0020\pmstd{0.0000}}$ & $\boldsymbol{0.0020\pmstd{0.0000}}$ & $\boldsymbol{0.0020\pmstd{0.0000}}$ & $\boldsymbol{0.0020\pmstd{0.0000}}$ & $\boldsymbol{0.0020\pmstd{0.0000}}$ & $\boldsymbol{0.0020\pmstd{0.0000}}$ & $0.0\%$ & $0.0\%$ & $0.0\%$ \\
SMARTS MMD$^2$ $\downarrow$ & $0.0023\pmstd{0.0001}$ & $\boldsymbol{0.0021\pmstd{0.0001}}$ & $0.0022\pmstd{0.0001}$ & $0.0023\pmstd{0.0000}$ & $\boldsymbol{0.0019\pmstd{0.0000}}$ & $0.0020\pmstd{0.0001}$ & $0.0\%$ & $+9.5\%$ & $+9.1\%$ \\
Ring MMD$^2$ $\downarrow$ & $0.0036\pmstd{0.0003}$ & $0.0024\pmstd{0.0003}$ & $\boldsymbol{0.0022\pmstd{0.0003}}$ & $0.0026\pmstd{0.0005}$ & $\boldsymbol{0.0021\pmstd{0.0003}}$ & $0.0023\pmstd{0.0004}$ & $+27.8\%$ & $+12.5\%$ & $-4.5\%$ \\
BRICS L2 $\downarrow$ & $0.0243\pmstd{0.0018}$ & $\boldsymbol{0.0184\pmstd{0.0008}}$ & $0.0193\pmstd{0.0008}$ & $0.0202\pmstd{0.0004}$ & $\boldsymbol{0.0185\pmstd{0.0004}}$ & $0.0193\pmstd{0.0018}$ & $+16.9\%$ & $-0.5\%$ & $0.0\%$ \\
\bottomrule
\end{tabular}
\end{adjustbox}
\end{table}

%% file: tables/appendix_regime_moses.tex
\begin{table}[!ht]
\caption{\textbf{Training-regime scaling on MOSES.} SENT, \emph{HDT}-HAC-R and \emph{HDT}-HAC-FR under both regimes, as mean\,$\pm$\,sample SD over generation seeds $42$--$44$; bold marks the best mean within each regime. The final columns give each serialization's change from GPT2-XS to GPT2-S, signed so positive is better. Model size, training population ($499{,}968$ to $1{,}584{,}640$ molecules), effective batch ($256$ to $512$) and schedule change together, so this is an efficiency observation rather than a controlled estimate of any one factor.}
\label{tab:appendix_regime_moses}
\centering
\scriptsize
\setlength{\tabcolsep}{3.5pt}
\begin{adjustbox}{max width=\textwidth}
\begin{tabular}{@{}l ccc ccc ccc@{}}
\toprule
 & \multicolumn{3}{c}{\textbf{GPT2-XS}} & \multicolumn{3}{c}{\textbf{GPT2-S}} & \multicolumn{3}{c}{\textbf{XS $\rightarrow$ S change}} \\
\cmidrule(lr){2-4}\cmidrule(lr){5-7}\cmidrule(lr){8-10}
 & \textbf{SENT} & \makecell{\emph{\textbf{HDT}}\textbf{-}\\\textbf{HAC-R}} & \makecell{\emph{\textbf{HDT}}\textbf{-}\\\textbf{HAC-FR}} & \textbf{SENT} & \makecell{\emph{\textbf{HDT}}\textbf{-}\\\textbf{HAC-R}} & \makecell{\emph{\textbf{HDT}}\textbf{-}\\\textbf{HAC-FR}} & \textbf{SENT} & \makecell{\emph{\textbf{HDT}}\textbf{-}\\\textbf{HAC-R}} & \makecell{\emph{\textbf{HDT}}\textbf{-}\\\textbf{HAC-FR}} \\
\midrule
Time (1 GPU, h) & $3.35$ & $4.50$ & $1.36$ & $30.60$ & $29.13$ & $30.55$ &  &  &  \\
Epochs & $51.2$ & $32.0$ & $32.0$ & $129.2$ & $65.6$ & $65.6$ &  &  &  \\
\midrule
Val $\uparrow$ & $0.977\pmstd{0.001}$ & $0.978\pmstd{0.001}$ & $\boldsymbol{0.985\pmstd{0.000}}$ & $0.991\pmstd{0.001}$ & $\boldsymbol{0.998\pmstd{0.000}}$ & $\boldsymbol{0.998\pmstd{0.001}}$ & $+1.4\%$ & $+2.0\%$ & $+1.3\%$ \\
Uniq $\uparrow$ & $\boldsymbol{1.000\pmstd{0.000}}$ & $0.999\pmstd{0.000}$ & $\boldsymbol{1.000\pmstd{0.000}}$ & $\boldsymbol{1.000\pmstd{0.000}}$ &  $0.998\pmstd{0.001}$ & $0.999\pmstd{0.000}$ & $0.0\%$ & $-0.1\%$ & $-0.1\%$ \\
Nov $\uparrow$ & $\boldsymbol{0.990\pmstd{0.002}}$ & $0.950\pmstd{0.002}$ & $0.971\pmstd{0.002}$ & $\boldsymbol{0.866\pmstd{0.002}}$ & $0.505\pmstd{0.002}$ & $0.743\pmstd{0.005}$ & $-12.5\%$ & $-46.8\%$ & $-23.5\%$ \\
V.U.N. $\uparrow$ & $\boldsymbol{0.967\pmstd{0.001}}$ & $0.929\pmstd{0.003}$ & $0.956\pmstd{0.002}$ & $\boldsymbol{0.857\pmstd{0.001}}$ & $0.503\pmstd{0.002}$ & $0.741\pmstd{0.006}$ & $-11.4\%$ & $-45.9\%$ & $-22.5\%$ \\
FCD $\downarrow$ & $0.696\pmstd{0.044}$ & $\boldsymbol{0.246\pmstd{0.003}}$ & $0.260\pmstd{0.008}$ & $0.310\pmstd{0.012}$ & $\boldsymbol{0.155\pmstd{0.006}}$ & $0.163\pmstd{0.004}$ & $+55.5\%$ & $+37.0\%$ & $+37.3\%$ \\
MoleculePGD $\downarrow$ & $0.217\pmstd{0.005}$ & $\boldsymbol{0.130\pmstd{0.003}}$ & $0.137\pmstd{0.004}$ & $0.141\pmstd{0.002}$ & $\boldsymbol{0.032\pmstd{0.012}}$ & $0.052\pmstd{0.016}$ & $+35.0\%$ & $+75.4\%$ & $+62.0\%$ \\
SNN $\uparrow$ & $0.506\pmstd{0.002}$ & $\boldsymbol{0.583\pmstd{0.000}}$ & $0.560\pmstd{0.003}$ & $0.580\pmstd{0.002}$ & $\boldsymbol{0.633\pmstd{0.001}}$ & $0.617\pmstd{0.001}$ & $+14.6\%$ & $+8.6\%$ & $+10.2\%$ \\
Frag $\uparrow$ & $0.998\pmstd{0.000}$ & $\boldsymbol{0.999\pmstd{0.000}}$ & $\boldsymbol{0.999\pmstd{0.000}}$ & $0.999\pmstd{0.000}$ & $0.999\pmstd{0.000}$ & $\boldsymbol{1.000\pmstd{0.000}}$ & $+0.1\%$ & $0.0\%$ & $+0.1\%$ \\
Scaff $\uparrow$ & $0.962\pmstd{0.004}$ & $\boldsymbol{0.981\pmstd{0.001}}$ & $0.979\pmstd{0.002}$ & $0.968\pmstd{0.002}$ & $\boldsymbol{0.984\pmstd{0.001}}$ & $0.981\pmstd{0.002}$ & $+0.6\%$ & $+0.3\%$ & $+0.2\%$ \\
IntDiv $\uparrow$ & $\boldsymbol{0.867\pmstd{0.002}}$ & $0.863\pmstd{0.001}$ & $0.864\pmstd{0.002}$ & $\boldsymbol{0.864\pmstd{0.000}}$ & $0.862\pmstd{0.000}$ & $0.863\pmstd{0.001}$ & $-0.3\%$ & $-0.1\%$ & $-0.1\%$ \\
FG MMD$^2$ $\downarrow$ & $\boldsymbol{0.0030\pmstd{0.0001}}$ & $0.0031\pmstd{0.0001}$ & $\boldsymbol{0.0030\pmstd{0.0000}}$ & $\boldsymbol{0.0030\pmstd{0.0000}}$ & $\boldsymbol{0.0030\pmstd{0.0001}}$ & $\boldsymbol{0.0030\pmstd{0.0000}}$ & $0.0\%$ & $+3.2\%$ & $0.0\%$ \\
SMARTS MMD$^2$ $\downarrow$ & $0.0145\pmstd{0.0007}$ & $0.0137\pmstd{0.0015}$ & $\boldsymbol{0.0126\pmstd{0.0003}}$ & $0.0127\pmstd{0.0015}$ & $0.0120\pmstd{0.0015}$ & $\boldsymbol{0.0113\pmstd{0.0008}}$ & $+12.4\%$ & $+12.4\%$ & $+10.3\%$ \\
Ring MMD$^2$ $\downarrow$ & $0.0201\pmstd{0.0018}$ & $0.0182\pmstd{0.0031}$ & $\boldsymbol{0.0171\pmstd{0.0001}}$ & $0.0168\pmstd{0.0005}$ & $0.0145\pmstd{0.0023}$ & $\boldsymbol{0.0143\pmstd{0.0016}}$ & $+16.4\%$ & $+20.3\%$ & $+16.4\%$ \\
BRICS L2 $\downarrow$ & $0.0521\pmstd{0.0020}$ & $0.0509\pmstd{0.0018}$ & $\boldsymbol{0.0498\pmstd{0.0033}}$ & $0.0505\pmstd{0.0002}$ & $\boldsymbol{0.0501\pmstd{0.0009}}$ & $0.0505\pmstd{0.0010}$ & $+3.1\%$ & $+1.6\%$ & $-1.4\%$ \\
\bottomrule
\end{tabular}
\end{adjustbox}
\end{table}

%% file: tables/appendix_resource_autograph.tex
\begin{table}[t]
\centering
\small
\caption{\textbf{Reproduction of the flat baseline under its own full protocol (sanity check).} Published SENT is transcribed from AutoGraph~\citep{chen2025autograph} where reported; our row is the current online-SENT GuacaMol reproduction under the full protocol. The purpose is narrow: our controlled comparisons are made against our SENT, so the flat baseline should not be weakened. \texttt{n/a} = not reported by the source. \textbf{Bold} = better where both are available.}
\label{tab:appendix_resource_autograph}
\setlength{\tabcolsep}{4pt}
\renewcommand{\arraystretch}{1.05}
\begin{adjustbox}{max width=\textwidth}
\begin{tabular}{@{}lccccccccc@{}}
\toprule
 & Val $\uparrow$ & Uniq $\uparrow$ & Nov $\uparrow$ & V.U.N. $\uparrow$ & FCD $\downarrow$ & KL score $\uparrow$ & PGD $\downarrow$ & SNN $\uparrow$ & Scaff $\uparrow$ \\
\midrule
Published SENT & $\boldsymbol{0.916}$ & $\boldsymbol{1.000}$ & $0.977$ & \texttt{n/a} & $1.170$ & $\boldsymbol{0.975}$ & \texttt{n/a} & \texttt{n/a} & \texttt{n/a} \\
Our SENT (repro) & $\boldsymbol{0.987 \pm 0.001}$ & $1.000 \pm 0.000$ & $\boldsymbol{0.984 \pm 0.000}$ & $0.971 \pm 0.001$ & $\boldsymbol{0.927 \pm 0.010}$ & $0.963 \pm 0.002$ & $0.154 \pm 0.005$ & $0.459 \pm 0.000$ & $0.914 \pm 0.006$ \\
\bottomrule
\end{tabular}
\end{adjustbox}
\end{table}

%% file: sections/conditional_motif_protocol.tex
\noindent\textbf{Task and checkpoints.}
SENT, HDT-HAC-R and HDT-HAC-FR complete the same isolated chemical motif
using existing unconditionally trained GPT2-XS checkpoints, without conditional
fine-tuning. Every attempt, including retries, serializes the prompt afresh
under the variant's native randomization policy. Hierarchical prefixes leave
the final leaf open for continuation.

\noindent\textbf{Motifs, references and sampling.}
We retain motifs with at least 1,000 full-training-population matches from a
fixed catalog of 29 (Table~\ref{tab:conditional_motif_populations}).
All variants share fixed, training-derived motif-specific references:
canonical-deduplicated unions of the earlier source and reference pools,
capped at 2,000 molecules per motif. Source molecules are allowed.
These pools differ from the unconditional held-out references;
novelty uses the entire corresponding training population.

Sampling uses top-$k=10$, temperature 1, FP32 without autocast,
highest matmul precision, matmul TF32 disabled, and batch cap 256.
Table~\ref{tab:conditional_motif_populations} lists contexts.
Each motif/variant/seed cell retains the first 500 chemically valid outputs
within 10,000 attempts; every cell reached this target.
Duplicates, prompt copies, disconnected outputs and outputs that lose the
motif remain accepted if chemically valid, without growth or novelty filters.
Batch overshoot after the 500th success is excluded from all scores.

\input{tables/conditional_motif_populations}

\noindent\textbf{Metrics and aggregation.}
Metric names follow Appendix~\ref{app:metrics}; FCD, IntDiv,
\texttt{MoleculePGD} and the GuacaMol KL score are omitted.
All metrics and variants use the same retained motifs within each dataset.
 FG, SMARTS and Ring MMD$^2$ use unit-bandwidth Gaussian-kernel V-statistics; BRICS L2 is the Euclidean distance between normalized fragment-frequency
vectors. These scores, SNN, Frag and Scaff compare all 500 selected outputs,
retaining duplicate multiplicity, against the motif-specific training reference.
They are therefore not comparable with unconditional scores.

\noindent\textbf{Validity, uniqueness, novelty and V.U.N.}
For each cell, let $A$ count attempts through the 500th valid output,
$B$ be the multiset of 500 selected valid molecules, $S$ its set of distinct
canonical molecules, and $T$ the full training-molecule set. We report
\begin{gather*}
\mathrm{Validity}=\frac{500}{A},\qquad \mathrm{Uniqueness}=\frac{|S|}{500},\\
\mathrm{Novelty}=\frac{\sum_{b\in B}\mathbf{1}[b\notin T]}{500},\qquad
\mathrm{V.U.N.}=\frac{|S\setminus T|}{A}.
\end{gather*}
Validity uses attempts through the stopping point rather than a fixed sample;
novelty counts repeated outputs each time. V.U.N.\ uses the joint set-based definition
in Appendix~\ref{app:metrics}, with the stopping-prefix attempt count as denominator. Deduplication is within
cells, not across motifs or seeds. These scores do not require prompt retention
or growth; a prompt copy absent from the full-molecule training set counts as novel.

\paragraph{Reporting conventions.}
Tables~\ref{tab:appendix_conditional_motif_guacamol}--\ref{tab:appendix_conditional_motif_macrocycle_db}
report absolute scores and percent change relative to SENT.
Values are mean\,$\pm$\,sample SD over generation seeds 42--44 after equal-motif
averaging within each seed, using one checkpoint per variant.
\textbf{Bold} and \underline{underline} mark the best and second-best generated
models; $^{*}$ marks ours.  Means and SDs are rounded directly from full-precision aggregates. Lower-panel changes use the displayed means,
are signed so positive is better, and are shaded separately within each row.

\clearpage
\subsection{GuacaMol}
\noindent\textbf{Motifs (23; context 512):} benzene, pyridine, pyrimidine, cyclohexane, thiophene, quinoline, thiazole, furan, pyrrole, cyclopropane, cyclopentane, naphthalene, indole, imidazole, pyrazine, cyclohexene, oxazole, benzofuran, cyclobutane, benzothiophene, cyclopentene, acridine, fluorene.

\input{tables/appendix_conditional_motif_guacamol}

\FloatBarrier
\subsection{MOSES}
\noindent\textbf{Motifs (20; context 512):} benzene, pyridine, pyrimidine, thiazole, thiophene, cyclopropane, furan, cyclohexane, cyclopentane, pyrrole, quinoline, pyrazine, oxazole, imidazole, indole, cyclobutane, naphthalene, benzofuran, cyclohexene, cyclopentene.

\input{tables/appendix_conditional_motif_moses}

\clearpage
\subsection{COCONUT 2.0}
\noindent\textbf{Motifs (20; context 1,024):} benzene, cyclohexane, cyclopentane, cyclohexene, pyridine, pyrrole, furan, indole, pyrimidine, cyclopentene, naphthalene, cyclopropane, benzofuran, quinoline, imidazole, thiazole, thiophene, cyclobutane, pyrazine, oxazole.

\input{tables/appendix_conditional_motif_coconut2}

\FloatBarrier
\subsection{GEOM-Drugs (disjoint)}
\noindent\textbf{Motifs (14; context 1,024):} benzene, pyridine, pyrimidine, furan, thiophene, cyclohexane, quinoline, thiazole, imidazole, cyclopentane, pyrrole, cyclohexene, indole, naphthalene.

\input{tables/appendix_conditional_motif_geom_drugs_disjoint}

\clearpage
\subsection{Macrocycle}
\noindent\textbf{Motifs (11; context 2,048):} benzene, pyridine, pyrrole, indole, cyclohexane, cyclopropane, pyrimidine, cyclobutane, cyclopentane, imidazole, naphthalene.

\input{tables/appendix_conditional_motif_macrocycle_db}

\FloatBarrier

%% file: tables/conditional_motif_populations.tex
\begin{table}[h]
\caption{\textbf{Motif-conditioned evaluation populations.} A cell is one motif, variant and generation seed, with 500 selected valid outputs. Reference sizes count distinct motif-matching molecules. Training counts identify the full populations used for novelty.}
\label{tab:conditional_motif_populations}
\centering\scriptsize\setlength{\tabcolsep}{4pt}
\begin{adjustbox}{max width=\textwidth}\begin{tabular}{@{}lrrrrrr@{}}\toprule
Dataset & Training & Motifs & Context & Reference/motif & Cells & Valid outputs\\\midrule
GuacaMol & 499,968 & 23 & 512 & 1,029--2,000 & 207 & 103,500\\
MOSES & 499,968 & 20 & 512 & 1,153--2,000 & 180 & 90,000\\
COCONUT 2.0 & 367,104 & 20 & 1,024 & 1,351--2,000 & 180 & 90,000\\
GEOM-Drugs (disjoint) & 62,720 & 14 & 1,024 & 1,076--2,000 & 126 & 63,000\\
Macrocycle & 27,392 & 11 & 2,048 & 1,140--2,000 & 99 & 49,500\\
\midrule Total & & 88 & & & 792 & 396,000\\
\bottomrule\end{tabular}\end{adjustbox}\end{table}

%% file: tables/appendix_conditional_motif_guacamol.tex
\begin{table}[!htbp]
\caption{\textbf{GuacaMol: motif-conditioned generation.} Protocol and reporting conventions: Appendix~\ref{app:conditional_generation}.}
\label{tab:appendix_conditional_motif_guacamol}
\centering
\scriptsize
\setlength{\tabcolsep}{2.2pt}
\renewcommand{\arraystretch}{0.96}
\begin{adjustbox}{max width=\textwidth}
\begin{tabular}{@{}lccc@{}}
\toprule
 & \textbf{SENT} & \textbf{HDT-HAC-R}$^{*}$ & \textbf{HDT-HAC-FR}$^{*}$ \\
\midrule
\multicolumn{4}{@{}l}{\emph{Absolute performance}} \\
Validity $\uparrow$ & $0.965 \pm 0.001$ & \underline{$0.972 \pm 0.002$} & $\boldsymbol{0.988 \pm 0.001}$ \\
Uniqueness $\uparrow$ & $\boldsymbol{0.986 \pm 0.001}$ & $0.446 \pm 0.002$ & \underline{$0.802 \pm 0.003$} \\
Novelty $\uparrow$ & $\boldsymbol{0.991 \pm 0.001}$ & $0.694 \pm 0.002$ &  \underline{$0.904 \pm 0.000$} \\
V.U.N. $\uparrow$ & $\boldsymbol{0.947 \pm 0.002}$ & $0.423 \pm 0.001$ & \underline{$0.769 \pm 0.003$} \\
SNN $\uparrow$ &  $0.339 \pm 0.001$ & \underline{$0.352 \pm 0.001$} & $\boldsymbol{0.372 \pm 0.000}$ \\
Frag $\uparrow$ & $\boldsymbol{0.943 \pm 0.003}$ & $0.240 \pm 0.002$ & \underline{$0.615 \pm 0.006$} \\
Scaff $\uparrow$ & $0.336 \pm 0.002$ & \underline{$0.433 \pm 0.002$} & $\boldsymbol{0.464 \pm 0.003}$ \\
FG MMD$^2$ $\downarrow$ & $\boldsymbol{0.0054 \pm 0.0003}$ & $0.3676 \pm 0.0004$ & \underline{$0.0679 \pm 0.0007$} \\
SMARTS MMD$^2$ $\downarrow$ & $\boldsymbol{0.0115 \pm 0.0007}$ & $0.5003 \pm 0.0039$ & \underline{$0.1656 \pm 0.0012$} \\
Ring MMD$^2$ $\downarrow$ & $\boldsymbol{0.0137 \pm 0.0004}$ & $0.3832 \pm 0.0031$ & \underline{$0.3685 \pm 0.0022$} \\
BRICS L2 $\downarrow$ & $\boldsymbol{0.0484 \pm 0.0008}$ & $0.4134 \pm 0.0013$ & \underline{$0.1299 \pm 0.0008$} \\
\midrule
\multicolumn{4}{@{}l}{\emph{Direction-normalized change relative to SENT}} \\
Validity $\uparrow$ & $0.00\%$ & \cellcolor{deltagreen!19} $+0.73\%$ & \cellcolor{deltagreen!62} $+2.38\%$ \\
Uniqueness $\uparrow$ & $0.00\%$ & \cellcolor{deltared!62} $-54.77\%$ & \cellcolor{deltared!21} $-18.66\%$ \\
Novelty $\uparrow$ & $0.00\%$ & \cellcolor{deltared!62} $-29.97\%$ & \cellcolor{deltared!18} $-8.78\%$ \\
V.U.N. $\uparrow$ & $0.00\%$ & \cellcolor{deltared!62} $-55.33\%$ & \cellcolor{deltared!21}$-18.80\%$ \\
SNN $\uparrow$ & $0.00\%$ & \cellcolor{deltagreen!24} $+3.83\%$ & \cellcolor{deltagreen!62} $+9.73\%$ \\
Frag $\uparrow$ & $0.00\%$ & \cellcolor{deltared!62}$-74.55\%$ & \cellcolor{deltared!29} $-34.78\%$ \\
Scaff $\uparrow$ & $0.00\%$ & \cellcolor{deltagreen!47} $+28.87\%$ & \cellcolor{deltagreen!62} $+38.10\%$ \\
FG MMD$^2$ $\downarrow$ & $0.00\%$ & \cellcolor{deltared!62}$-6707.41\%$ & \cellcolor{deltared!11}$-1157.41\%$ \\
SMARTS MMD$^2$ $\downarrow$ & $0.00\%$ & \cellcolor{deltared!62}$-4250.43\%$ & \cellcolor{deltared!20}$-1340.00\%$ \\
Ring MMD$^2$ $\downarrow$ & $0.00\%$ & \cellcolor{deltared!62}$-2697.08\%$ & \cellcolor{deltared!60}$-2589.78\%$ \\
BRICS L2 $\downarrow$ & $0.00\%$ & \cellcolor{deltared!62}$-754.13\%$ & \cellcolor{deltared!14}$-168.39\%$ \\
\bottomrule
\end{tabular}
\end{adjustbox}
\end{table}

%% file: tables/appendix_conditional_motif_moses.tex
\begin{table}[!htbp]
\caption{\textbf{MOSES: motif-conditioned generation.} Protocol and reporting conventions: Appendix~\ref{app:conditional_generation}.}
\label{tab:appendix_conditional_motif_moses}
\centering
\scriptsize
\setlength{\tabcolsep}{2.2pt}
\renewcommand{\arraystretch}{0.96}
\begin{adjustbox}{max width=\textwidth}
\begin{tabular}{@{}lccc@{}}
\toprule
 & \textbf{SENT} & \textbf{HDT-HAC-R}$^{*}$ & \textbf{HDT-HAC-FR}$^{*}$ \\
\midrule
\multicolumn{4}{@{}l}{\emph{Absolute performance}} \\
Validity $\uparrow$ & $\boldsymbol{0.971 \pm 0.002}$ & $0.785 \pm 0.005$ &  \underline{$0.956 \pm 0.003$} \\
Uniqueness $\uparrow$ & $\boldsymbol{1.000 \pm 0.000}$ & $0.897 \pm 0.001$ & \underline{$0.990 \pm 0.001$} \\
Novelty $\uparrow$ & $\boldsymbol{0.993 \pm 0.001}$ & \underline{$0.992 \pm 0.001$} & $0.988 \pm 0.002$ \\
V.U.N. $\uparrow$ & $\boldsymbol{0.964 \pm 0.002}$ & $0.711 \pm 0.004$ & \underline{$0.936 \pm 0.003$} \\
SNN $\uparrow$ & $\boldsymbol{0.406 \pm 0.000}$ & $0.341 \pm 0.001$ & \underline{$0.373 \pm 0.000$} \\
Frag $\uparrow$ & $\boldsymbol{0.967 \pm 0.000}$ & $0.665 \pm 0.005$ & \underline{$0.873 \pm 0.001$} \\
Scaff $\uparrow$ & \underline{$0.410 \pm 0.011$} & $0.300 \pm 0.013$ & $\boldsymbol{0.470 \pm 0.003}$ \\
FG MMD$^2$ $\downarrow$ & $\boldsymbol{0.0043 \pm 0.0000}$ & $0.0274 \pm 0.0004$ & \underline{$0.0132 \pm 0.0002$} \\
SMARTS MMD$^2$ $\downarrow$ & $\boldsymbol{0.0106 \pm 0.0005}$ & $0.1011 \pm 0.0012$ & \underline{$0.0441 \pm 0.0003$} \\
Ring MMD$^2$ $\downarrow$ & $\boldsymbol{0.0106 \pm 0.0008}$ & \underline{$0.0604 \pm 0.0004$} & $0.1703 \pm 0.0017$ \\
BRICS L2 $\downarrow$ & $\boldsymbol{0.0467 \pm 0.0003}$ & $0.1495 \pm 0.0011$ & \underline{$0.0879 \pm 0.0003$} \\
\midrule
\multicolumn{4}{@{}l}{\emph{Direction-normalized change relative to SENT}} \\
Validity $\uparrow$ & $0.00\%$ & \cellcolor{deltared!62} $-19.16\%$ & \cellcolor{deltared!5} $-1.54\%$ \\
Uniqueness $\uparrow$ & $0.00\%$ & \cellcolor{deltared!62} $-10.30\%$ & \cellcolor{deltared!6} $-1.00\%$ \\
Novelty $\uparrow$ & $0.00\%$ & \cellcolor{deltared!12} $-0.10\%$ & \cellcolor{deltared!62} $-0.50\%$ \\
V.U.N. $\uparrow$ & $0.00\%$ & \cellcolor{deltared!62}$-26.24\%$ & \cellcolor{deltared!7} $-2.90\%$ \\
SNN $\uparrow$ & $0.00\%$ & \cellcolor{deltared!62} $-16.01\%$ & \cellcolor{deltared!31} $-8.13\%$ \\
Frag $\uparrow$ & $0.00\%$ & \cellcolor{deltared!62} $-31.23\%$ & \cellcolor{deltared!19} $-9.72\%$ \\
Scaff $\uparrow$ & $0.00\%$ & \cellcolor{deltared!62} $-26.83\%$ & \cellcolor{deltagreen!62} $+14.63\%$ \\
FG MMD$^2$ $\downarrow$ & $0.00\%$ & \cellcolor{deltared!62}$-537.21\%$ & \cellcolor{deltared!24}$-206.98\%$ \\
SMARTS MMD$^2$ $\downarrow$ & $0.00\%$ & \cellcolor{deltared!62}$-853.77\%$ & \cellcolor{deltared!23}$-316.04\%$ \\
Ring MMD$^2$ $\downarrow$ & $0.00\%$ & \cellcolor{deltared!19}$-469.81\%$ & \cellcolor{deltared!62}$-1506.60\%$ \\
BRICS L2 $\downarrow$ & $0.00\%$ & \cellcolor{deltared!62}$-220.13\%$ & \cellcolor{deltared!25}$-88.22\%$ \\
\bottomrule
\end{tabular}
\end{adjustbox}
\end{table}

%% file: tables/appendix_conditional_motif_coconut2.tex
\begin{table}[!htbp]
\caption{\textbf{COCONUT 2.0: motif-conditioned generation.} Protocol and reporting conventions: Appendix~\ref{app:conditional_generation}.}
\label{tab:appendix_conditional_motif_coconut2}
\centering
\scriptsize
\setlength{\tabcolsep}{2.2pt}
\renewcommand{\arraystretch}{0.96}
\begin{adjustbox}{max width=\textwidth}
\begin{tabular}{@{}lccc@{}}
\toprule
 & \textbf{SENT} & \textbf{HDT-HAC-R}$^{*}$ & \textbf{HDT-HAC-FR}$^{*}$ \\
\midrule
\multicolumn{4}{@{}l}{\emph{Absolute performance}} \\
Validity $\uparrow$ & \underline{$0.929 \pm 0.002$} &  $0.899 \pm 0.003$ & $\boldsymbol{0.967 \pm 0.001}$ \\
Uniqueness $\uparrow$ & $\boldsymbol{0.989 \pm 0.001}$ & $0.378 \pm 0.001$ & \underline{$0.692 \pm 0.002$} \\
Novelty $\uparrow$ & $\boldsymbol{0.977 \pm 0.001}$ & $0.602 \pm 0.003$ & \underline{$0.767 \pm 0.003$} \\
V.U.N. $\uparrow$ & $\boldsymbol{0.905 \pm 0.002}$ & $0.313 \pm 0.002$ & \underline{$0.608 \pm 0.004$} \\
SNN $\uparrow$ & $0.315 \pm 0.001$ & \underline{$0.378 \pm 0.002$} & $\boldsymbol{0.405 \pm 0.002}$ \\
Frag $\uparrow$ & $\boldsymbol{0.928 \pm 0.001}$ & $0.113 \pm 0.004$ & \underline{$0.419 \pm 0.002$} \\
Scaff $\uparrow$ & $0.441 \pm 0.004$ & \underline{$0.492 \pm 0.001$} & $\boldsymbol{0.506 \pm 0.001}$ \\
FG MMD$^2$ $\downarrow$ & $\boldsymbol{0.0049 \pm 0.0001}$ & $0.3848 \pm 0.0033$ & \underline{$0.1301 \pm 0.0021$} \\
SMARTS MMD$^2$ $\downarrow$ & $\boldsymbol{0.0108 \pm 0.0003}$ & $0.5474 \pm 0.0017$ & \underline{$0.2862 \pm 0.0021$} \\
Ring MMD$^2$ $\downarrow$ & $\boldsymbol{0.0205 \pm 0.0013}$ & $0.4542 \pm 0.0054$ & \underline{$0.4071 \pm 0.0019$} \\
BRICS L2 $\downarrow$ & $\boldsymbol{0.0530 \pm 0.0004}$ & $0.4407 \pm 0.0025$ & \underline{$0.1715 \pm 0.0014$} \\
\midrule
\multicolumn{4}{@{}l}{\emph{Direction-normalized change relative to SENT}} \\
Validity $\uparrow$ & $0.00\%$ & \cellcolor{deltared!62} $-3.23\%$ & \cellcolor{deltagreen!62} $+4.09\%$ \\
Uniqueness $\uparrow$ & $0.00\%$ & \cellcolor{deltared!62} $-61.78\%$ & \cellcolor{deltared!30} $-30.03\%$ \\
Novelty $\uparrow$ & $0.00\%$ & \cellcolor{deltared!62}$-38.38\%$ & \cellcolor{deltared!35} $-21.49\%$ \\
V.U.N. $\uparrow$ & $0.00\%$ & \cellcolor{deltared!62} $-65.41\%$ & \cellcolor{deltared!31} $-32.82\%$ \\
SNN $\uparrow$ & $0.00\%$ & \cellcolor{deltagreen!43} $+20.00\%$ & \cellcolor{deltagreen!62} $+28.57\%$ \\
Frag $\uparrow$ & $0.00\%$ & \cellcolor{deltared!62} $-87.82\%$ & \cellcolor{deltared!39} $-54.85\%$ \\
Scaff $\uparrow$ & $0.00\%$ & \cellcolor{deltagreen!49} $+11.56\%$ & \cellcolor{deltagreen!62} $+14.74\%$ \\
FG MMD$^2$ $\downarrow$ & $0.00\%$ & \cellcolor{deltared!62}$-7753.06\%$ & \cellcolor{deltared!20}$-2555.10\%$ \\
SMARTS MMD$^2$ $\downarrow$ & $0.00\%$ & \cellcolor{deltared!62}$-4968.52\%$ & \cellcolor{deltared!32}$-2550.00\%$ \\
Ring MMD$^2$ $\downarrow$ & $0.00\%$ & \cellcolor{deltared!62}$-2115.61\%$ & \cellcolor{deltared!55}$-1885.85\%$ \\
BRICS L2 $\downarrow$ & $0.00\%$ & \cellcolor{deltared!62}$-731.51\%$ & \cellcolor{deltared!19}$-223.58\%$ \\
\bottomrule
\end{tabular}
\end{adjustbox}
\end{table}

%% file: tables/appendix_conditional_motif_geom_drugs_disjoint.tex
\begin{table}[!htbp]
\caption{\textbf{GEOM-Drugs (disjoint): motif-conditioned generation.} Protocol and reporting conventions: Appendix~\ref{app:conditional_generation}.}
\label{tab:appendix_conditional_motif_geom_drugs_disjoint}
\centering
\scriptsize
\setlength{\tabcolsep}{2.2pt}
\renewcommand{\arraystretch}{0.96}
\begin{adjustbox}{max width=\textwidth}
\begin{tabular}{@{}lccc@{}}
\toprule
 & \textbf{SENT} & \textbf{HDT-HAC-R}$^{*}$ & \textbf{HDT-HAC-FR}$^{*}$ \\
\midrule
\multicolumn{4}{@{}l}{\emph{Absolute performance}} \\
Validity $\uparrow$ & \underline{$0.946 \pm 0.002$} & $0.893 \pm 0.002$ & $\boldsymbol{0.984 \pm 0.001}$ \\
Uniqueness $\uparrow$ &  $\boldsymbol{0.999 \pm 0.000}$ & $0.456 \pm 0.003$ & \underline{$0.857 \pm 0.008$} \\
Novelty $\uparrow$ & $\boldsymbol{0.998 \pm 0.000}$ & \underline{$0.989 \pm 0.001$} & $0.986 \pm 0.001$ \\
V.U.N. $\uparrow$ & $\boldsymbol{0.943 \pm 0.002}$ & $0.387 \pm 0.004$ & \underline{$0.835 \pm 0.009$} \\
SNN $\uparrow$ & \underline{$0.358 \pm 0.001$} &  $0.323 \pm 0.001$ & $\boldsymbol{0.367 \pm 0.001}$ \\
Frag $\uparrow$ & $\boldsymbol{0.938 \pm 0.004}$ & $0.354 \pm 0.004$ & \underline{$0.698 \pm 0.005$} \\
Scaff $\uparrow$ & $0.269 \pm 0.010$ & \underline{$0.314 \pm 0.002$} &  $\boldsymbol{0.388 \pm 0.001}$ \\
FG MMD$^2$ $\downarrow$ & $\boldsymbol{0.0041 \pm 0.0000}$ & $0.3107 \pm 0.0008$ & \underline{$0.0447 \pm 0.0016$} \\
SMARTS MMD$^2$ $\downarrow$ & $\boldsymbol{0.0099 \pm 0.0004}$ & $0.4314 \pm 0.0013$ & \underline{$0.1357 \pm 0.0019$} \\
Ring MMD$^2$ $\downarrow$ & $\boldsymbol{0.0169 \pm 0.0001}$ & $0.3800 \pm 0.0006$ & \underline{$0.3151 \pm 0.0030$} \\
BRICS L2 $\downarrow$ & $\boldsymbol{0.0526 \pm 0.0017}$ & $0.3774 \pm 0.0050$ & \underline{$0.1176 \pm 0.0014$} \\
\midrule
\multicolumn{4}{@{}l}{\emph{Direction-normalized change relative to SENT}} \\
Validity $\uparrow$ & $0.00\%$ & \cellcolor{deltared!62} $-5.60\%$ & \cellcolor{deltagreen!62}$+4.02\%$ \\
Uniqueness $\uparrow$ & $0.00\%$ & \cellcolor{deltared!62} $-54.35\%$ & \cellcolor{deltared!16} $-14.21\%$ \\
Novelty $\uparrow$ & $0.00\%$ & \cellcolor{deltared!47}$-0.90\%$ & \cellcolor{deltared!62} $-1.20\%$ \\
V.U.N. $\uparrow$ & $0.00\%$ & \cellcolor{deltared!62} $-58.96\%$ & \cellcolor{deltared!12} $-11.45\%$ \\
SNN $\uparrow$ & $0.00\%$ & \cellcolor{deltared!62} $-9.78\%$ & \cellcolor{deltagreen!62} $+2.51\%$ \\
Frag $\uparrow$ & $0.00\%$ & \cellcolor{deltared!62} $-62.26\%$ & \cellcolor{deltared!25} $-25.59\%$ \\
Scaff $\uparrow$ & $0.00\%$ & \cellcolor{deltagreen!23} $+16.73\%$ & \cellcolor{deltagreen!62} $+44.24\%$ \\
FG MMD$^2$ $\downarrow$ & $0.00\%$ & \cellcolor{deltared!62}$-7478.05\%$ & \cellcolor{deltared!8}$-990.24\%$ \\
SMARTS MMD$^2$ $\downarrow$ & $0.00\%$ & \cellcolor{deltared!62}$-4257.58\%$ & \cellcolor{deltared!19}$-1270.71\%$ \\
Ring MMD$^2$ $\downarrow$ & $0.00\%$ & \cellcolor{deltared!62}$-2148.52\%$ & \cellcolor{deltared!51}$-1764.50\%$ \\
BRICS L2 $\downarrow$ & $0.00\%$ & \cellcolor{deltared!62}$-617.49\%$ & \cellcolor{deltared!12}$-123.57\%$ \\
\bottomrule
\end{tabular}
\end{adjustbox}
\end{table}

%% file: tables/appendix_conditional_motif_macrocycle_db.tex
\begin{table}[!htbp]
\caption{\textbf{Macrocycle: motif-conditioned generation.} Protocol and reporting conventions: Appendix~\ref{app:conditional_generation}.}
\label{tab:appendix_conditional_motif_macrocycle_db}
\centering
\scriptsize
\setlength{\tabcolsep}{2.2pt}
\renewcommand{\arraystretch}{0.96}
\begin{adjustbox}{max width=\textwidth}
\begin{tabular}{@{}lccc@{}}
\toprule
 & \textbf{SENT} & \textbf{HDT-HAC-R}$^{*}$ & \textbf{HDT-HAC-FR}$^{*}$ \\
\midrule
\multicolumn{4}{@{}l}{\emph{Absolute performance}} \\
Validity $\uparrow$ & \underline{$0.782 \pm 0.002$} & $0.578 \pm 0.007$ & $\boldsymbol{0.801 \pm 0.003}$ \\
Uniqueness $\uparrow$ & $\boldsymbol{1.000 \pm 0.000}$ & $0.919 \pm 0.006$ & \underline{$0.963 \pm 0.001$} \\
Novelty $\uparrow$ & $\boldsymbol{1.000 \pm 0.000}$ & \underline{$0.998 \pm 0.000$} & $0.995 \pm 0.000$ \\
V.U.N. $\uparrow$ & $\boldsymbol{0.782 \pm 0.002}$ & $0.528 \pm 0.011$ &  \underline{$0.768 \pm 0.002$} \\
SNN $\uparrow$ & $\boldsymbol{0.338 \pm 0.002}$ & $0.252 \pm 0.001$ & \underline{$0.279 \pm 0.002$} \\
Frag $\uparrow$ & $\boldsymbol{0.864 \pm 0.001}$ & $0.426 \pm 0.006$ & \underline{$0.608 \pm 0.004$} \\
Scaff $\uparrow$ & \underline{$0.005 \pm 0.002$} & $0.000 \pm 0.000$ & $\boldsymbol{0.020 \pm 0.001}$ \\
FG MMD$^2$ $\downarrow$ & $\boldsymbol{0.0071 \pm 0.0000}$ & $0.0221 \pm 0.0005$ & \underline{$0.0178 \pm 0.0010$} \\
SMARTS MMD$^2$ $\downarrow$ & $\boldsymbol{0.0128 \pm 0.0003}$ & \underline{$0.0863 \pm 0.0016$} & $0.1062 \pm 0.0022$ \\
Ring MMD$^2$ $\downarrow$ & $\boldsymbol{0.0318 \pm 0.0006}$ & \underline{$0.0923 \pm 0.0012$} & $0.1151 \pm 0.0020$ \\
BRICS L2 $\downarrow$ & $\boldsymbol{0.0765 \pm 0.0001}$ & $0.1721 \pm 0.0014$ & \underline{$0.1234 \pm 0.0004$} \\
\midrule
\multicolumn{4}{@{}l}{\emph{Direction-normalized change relative to SENT}} \\
Validity $\uparrow$ & $0.00\%$ & \cellcolor{deltared!62} $-26.09\%$ & \cellcolor{deltagreen!62} $+2.43\%$ \\
Uniqueness $\uparrow$ & $0.00\%$ & \cellcolor{deltared!62} $-8.10\%$ & \cellcolor{deltared!28} $-3.70\%$ \\
Novelty $\uparrow$ & $0.00\%$ & \cellcolor{deltared!25}$-0.20\%$ & \cellcolor{deltared!62} $-0.50\%$ \\
V.U.N. $\uparrow$ & $0.00\%$ & \cellcolor{deltared!62} $-32.48\%$ & \cellcolor{deltared!3} $-1.79\%$ \\
SNN $\uparrow$ & $0.00\%$ & \cellcolor{deltared!62} $-25.44\%$ & \cellcolor{deltared!43} $-17.46\%$ \\
Frag $\uparrow$ & $0.00\%$ & \cellcolor{deltared!62} $-50.69\%$ & \cellcolor{deltared!36} $-29.63\%$ \\
Scaff $\uparrow$ & $0.00\%$ & \cellcolor{deltared!62} $-100.00\%$ & \cellcolor{deltagreen!62} $+300.00\%$ \\
FG MMD$^2$ $\downarrow$ & $0.00\%$ & \cellcolor{deltared!62}$-211.27\%$ & \cellcolor{deltared!44}$-150.70\%$ \\
SMARTS MMD$^2$ $\downarrow$ & $0.00\%$ & \cellcolor{deltared!49}$-574.22\%$ & \cellcolor{deltared!62}$-729.69\%$ \\
Ring MMD$^2$ $\downarrow$ & $0.00\%$ & \cellcolor{deltared!45}$-190.25\%$ & \cellcolor{deltared!62}$-261.95\%$ \\
BRICS L2 $\downarrow$ & $0.00\%$ & \cellcolor{deltared!62}$-124.97\%$ & \cellcolor{deltared!30}$-61.31\%$ \\
\bottomrule
\end{tabular}
\end{adjustbox}
\end{table}

%% file: checklist.tex
\section*{NeurIPS Paper Checklist}

\begin{enumerate}

\item {\bf Claims}
    \item[] Question: Do the main claims made in the abstract and introduction accurately reflect the paper's contributions and scope?
    \item[] Answer: \answerYes{} 
    \item[] Justification: Sections~\ref{sec:intro} and~\ref{sec:results} state the empirical contributions and their scope; Section~\ref{sec:results:limitations} and Appendix~\ref{app:limitations} discuss their limitations.

\item {\bf Limitations}
    \item[] Question: Does the paper discuss the limitations of the work performed by the authors?
    \item[] Answer: \answerYes{} 
    \item[] Justification:  Section~\ref{sec:results:limitations} and Appendix~\ref{app:limitations} discuss sequence costs, novelty and the dependence on coarsening and dataset.

\item {\bf Theory assumptions and proofs}
    \item[] Question: For each theoretical result, does the paper provide the full set of assumptions and a complete (and correct) proof?
    \item[] Answer: \answerNA{} 
    \item[] Justification: N/A

    \item {\bf Experimental result reproducibility}
    \item[] Question: Does the paper fully disclose all the information needed to reproduce the main experimental results of the paper to the extent that it affects the main claims and/or conclusions of the paper (regardless of whether the code and data are provided or not)?
    \item[] Answer: \answerYes{} 
    \item[] Justification:  Sections~\ref{sec:methods} and~\ref{sec:results:setup}, together with Appendices~\ref{app:hyperparameters}, \ref{app:metrics} and~\ref{app:conditional_generation}, specify the serialization, training and evaluation protocols.

\item {\bf Open access to data and code}
    \item[] Question: Does the paper provide open access to the data and code, with sufficient instructions to faithfully reproduce the main experimental results, as described in supplemental material?
    \item[] Answer: \answerYes{} 
    \item[] Justification: We will release the code upon acceptance.

\item {\bf Experimental setting/details}
    \item[] Question: Does the paper specify all the training and test details (e.g., data splits, hyperparameters, how they were chosen, type of optimizer) necessary to understand the results?
    \item[] Answer: \answerYes{} 
    \item[] Justification: Section~\ref{sec:results:setup} and Appendices~\ref{app:dataset_geometry}, \ref{app:hyperparameters} and~\ref{app:generation_details} give the populations, splits, optimizer, schedules, context caps and generation settings; hierarchy-selection rules are in Appendix~\ref{app:sbm_full}.

\item {\bf Experiment statistical significance}
    \item[] Question: Does the paper report error bars suitably and correctly defined or other appropriate information about the statistical significance of the experiments?
    \item[] Answer:  \answerYes{} 
    \item[] Justification:  Our graph generators are trained from scratch. For each evaluated checkpoint, we report the mean and sample SD of its three generation-seed scores ($42$--$44$), with variance denominator $2$. These SDs describe generation and, where applicable, reference-sampling variability, not variation across independently trained models; we do not claim significance across training runs.

\item {\bf Experiments compute resources}
    \item[] Question: For each experiment, does the paper provide sufficient information on the computer resources (type of compute workers, memory, time of execution) needed to reproduce the experiments?
    \item[] Answer: \answerYes{} 
    \item[] Justification:  Appendix~\ref{app:hyperparameters} reports the hardware, memory capacity and achieved training times. Scaling times are stated on a one-GPU basis in Appendix~\ref{app:regime_delta}.
    
\item {\bf Code of ethics}
    \item[] Question: Does the research conducted in the paper conform, in every respect, with the NeurIPS Code of Ethics \url{https://neurips.cc/public/EthicsGuidelines}?
    \item[] Answer: \answerYes{} 
    \item[] Justification: The research adheres to ethics guidelines.

\item {\bf Broader impacts}
    \item[] Question: Does the paper discuss both potential positive societal impacts and negative societal impacts of the work performed?
    \item[] Answer: \answerYes{} 
    \item[] Justification:  This paper studies graph generation as a machine-learning method. Potential applications include molecular design and biological modeling; we do not deploy the models or make clinical claims.
    
\item {\bf Safeguards}
    \item[] Question: Does the paper describe safeguards that have been put in place for responsible release of data or models that have a high risk for misuse (e.g., pre-trained language models, image generators, or scraped datasets)?
    \item[] Answer: \answerNA{} 
    \item[] Justification: N/A

\item {\bf Licenses for existing assets}
    \item[] Question: Are the creators or original owners of assets (e.g., code, data, models), used in the paper, properly credited and are the license and terms of use explicitly mentioned and properly respected?
    \item[] Answer: \answerYes{} 
    \item[] Justification:  We credit the original datasets and software in Sections~\ref{sec:background} and~\ref{sec:results:setup}, with dataset provenance and preprocessing in Appendix~\ref{app:dataset_geometry}.

\item {\bf New assets}
    \item[] Question: Are new assets introduced in the paper well documented and is the documentation provided alongside the assets?
    \item[] Answer:  \answerYes{} 
    \item[] Justification:  Public archives of \href{https://huggingface.co/datasets/MOSAIC-UCSD/MOSAIC_NeurIPS2026_data}{dataset caches}, \href{https://huggingface.co/datasets/MOSAIC-UCSD/MOSAIC_NeurIPS2026_checkpoints}{selected final checkpoints} and \href{https://huggingface.co/datasets/MOSAIC-UCSD/MOSAIC_NeurIPS2026_results}{raw results} provide repository cards, scope manifests and restore metadata. Generator laws are documented in Appendix~\ref{app:block_datasets}.

\item {\bf Crowdsourcing and research with human subjects}
    \item[] Question: For crowdsourcing experiments and research with human subjects, does the paper include the full text of instructions given to participants and screenshots, if applicable, as well as details about compensation (if any)? 
    \item[] Answer: \answerNA{} 
    \item[] Justification: N/A

\item {\bf Institutional review board (IRB) approvals or equivalent for research with human subjects}
    \item[] Question: Does the paper describe potential risks incurred by study participants, whether such risks were disclosed to the subjects, and whether Institutional Review Board (IRB) approvals (or an equivalent approval/review based on the requirements of your country or institution) were obtained?
    \item[] Answer: \answerNA{} 
    \item[] Justification: N/A

\item {\bf Declaration of LLM usage}
    \item[] Question: Does the paper describe the usage of LLMs if it is an important, original, or non-standard component of the core methods in this research? Note that if the LLM is used only for writing, editing, or formatting purposes and does \emph{not} impact the core methodology, scientific rigor, or originality of the research, declaration is not required.
    \item[] Answer: \answerYes{} 
    \item[] Justification: We used LLM coding agents to organize code, reimplement prior SOTA baselines, and generate figures, tables, and plots.

\end{enumerate}

%% file: references.bib
@article{weininger1988smiles,
  title={SMILES, a chemical language and information system. 1. Introduction to methodology and encoding rules},
  author={Weininger, David},
  journal={Journal of Chemical Information and Computer Sciences},
  volume={28},
  number={1},
  pages={31--36},
  year={1988},
  publisher={American Chemical Society}
}

@inproceedings{chen2025autograph,
  title={Flatten Graphs as Sequences: Transformers are Scalable Graph Generators},
  author={Chen, Dexiong and Krimmel, Markus and Borgwardt, Karsten},
  booktitle={Advances in Neural Information Processing Systems},
  year={2025},
  note={arXiv:2502.02216}
}

@article{polykovskiy2020moses,
	title = {Molecular {Sets} ({MOSES}): {A} {Benchmarking} {Platform} for {Molecular} {Generation} {Models}},
	volume = {Volume 11 - 2020},
	issn = {1663-9812},
	doi = {10.3389/fphar.2020.565644},
	journal = {Frontiers in Pharmacology},
	author = {Polykovskiy, Daniil and Zhebrak, Alexander and Sanchez-Lengeling, Benjamin and Golovanov, Sergey and Tatanov, Oktai and Belyaev, Stanislav and Kurbanov, Rauf and Artamonov, Aleksey and Aladinskiy, Vladimir and Veselov, Mark and Kadurin, Artur and Johansson, Simon and Chen, Hongming and Nikolenko, Sergey and Aspuru-Guzik, Alán and Zhavoronkov, Alex},
	year = {2020},
}

@inproceedings{rombach2022high,
  title={High-resolution image synthesis with latent diffusion models},
  author={Rombach, Robin and Blattmann, Andreas and Lorenz, Dominik and Esser, Patrick and Ommer, Bj{\"o}rn},
  booktitle={Proceedings of the IEEE/CVF conference on computer vision and pattern recognition},
  pages={10684--10695},
  year={2022}
}

@article{gpt3,
  title={Language models are few-shot learners},
  author={Brown, Tom and Mann, Benjamin and Ryder, Nick and Subbiah, Melanie and Kaplan, Jared D and Dhariwal, Prafulla and Neelakantan, Arvind and Shyam, Pranav and Sastry, Girish and Askell, Amanda and others},
  journal={Advances in neural information processing systems},
  volume={33},
  pages={1877--1901},
  year={2020}
}

@inproceedings{huang2024audiogpt,
  title={Audiogpt: Understanding and generating speech, music, sound, and talking head},
  author={Huang, Rongjie and Li, Mingze and Yang, Dongchao and Shi, Jiatong and Chang, Xuankai and Ye, Zhenhui and Wu, Yuning and Hong, Zhiqing and Huang, Jiawei and Liu, Jinglin and others},
  booktitle={Proceedings of the AAAI Conference on Artificial Intelligence},
  pages={23802--23804},
  year={2024}
}

@article{ingraham2019generative,
  title={Generative models for graph-based protein design},
  author={Ingraham, John and Garg, Vikas and Barzilay, Regina and Jaakkola, Tommi},
  journal={Advances in neural information processing systems},
  volume={32},
  year={2019}
}

@article{jo2023graph,
  title={Graph generation with diffusion mixture},
  author={Jo, Jaehyeong and Kim, Dongki and Hwang, Sung Ju},
  journal={arXiv preprint arXiv:2302.03596},
  year={2023}
}

@article{brockschmidt2018generative,
  title={Generative code modeling with graphs},
  author={Brockschmidt, Marc and Allamanis, Miltiadis and Gaunt, Alexander L and Polozov, Oleksandr},
  journal={arXiv preprint arXiv:1805.08490},
  year={2018}
}

@article{lim2020scaffold,
  title={Scaffold-based molecular design with a graph generative model},
  author={Lim, Jaechang and Hwang, Sang-Yeon and Moon, Seokhyun and Kim, Seungsu and Kim, Woo Youn},
  journal={Chemical science},
  volume={11},
  number={4},
  pages={1153--1164},
  year={2020},
  publisher={Royal Society of Chemistry}
}

@article{touvron2023llama,
  title={Llama: Open and efficient foundation language models},
  author={Touvron, Hugo and Lavril, Thibaut and Izacard, Gautier and Martinet, Xavier and Lachaux, Marie-Anne and Lacroix, Timoth{\'e}e and Rozi{\`e}re, Baptiste and Goyal, Naman and Hambro, Eric and Azhar, Faisal and others},
  journal={arXiv preprint arXiv:2302.13971},
  year={2023}
}

@article{dhariwal2020jukebox,
  title={Jukebox: A generative model for music},
  author={Dhariwal, Prafulla and Jun, Heewoo and Payne, Christine and Kim, Jong Wook and Radford, Alec and Sutskever, Ilya},
  journal={arXiv preprint arXiv:2005.00341},
  year={2020}
}

@article{touvron2023llama2,
  title={Llama 2: Open foundation and fine-tuned chat models},
  author={Touvron, Hugo and Martin, Louis and Stone, Kevin and Albert, Peter and Almahairi, Amjad and Babaei, Yasmine and Bashlykov, Nikolay and Batra, Soumya and Bhargava, Prajjwal and Bhosale, Shruti and others},
  journal={arXiv preprint arXiv:2307.09288},
  year={2023}
}

@article{chandrasekhar2025coconut2,
author = {Nainala, Venkata and Rajan, Kohulan and Kanakam, Sri and Sharma, Nisha and Weißenborn, Viktor and Schaub, Jonas and Steinbeck, Christoph},
year = {2024},
month = {10},
pages = {},
title = {COCONUT 2.0: A comprehensive overhaul and curation of the collection of open natural products database},
doi = {10.26434/chemrxiv-2024-fxq2s-v2}
}

@article{jiang2026macrocycle,
	title = {Macrocycle-{DB}: a comprehensive database for macrocycle-based drug discovery},
	volume = {54},
	issn = {1362-4962},
	url = {https://doi.org/10.1093/nar/gkaf1107},
	doi = {10.1093/nar/gkaf1107},
	number = {D1},
	journal = {Nucleic Acids Research},
	author = {Jiang, Minchuan and Liu, Tianyue and Hussain, Muzammal and Luo, Yixin and Zheng, Rui and Hou, Tingjun and Lu, Xiaoyun and Zhou, Yang},
	month = jan,
	year = {2026},
	pages = {D1469--D1476},
}

@article{axelrod2022geom,
  title={{GEOM}, energy-annotated molecular conformations for property prediction and molecular generation},
  author={Axelrod, Simon and G{\'o}mez-Bombarelli, Rafael},
  journal={Scientific Data},
  volume={9},
  number={1},
  pages={185},
  year={2022}
}

@article{dobson2003ddprotein,
  title={Distinguishing enzyme structures from non-enzymes without alignments},
  author={Dobson, Paul D. and Doig, Andrew J.},
  journal={Journal of Molecular Biology},
  volume={330},
  number={4},
  pages={771--783},
  year={2003}
}

@inproceedings{karami2024higen,
  title={{HiGen}: Hierarchical Graph Generative Networks},
  author={Karami, Mahdi},
  booktitle={The Twelfth International Conference on Learning Representations},
  year={2024}
}

@inproceedings{jin2018jtvae,
  title={Junction Tree Variational Autoencoder for Molecular Graph Generation},
  author={Jin, Wengong and Barzilay, Regina and Jaakkola, Tommi},
  booktitle={Proceedings of the 35th International Conference on Machine Learning},
  pages={2323--2332},
  year={2018}
}

@article{newman2004modularity,
  title={Finding and evaluating community structure in networks},
  author={Newman, Mark E. J. and Girvan, Michelle},
  journal={Physical Review E},
  volume={69},
  number={2},
  pages={026113},
  year={2004}
}

@article{preuer2018frechet,
  title={{Fr\'echet} {ChemNet} Distance: A metric for generative models for molecules in drug discovery},
  author={Preuer, Kristina and Renz, Philipp and Unterthiner, Thomas and Hochreiter, Sepp and Klambauer, G{\"u}nter},
  journal={Journal of Chemical Information and Modeling},
  volume={58},
  number={9},
  pages={1736--1741},
  year={2018}
}

@article{krimmel2025polygraph,
  title={{PolyGraph} Discrepancy: A classifier-based metric for graph generation},
  author={Krimmel, Markus and Hartout, Philip and Borgwardt, Karsten and Chen, Dexiong},
  journal={arXiv preprint arXiv:2510.06122},
  year={2025}
}

@inproceedings{vignac2023digress,
  title={{DiGress}: Discrete Denoising Diffusion for Graph Generation},
  author={Vignac, Cl{\'e}ment and Krawczuk, Igor and Siraudin, Antoine and Wang, Bohan and Cevher, Volkan and Frossard, Pascal},
  booktitle={International Conference on Learning Representations},
  year={2023}
}

@inproceedings{martinkus2022spectre,
  title={{SPECTRE}: Spectral Conditioning Helps to Overcome the Expressivity Limits of One-shot Graph Generators},
  author={Martinkus, Karolis and Loukas, Andreas and Perraudin, Nathana{\"e}l and Wattenhofer, Roger},
  booktitle={Proceedings of the 39th International Conference on Machine Learning},
  pages={15159--15179},
  year={2022}
}

@inproceedings{you2018graphrnn,
  title={{GraphRNN}: Generating Realistic Graphs with Deep Auto-regressive Models},
  author={You, Jiaxuan and Ying, Rex and Ren, Xiang and Hamilton, William L. and Leskovec, Jure},
  booktitle={Proceedings of the 35th International Conference on Machine Learning},
  pages={5708--5717},
  year={2018}
}

@article{liao2019efficient,
  title={Efficient graph generation with graph recurrent attention networks},
  author={Liao, Renjie and Li, Yujia and Song, Yang and Wang, Shenlong and Hamilton, Will and Duvenaud, David K and Urtasun, Raquel and Zemel, Richard},
  journal={Advances in neural information processing systems},
  volume={32},
  year={2019}
}

@inproceedings{goyal2020graphgen,
  title={{GraphGen}: A Scalable Approach to Domain-agnostic Labeled Graph Generation},
  author={Goyal, Nikhil and Jain, Harsh Vardhan and Ranu, Sayan},
  booktitle={Proceedings of The Web Conference 2020},
  pages={1253--1263},
  year={2020}
}

@inproceedings{bojchevski2018netgan,
  title={{NetGAN}: Generating graphs via random walks},
  author={Bojchevski, Aleksandar and Shchur, Oleksandr and Z{\"u}gner, Daniel and G{\"u}nnemann, Stephan},
  booktitle={International Conference on Machine Learning},
  pages={610--619},
  year={2018}
}

@inproceedings{niu2020scoregraph,
  title={Permutation Invariant Graph Generation via Score-Based Generative Modeling},
  author={Niu, Chenhao and Song, Yang and Song, Jiaming and Zhao, Shengjia and Grover, Aditya and Ermon, Stefano},
  booktitle={Proceedings of the Twenty Third International Conference on Artificial Intelligence and Statistics},
  pages={4474--4484},
  year={2020}
}

@inproceedings{kong2023autoregressivediffusion,
  title={Autoregressive Diffusion Model for Graph Generation},
  author={Kong, Lingkai and Cui, Jiaming and Sun, Haotian and Zhuang, Yuchen and Prakash, B. Aditya and Zhang, Chao},
  booktitle={Proceedings of the 40th International Conference on Machine Learning},
  pages={17391--17408},
  year={2023}
}

@misc{rdkit,
  title={{RDKit}: Open-source cheminformatics},
  author={{RDKit contributors}},
  howpublished={\url{https://www.rdkit.org}},
  year={2024},
  note={Version 2024.03.6},
  url={https://github.com/rdkit/rdkit/releases/tag/Release_2024_03_6}
 }

@article{wu2024tsmiles,
  title={{t-SMILES}: A fragment-based molecular representation framework for de novo ligand design},
  author={Wu, Juan-Ni and Wang, Tong and Chen, Yue and Tang, Li-Juan and Wu, Hai-Long and Yu, Ru-Qin},
  journal={Nature Communications},
  volume={15},
  number={1},
  pages={4993},
  year={2024},
  doi={10.1038/s41467-024-49388-6}
}

@article{10.1093/bib/bbae142,
    author = {Wang, Conghao and Ong, Hiok Hian and Chiba, Shunsuke and Rajapakse, Jagath C},
    eprint = {https://academic.oup.com/bib/article-pdf/25/3/bbae142/57160111/bbae142.pdf},
    issn = {1477-4054},
    journal = {Briefings in Bioinformatics},
    number = {3},
    volume = {25},
    title = {GLDM: hit molecule generation with constrained graph latent diffusion model},
    year = {2024}
}

@inproceedings{10.1145/3627673.3679547,
    author = {Bian, Tian and Niu, Yifan and Chang, Heng and Yan, Divin and Huang, Junzhou and Rong, Yu and Xu, Tingyang and Li, Jia and Cheng, Hong},
    booktitle = {Proceedings of the 33rd ACM International Conference on Information and Knowledge Management},
    isbn = {9798400704369},
    location = {Boise, ID, USA},
    numpages = {11},
    series = {CIKM '24},
    title = {Hierarchical Graph Latent Diffusion Model for Conditional Molecule Generation},
    year = {2024}
}

@article{10508504,
    author = {Yang, Ling and Huang, Zhilin and Zhang, Zhilong and Liu, Zhongyi and Hong, Shenda and Zhang, Wentao and Yang, Wenming and Cui, Bin and Zhang, Luxia},
    journal = {TKDE},
    number = {11},
    volume={36},
    title = {Graphusion: Latent Diffusion for Graph Generation},
    year = {2024}
}

@article{brown2019guacamol,
  title={{GuacaMol}: Benchmarking models for de novo molecular design},
  author={Brown, Nathan and Fiscato, Marco and Segler, Marwin H. S. and Vaucher, Alain C.},
  journal={Journal of Chemical Information and Modeling},
  volume={59},
  number={3},
  pages={1096--1108},
  year={2019}
}

@inproceedings{esser2024scaling,
    author = {Patrick Esser and
Sumith Kulal and
Andreas Blattmann and
Rahim Entezari and
Jonas M{\"{u}}ller and
Harry Saini and
Yam Levi and
Dominik Lorenz and
Axel Sauer and
Frederic Boesel and
Dustin Podell and
Tim Dockhorn and
Zion English and
Robin Rombach},
    booktitle = {International Conference on Machine Learning},
    title = {Scaling Rectified Flow Transformers for High-Resolution Image Synthesis},
    year = {2024}
}

@inproceedings{fu2024hyperbolic,
    author = {Xingcheng Fu and
Yisen Gao and
Yuecen Wei and
Qingyun Sun and
Hao Peng and
Jianxin Li and
Xianxian Li},
    booktitle = {International Conference on Machine Learning},
    title = {Hyperbolic Geometric Latent Diffusion Model for Graph Generation},
    year = {2024}
}

@inproceedings{ho2020denoising,
    author = {Jonathan Ho and
Ajay Jain and
Pieter Abbeel},
    booktitle = {Advances in Neural Information Processing Systems 33},
    title = {Denoising Diffusion Probabilistic Models},
    year = {2020}
}

@inproceedings{jangsimple,
    author = {Yunhui Jang and
Seul Lee and
Sungsoo Ahn},
    booktitle = {International Conference on Learning Representations},
    title = {A Simple and Scalable Representation for Graph Generation},
    year = {2024}
}

@inproceedings{jo2022scoregraph,
  title={Score-based Generative Modeling of Graphs via the System of Stochastic Differential Equations},
  author={Jo, Jaehyeong and Lee, Seul and Hwang, Sung Ju},
  booktitle={Proceedings of the 39th International Conference on Machine Learning},
  pages={10362--10383},
  year={2022}
}

@inproceedings{joshiall,
    author = {Joshi, Chaitanya K and Fu, Xiang and Liao, Yi-Lun and Gharakhanyan, Vahe and Miller, Benjamin Kurt and Sriram, Anuroop and Ulissi, Zachary Ward},
    booktitle = {International Conference on Machine Learning},
    title = {All-atom Diffusion Transformers: Unified generative modelling of molecules and materials},
    year = {2025}
}

@inproceedings{ketatalift,
    author = {Ketata, Mohamed Amine and Gao, Nicholas and Sommer, Johanna and Wollschl{\"a}ger, Tom and G{\"u}nnemann, Stephan},
    booktitle = {International Conference on Learning Representations},
    title = {Lift Your Molecules: Molecular Graph Generation in Latent Euclidean Space},
    year = {2025}
}

@article{kipf2016variational,
    author = {Kipf, Thomas N and Welling, Max},
    journal = {ArXiv preprint},
    title = {Variational graph auto-encoders},
    year = {2016}
}

@inproceedings{lipmanflow,
    author = {Yaron Lipman and
Ricky T. Q. Chen and
Heli Ben{-}Hamu and
Maximilian Nickel and
Matthew Le},
    booktitle = {International Conference on Learning Representations},
    title = {Flow Matching for Generative Modeling},
    year = {2023}
}

@inproceedings{nguyen2024glad,
    author = {Nguyen, Van Khoa and Boget, Yoann and Lavda, Frantzeska and Kalousis, Alexandros},
    booktitle = {Proceedings of the ICML 2024 Workshop SPIGM},
    organization = {OpenReview. net},
    title = {GLAD: improving latent graph generative modeling with simple quantization},
    year = {2024}
}

@article{parviz2026nonlinear,
  title={Nonlinear Laplacians Improve Signed-Directed Graph Learning},
  author={Parviz, Ali and Yoshida, Yuichi},
  journal={arXiv preprint arXiv:2608.00836},
  year={2026}
}

@article{sghaier2026probabilistic,
  title={Probabilistic Tiny Recursive Model},
  author={Sghaier, Amin and Parviz, Ali and Jolicoeur-Martineau, Alexia},
  journal={arXiv preprint arXiv:2605.19943},
  year={2026}
}

@article{huang2025large,
  title={Are large language models good temporal graph learners?},
  author={Huang, Shenyang and Parviz, Ali and Kondrup, Emma and Yang, Zachary and Ding, Zifeng and Bronstein, Michael and Rabbany, Reihaneh and Rabusseau, Guillaume},
  journal={arXiv preprint arXiv:2506.05393},
  year={2025}
}

@inproceedings{peebles2023scalable,
    author = {William Peebles and
Saining Xie},
    booktitle = {{IEEE/CVF} International Conference on Computer Vision, {ICCV} 2023,
Paris, France, October 1-6, 2023},
    title = {Scalable Diffusion Models with Transformers},
    year = {2023}
}

@article{mathiasen2024reducing,
  title={Reducing the cost of quantum chemical data by backpropagating through density functional theory},
  author={Mathiasen, Alexander and Helal, Hatem and Balanca, Paul and Krzywaniak, Adam and Parviz, Ali and Hvilsh{\o}j, Frederik and Banaszewski, Blazej and Luschi, Carlo and Fitzgibbon, Andrew William},
  journal={arXiv preprint arXiv:2402.04030},
  year={2024}
}

@inproceedings{gainski2026scalable,
	title = {Scalable and {Cost}-{Efficient} de {Novo} {Template}-{Based} {Molecular} {Generation}},
	volume = {38, Main Conference},
	doi = {10.52202/085713-1202},
	booktitle = {Advances in {Neural} {Information} {Processing} {Systems}},
	publisher = {Curran Associates, Inc.},
	author = {Gaiński, Piotr and Boussif, Oussama and Rekesh, Andrei and Shevchuk, Dmytro and Parviz, Ali and Tyers, Mike and Batey, Robert and Koziarski, Michał},
	editor = {Belgrave, D. and Zhang, C. and Lin, H. and Pascanu, R. and Koniusz, P. and Ghassemi, M. and Chen, N.},
	year = {2025},
	pages = {35787--35822},
}

@inproceedings{songscore,
    author = {Yang Song and
Jascha Sohl{-}Dickstein and
Diederik P. Kingma and
Abhishek Kumar and
Stefano Ermon and
Ben Poole},
    booktitle = {International Conference on Learning Representations},
    title = {Score-Based Generative Modeling through Stochastic Differential Equations},
    year = {2021}
}

@inproceedings{
  zhou2024unifying,
  title={Unifying Generation and Prediction on Graphs with Latent Graph Diffusion},
  author={Cai Zhou and Xiyuan Wang and Muhan Zhang},
  booktitle={The Thirty-eighth Annual Conference on Neural Information Processing Systems},
  year={2024},
  url={https://openreview.net/forum?id=lvibangnAs}
}

@inproceedings{maziarz2022moler,
  title={Learning to Extend Molecular Scaffolds with Structural Motifs},
  author={Maziarz, Krzysztof and Jackson-Flux, Henry and Cameron, Pashmina and Sirockin, Finton and Schneider, Nadine and Stiefl, Nikolaus and Segler, Marwin and Brockschmidt, Marc},
  booktitle={International Conference on Learning Representations},
  year={2022}
}
